\documentclass{article}

\usepackage{arxiv}

\usepackage[utf8]{inputenc} %
\usepackage[T1]{fontenc}    %
\usepackage{hyperref}       %
\usepackage{url}            %
\usepackage{booktabs}       %
\usepackage{amsfonts}       %
\usepackage{nicefrac}       %
\usepackage{microtype}      %
\usepackage{lipsum}         %

\newcommand{\reals}{{\mathbb{R}}}

\newcommand{\Tr}{\mathop{\bf Tr}}

\newcommand{\argmin}{\mathop{\rm argmin}}

\newcommand{\dquote}[1]{``#1''}

\usepackage{url}            %
\usepackage{booktabs}       %
\usepackage{multirow}    
\usepackage{amsfonts}       %
\usepackage{nicefrac}       %
\usepackage{microtype}      %
\usepackage{natbib}
\usepackage{enumerate}
\usepackage{hhline}
\usepackage{makecell}
\usepackage{pifont}

\usepackage{graphicx} %
\usepackage{caption}
\usepackage{subcaption}
\usepackage{amsmath}
\usepackage{amsthm}
\usepackage{amssymb}
\usepackage{tikz}
\usepackage{xcolor}
\usetikzlibrary{arrows}

\allowdisplaybreaks

\usepackage{mathrsfs}

\usepackage{algorithm}
\usepackage{algorithmic}
\usepackage{hyperref}
\usepackage{bm}

\allowdisplaybreaks

\newcommand{\labs}{\left\vert}
\newcommand{\rabs}{\right\vert}
\newcommand{\lnorm}{\left\Vert}
\newcommand{\rnorm}{\right\Vert}

\newcommand{\tr}{\operatorname{tr}}
\newcommand{\opt}{\mathrm{opt}}

\newcommand{\real}{\mathbb{R}}

\newcommand{\expect}{\mathbb{E}}

\newcommand{\indict}{\mathbb{I}}

\newtheorem{thm}{Theorem}
\newtheorem{lem}{Lemma}

\newtheorem{prop}{Proposition}
\newtheorem{asmp}{Assumption}
\newtheorem{defn}{Definition}

\newtheorem{fact}{Fact}
\newtheorem{conj}{Conjecture}
\newtheorem{rem}{Remark}
\newtheorem{example}{Example}

\newtheorem{claim}{Claim}

\usepackage[capitalize,noabbrev]{cleveref}
\crefname{thm}{Theorem}{Theorems}
\crefname{lem}{Lemma}{Lemmas}
\crefname{cor}{Corollary}{Corollaries}
\crefname{prop}{Proposition}{Propositions}
\crefname{asmp}{Assumption}{Assumptions}
\crefname{defn}{Definition}{Definitions}
\crefname{oracle}{Oracle}{Oracles}
\crefname{fact}{Fact}{Facts}
\crefname{conj}{Conjecture}{Conjectures}
\crefname{rem}{Remark}{Remarks}
\crefname{claim}{Claim}{Claims}

\renewcommand{\algorithmicrequire}{\textbf{Input:}}
\renewcommand{\algorithmicensure}{\textbf{Output:}}

\definecolor{red}{rgb}{1, 0, 0}
\newcommand{\RED}[1]{{\color{red} #1}}

\definecolor{green}{rgb}{0, 1, 0}

\definecolor{blue}{rgb}{0, 0, 1}
\newcommand{\BLUE}[1]{{\color{blue} #1}}

\definecolor{orange}{rgb}{1, 0.4, 0.0}

\input{packages/math_commands}

\usepackage{todonotes}

\title{On Generalization of Adversarial Imitation Learning and Beyond}

\AtBeginDocument{\setlength\abovedisplayskip{4pt}}
\AtBeginDocument{\setlength\belowdisplayskip{4pt}}

\newcommand{\piE}{\pi^{\operatorname{E}}}
\newcommand{\bc}{\operatorname{BC}}
\newcommand{\pibc}{\pi^{\bc}}
\newcommand{\ail}{\operatorname{AIL}}
\newcommand{\piail}{\pi^{\ail}}
\newcommand{\goodS}{\gS^{\operatorname{G}}}
\newcommand{\badS}{\gS^{\operatorname{B}}}

\renewcommand{\textsf}[1]{#1}

\usepackage{framed}
\colorlet{shadecolor}{orange!15}

\usepackage{enumitem}
\setlist{topsep=0pt,parsep=0pt,partopsep=0pt}

\usepackage{tabulary}

\hypersetup{
    colorlinks=true,
    citecolor=blue,
    linkcolor=blue,
}

\theoremstyle{plain}
\newtheorem{theorem}{Theorem}[section]

\newtheorem{corollary}[theorem]{Corollary}
\theoremstyle{definition}

\theoremstyle{remark}

\usepackage{authblk}
\newcommand\nnfootnote[1]{%
  \begin{NoHyper}
  \renewcommand\thefootnote{}\footnote{#1}%
  \addtocounter{footnote}{-1}%
  \end{NoHyper}
}
\author[1]{\textbf{Tian Xu}\textsuperscript{*}}
\author[2]{\textbf{Ziniu Li}\textsuperscript{*}}
\author[1]{\textbf{Yang Yu}\textsuperscript{\dag}}
\author[2]{\textbf{Zhi-Quan Luo}\textsuperscript{\dag}}
\affil[1]{National Key Laboratory for Novel Software Technology, Nanjing University, Nanjing 210023, China}
\affil[2]{Shenzhen Research Institute of Big Data, The Chinese University of Hong Kong, Shenzhen, Shenzhen 518172, China}
\affil[ ]{\texttt{xut@lamda.nju.edu.cn, ziniuli@link.cuhk.edu.cn, yuy@nju.edu.cn, luozq@cuhk.edu.cn}}

\renewcommand{\headeright}{}
\renewcommand{\shorttitle}{On Generalization of Adversarial Imitation Learning and Beyond}

\begin{document}
\maketitle

\begin{abstract}
Despite massive empirical evaluations, one of the fundamental questions in imitation learning is still not fully settled: does AIL (adversarial imitation learning) provably generalize better than BC (behavioral cloning)? We study this open problem with tabular and episodic MDPs. For vanilla AIL that uses the direct maximum likelihood estimation, we provide both negative and positive answers under the \emph{known} transition setting. For some MDPs, we show that vanilla AIL has a worse sample complexity than BC. The key insight is that the state-action distribution matching principle is weak so that AIL may generalize poorly even on \emph{visited} states from the expert demonstrations. For another class of MDPs, vanilla AIL is proved to generalize well even on \emph{non-visited} states. Interestingly, its sample complexity is horizon-free, which provably beats BC by a wide margin. Finally, we establish a framework in the \emph{unknown} transition scenario, which allows AIL to explore via reward-free exploration strategies. Compared with the best-known online apprenticeship learning algorithm, the resulting algorithm improves the sample complexity and interaction complexity.

\end{abstract}

\nnfootnote{*: Equal contribution.}
\nnfootnote{\dag: Corresponding authors.}

\section{Introduction}

Imitation learning approaches obtain the optimal policy from expert demonstrations \citep{argall2009survey, hussein2017survey, osa2018survey}. The classical approach Behavioral cloning (BC) performs imitation via supervised learning \citep{Pomerleau91bc}. This approach is simple and is widely applied in all kinds of applications \citep{ross11dagger, silver2016mastering, levin16_end_to_end}. However, BC is known to suffer compounding errors \citep{ross11dagger}. That is, the decision errors (due to imperfect imitation) accumulate over time steps under the sequential decision-making framework. This often explains the poor performance of BC when expert demonstrations are scarce.

Later on, generative adversarial imitation learning (GAIL) \citep{ho2016gail} is introduced. Different from BC, GAIL performs state-action distribution matching in an adversarial manner (i.e., min-max optimization). This idea is not novel as it has appeared in apprenticeship learning algorithms \citep{pieter04apprentice, syed07game}. With powerful neural networks, GAIL empirically outperforms BC by a wide margin. This motivates lots of practical advances \citep{fu2018airl, Kostrikov19dac, Brantley20disagreement, Kostrikov20value_dice, paul2020adversarial, dadashi2021primal, garg2021iq} and theoretical analysis \citep{sun2019provably, wang2020computation, zhang2020generative, rajaraman2020fundamental, nived2021provably, xu2021error, swamy2021moments, liu2021provably}. However, the following fundamental question is still not fully settled:

\begin{snugshade}
\begin{center}
    Does AIL provably generalize better than BC?
\end{center}
\end{snugshade}

\subsection{Problem Formulation}

Let us investigate this research question under the episodic Markov Decision Processes (MDPs) with finite states and actions. We study the generalization through the lens of sample complexity: the number of expert trajectories required to achieve a small policy value gap $\varepsilon$ between the expert policy $\piE$ and the learner's policy $\pi$, i.e.,  $\varepsilon := V^{\piE} - V^{\pi}$.

In the first place, let us consider the generalization of BC.  From a learning theory perspective, BC solves the empirical risk minimization (ERM) problem:
\begin{align}   \label{eq:bc}
    \min_{\pi \in \Pi} \sum_{h=1}^{H} \sum_{(s_h, a_h) \in \gD} | \pi_h(a_h |s_h) - \widehat{\piE_h} (a_h | s_h) |,
\end{align}
where $\gD$ is the expert dataset, $\Pi$ is the set of all stochastic policies, and $\widehat{\piE_h}$ is the empirical estimation for $\piE_h$ (refer to \cref{sec:preliminary}). One important fact is that BC performs ERM only on \emph{visited} states from the expert dataset. As a result, it is not expected to perform well on \emph{non-visited} states. Fundamentally, $\pibc_h(\cdot|s)$ for $s \notin \gD$ is not well defined by \eqref{eq:bc}. Conventionally, we set $\pibc_h(a | s) = 1/|\gA|$ for $s \notin \gD$. Thus, BC selects a wrong action with probability $1- 1/|\gA|$ on non-visited states and suffers a poor sample complexity $\gO(|\gS| H^2 / \varepsilon)$ \citep{rajaraman2020fundamental}.

As mentioned, the main punchline of AIL approaches is the state-action distribution matching principle. In particular, we consider the total variation distance to measure the state-action distribution discrepancy\footnote{One reason why we do not consider the KL divergence (i.e., $\KL(P^{\pi}_h || \widehat{P}^{\piE}_h)$) is that the problem becomes ill-conditioned when $P^{\pi}_h(s, a) = 0$.}, which leads to the following formulation:
\begin{align}  \label{eq:ail}
  \min_{\pi \in \Pi} \sum_{h=1}^{H} \sum_{(s, a) \in \gS \times \gA} | P^{\pi}_h(s, a) - \widehat{P}^{\piE}_h(s, a) |,
\end{align}
where $\widehat{P}^{\piE}_h$ is the maximum likelihood estimation for the state-action distribution $P^{\piE}_h$ (refer to \cref{sec:generalization_of_ail}). To avoid the confusion with advanced AIL algorithms introduced later, we call such an approach as \textsf{VAIL} (Vanilla AIL).

\textbf{Challenges.} Compared with BC's objective, \textsf{VAIL}'s objective has two differences: 1) it performs optimization on \emph{all} states; 2) it involves a multi-stage optimization. As such, \textsf{VAIL} has indirect guidance on \emph{non-visited} states. The bad news is that BC's objective is convex but \textsf{VAIL}'s objective is non-convex\footnote{Policy optimization is a non-convex problem for tabular MDPs \citep{agarwal2020pg} so it is not surprising that \textsf{VAIL}'s objective is also non-convex. Refer to the discussion in \cref{appendix:vail_objective_is_non_convex}.}. Thus, it is challenging to study the generalization performance of optimal policies obtained by \textsf{VAIL}.  Besides, even though we have obtained the upper bound of \textsf{VAIL}'s sample complexity,  the direct comparison with BC may not be insightful if the upper bounds are not sharp\footnote{\label{footnote:algorithm_comparison} In general, even if algorithm \textsf{A}'s sample complexity is better than algorithm \textsf{B}'s, we do not know whether \textsf{A} can beat \textsf{B} on a specific instance because the upper bounds may be loose. However, this comparison becomes meaningful if algorithm \textsf{B}'s sample complexity could be tight. For this case, we know there exists at least an instance that \textsf{A} is better than \textsf{B}.}.

\subsection{Our Contributions}

Our answers to the introduced open problem are summarized below (see \cref{table:summary_ail} and \cref{table:summary_tail} for a quick overview).

First, we prove that \textsf{VAIL}'s worst-case sample complexity is $\gO(|\gS| H^2/\varepsilon^2)$ by the reduction framework in \citep{xu2020error, rajaraman2020fundamental}, which provides a simple way of bypassing the non-convex difficulty. We also prove that this upper bound is sharp on some special instances\footnote{\dquote{Instance} means the underlying MDP and expert policy.} called \textsf{Standard Imitation}. Under these instances, BC has a refined sample complexity $\gO(|\gS| H /\varepsilon)$ due to the deterministic transitions of \textsf{Standard Imitation}. Accordingly, we obtain a negative answer that \textsf{VAIL} is inferior to BC on \textsf{Standard Imitation} in the worst case.

For the policy optimization in each time step, our analysis discloses two fundamental issues of \textsf{VAIL}: \emph{weak convergence} and \emph{sample barrier}. In particular, the weak convergence issue suggests that \textsf{VAIL} may make a wrong decision even on \emph{visited} states. In contrast, BC has no such an issue: it exactly recovers the expert action on \emph{visited} states. Furthermore, the sample barrier issue refers to that the statistical estimation error (i.e., $\sum_{(s, a)}\vert P^{\piE}_h(s, a) - \widehat{P}^{\piE}_h(s, a) \vert$ ) in \textsf{VAIL} diminishes at a rate $\gO(\sqrt{|\gS|/|\gD|})$, which is slower than BC's $\gO(|\gS|/|\gD|)$, where $|\gD|$ is the number of expert trajectories. Importantly, the above two issues definitely hold for \textsf{VAIL} on \emph{all} instances for the policy optimization in the last time step, where no future guidance is provided to mitigate the above issues.

Second, we demonstrate that \textsf{VAIL} can generalize well even on \emph{non-visited} states for another class of instances called \textsf{Reset Cliff}, which extends the lower bound instances for BC under the offline setting \citep{rajaraman2020fundamental}. In this scenario, a non-expert action would lead to a bad terminal state with reward 0. As a result, BC has compounding errors and its sample complexity $\gO(|\gS|H^2/\varepsilon)$ is sharp.  Surprisingly, we prove that \textsf{VAIL} has a \emph{horizon-free} sample complexity ${\gO}(|\gS|/\varepsilon^2)$, which is much better than BC in the regime $\varepsilon \geq {\gO}(1/H^2)$. Interestingly, even with one expert trajectory, \textsf{VAIL} can exactly recover the expert action on \emph{all} states in time steps before $H$. This is because \textsf{VAIL}'s objective in future time steps can provide effective guidance to resolve the weak convergence issue. Correspondingly, its optimality gap only comes from decision errors in the last time step, which is inevitable as discussed.

\vspace{-0.1cm}
\begin{table}[htbp]
    \centering
    \caption{Sample complexity of BC and \textsf{VAIL} on two types of MDPs: \textsf{Standard Imitation} and \textsf{Reset Cliff}. \textsf{VAIL} is worse than BC on \textsf{Standard Imitation} but is much better on \textsf{Reset Cliff}.}
    \label{table:summary_ail}

    {
    \begin{tabulary}{\linewidth}{l|C|C}
         &  \textsf{Standard Imitation}   & \textsf{Reset Cliff} \\
    \hline 
    BC  &  ${\gO}(|\gS| H/\varepsilon)$ & {$\Theta(|\gS| H^2/\varepsilon)$} \\ 
    \hline 
    \textsf{VAIL} &   $\Theta(|\gS| H^2/\varepsilon^2)$ & $\gO(|\gS|/\varepsilon^2)$
    \end{tabulary}
    }
\end{table}

To interpret our result, we discuss the MIMIC-MD algorithm in \citep{rajaraman2020fundamental}. Particularly, MIMIC-MD is also based on state-action distribution matching but it improves the generalization by mainly addressing the sample barrier issue. Concretely, MIMIC-MD has a sample complexity $\gO(|\gS|H^{3/2}/\varepsilon)$. Thus, the analysis in \citep{rajaraman2020fundamental} suggests that MIMIC-MD beats BC by $\gO(\sqrt{H})$. Instead, our analysis demonstrates that MIMIC-MD also enjoys a horizon-free sample complexity on Reset Cliff. To this end, we are the first to theoretically validate that AIL approaches (including \textsf{VAIL} and MIMIC-MD) could outperform BC by a wide margin (i.e., $\gO(H^2)$). Furthermore, our analysis is important to understand the empirical observation that AIL-style algorithms (e.g., GAIL) work well with a few expert demonstrations (e.g., 4 trajectories) for long-horizon tasks (e.g., $H=1000$) \citep{ho2016gail}. Technically, we overcome the non-convex difficulty by a novel dynamic programming based analysis rather than the reduction framework used before.

\vspace{-0.1cm}
\begin{table}[htbp]
\centering
\caption{Sample complexity and interaction complexity of OAL \citep{shani21online-al} and \textsf{MB-TAIL} in the unknown transition setting.  Compared with OAL, \textsf{MB-TAIL} improves both complexities.}
\label{table:summary_tail}
{
\begin{tabulary}{\linewidth}{l|C|C}
     & Sample Complexity   & Interaction Complexity \\
\hline 
OAL  &  $\widetilde{\gO}(|\gS|H^2/\varepsilon^2)$ & $\widetilde{\gO} (|\gS|^2 |\gA| H^4 / \varepsilon^2)$ \\
\hline 
\textsf{MB-TAIL} &   $\widetilde{\gO} ( |\gS|H^{3/2}  / \varepsilon )$  & $\widetilde{\gO} ( |\gS|^2 |\gA|H^3  / \varepsilon^2 )$
\end{tabulary}
}
\end{table}

Finally, we establish a framework to address the exploration issue in the unknown transition scenario. Previously, we implicitly assume the transition function is known for AIL algorithms so that they can directly calculate $P^{\pi}_h$ to perform optimization. Now, our framework allows these known-transition AIL approaches (e.g., VAIL and MIMIC-MD) to efficiently explore and imitate via the reward-free exploration strategies \citep{chi20reward-free}. In particular, we combine MIMIC-MD and the reward-free exploration method RF-Express \citep{menard20fast-active-learning} to obtain a new algorithm named \textsf{MB-TAIL}. The sample complexity of \textsf{MB-TAIL} is $\widetilde{\gO}(|\gS|H^{3/2}  / \varepsilon )$ while its interaction complexity is $\widetilde{\gO}( |\gS|^2 |\gA|H^3 / \varepsilon^2)$. Compared with the best-known online apprenticeship learning (OAL) algorithm \citep{shani21online-al} under the same setting, \textsf{MB-TAIL} has improvements in both complexities; refer to the summary in \cref{table:summary_tail}.

Proofs, related discussion, and empirical verification of our theoretical results can be found in Appendix.

\subsection{Related Work}  \label{sec:related_work}

As mentioned, apprenticeship learning algorithms such as FEM \citep{pieter04apprentice} and GTAL \citep{syed07game} amount to state-action distribution matching when the feature is selected as one-hot under tabular MDPs. Instead of the $\ell_1$-norm metric in \textsf{VAIL}, FEM and GTAL choose $\ell_2$-norm and $\ell_{\infty}$-norm metrics, respectively. Moreover, the sample complexity of both FEM and GTAL translates to ${\gO}(|\gS| H^2/\varepsilon^2)$, which is identical to the worst-case performance of \textsf{VAIL}. This supports that the metric for distribution discrepancy is not very essential under tabular MDPs. 

In contrast to the limited understanding of AIL approaches, the theoretical results of BC are adequate \citep{ross2010efficient, xu2020error, rajaraman2021value}. In particular, BC is shown to have compounding errors in \citep{ross11dagger}. Recently, \citet{rajaraman2020fundamental} derived the sample complexity ${\gO}(|\gS| H^2/ \varepsilon)$ for BC, which is minimax optimal in the offline setting.  Hence, we investigate the introduced open question under the case where the transition function is known or environment interaction is allowed.

We notice that a few attempts have been made to answer the introduced open question. For instance,  \citet{ghasemipour2019divergence} empirically validated that state-marginal matching objective in AIL matters for Gym MuJoCo locomotion tasks. However, their empirical study is unable to provide a satisfying answer due to the subsampling procedure; see the discussion in \citep{li2022rethinking}. In addition, the error bound analysis in \citep{xu2020error} indicates that AIL approaches enjoy a better dependence on the planning horizon than BC. Unfortunately, their analysis is at a \emph{population} level, which cannot tell us the generalization performance of \emph{empirical} minimizers. The information-theoretic results in \citep{rajaraman2020fundamental, nived2021provably, xu2021error} provide many insights but these results hold for all tabular MDPs, which fails to provide a fine-grained understanding.

Additional related work is reviewed in \cref{appendix:review_of_previous_work}.

\section{Preliminary}
\label{sec:preliminary}

\textbf{Episodic Markov Decision Process.} In this paper, we consider episodic Markov decision process (MDP), which can be described by the tuple $\gM = (\gS, \gA, \gP, r, H, \rho)$. Here $\gS$ and $\gA$ are the state and action space, respectively. $H$ is the planning horizon and $\rho$ is the initial state distribution. $\gP = \{P_1, \cdots, P_{H}\}$ specifies the non-stationary transition function of this MDP; concretely, $P_h(s_{h+1}|s_h, a_h)$ determines the probability of transiting to state $s_{h+1}$ conditioned on state $s_h$ and action $a_h$ in time step $h$, for $h \in [H]$\footnote{$[x]$ denotes the set of integers from $1$ to $x$.}. Similarly, $r = \{r_1, \cdots, r_{H}\}$ specifies the reward function of this MDP; without loss of generality, we assume that $r_h: \gS \times \gA \rar [0, 1]$, for $h \in [H]$. A non-stationary policy $\pi = \lb \pi_1, \cdots, \pi_h \rb$, where $\pi_h: \gS \rar \Delta(\gA)$ and $\Delta(\gA)$ is the probability simplex, $\pi_h (a|s)$ gives the probability of selecting action $a$ on state $s$ in time step $h$, for $h \in [H]$.

The quality of a policy is measured by its \emph{policy value} (i.e., the expected long-term return): $V^{\pi} := \expect[ \sum_{h=1}^{H} r_h(s_h, a_h) | s_1\sim \rho; a_h \sim \pi_h (\cdot|s_h), s_{h+1} \sim P_h(\cdot|s_h, a_h), \forall h \in [H] ]$.
To facilitate later analysis, we introduce the state-action distribution induced by a policy $\pi$: $P_h^{\pi}(s, a) := \sP^{\pi} ( s_h = s, a_h = a  )$. In other words, $P_h^{\pi}(s, a)$ quantifies the visitation probability of state-action pair $(s, a)$ in time step $h$. 

\textbf{Imitation Learning.} The goal of imitation learning is to learn a high quality policy directly from expert demonstrations. To this end, we often assume there is a nearly optimal expert policy $\piE$ that could interact with the environment to generate a dataset (i.e., $m$ trajectories of length $H$):
\begin{align*}
    \gD& = \{ \tr = \lp s_1, a_1, s_2, a_2, \cdots, s_H, a_H \rp; s_1 \sim \rho, a_h \sim \piE_h(\cdot|s_h), s_{h+1} \sim P_h(\cdot|s_h, a_h), \forall h \in [H] \}.
\end{align*}
Then, the learner can use the dataset $\gD$ to mimic the expert and to obtain a good policy. The quality of imitation is measured by the \emph{policy value gap}: $ V^{\piE} - V^{\pi}$. Following \citep{xu2020error, rajaraman2020fundamental}, we assume the expert policy is deterministic. 

\textbf{Notation.} We denote $\Pi$ as the set of all stochastic policies. For a trajectory $\tr$ in expert demonstrations $\gD$, $\tr (s_h)$ and $\tr(s_h, a_h)$ denote the specific state and state-action pair in time step $h$ in the trajectory $\tr$, respectively. Furthermore, $|\gD|$ is the number of trajectories in $\gD$.

With the above notations, we can write down the empirical estimation in BC: $\widehat{\piE_h}(a|s) = \sum_{\tr \in \gD}\mathbb{I}(\tr(s_h, a_h) = (s, a)) / \sum_{\tr \in \gD}\mathbb{I}(\tr(s_h) = s)$ if $\sum_{\tr \in \gD}\mathbb{I}(\tr(s_h) = s) > 0 $ and $\widehat{\piE_h}(a|s) = 1/|\gA|$ otherwise. In other words, it is  a \dquote{counting} based estimation. It is easy to see that $\widehat{\piE}$ is the optimal solution to \eqref{eq:bc}.

\section{Generalization of AIL}
\label{sec:generalization_of_ail}

In this section, we study the generalization performance of  adversarial imitation approaches. We mainly focus on conventional methods with the following maximum likelihood estimation for $P^{\piE}_h (s, a)$:
\begin{align}   \label{eq:estimate_by_count}
   \widehat{P}_h^{\piE} (s, a) := \frac{  \sum_{\tr \in \gD}  \indict\lb \tr(s_h, a_h) = (s, a) \rb }{|\gD|}.
\end{align}    
Such methods are widely applied in practice \citep{pieter04apprentice, syed07game, ho2016gail} and one archetype is  \textsf{VAIL} in \eqref{eq:ail}. To gain a complete understanding, we will discuss another type of AIL methods in \cref{subsec:beyond_vail}. We assume the transition function is known in this section\footnote{When the transition function is known, we additionally assume the initial state distribution is also known. In fact, this assumption is acceptable. To see this, without loss of generality, we can add an artificial state and its next state is sampled according to the initial state distribution. Then, we set this state as the fixed initial state. Under this setting, the new transition function contains the original initial state distribution information.}.

\subsection{When Does VAIL Generalize Poorly?}  \label{subsec:when_does_vail_generalize_poorly}

We can study the worst-case generalization performance of \textsf{VAIL} in the following way.  
\begin{itemize} 
    \item First, suppose $\piail$ is the minimizer of \eqref{eq:ail}, we show that the policy value gap $V^{\piE} - V^{\piail}$ is upper bounded by the estimation error, i.e., 
    \begin{align*}
        \labs  V^{\piE} - V^{\piail} \rabs &= \sum_{h=1}^{H}  \sum_{(s, a) \in \gS \times \gA} \labs P^{\piE}_h(s, a) r_h(s, a) - P^{\piail}_h(s, a) r_h(s, a) \rabs 
        \\ &\leq \sum_{h=1}^{H}  \sum_{(s, a) \in \gS \times \gA} \labs P^{\piE}_h(s, a) - P^{\piail}_h(s, a) \rabs \\
        &\leq \sum_{h=1}^{H}  \sum_{(s, a) \in \gS \times \gA} \labs P^{\piE}_h(s, a) - \widehat{P}^{\piE}_h(s, a) \rabs + \sum_{h=1}^{H}  \sum_{(s, a) \in \gS \times \gA} \labs  \widehat{P}^{\piE}_h(s, a) - P^{\piail}_h(s, a) \rabs \\
        &\leq 2 \sum_{h=1}^{H}  \sum_{(s, a) \in \gS \times \gA} \labs P^{\piE}_h(s, a) - \widehat{P}^{\piE}_h(s, a) \rabs,
    \end{align*}
    where the last inequality holds because $\piail$ is the optimal solution. 
    \item Second, we show that in expectation, the estimation error is well-controlled, i.e.,
    \begin{align}
    \label{eq:l1_concentration}
    \forall h \in [H], \, \expect\ls \lnorm \widehat{P}^{\piE}_h - P^{\piE}_h   \rnorm_{1} \rs \leq \sqrt{ \frac{|\gS| - 1}{|\gD|}},
    \end{align}
    which is an application of $\ell_1$ risk of the empirical distribution (see e.g., \citep[Theorem 1]{han2015minimax}). 
\end{itemize}

\begin{thm}[Sample Complexity of \textsf{VAIL}] \label{theorem:worst_case_sample_complexity_of_vail}
For any tabular and episodic MDP, to obtain an $\varepsilon$-optimal policy (i.e., $V^{\piE} - \expect[ V^{\piail}] \leq \varepsilon$), in expectation,  \textsf{VAIL} in \eqref{eq:ail} requires at most $\gO(|\gS|H^2/\varepsilon^2)$ expert trajectories. 
\end{thm}

Based on \cref{theorem:worst_case_sample_complexity_of_vail}, we want to know whether \textsf{VAIL} generalizes better than BC. In fact, we observe that \textsf{VAIL} tends to have a worse dependence on $\varepsilon$ in the sense that in the regime $\varepsilon < {\gO}(1)$, we have ${\gO}(|\gS|H^2/\varepsilon^2) > {\gO}(|\gS| H^2/\varepsilon)$, where the latter is the sample complexity for BC.  However, this does not directly imply \textsf{VAIL} is worse than BC; refer to the discussion in \cref{footnote:algorithm_comparison}. Instead, if we want the above comparison meaningful, we need to show that the sample complexity ${\gO}(|\gS| H^2/\varepsilon^2)$ is sharp for \textsf{VAIL}. To this end, we need two claims.
\begin{itemize} 
    \item \textbf{C1}: On certain MDPs, $\vert V^{\piE} - V^{\piail} \vert = c \cdot \sum_{h=1}^{H} \Vert P^{\piE}_h - \widehat{P}^{\piE}_h \Vert_1$, where $c > 0$ is some constant. 
    \item \textbf{C2}:  On the same MDPs, the convergence rate in \eqref{eq:l1_concentration} cannot be improved.
\end{itemize}
From the literature, we know \textbf{C2} is true (see e.g., \citep[Corollary 2]{han2015minimax} or \citep[Lemma 8]{kamath2015learning}). Hence, we only need to focus on \textbf{C1}. In particular, we show it is true on a class of instances called \textsf{Standard Imitation}, which corresponds to the lower bound instances for known-transition algorithms in \citep{rajaraman2020fundamental}.

\begin{asmp}[Standard Imitation]    \label{asmp:standard_imitation}
For a tabular and episodic MDP, we assume that 
\begin{itemize}
    \item Each state is absorbing and each action has the same transitions. i.e., $\forall (s, a) \in \gS \times \gA, h \in [H]$, we have $ P_h(s|s, a) = 1$.
    \item For any state, $a^{1}$ is the expert action with reward 1 and the others are non-expert actions with reward 0.
\end{itemize}
\end{asmp}

In the following \cref{proposition:ail_policy_value_gap_standard_imitation}, we prove that \textbf{C1} holds with $c=0.5$ for MDPs satisfying \cref{asmp:standard_imitation}.

\begin{prop}    
\label{proposition:ail_policy_value_gap_standard_imitation}
For any tabular and episodic MDP satisfying \cref{asmp:standard_imitation},  for each time step $h$, we define a set of states $\gS_h^1 := \{s \in \gS: \widehat{P}^{\piE}_h (s) < \rho (s)   \}$. Then, for each time step $h$, $\piail_h (a^1|s) \in [\widehat{P}^{\piE}_h(s) / \rho (s), 1], \forall s \in \gS_h^1$ and $\piail_h (a^1|s) = 1, \forall s \in (\gS_h^1)^c$ are all optimal solutions of \textsf{VAIL}'s objective \eqref{eq:ail}. Furthermore, the largest policy value gap among all globally optimal solutions is
\begin{align*}
    V^{\piE} - \expect[ V^{\piail}] = \frac{1}{2} \expect \ls \sum_{h=1}^H \lnorm \widehat{P}^{\piE}_h - P^{\piE}_h   \rnorm_1 \rs.  
\end{align*}
The optimal policy that incurs the largest policy value gap is $\piail_h (a^1|s) = \widehat{P}^{\piE}_h(s) / \rho (s), \forall s \in \gS_h^1$ and $\piail_h (a^1|s) = 1, \forall s \in (\gS_h^1)^c$.
\end{prop}

Based on \cref{proposition:ail_policy_value_gap_standard_imitation}, we immediately obtain the following lower bound.
\begin{prop}[Lower Bound for \textsf{VAIL}'s Sample Complexity]  \label{prop:lower_bound_vail}
To break the tie, suppose that \textsf{VAIL} outputs a policy $\piail$ uniformly sampled from all optimal solutions of \textsf{VAIL's} objective \eqref{eq:ail}. There exists a tabular and episodic MDP satisfying \cref{asmp:standard_imitation} such that to obtain an $\varepsilon$-optimal policy (i.e., $V^{\piE} - \expect[ V^{\piail}] \leq \varepsilon$), in expectation, \textsf{VAIL} in \eqref{eq:ail} requires at least $\Omega(|\gS| H^2/\varepsilon^2)$ expert trajectories. 
\end{prop}

Now, we can claim that \textsf{VAIL} generalizes worse than \textsf{BC} in the regime $\varepsilon < \gO(1)$ on certain instances satisfying \cref{asmp:standard_imitation}. But we realize that the lower bound MDPs for \textsf{VAIL} have deterministic transitions. For such MDPs, we actually can prove that \textsf{BC} has a better sample complexity. 

\begin{thm}[Sample Complexity of BC for Deterministic MDPs]  \label{theorem:bc_deterministic}
For any tabular and episodic MDP with deterministic transitions, to obtain an $\varepsilon$-optimal policy (i.e., $V^{\piE} - \expect[V^{\pibc}] \leq \varepsilon$), in expectation, BC in \eqref{eq:bc} requires at most ${\gO}(|\gS| H/\varepsilon)$ expert trajectories.  
\end{thm}

According to \cref{theorem:bc_deterministic}, we have a strong result that \textsf{VAIL} generalizes worse than BC in the whole regime $\varepsilon \in (0, H]$ on Standard Imitation. Next, we explain the underlying intuitions without delving into mathematical proofs.

\begin{rem}  \label{remark:standard_imtaition_issues}
We attribute the failure of \textsf{VAIL} to two issues: \underline{weak convergence} and \underline{sample barrier}. We explain them in plain language. 
\begin{itemize}  
    \item (Weak convergence)  \cref{proposition:ail_policy_value_gap_standard_imitation} shows that certain optimal solution of \textsf{VAIL} has a policy value gap, which is proportional to the state-action distribution estimation error. We mention that the estimation error is defined over all state-action pairs. This implies that \textsf{VAIL} could make a wrong decision even on \underline{visited} states from the expert demonstrations. In contrast, \textsf{BC} directly copies the expert action, which means \textsf{BC} never makes a mistake on a visited state.
    \item (Sample barrier) The statistical estimation error (i.e., $\Vert P^{\piE}_h - \widehat{P}^{\piE}_h \Vert_1 $) in \textsf{VAIL} diminishes at a slow rate $\gO(\sqrt{|\gS|/|\gD|})$. In contrast, BC's estimation error diminishes at a quicker speed, i.e., $\gO(|\gS|/|\gD|)$ \citep[Lemma A.1]{rajaraman2020fundamental}. This issue is also identified in \citep{rajaraman2020fundamental}.
\end{itemize}

The above two issues explain why \textsf{VAIL} could generalize poorly in a single time step. Finally, since each state is absorbing in the hard instance of \textsf{Standard Imitation}, future objectives cannot provide effective guidance for \textsf{VAIL}. Accordingly, \textsf{VAIL}'s sample complexity also has a quadratic dependence on $H$.
\end{rem}

We use the following  example to help readers better understand the issues of \textsf{VAIL}.

\begin{figure}[t]
\begin{minipage}[t]{0.45\textwidth}
\centering
\includegraphics[width=0.9\linewidth]{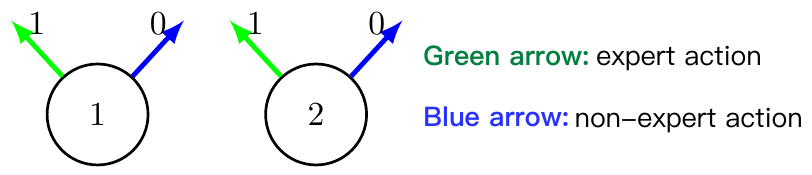}  
\caption{A simple MDP corresponding to \cref{example:ail_fail}.}
\label{fig:toy_bandit}
\end{minipage}
\hfill
\begin{minipage}[t]{0.45\textwidth}
\centering
\includegraphics[width=\linewidth]{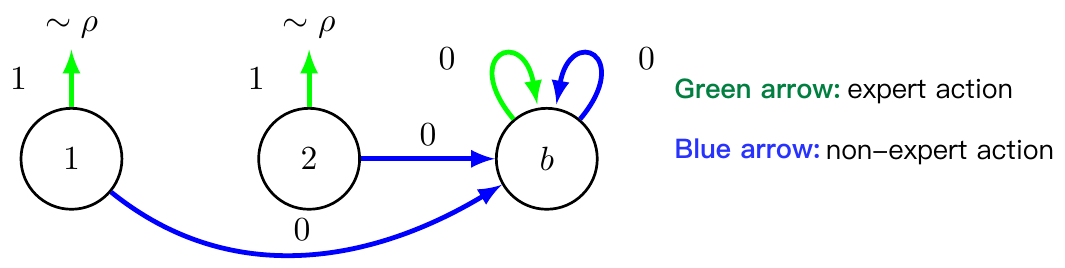}
\caption{A simple MDP corresponding to \cref{example:ail_success}.}
\label{fig:toy_reset_cliff}
\end{minipage}
\end{figure}

\begin{example}    \label{example:ail_fail}
Consider a simple MDP where $\gS = \{ s^{1}, s^{2} \}$ and $\gA = \{ a^{1}, a^{2} \}$; see \cref{fig:toy_bandit}. The reward information is $r (\cdot, a^{1}) = 1$ and $r (\cdot, a^{2}) = 0$. For simplicity, we let $H=1$ and omit the subscript. 
Suppose the expert takes action $a^{1}$ on each state and the initial state distribution $\rho = (0.5, 0.5)$. The agent is provided with 10 trajectories: 4 trajectories start from $s^{1}$ and the others start from $s^{2}$.

For BC, we  obtain that $\pibc (a^{1}|s^{1}) = \pibc (a^{1} | s^{2}) = 1$, which exactly recovers the expert policy. Thus, BC's policy value gap is 0.

For \textsf{VAIL}, it is easy to calculate the empirical distribution:
\begin{align*}
    \widehat{P}^{\piE} (s^{1}, a^{1}) = 0.4, \widehat{P}^{\piE} (s^{1}, a^{2}) = 0.0, \\ 
    \widehat{P}^{\piE} (s^{2}, a^{1}) = 0.6, \widehat{P}^{\piE} (s^{2}, a^{2}) = 0.0.
\end{align*}
Note that there are multiple globally optimal solutions for \textsf{VAIL}. For instance, $\pi(a^1 | s^1) = 0.8, \pi(a^2 | s^1) = 0.2, \pi(a^1 | s^2) = 1.0$, and
\begin{align*}
    P^{\pi}(s^1, a^1) = 0.4, P^{\pi}(s^1, a^2) = 0.1, \\
    P^{\pi}(s^2, a^1) = 0.5, P^{\pi}(s^2, a^2) = 0.0.
\end{align*}
For such an optimal policy, the empirical loss is $0.2$ and its policy value gap is $0.1$. 
\end{example}

Before we finish the discussion, we clarify the relationship between our results and the information-theoretic analysis in \citep{rajaraman2020fundamental}. First, \citet{rajaraman2020fundamental} conjectured that the convention AIL approaches cannot achieve a better sample complexity $\gO(|\gS|H^{3/2}/\varepsilon)$ as MIMIC-MD does; refer to Remark 5.3 in their paper. In fact, our upper bound and lower bound suggest this conjecture is true.  Second,  \textsf{Standard Imitation} instances are first developed by \citet{rajaraman2020fundamental} to derive the algorithm-independent lower bound $\Omega(|\gS| H/\varepsilon)$ while our lower bound in \cref{prop:lower_bound_vail} only holds for \textsf{VAIL} but it is tighter.

\subsection{When Does VAIL Generalize Well?}
\label{subsec:ail_generalize_well}

In the previous part, we provide a negative answer that on some special MDPs, \textsf{VAIL} is worse than BC. However, massive empirical evaluations suggest that AIL methods can surpass BC. In an attempt to obtain a positive answer, we revisit the lower bound instances for BC under the offline setting  \citep{rajaraman2020fundamental}. We use the following simple example to present our idea.

\begin{example} \label{example:ail_success}
Consider a simple MDP, where there are three states $(s^{1}, s^{2}, s^{3})$ and two actions $(a^{1}, a^{2})$; see \cref{fig:toy_reset_cliff}.  In particular, $s^{1}$ and $s^{2}$ are good states while $s^{3}$ is a bad absorbing state. Suppose $H=2$ and $\rho = (0.5, 0.5, 0.0)$. Other information: action $a^{1}$ resets the next state according to $\rho$ on $s^{1}$ and $s^{2}$ (with reward 1); action $a^{2}$ deterministically leads to the absorbing state $s^{3}$ on $s^{1}$ and $s^{2}$ (with reward 0); any action yields reward 0 on the absorbing state $s^{3}$. Moreover, the expert policy always takes action $a^{1}$. The agent is provided only 2 expert trajectories: $\tr_1 = (s^{1}, a^{1}) \rar (s^{1}, a^{1}) $ and $\tr_2 = (s^{1}, a^{1}) \rar (s^{2}, a^{1})$.

For BC, it exactly recovers the expert action on visited states but it poses a uniform policy on the \underline{non-visited} $s^{2}$ in time step $h = 1$. As a result, BC makes a mistake with probability $\rho(s^{2}) \cdot \pi_1^{\bc}(a^{2} | s^{2}) = 0.25$. Accordingly, its policy value gap is $0.25 \cdot 2 = 0.5$.

For \textsf{VAIL}, it make senses to guess that the expert action is recovered on visited states. Interestingly, we argue that \textsf{VAIL} exactly recovers the expert action even on \underline{non-visited} state $s^{2}$ in time step $h=1$. Consequently, the policy value gap of \textsf{VAIL} is 0, which is smaller than BC. The formal proof of the above argument is a little tricky (refer to \cref{appendix:warm_up_of_proposition_reset_cliff}) and we explain the intuition here.

Assume $\pi_{1}(a^{2} | s^{2}) = 1 - \beta$, where $\beta \in [0, 1]$. We note that a positive $\beta$ makes no difference for the loss function in time step $h = 1$, since 
\begin{align*}
    &|P^{\pi}_1(s^{2}, a^{1}) - \widehat{P}^{\piE}_1(s^{2}, a^{1}) | + |P^{\pi}_1(s^{2}, a^{2}) - \widehat{P}^{\piE}_1(s^{2}, a^{2}) | = |P^{\pi}_1(s^{2}, a^{1}) - 0 | + |P^{\pi}_1(s^{2}, a^{2}) - 0 | = \rho(s^{2}).
\end{align*}
However, it matters for the loss function in time step $h=2$. Recall the \dquote{transition flow equation}:
\begin{align*}
    P^{\pi}_h(s) = \sum_{(s^\prime, a^\prime)} P^{\pi}_{h-1} (s^\prime) \pi_{h-1} (a^\prime|s^\prime) P_{h-1}(s | s^\prime, a^\prime).
\end{align*}
Provided $\pi$ takes the expert action elsewhere, we can calculate the loss function for $\beta$: 
\begin{align*}
    \text{Loss} (\beta) &= \sum_{h=1}^{2} \sum_{(s, a)} | P^{\pi}_h(s, a) - \widehat{P}^{\piE}_h(s, a) | 
    = 2 - \beta,
\end{align*}
which has a unique globally optimal solution at $\beta = 1$.

\end{example}

Let us further explain the phenomenon in \cref{example:ail_success}. The empirical state-action distribution of the expert policy concentrates on \dquote{good} states (i.e., non-absorbing states). However, the estimation error is still non-negative due to some non-visited \dquote{good} states. The magic is that the objective of \textsf{VAIL} in \eqref{eq:ail} prefers policies that can match these visited good states. Importantly, to maintain its status on \dquote{good} states, \textsf{VAIL} has to select the expert action even on non-visited states. Otherwise, it goes to the bad absorbing state and suffers a large loss. In this way, \textsf{VAIL} recovers the expert policy on almost all states. In contrast, BC makes a mistake on \emph{non-visited} states with a positive probability. Hence, these two approaches have dramatically different generalization performances in the above example.

We can extend the simple setting in \cref{example:ail_success} by the following conditions; see  \cref{fig:reset_cliff} in \cref{appendix:reset_cliff_and_ail_properties}.

\begin{asmp}[Reset Cliff]   \label{asmp:reset_cliff}
For a tabular and episodic MDP, we assume that 
\begin{itemize}   
    \item State space is divided into the sets of \dquote{good} states and \dquote{bad} states, i.e., $\gS = \goodS \cup \badS, \goodS \cap \badS = \emptyset$.
    \item For any good state, $a^1$ is the expert action with reward 1 and the others are non-expert actions with reward 0. For any bad state, all actions have reward 0.
    \item For action $a_1$, we have for any state $s, s^\prime \in \goodS$ and $h \in [H]$, $P_h(s^\prime|s, a^{1}) > 0$. 
    \item For action $a \ne a^{1}$, we have for any state $s, s^\prime \in \goodS$ and $h \in [H]$, $P_h(s^\prime|s, a) = 0$.
    \item All bad states only have transitions to themselves. 
\end{itemize}
\end{asmp}

Since instances satisfying \cref{asmp:reset_cliff} include the lower bound instances for BC \citep{rajaraman2020fundamental}, we know that the sample complexity ${\gO}(|\gS| H^2/\varepsilon)$ is tight for BC. How about \textsf{VAIL}?  Surprisingly, we prove that even only with 1 expert trajectory, \textsf{VAIL} exactly recovers the expert action in the first $H - 1$ time steps.

\begin{prop}
\label{prop:ail_general_reset_cliff}
For any tabular and episodic MDP satisfying \cref{asmp:reset_cliff}, with any fixed unbiased estimation $\widehat{P}^{\piE}_h (s, a)$ established from $\gD$, suppose that $\piail$ is any minimizer of \eqref{eq:ail}. When $\vert \gD \vert \geq 1$, we have the following optimality condition almost surely:
\begin{align*}
    \piail_h (a^1|s) = 1, \forall s \in \goodS, h \in [H-1].
\end{align*}
\end{prop}

Despite the fact that policy optimization for tabular MDPs is a non-convex problem \citep{agarwal2020pg}, \cref{prop:ail_general_reset_cliff} states that the optimal solution of \textsf{VAIL} in the first $H-1$ time steps is unique. In fact, our analysis overcomes the non-convex difficulty by the dynamic programming technique \citep{bertsekas2012dynamic}. To the best of our knowledge, this kind of analysis is new for the imitation learning area.

We remark that in the last time step $h = H$, \textsf{VAIL} still suffers issues as discussed in \cref{remark:standard_imtaition_issues}. However, since the transitions are non-trivial on \textsf{Reset Cliff}, we obtain a \emph{horizon-free} sample complexity for \textsf{VAIL}. 

\begin{thm}[Horizon-free Sample Complexity of \textsf{VAIL} on \textsf{Reset Cliff}]     \label{theorem:ail_reset_cliff}
For any tabular and episodic MDP satisfying \cref{asmp:reset_cliff}, to obtain an $\varepsilon$-optimal policy (i.e., $V^{\piE} - \expect[V^{\piail}] \leq \varepsilon$), in expectation, \textsf{VAIL} requires at most ${\gO}(|\gS|/\varepsilon^2)$ expert trajectories.
\end{thm}

\cref{theorem:ail_reset_cliff} indicates that \textsf{VAIL} beats BC in the regime $\varepsilon \geq {O}(1/H^2)$, which provides a positive answer for the introduced open problem. This comparison is meaningful since BC's sample complexity is sharp under this setting. Note that this horizon-free sample complexity is not obtained by the conventional reduction based analysis in \cref{subsec:when_does_vail_generalize_poorly}. Instead, it is based on the optimality condition in \cref{prop:ail_general_reset_cliff}.

We mention that many continuous control tasks as in \citep{duan2016benchmarking} are similar to \textsf{Reset Cliff}: once taking a non-expert/wrong action, the robot falls down and goes into a bad terminal/absorbing state. Therefore, \cref{theorem:ail_reset_cliff} is important to help us understand the empirical observation \citep{ho2016gail, ghasemipour2019divergence, Kostrikov19dac} that AIL-type algorithms (e.g., GAIL and DAC) work well with a few expert demonstrations (e.g., 4 trajectories) for long-horizon tasks (e.g., $H=1000$) and increasing the number of samples does not improve the performance a lot\footnote{In fact, this empirical observation is based on \emph{subsampled} trajectories, which is slightly different from the \emph{complete} trajectory setting in \cref{theorem:ail_reset_cliff}. However, the implication
is expected to hold for subsampled trajectories; see the discussion in \cref{appendix:further_discssuion_of_example_success_subsampling}. }.

In the last, readers may notice that we have put much effort in discussing the generalization performance of exactly optimal solutions. In practice, people usually use gradient based methods to obtain an approximately optimal solution. However, the conclusions do not change if the optimization error is small; see \cref{appendix:proof_of_the_approximate_solution_in_VAIL}.

\subsection{Beyond VAIL}    \label{subsec:beyond_vail}

In this section, we discuss the MIMIC-MD algorithm in \citep{rajaraman2020fundamental}. MIMIC-MD is also a state-action distribution matching based method, which improves the generalization via \emph{estimation}. Specifically, this approach leverages the transition function to obtain a more accurate estimation for the expert state-action distribution. Though MIMIC-MD is already analyzed in \citep{rajaraman2020fundamental}, we provide some new conclusions.

For ease of exposition, let us introduce several notations.
\begin{itemize}  
    \item $\tr_h$: the truncated trajectory up to time step $h$, i.e., $\tr_h = (s_1, a_1, \cdots, s_h, a_h)$.
    \item $\gS_{h}(\gD)$: the set of states visited in time step $h$ in $\gD$. 
    \item $\Tr_h^{\gD} = \lb \tr_h: \tr_h (s_\ell) \in \gS_{\ell} (\gD), \forall \ell \in [h] \rb$: the trajectories along which each state has been visited in $\mathcal{D}$ up to time step $h$.
\end{itemize}

Now, consider the dataset $\gD$ is randomly divided into two parts, i.e., $\gD = \gD_1 \cup \gD_1^{c}$. The estimator in \citep{rajaraman2020fundamental} is:
\begin{align} 
& \widetilde{P}_h^{\piE}  (s, a) =   { \sum_{\tr_h \in \Tr_h^{\gD_1} } \sP^{\piE}(\tr_h) \indict\lb \tr_h(s_h, a_h) = (s, a)\rb} + {\frac{  \sum_{\tr_h \in \gD_1^c}  \indict\{ \tr_h (s_h, a_h) = (s, a), \tr_h \not\in \Tr_h^{\gD_1}  \} }{|\gD_1^c|}}, \label{eq:new_estimator}
\end{align}
where $\sP^{\piE} (\tr_h)$ is the probability of the truncated trajectory $\tr_h$ induced by the deterministic expert policy $\piE$. 
To get a better intuition, consider the following key decomposition of $P^{\piE}_h (s, a)$ from a trajectory space view:
\begin{align} 
&P_h^{\piE}(s, a) = {\sum_{\tr_h \in \Tr_h^{\gD_1}} \sP^{\piE}(\tr_h) \indict\lb \tr_h(s_h, a_h) = (s, a) \rb} + {\sum_{\tr_h \notin \Tr_h^{\gD_1}} \sP^{\piE}(\tr_h) \indict\lb \tr_h(s_h, a_h) = (s, a) \rb}. \label{eq:key_decomposition}
\end{align}
First of all, we see that the first term in the estimator \eqref{eq:new_estimator} is exactly the first part in \eqref{eq:key_decomposition}. For this part, all state-action pairs up to time step $h$ are known from $\gD_1$. Therefore, we can compute $\sP^{\piE} (\tr_h)$ exactly as the transition function is known\footnote{$\sP^{\piE}(\tr_h) = \rho(\tr(s_1)) \prod_{\ell=1}^{h-1} P_{\ell}(\tr(s_{\ell+1})|\tr(s_\ell), \tr(a_\ell))$.}. Secondly, for the second term in \eqref{eq:key_decomposition} (i.e., $\tr_h \not\in \Tr_h^{\gD_1}$), we cannot exactly calculate $\sP^{\piE} (\tr_h)$ since we do not know some actions in $\tr_h$ from $\gD_1$. Fortunately, we can leverage the complementary dataset $\gD^c_1$ to establish an estimator. In fact, the second term in \eqref{eq:new_estimator} is a maximum likelihood estimation for the associated part in \eqref{eq:key_decomposition}. Finally, since the estimator in \eqref{eq:new_estimator} utilizes the transition function information explicitly, it has a better statistical guarantee.  

\begin{thm}[\citep{rajaraman2020fundamental}]   \label{lemma:sample_complexity_improved_estimator_in_expectation}
For any tabular and episodic MDP, assume $\gD$ is randomly divided into two subsets, i.e., $\gD = \gD_{1} \cup \gD_{1}^c$ with $| \gD_1 | = | \gD_1^{c} | = |\gD| / 2$. To obtain an $\varepsilon$-optimal estimation (i.e., $\sum_{h=1}^{H} \Vert P^{\piE}_h - \widetilde{P}^{\piE}_h \Vert_1 \leq \varepsilon$), in expectation, the estimator in \eqref{eq:new_estimator} requires at most $\min\{ \gO(|\gS|H^2/\varepsilon^2), \gO(|\gS|H^{3/2}/\varepsilon)\}$ expert trajectories.
\end{thm}

When $\varepsilon$ is small, the second term dominates in \cref{lemma:sample_complexity_improved_estimator_in_expectation}. As a result, the estimator in \eqref{eq:new_estimator} has improvements in both $\varepsilon$ and $H$ compared with the direct maximum likelihood estimation in \eqref{eq:l1_concentration}. The improvement in $\varepsilon$ is easy to explain under the setting $H=1$; refer to the following example. 

\begin{example}[\textsf{Standard Imitation} Revisited]
We again consider the case in \cref{example:ail_fail}. To apply the estimator in \eqref{eq:new_estimator}, we need to randomly split the dataset, i.e., $\gD = \gD_1 \cup \gD_1^{c}$. There are three cases.
\begin{itemize}
    \item If $\gD_1$ contains both $(s^{1}, a^{1})$ and $(s^{2}, a^{1})$, the second term in \eqref{eq:new_estimator} disappears and the first term dominates. Accordingly, we have the exact estimation by leveraging the transition function and initial state distribution:  
    \begin{align*}
        \widetilde{P}^{\piE} (s^{1}, a^{1}) = 0.5, \widetilde{P}^{\piE} (s^{1}, a^{2}) = 0, \\
        \widetilde{P}^{\piE} (s^{2}, a^{1}) = 0.5, \widetilde{P}^{\piE} (s^{2}, a^{2}) = 0.
    \end{align*}
    This implies that the estimator error is 0 and the policy value gap is also 0. 
    \item  If $\gD_1$ only contains $(s^{2}, a^{1})$, we obtain an exact estimation $\widetilde{P}^{\piE}(s^{2}, a^{1}) = 0.5$. Furthermore, we have an inaccurate estimation $\widetilde{P}^{\piE}(s^{1}, a^{1}) = 0.8$. Thus, the estimation error is $0.3$ and policy value gap is at most $0.15$.
    \item It is impossible that $\gD_1$ only contains $(s^{1}, a^{1})$ in this example. 
\end{itemize}
By random split, we know that the second case happens with a small probability $\binom{6}{5}/\binom{10}{5} = 1/42$ and the first case happens with a large probability $1 - 1/42 = 41/42$. Consequently, we have that $\expect[\Vert \widetilde{P}^{\piE} - P^{\piE} \Vert_1] = 1/140$, which is smaller than the estimation error $0.2$ of the direct maximum likelihood estimation as shown in \cref{example:ail_fail}. Here the expectation is taken over the random split process. Moreover, in expectation, we have that $V^{\piE} - \expect[V^{\pi}] \leq 1/280$.

\end{example}

With the same reduction framework used for \textsf{VAIL}'s analysis, \citet{rajaraman2020fundamental} derived the sample complexity $\gO(H^{3/2}|\gS|/\varepsilon)$ for MIMIC-MD when $\varepsilon$ is small. This further hints that MIMIC-MD improves over BC in the worst case by $\gO(\sqrt{H})$. However, our analysis based on \cref{prop:ail_general_reset_cliff} would disclose that MIMIC-MD also achieves the horizon-free sample complexity on Reset Cliff; refer to \cref{theorem:ail_mimic_md}. In that case, both MIMIC-MD and VAIL provably perform better than BC by a wide margin $\gO(H^2)$. To this end, we believe other AIL methods can also achieve the horizon-free sample complexity by our analysis; see the empirical evidence in \cref{appendix:experiment_known_transition}.

\begin{thm}[Horizon-free Sample Complexity of MIMIC-MD on \textsf{Reset Cliff}]     \label{theorem:ail_mimic_md}
For each tabular and episodic MDP satisfying \cref{asmp:reset_cliff}, suppose that $\piail$ is the optimal solution to the above problem and $\vert \gD \vert \geq 2$, to obtain an $\varepsilon$-optimal policy (i.e., $V^{\piE} - \expect [ V^{\piail} ] \leq \varepsilon$), in expectation, MIMIC-MD requires at most ${\gO}(|\gS|/\varepsilon^2)$ expert trajectories.
\end{thm}

In fact, our analysis would suggest that the sample complexity of MIMIC-MD is $\min \{{\gO}(|\gS|/\varepsilon^2), {\gO}(|\gS| \sqrt{H}/\varepsilon)\}$ but we show the horizon-free one in \cref{theorem:ail_mimic_md}. Hence, \cref{theorem:ail_mimic_md} does not contradict the fact that MIMIC-MD addresses the sample barrier issue.

\section{Beyond Known Transition Algorithms}
\label{sec:beyond_vanilla_ail}

Previously, we assume AIL approaches know the transition function so that they can calculate $P^{\pi}_h$ to minimize the state-action distribution discrepancy. In this part, we remove this assumption and assume the transition function is unknown but environment interaction is allowed. We hope to design methods that can efficiently explore and imitate. Specifically, in addition to the number of expert demonstrations, we also care about the number of environment interactions. Here we refer to the above two measures as \emph{(expert) sample complexity} and \emph{interaction complexity}, respectively. For simplicity, we mainly focus on the worst-case analysis.

Our framework builds on recent advances in reward-free exploration \citep{chi20reward-free, menard20fast-active-learning}.

\begin{defn}  \label{defn:reward_free}
An algorithm is $(\varepsilon, \delta)$-PAC for reward-free exploration \citep{menard20fast-active-learning} if 
\begin{align*}
    \sP(&\text{for any reward function $r$}, |V^{\pi^*_{r}} - V^{\widehat{\pi}_{r}^*}| \leq \varepsilon) \geq 1 - \delta,
\end{align*}
where $\pi^*_{r}$ is the optimal policy in the MDP with the reward function $r$, and $\widehat{\pi}_{r}^*$ is the optimal policy in the MDP with the learned transition model $\widehat{\gP}$ and the reward function $r$.
\end{defn}

This definition suggests the reward-free exploration methods could achieve \emph{uniform policy evaluation} after the exploration. Formally, a reward-free exploration method ensures 
\begin{align*}
  \forall r: \gS \times \gA \rightarrow [0, 1], \pi \in \Pi: \vert V^{\pi, \gP,r} - V^{\pi, \widehat{\gP}, r} \vert \leq \varepsilon,  
\end{align*}
where $V^{\pi, \gP, r}$ is the policy value under transition $\gP$ and reward $r$ \citep{chi20reward-free}. Based on such a learned transition model, an AIL algorithm can perform policy optimization as if this empirical transition is the same as the true transition function. We outline such a general idea in Algorithm \ref{algo:framework}.

\begin{algorithm}[htbp]
\caption{AIL With Unknown Transitions}
\label{algo:framework}
\begin{algorithmic}[1]
\REQUIRE{expert demonstrations $\gD$.}
\STATE{Establish the state-action distribution estimation $\widehat{P}_h^{\piE}$.}
\STATE{$\widehat{\gP} \lar$ invoke a reward-free method to collect interaction trajectories and learn a transition model.}
\STATE{${\pi} \lar$ apply an AIL approach to perform imitation with the estimation $\widehat{P}_h^{\piE}$ under transition model $\widehat{\gP}$.}
\ENSURE{policy ${\pi}$.}
\end{algorithmic}
\end{algorithm}

\begin{prop}   \label{prop:connection}
Suppose that 
\begin{itemize}  
    \item[(a)] An algorithm A solves the reward-free exploration problem (see Definition \ref{defn:reward_free}) up to an error $\varepsilon_{\mathrm{RFE}}$ with probability at least $1-\delta_{\mathrm{RFE}}$;
    \item[(b)] An algorithm B has a state-action distribution estimator for $P^{\piE}_h$, which satisfies $\sum_{h=1}^H \Vert \widehat{P}^{\piE}_h - P^{\piE}_h  \Vert_{1} \leq \varepsilon_{\mathrm{EST}}$, with probability at least $1-\delta_{\mathrm{EST}}$;
    \item[(c)] With the estimator in (b), the algorithm B solves the optimization problem in \eqref{eq:ail} up to an error $\varepsilon_{\mathrm{AIL}}$.
\end{itemize}

Then applying algorithm A and B under the framework in Algorithm \ref{algo:framework} could return a policy ${\pi}$, which has a policy value gap (i.e., $V^{\piE} - V^{{\pi}}$) at most $2 \varepsilon_{\mathrm{EST}} + 2 \varepsilon_{\mathrm{RFE}} + \varepsilon_{\mathrm{AIL}}$, with probability at least $1-\delta_{\mathrm{EST}} - \delta_{\mathrm{RFE}}$.
\end{prop}

As an application of \cref{prop:connection}, we can combine the adversarial imitation learning method \textsf{VAIL} and the reward-free exploration method \textsf{RF-Express} \citep{menard20fast-active-learning} under the unknown transition setting.

\begin{corollary}
For any tabular and episodic MDP, with probability $1-\delta$, to obtain an $\varepsilon$-optimal policy, there exists an algorithm that combines \textsf{VAIL} and RF-Express \citep{menard20fast-active-learning} requiring at most $\widetilde{\gO} (|\gS| H^{2}/\varepsilon^2)$ expert trajectories and $\widetilde{\gO}(|\gS|^2 |\gA|H^3/\varepsilon^2)$ interaction trajectories if the transition is unknown. 
\end{corollary}

In addition, even though we use the total variation distance for assumptions $(b)$ and $(c)$ in \cref{prop:connection}, our result is general in the sense that other discrepancy metrics are applicable. For instance, FEM can also be applied; refer to the discussion in \cref{appendix:discussion_of_prop:connection}.

Finally, we may also want to extend MIMIC-MD to have a better guarantee in the worst case. However, the direct application of \cref{prop:connection} fails. This is because MIMIC-MD uses the transition function in two parts: the estimation and the optimization. However, \cref{prop:connection} only addresses the issue in the optimization part. Fortunately, we can tackle this difficulty with a new technique. We do not involve this part; see \cref{subsection:mb-tail} for details.

\begin{thm}[Informal version of \cref{theorem:sample-complexity-unknown-transition} in Appendix]    \label{theorem:informal_mbtail}
For any tabular and episodic MDP, with probability 1-$\delta$, to obtain an $\varepsilon$-optimal policy, there exists an algorithm called \textsf{MB-TAIL} (see \cref{algo:mbtail-abstract} in Appendix) requiring at most $\widetilde{\gO}(|\gS|H^{3/2}/\varepsilon)$ expert trajectories and $\widetilde{\gO}(|\gS|^2 |\gA|H^3/\varepsilon^2)$ interaction trajectories if the transition is unknown. 
\end{thm}

To interpret our result, we compare MB-TAIL with the best-known online apprenticeship learning algorithm (OAL) in \citep{shani21online-al}. OAL addresses the exploration issue by two mirror descent based no-regret algorithms. Concretely, OAL algorithm  has the sample complexity $\widetilde{\gO} ( |\gS| H^{2} / \varepsilon^2 )$ and interaction complexity $\widetilde{\gO} ( |\gS|^2 |\gA| H^4  / \varepsilon^2 )$ under the same setting\footnote{In \citep{shani21online-al}, a regret guarantee is given. We convert it into the sample complexity guarantee; see  \cref{appendix:from_regret_to_pac}. For our purpose, we hide the polynomial factor about $\delta$.}. \cref{theorem:informal_mbtail} implies our approach has improvements over the OAL algorithm in both bounds.

Readers may notice that theoretical guarantees in \cref{prop:connection} and \cref{theorem:informal_mbtail} are in high probability forms while we state many in expectation bounds in \cref{sec:generalization_of_ail}. We note that this is not a big issue since two kinds of guarantees can be transformed to each other and we also have high probability bounds for algorithms in \cref{sec:generalization_of_ail}; see \cref{appendix:proof_generalization_ail}.

\section{Conclusion}
\label{sec:conclusion}

In this paper, we disclose when/why AIL approaches could generalize better than BC and when/why  AIL approaches would fail. In particular, we are the first to validate that AIL-type algorithms provably generalize better than BC by a wide margin in certain cases.  Furthermore, we present how to enable known-transition AIL approaches to efficiently explore and imitate when transitions are unknown. For future studies, we discuss the relation between our results and related open problems in \cref{appendix:open_problem}. We hope our results could provide a better understanding of AIL and BC in both theory and practice.

\bibliographystyle{abbrvnat}
\bibliography{reference}  

\newpage
\appendix
\onecolumn

\section*{\Large Appendix: On Generalization of Adversarial Imitation Learning and Beyond}

{\footnotesize 
\tableofcontents
}

\newpage 
\begin{table}[H]
\caption{Notations}
\label{table:notations}
\centering
\begin{tabular}{@{}ll@{}}
\toprule
Symbol                     & Meaning \\ \midrule
$\gS$                & the state space  \\
$\gA$                   & the action space   \\
$\gP = \lb P_1, \cdots, P_{H} \rb$ & the transition function \\
$H$              & the planning horizon  \\
$\rho$                & the initial state distribution    \\
$r= \lb r_1, \cdots, r_{H} \rb$ & the reward function    \\
$\pi = \lb \pi_1, \cdots, \pi_{h} \rb$ & non-stationary policy \\
$\piE$ & the expert policy \\
$V^{\pi, \gP, r}$ & policy value under the transition $\gP$ and reward $r$ \\
$\varepsilon$ & the policy value gap \\
$\delta$ & failure probability \\
$P^{\pi}_h (s)$ & state distribution \\
$P^{\pi}_h (s, a)$ & state-action distribution \\
$\tr = \lp s_1, a_1, \cdots, s_{H}, a_{H} \rp$ & the trajectory \\
$\tr_h = \lp s_1, a_1, \cdots, s_{h}, a_{h} \rp$ & the truncated trajectory \\
$\tr (s_h)$ & the state in time step $h$ in $\tr$ \\
$\tr (a_h)$ & the action in time step $h$ in $\tr$ \\
$\tr (s_h, a_h)$ & the state-action pair in time step $h$ in $\tr$ \\
$\gD$ & expert demonstrations \\
$m$   & number of expert demonstrations \\
$\widehat{P}_h^{\piE} (s, a)$ & maximum likelihood estimator in $\gD$
\\
$\widetilde{P}_h^{\piE} (s, a)$ & transition-aware estimator used in TAIL and MB-TAIL
\\
$\sP^{\piE}(\tr)$ & probability of the trajectory $\tr$ under the expert policy $\piE$
\\
$\sP^{\piE}(\tr_h)$ & probability of the truncated trajectory $\tr_h$ under the expert policy $\piE$
\\
$\gS_h (\gD)$ & the set of states visited in time step $h$ in dataset $\gD$
\\
$\Tr_h^{\gD}$ & the trajectories along which each state has been visited in $\gD$ up to time step $h$
\\
$\pi^{(t)}$ & the policy obtained in the iteration $t$ \\
$w^{(t)}$ & the reward function learned in the iteration $t$ \\
$\eta^{(t)}$ & the step size in the iteration $t$ \\
$f^{(t)} (w)$ & the objective function in the iteration $t$ \\
$\widebar{P}_h (s, a)$ & the mean state-action distribution \\
$\widebar{\pi}$ & the policy derived by the mean state-action distribution \\
$\Pi_{\text{BC}} \lp \gD_{1} \rp$ & the set of policies which take the expert action on states covered in $\gD_{1}$
\\
$\widehat{\gP}$ & the empirical transition function
\\
$P^{\pi, \gP}_h (s, a)$ & the state-action distribution of $\pi$ under the empirical transition function $\widehat{P}$
\\\bottomrule
\end{tabular}%
\end{table}

\newpage 
\section{Review of Previous Work}
\label{appendix:review_of_previous_work}

In addition to the related work discussed in \cref{sec:related_work}, we provide a more detailed overview of other works in this section. %

\textbf{Function Approximation.} Beyond the tabular setting, researchers also have considered the statistical guarantees for imitation learning algorithms with function approximation. For instance, \citet{cai2019lqr} and \citet{liu2021provably} considered GAIL with linear function approximation setting while the neural network approximation case is studied in \citep{wang2020computation, zhang2020generative, xu2021error}. In addition, \citet{rajaraman2021value} studied BC and MIMIC-MD with linear function approximation. The main message is that under mild assumptions, the dependence on $|\gS|$ can be improved to the inherent dimension $d$ with function approximation. This direction is orthogonal to us since we mainly focus on the improvement/comparison in the planning horizon $H$, which is usually unrelated to function approximation.

\textbf{Active Imitation Learning.} In this paper, we mainly focus on the case where the expert demonstrations are fixed over the learning process. There is another setting where the agent can actively query the expert policy to obtain guidance in an online way. For this setting, DAgger \citep{ross11dagger} and AggraVaTe \citep{ross2014reinforcement} are two famous methods using the no-regret online learning. For general tabular and episodic MDPs, DAgger cannot improve the sample complexity compared with BC; see the theoretical result and explanation in \citep{rajaraman2020fundamental}. However, under the $\mu$-recoverability assumption, \citet{rajaraman2021value} proved that there is a clear boundary between the active setting and offline setting. That is, under the $\mu$-recoverability assumption, the agent can improve its sample complexity if it can actively query the expert policy.

\textbf{Inverse Reinforcement Learning.} Given the expert demonstrations, one of the interesting questions is to recover the reward function used by the expert policy. This field is often called inverse reinforcement learning (IRL) \citep{ng00irl, ziebart2008maximum}. Adversarial imitation learning algorithms are closely related to IRL (see e.g., \citep{pieter04apprentice, ho2016gail, fu2018airl}). However, without any assumption, the recovered reward function by adversarial imitation learning algorithms is not the true environment reward function.

\textbf{Reward-free Exploration.} The reward-free exploration framework is firstly proposed in \citep{chi20reward-free} with the goal of 1) isolating the exploration issue and planning issue under a standard RL framework and 2) learning a \dquote{robust} environment to cover all possible training scenarios. Following \citep{chi20reward-free}, there are many advances in this direction \citep{wang2020rewardfree, menard20fast-active-learning, Kaufmann21adaptive-rfe, zhang2021reward, chen2021near}, in which the minimax rate under the tabular setting is achieved by \citep{menard20fast-active-learning}. Our framework in Section \ref{sec:beyond_vanilla_ail} connects the reward-free exploration and adversarial imitation learning.
\section{Extension of MIMIC-MD}
\label{appendix:extension_of_mimic_md}

In this section, we develop an extension of MIMIC-MD under the unknown transition setting. Before presentation, we note that the exact optimal solutions of the original MIMIC-MD formulation cannot be solved in a polynomial time as stated in \citep{rajaraman2020fundamental}. To this end, we will first establish an adversarial formulation for MIMIC-MD, which allows us to use gradient based methods to efficiently obtain an approximate solution. We call such a method TAIL (Transition-aware AIL). Then, we show how to combine TAIL with reward-free exploration methods in the unknown transition scenario. 

\subsection{TAIL}

With the improved estimator in \eqref{eq:new_estimator}, we arrive at the following state-action distribution matching problem:
\begin{align} \label{eq:l1_norm_imitation_with_estimator} 
   \min_{\pi \in \Pi} \sum_{h=1}^{H} \lnorm P^{\pi}_h - \widetilde{P}^{\piE}_h \rnorm_1.
\end{align}
We highlight that \eqref{eq:l1_norm_imitation_with_estimator} is slightly different from MIMIC-MD's objective in \citep{rajaraman2020fundamental}. Specifically, MIMIC-MD restricts candidate policies to $\Pi_{\text{BC}} (\gD_1) = \{ \pi \in \Pi: \pi_h (s) = \piE_h (s), \forall h \in [H], s \in \gS_h (\gD_1) \}$, which is the set of BC policies on $\gD_1$. The intuition in MIMIC-MD is that the expert actions are known on $\gD_1$ so that direct projection is feasible.

Now, we would like to develop an adversarial formulation for \eqref{eq:l1_norm_imitation_with_estimator}.  With the famous min-max theorem~\citep{bertsekas2016nonlinear}, we transform \eqref{eq:l1_norm_imitation_with_estimator} to:
\begin{align}   \label{eq:new_algo_max_min}
     \max_{w \in \gW} \min_{\pi \in \Pi} \sum_{h=1}^H \sum_{(s, a)} w_h (s, a) \lp \widetilde{P}^{\piE}_h (s, a) -  P^{\pi}_h (s, a)  \rp.
\end{align}
where $\gW = \{w: \lnorm w \rnorm_\infty \leq 1 \}$ is the unit ball. We see that the inner problem in \eqref{eq:new_algo_max_min} is to maximize the policy value of $\pi$ given the reward function $w_h(s, a)$. For the outer optimization problem, we can use online gradient descent methods \citep{shalev12online-learning} so that we can finally reach an approximate saddle point. Formally, let us define the objective $f^{(t)}(w)$:
\begin{align}   
    & \sum_{h=1}^{H} \sum_{(s, a)} w_h(s, a) \lp P^{\pi^{(t)}}_h (s, a) - \widetilde{P}^{\piE}_h (s, a)  \rp, \label{eq:objective_w}
\end{align}
where $\pi^{(t)}$ is the optimized policy at iteration $t$. Then the update rule for $w$ is:
\begin{align*}   
    w^{(t+1)} := \gP_{\gW}\lp w^{(t)} - \eta^{(t)}  \nabla f^{(t)}(w^{(t)}) \rp, 
\end{align*}
where $\eta^{(t)} > 0$ is the stepsize to be chosen later, and $\gP_{\gW}$ is the Euclidean projection on the unit  ball $\gW$, i.e., $\gP_{\gW}(w) := \argmin_{z \in \gW} \lnorm z- w \rnorm_2$. The procedure for solving \eqref{eq:new_algo_max_min} is outlined in \cref{algo:main_aglorithm}.

\begin{algorithm}[htbp]
\caption{\textsf{TAIL}}
\label{algo:main_aglorithm}
{
\begin{algorithmic}[1]
\REQUIRE{expert demonstrations $\gD$, number of iterations $T$, step size $\eta^{(t)}$, and initialization $w^{(1)}$.}
\STATE{Randomly split $\gD$ into two equal parts: $\gD = \gD_1 \cup \gD_1^{c}$ and obtain the estimation $\widetilde{P}_h^{\piE}$ in \eqref{eq:new_estimator}.}
\FOR{$t = 1, 2, \cdots, T$}
\STATE{$\pi^{(t)} \lar $ solve the optimal policy with the reward function $w^{(t)}$ up to an error of $\varepsilon_{\mathrm{opt}}$.}
\STATE{Compute the state-action distribution $P^{\pi^{(t)}}_h$ for $\pi^{(t)}$.}
\STATE{Update $ w^{(t+1)} := \gP_{\gW}\lp w^{(t)} - \eta^{(t)}  \nabla f^{(t)}(w^{(t)}) \rp$ with $f^{(t)}(w)$ defined in \eqref{eq:objective_w}.}
\ENDFOR
\STATE{Compute the mean state-action distribution $\widebar{P}_h(s, a) = \sum_{t=1}^{T} P^{\pi^{(t)}}_h(s, a) / T$.}
\STATE{Derive $\widebar{\pi}_h (a|s) \lar \widebar{P}_h(s, a) / \sum_{a} \widebar{P}_h(s, a)$.}
\ENSURE{policy $\widebar{\pi}$.}
\end{algorithmic}
}
\end{algorithm}

\begin{thm} \label{theorem:final_sample_complexity} 
Fix $\varepsilon \in \lp 0, H \rp$ and $\delta \in (0, 1)$; suppose $H \geq 5$. Consider the approach \textsf{TAIL} in Algorithm \ref{algo:main_aglorithm} with $\widebar{\pi}$ being the output policy. Assume that the optimization error $\varepsilon_{\mathrm{opt}} \leq \varepsilon / 2$, the number of iterations $T \succsim  |\gS|  |\gA| H^2 / \varepsilon^2$, and the step size $\eta^{(t)} :=  \sqrt{|\gS||\gA| / (8T)}$. If the number of expert trajectories ($m$) satisfies
\begin{align*}
m \succsim  \frac{  |\gS| H^{3/2}}{\varepsilon} \log\lp\frac{ |\gS| H}{\delta} \rp,
\end{align*}
then with probability at least $1-\delta$, we have $V^{\piE} - V^{\widebar{\pi}} \leq \varepsilon$.
\end{thm}
See \cref{appendix:proof_of_theorem:final_sample_complexity} for the proof. Let us briefly discuss the computation details of \textsf{TAIL}. For the optimization problem in Line 3 of \cref{algo:main_aglorithm}, we can use value iteration or policy gradient methods \citep{agarwal2020pg}. Specifically, if we use value iteration, it is clear that $\varepsilon_{\opt} = 0$ and this procedure can be done in $H$ iterations. For each iteration of the value iteration algorithm, the computation complexity is $\gO(|\gS||\gA| \times |\gS| + |\gS| |\gA|)$ for computing the target $Q$-values and  greedy actions for each state-action pairs. Since the total number of iterations of \cref{algo:main_aglorithm} is $\gO( |\gS| |\gA| H^2/\varepsilon^2)$, we have the following total computation complexity:
\begin{align*}
    \gO(|\gS|^2|\gA|) \times H  \times \gO( |\gS| |\gA| H^2/\varepsilon^2) = \gO(|\gS|^3 |\gA|^2 H^{3}/\varepsilon^2).
\end{align*}
On the other hand, the space complexity of \cref{algo:main_aglorithm} is $\gO(|\gS||\gA|H)$ for storing $w^{(t)}, \pi^{(t)}$ and $P^{\pi^{(t)}}_h$.

We notice that in \citep{nived2021provably}, a linear programming (LP) formulation is proposed to solve the exactly optimal solutions of MIMIC-MD. The computation complexity of this method is about $\widetilde{\gO}(d^{2.5})$ where $d = 2|\gS| |\gA| H$. However, the space complexity of this method is $\gO(|\gS|^2 |\gA|^2 H^2)$, which is unbearable in practice; see the evidence in \cref{appendix:experiments}.

\subsection{MB-TAIL}
\label{subsection:mb-tail}

In the following part, we present how to apply \textsf{TAIL} in Algorithm \ref{algo:main_aglorithm} under our framework. As mentioned, the main challenge is that the refined estimation in \eqref{eq:new_estimator} requires the knowledge of the true transition function. Unfortunately, we cannot utilize the biased empirical model instead of the true transition, since the induced estimation error is difficult to control.

Technically, the term ${\sum_{\tr_h \in \Tr_h^{\gD_1} } \sP^{\piE}(\tr_h) \indict \{ \tr_h(s_h, a_h) = (s, a)\}}$ in \eqref{eq:new_estimator} relies on the exact transition function. To address the mentioned issue, we present a key observation in \cref{lemma:unknown-transition-unbiased-estimation} in \cref{subsection:proof-of-theorem-sample-complexity-unknown-transition}. In particular, for a BC policy $\pi \in \Pi_{\text{BC}} \lp \gD_{1} \rp$, for all trajectories $\tr_h \in \Tr_h^{\gD_1}$, the trajectory probabilities induced by $\pi$ and $\piE$ are identical up to time step $h$. Based on this observation, we can estimate this term with a dataset $\gD^{\prime}_{\text{env}}$ collected by rolling out a BC policy $\pi \in \Pi_{\text{BC}} \lp \gD_{1} \rp$ with the environment. The new estimator is formulated as
\begin{align}
  \widetilde{P}_h^{\piE} (s, a) = {\frac{\sum_{\tr_h \in \gD_{\text{env}}^\prime} \indict \{ \tr_h (s_h, a_h) = (s, a), \tr_h \in \Tr_h^{\gD_1} \}}{|\gD^\prime_{\text{env}}|}} + {\frac{  \sum_{\tr_h \in \gD_1^c}  \indict\{ \tr_h (s_h, a_h) = (s, a), \tr_h \not\in \Tr_h^{\gD_1}  \} }{|\gD_1^c|}}. \label{eq:new_estimator_unknown_transition}
\end{align}
With the estimator in \eqref{eq:new_estimator_unknown_transition}, we develop an extension of TAIL named \textsf{MB-TAIL} presented in Algorithm \ref{algo:mbtail-abstract}. 

\begin{algorithm}[htbp]
\caption{\textsf{MB-TAIL}}
\label{algo:mbtail-abstract}
\begin{algorithmic}[1]
\REQUIRE{expert demonstrations $\gD$.}
\STATE{Randomly split $\gD$ into two equal parts: $\gD = \gD_1 \cup \gD_1^{c}$.}
\STATE{Learn $\pi \in \Pi_{\text{BC}} \lp \gD_{1} \rp$ by BC and roll out $\pi$ to obtain dataset $\gD_{\text{env}}^\prime$ with $|\gD_{\text{env}}^\prime| = n^{\prime}$}.
\STATE{Obtain the estimator $\widetilde{P}_h^{\piE}$ in \eqref{eq:new_estimator_unknown_transition} with $\gD$ and $\gD_{\text{env}}^\prime$.}
\STATE{Invoke \textnormal{RF-Express} to collect $n$ trajectories and learn an empirical transition function $\widehat{\gP}$.}
\STATE{$\widebar{\pi} \lar$ apply TAIL to perform imitation with the estimation $\widetilde{P}_h^{\piE}$ under transition model $\widehat{\gP}$.}
\ENSURE{policy $\widebar{\pi}$.}
\end{algorithmic}
\end{algorithm}

\begin{thm}\label{theorem:sample-complexity-unknown-transition}
Fix $\varepsilon \in \lp 0, 1 \rp$ and $\delta \in (0, 1)$; suppose $H \geq 5$. Under the unknown transition setting, consider \textsf{MB-TAIL} displayed in Algorithm \ref{algo:mbtail-abstract} and $\widebar{\pi}$ is output policy, assume that the optimization error $\varepsilon_{\mathrm{opt}} \leq \varepsilon / 2$, the number of iterations and the step size are the same as in Theorem \ref{theorem:final_sample_complexity}, if the number of expert trajectories ($m$), the number of interaction trajectories for estimation ($n^\prime$), and the number of interaction trajectories for reward-free exploration ($n$) satisfy
\begin{align*}
&m \succsim  \frac{  |\gS| H^{3/2}}{\varepsilon} \log\lp\frac{ |\gS| H}{\delta} \rp, n^{\prime} \succsim  \frac{  |\gS| H^2}{\varepsilon^2} \log \lp \frac{ |\gS| H }{\delta} \rp,
\\
& n \succsim \frac{ |\gS| |\gA| H^3}{\varepsilon^2} \lp |\gS| + \log \lp \frac{ |\gS| |\gA| H}{\delta \varepsilon} \rp \rp
\end{align*}
Then with probability at least $1-\delta$, we have $V^{\piE} - V^{\widebar{\pi}} \leq \varepsilon $.
\end{thm}
See \cref{subsection:proof-of-theorem-sample-complexity-unknown-transition} for the proof.

\section{Experiments}
\label{appendix:experiments}

In this section, we present experiment results to help verify and understand our theoretical claims.

\subsection{Baselines}

We consider the following baselines on two MDPs: \emph{Standard Imitation} and \emph{Reset Cliff} introduced in \cref{sec:generalization_of_ail}. 
\begin{itemize}
    \item Behavioral Cloning (BC) \citep{Pomerleau91bc}. 
    \item Feature Expectation Matching (FEM) \citep{pieter04apprentice}.
    \item Game-theoretic Apprenticeship Learning (GTAL) \citep{syed07game}. 
    \item Vanilla Adversarial Imitation Learning (VAIL) (refer to \eqref{eq:ail}).
    \item Transition-aware Adversarial Imitation Learning (TAIL) (refer to \cref{algo:main_aglorithm}) .
    \item Model-based Transition-aware Adversarial Imitation Learning (MB-TAIL) (refer to \cref{algo:mbtail-abstract}).
    \item Online apprenticeship learning (OAL) \citep{shani21online-al}. 
\end{itemize}

Algorithm configurations are given in \cref{subsection:experiment_details}. We do not involve MIMIC-MD \citep{rajaraman2020fundamental} because its LP formulation in \citep{nived2021provably}  runs out of memory on a machine with $128$GB RAM when $H \geq 100$.  GAIL is not considered as it does not have a formal convergent algorithm. Furthermore, GAIL differs from \textsf{VAIL} in terms of the discrepancy metric, which does not matter under tabular MDPs. For completeness, we provide a variant of GAIL \citep{ho2016gail} and investigate its performance in Appendix \ref{subsection:additional_results_gail}.

\subsection{Known Transition Setting}
\label{appendix:experiment_known_transition}

We aim to study the dependence on $H$ and $\varepsilon$ appeared in the sample complexity. To achieve this goal, figures have used \emph{logarithmic} scales so that we can read the order dependence from slopes of curves. Specifically, a worst-case sample complexity $m \succsim |\gS| H^{\alpha}/\varepsilon^{\beta}$ implies the policy value gap $ V^{\piE} - V^{\pi} \precsim |\gS|^{1/\beta} H^{\alpha/\beta} /m^{1 / \beta}$. Then,
\begin{align*}
\log(V^{\piE} - V^{\pi}) \precsim & (\alpha/\beta) \log (H) - 1/\beta \log (m) +1/\beta \log (|\gS|).
\end{align*}
For example, the sample complexity ${\gO}(|\gS| H^2/\varepsilon)$ of \textsf{VAIL} similarly suggests slope $1$ w.r.t. $\log(H)$ and slope $-1/2$ w.r.t. $\log(m)$ for its $\log$ policy value gap. It is worth mentioning that these implications are true only on the worst instances.

\begin{figure}[htbp]
     \centering
     \begin{subfigure}[b]{0.45\textwidth}
         \centering
         \includegraphics[width=\textwidth]{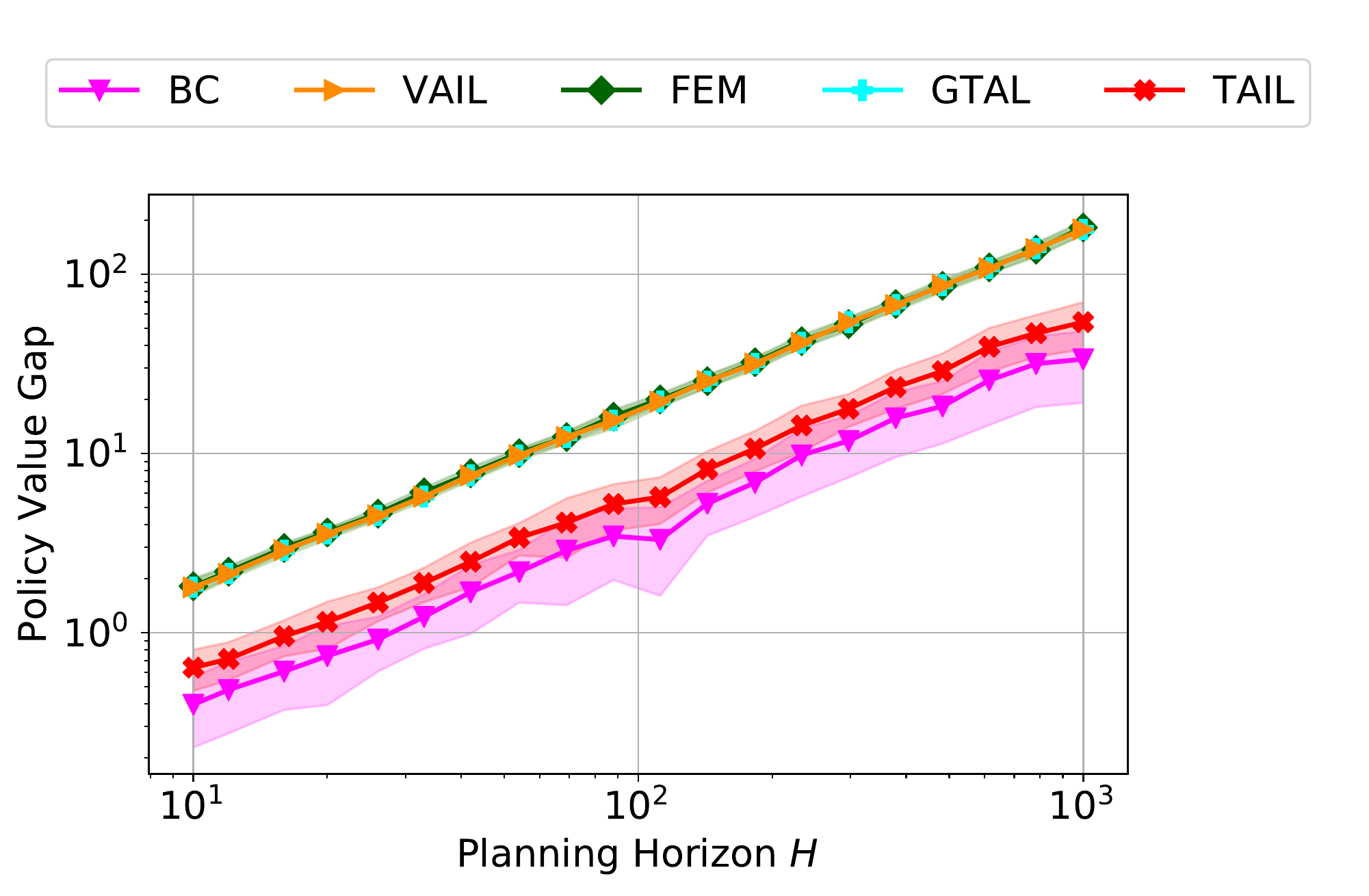}
         \caption{On the planning horizon on Standard Imitation.}
         \label{fig:bandit_h_log_result}
     \end{subfigure}
     \hfill
     \begin{subfigure}[b]{0.45\textwidth}
         \centering
         \includegraphics[width=\textwidth]{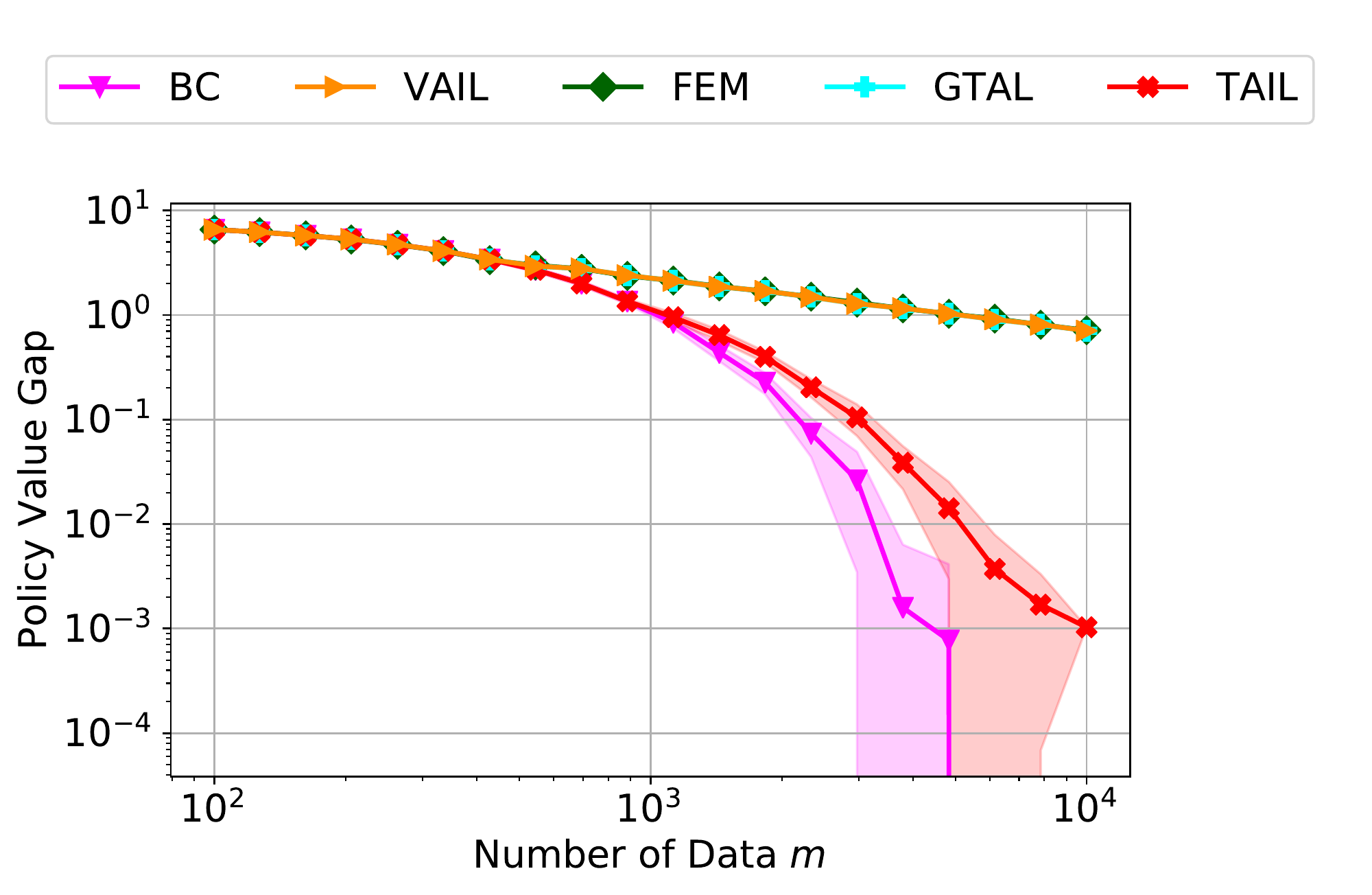}
         \caption{On the expert sample size on Standard Imitation.}
         \label{fig:bandit_m_log_result}
     \end{subfigure}
     \hfill
     \vskip\baselineskip
     \begin{subfigure}[b]{0.45\textwidth}
         \centering
         \includegraphics[width=\textwidth]{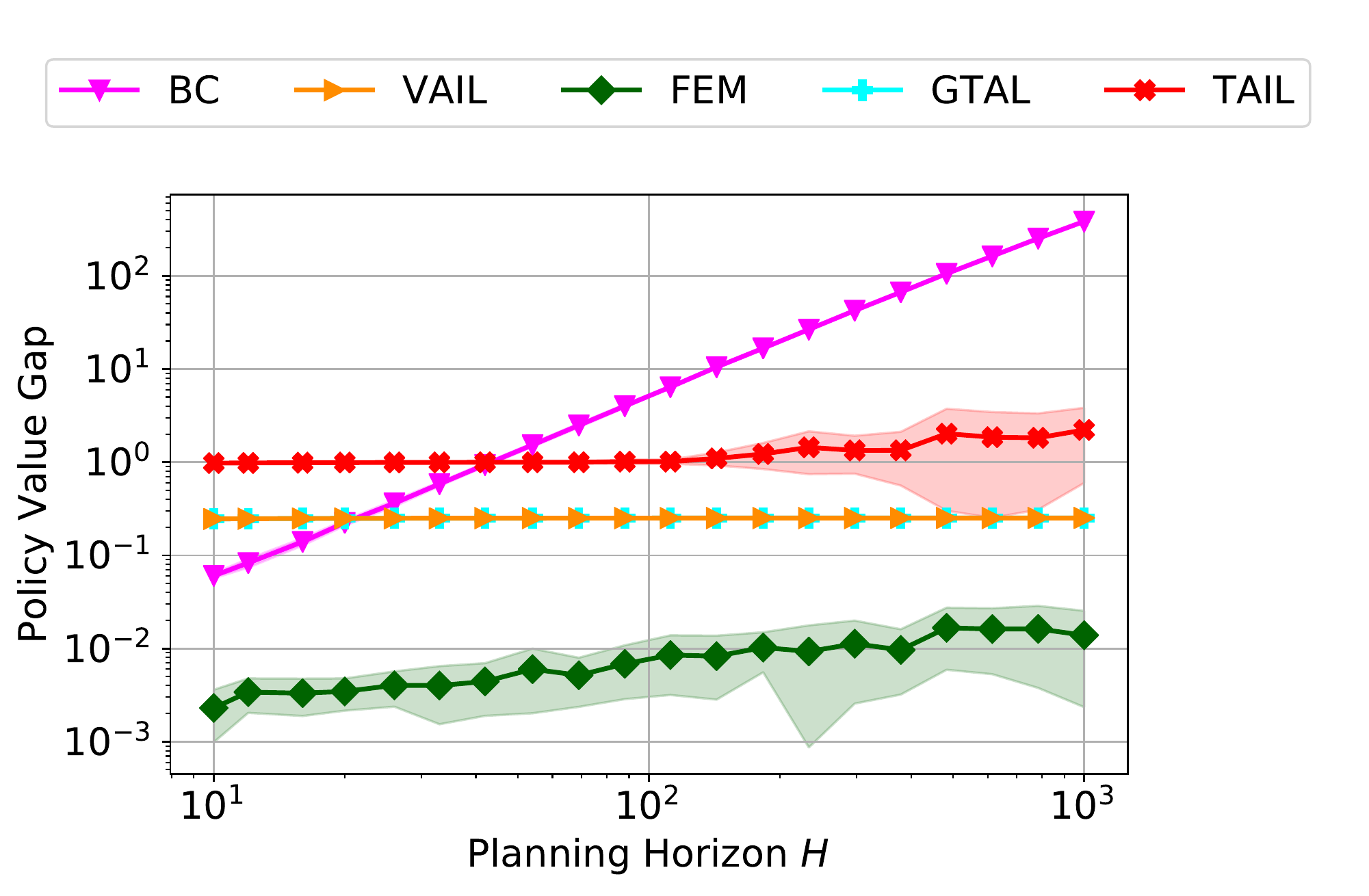}
         \caption{On the planning horizon on Reset Cliff.}
         \label{fig:cliffwalking_h_log_result}
     \end{subfigure}
     \hfill
    \begin{subfigure}[b]{0.45\textwidth}
      \centering
      \includegraphics[width=\textwidth]{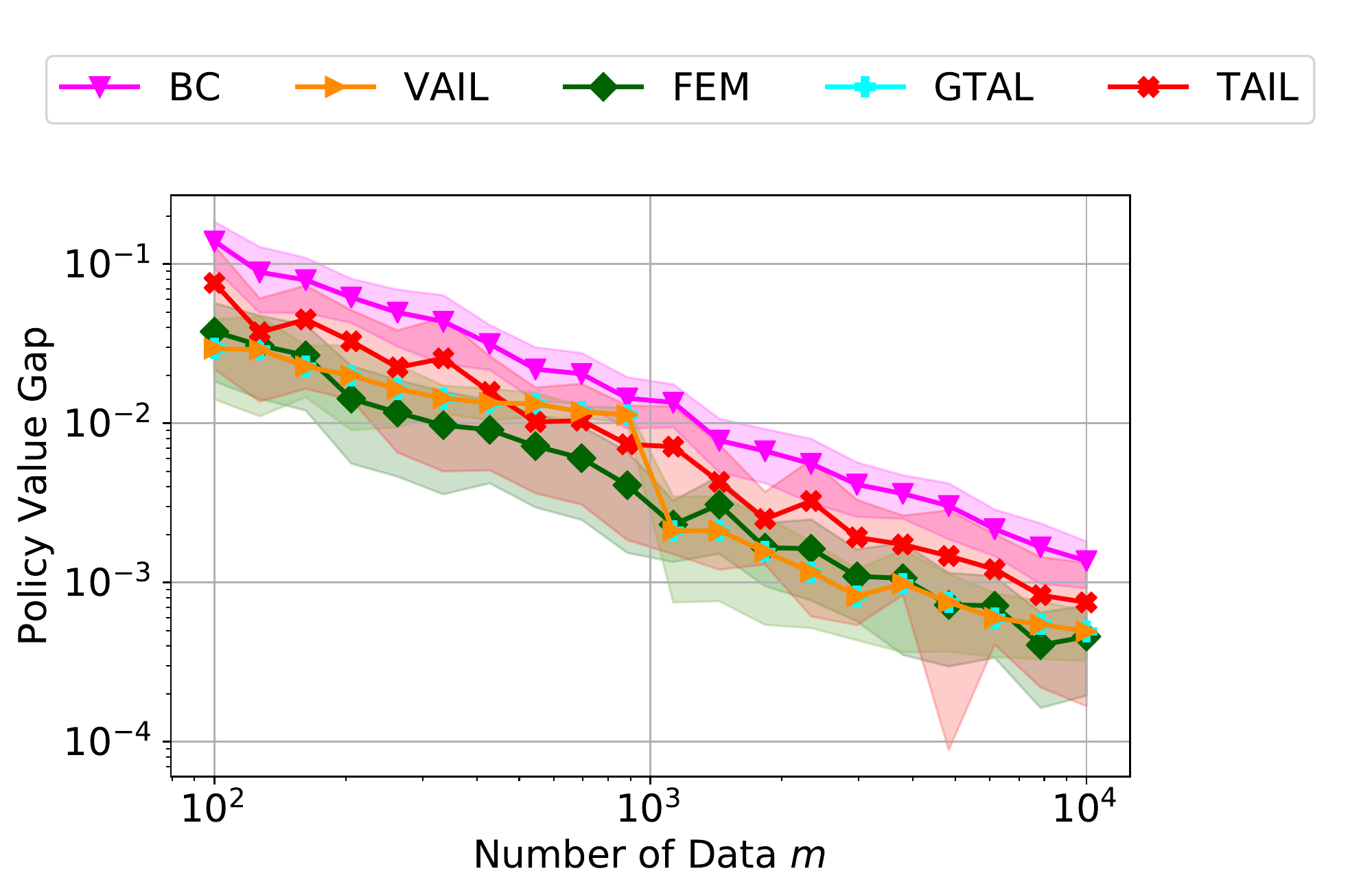}
      \caption{On the expert sample size on Reset Cliff.}
      \label{fig:cliffwalking_m_log_result}
    \end{subfigure}

     \caption{The policy value gap (i.e., $V^{\piE} - V^{\pi}$) on Standard Imitation and Reset Cliff. The solid lines are mean of results and the shaded region corresponds to the standard deviation over $20$ random seeds (same with the following figure). \dquote{sample size} refers to the number of expert trajectories. }
     \label{figure:main_results}
\end{figure}

\textbf{Case Study on Standard Imitation.} For Standard Imitation (Figure \ref{fig:bandit}), each state is absorbing and the agent gets $+1$ reward only by taking the expert action (shown in green). Different from \citep{rajaraman2020fundamental}, the initial state distribution $\rho$ is $({1}/{|\gS|}, \cdots, {1}/{|\gS|})$ to better disclose the sample barrier issue of AIL methods discussed in Section \ref{sec:generalization_of_ail}.

First, we focus on the planning horizon dependence issue; see the result in Figure~\ref{fig:bandit_h_log_result}. In particular, the numerical result shows that the policy value gap of all methods grows linearly with respect to the planning horizon. This is reasonable since each state on Standard Imitation is absorbing. As suggested in \cref{theorem:bc_deterministic}, Standard Imitation is not the worst-case MDP for BC due to its absorbing structure. However, Standard Imitation is challenging for conventional AIL approaches (VAIL, FEM, and GTAL) and thus can be used to validate the tightness of their sample complexity.

Second, we display the result regarding the number of expert demonstrations in Figure~\ref{fig:bandit_m_log_result}. Under Standard Imitation, the state distribution of every policy is a uniform distribution at every time step, which raises a statistical estimation challenge for conventional AIL. Specifically, the $\ell_1$-norm estimation error of maximum likelihood estimation is highest at uniform distribution (refer to the discussion below \citep[Lemma 8]{kamath2015learning}). From Figure~\ref{fig:bandit_m_log_result}, we clearly see that the slopes of VAIL, FEM and GTAL with respect to $\log \lp m \rp$ are around $-1/2$. This can be explained by their sample complexity ${\gO}( |\gS| H^2/\varepsilon^2)$, which implies $\log(V^{\piE} - V^{\pi}) \precsim -1/2 \log(m) + \text{constant}$. This empirical result demonstrates the sample barrier issue of VAIL discussed in \cref{subsec:when_does_vail_generalize_poorly}. Combined with the observation on the horizon dependence, these results verify the sample complexity lower bound of VAIL in \cref{prop:lower_bound_vail} and further indicate that its worst-case sample complexity in \cref{theorem:worst_case_sample_complexity_of_vail} is tight. As for TAIL, as shown in Figure~\ref{fig:bandit_m_log_result}, the policy value gap of TAIL diminishes substantially faster than VAIL, FEM, and GTAL, which verifies the sample efficiency of TAIL. The fast diminishing rate of BC is due to the quick concentration rate of missing mass; see \citep{rajaraman2020fundamental} for more explanation.

\textbf{Case Study on Reset Cliff.} Next, we consider the Reset Cliff MDP (\cref{fig:reset_cliff}) with 1 bad absorbing state and 19 good states. For Reset Cliff, the agent gets $+1$ reward by taking the expert action (shown in green) on states except the bad state $b$, then the next state is renewed according to the initial state distribution $\rho$. Here, $\rho = ( {1}/({m+1}), \cdots, {1}/({m+1}), 1 - {(\vert \mathcal{S} \vert -2)}/{(m+1)}, 0 )$ \citep{rajaraman2020fundamental}. Once taking a non-expert action (shown in blue), the agent goes to the absorbing state $b$ and gets $0$ reward.

On the one hand, Reset Cliff highlights the compounding errors issue and recovers the key characteristics of many practical tasks. Take the Gym MuJoCo locomotion task as an example, once the robot takes a wrong action, it would go to the terminate state and obtain $0$ reward forever. The numerical result about the planning horizon is given in Figure \ref{fig:cliffwalking_h_log_result}. From \cref{fig:cliffwalking_h_log_result}, we clearly see that the slope of BC w.r.t $\log (H)$ is around $2$, indicating the compounding errors issue of BC. As for conventional AIL methods, especially VAIL, their policy value gaps almost keep constant as the planning horizon increases. This result validates the horizon-free sample complexity of AIL approaches on Reset Cliff.

\begin{figure}[htbp]
\begin{subfigure}{.42\textwidth}
  \centering
  \includegraphics[width=0.8\textwidth]{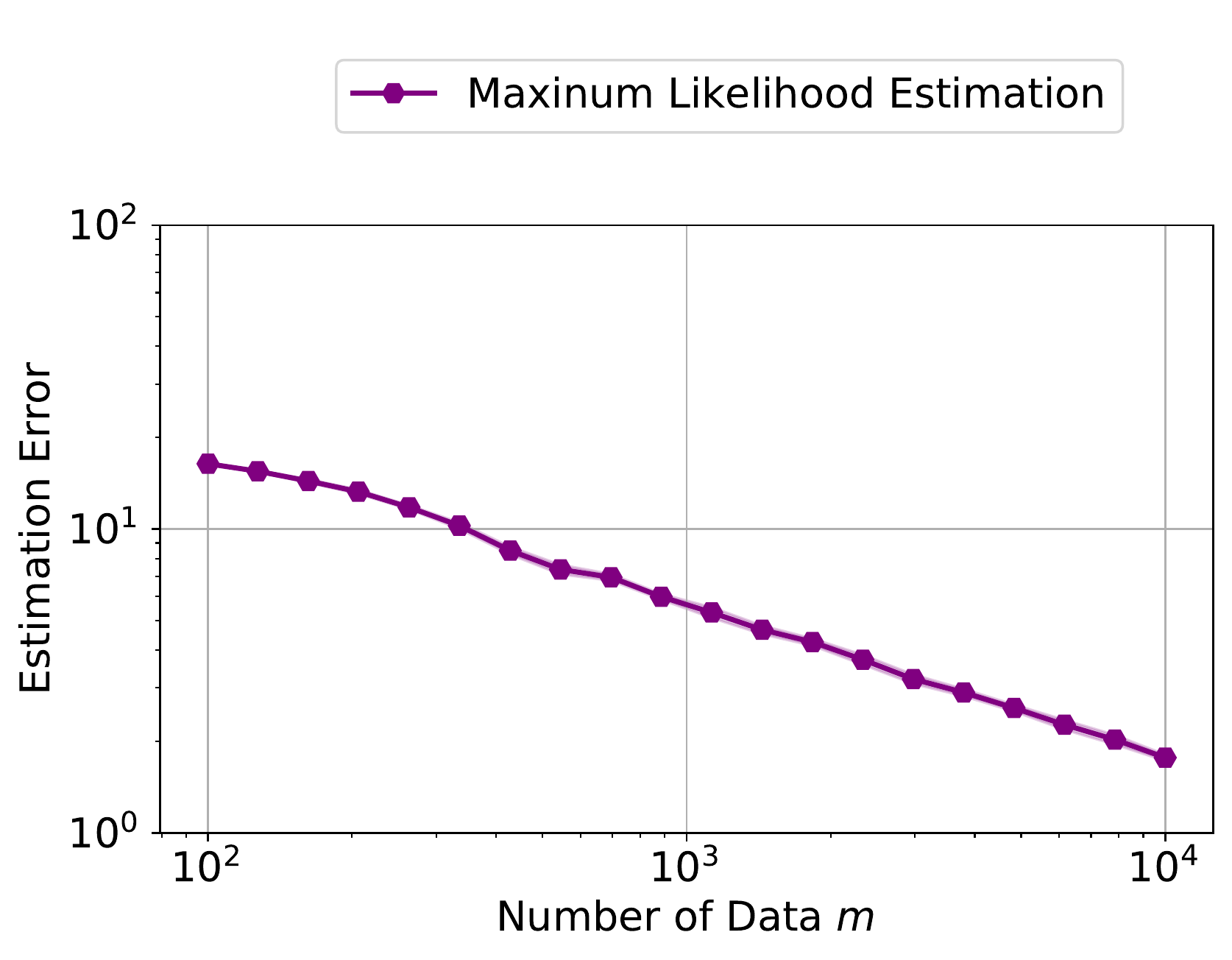}
  \caption{The $\ell_1$-norm estimation error on Standard Imitation.}
  \label{fig:mle_estimation_error_bandit}
\end{subfigure}
\hfill
\begin{subfigure}{.42\textwidth}
  \centering
  \includegraphics[width=0.8\textwidth]{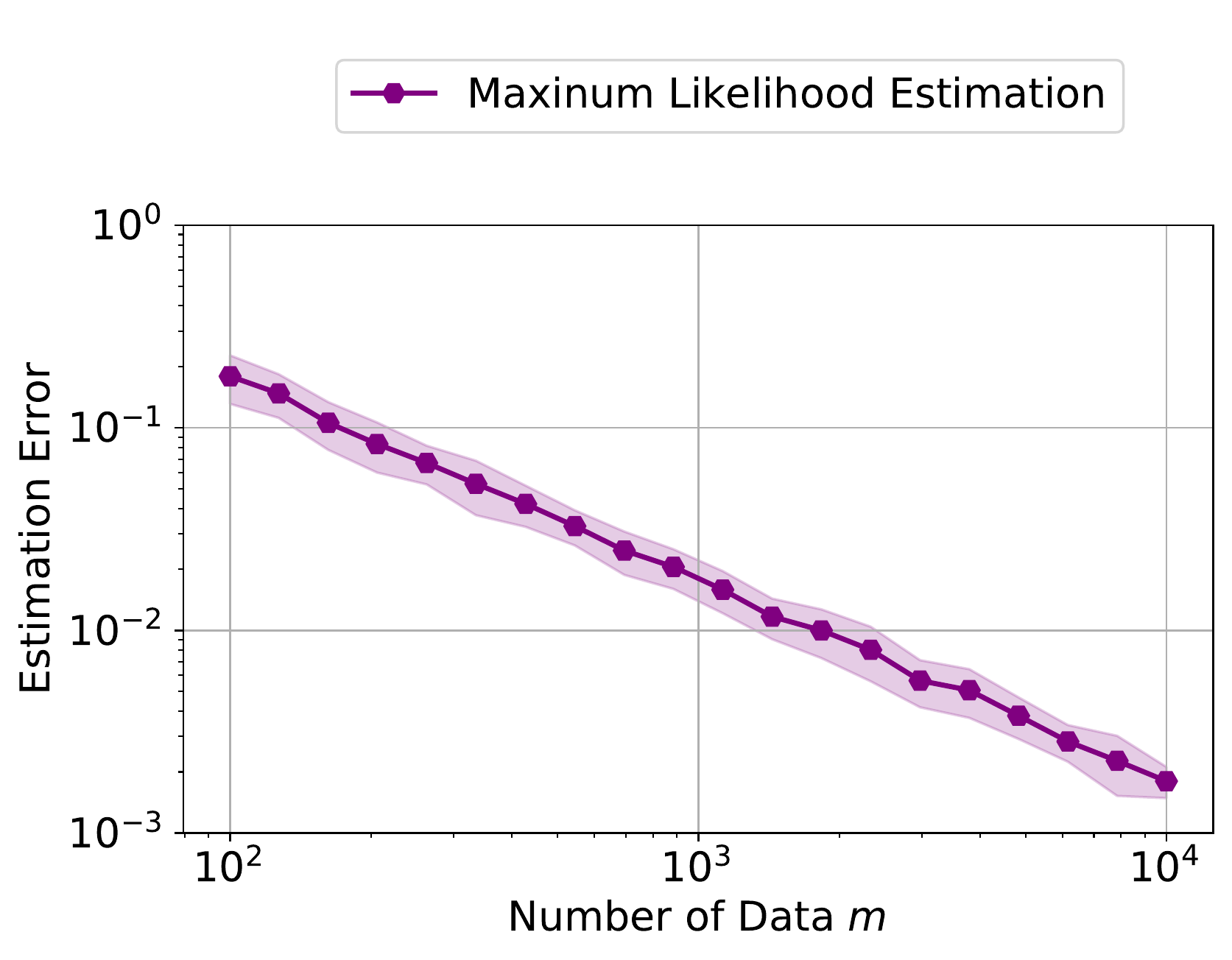}
  \caption{The $\ell_1$-norm estimation error on Reset Cliff.}
  \label{fig:mle_estimation_error_cliffwalking}
\end{subfigure}
\caption{The $\ell_1$-norm estimation error of maximum likelihood estimation $\sum_{h=1}^{H} \Vert P^{\piE}_h - \widehat{P}_h^{\piE} \Vert_1$ on Standard Imitation and Reset Cliff with different number of expert demonstrations.}
\label{fig:estimation_result}
\end{figure}

On the other hand, we consider the dependence on the number of expert demonstrations; the corresponding numerical result is shown in Figure \ref{fig:cliffwalking_m_log_result}. From Figure \ref{fig:cliffwalking_m_log_result}, we see that the slopes of all methods are around $-1$. Combined with the quadratic horizon dependency of BC, we empirically validate that the sample complexity analysis of BC is tight. Notice that we do not empirically observe the sample barrier issue of VAIL on Reset Cliff. The reason is that there is no statistical difficulty in estimating the state distribution of the expert policy on Reset Cliff. More specifically, instead of the uniform distribution on Standard Imitation, the state distribution of the expert policy on Reset Cliff is $( {1}/({m+1}), \cdots, {1}/({m+1}), 1 - {(\vert \mathcal{S} \vert -2)}/{(m+1)}, 0)$ in each step. When $m$ is large, this distribution assigns all probability mass on the 
penultimate state and thus the estimation problem is easy. We empirically validate this claim. In particular, the $\ell_1$-norm estimation error of maximum likelihood estimation is illustrated in \cref{fig:estimation_result}. We see that the slope on Standard Imitation is about $-1/2$ while the slope on Reset Cliff is about $-1$. This result is consistent with the policy value gap of VAIL on Standard Imitation (\cref{fig:bandit_m_log_result}) and Reset Cliff (\cref{fig:cliffwalking_m_log_result}).

\subsection{Unknown Transitions Setting}
\label{appendix:experiment_unknown_transition}

In this part, we study the interaction complexity under the unknown transition setting. We still use the above two MDPs, but they may not be hard instances. Hence, we do not verify the tightness of order dependency. The comparison involves BC~\citep{Pomerleau91bc}, OAL~\citep{shani21online-al} and MB-TAIL (see Algorithm \ref{algo:mbtail-abstract}). All algorithms are provided with the same expert demonstrations.

\begin{figure}[htbp]
\begin{subfigure}{.42\textwidth}
  \centering
  \includegraphics[width=0.8\textwidth]{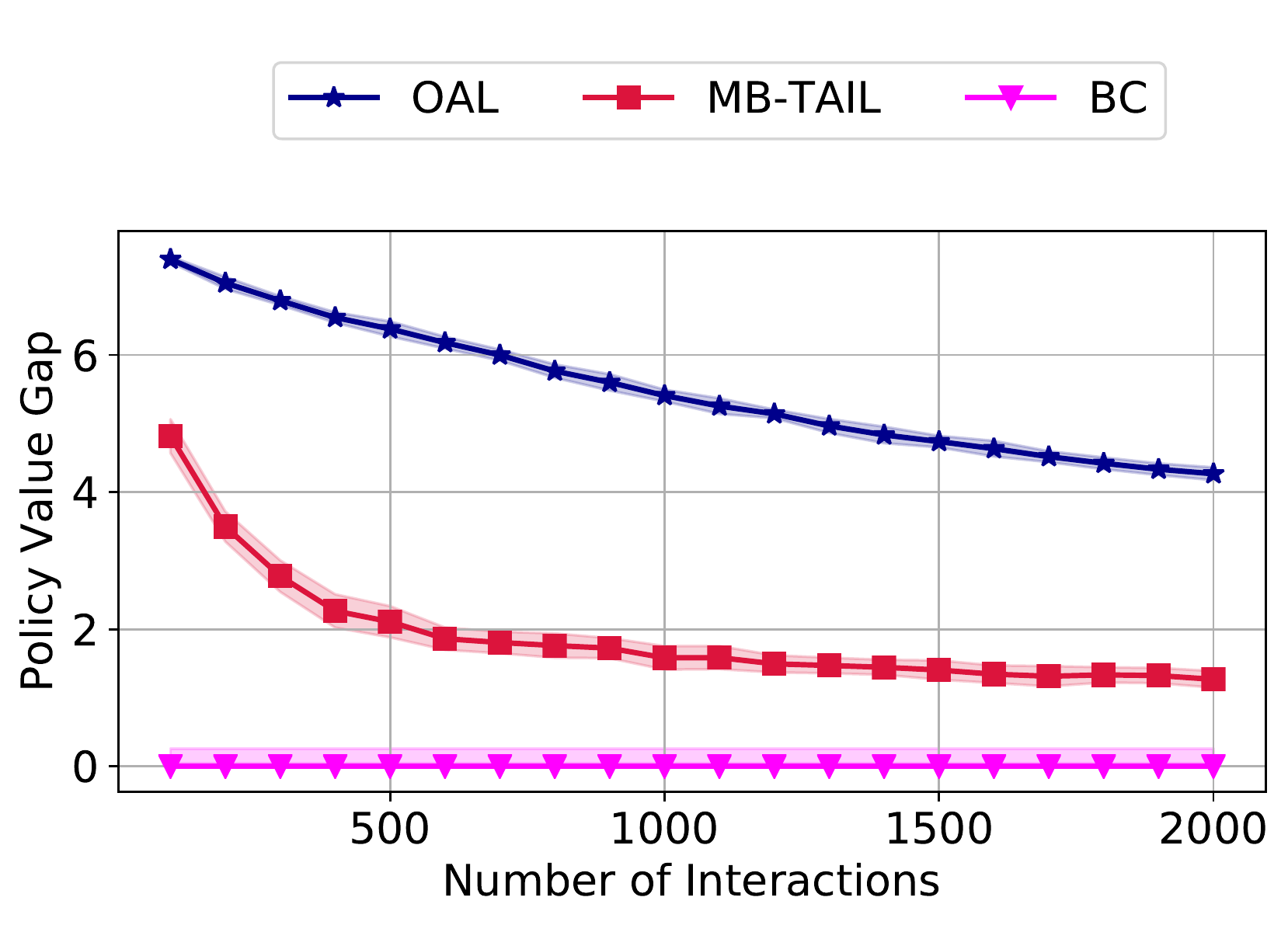}
  \caption{The policy value gap on Standard Imitation.}
  \label{fig:unknown_transition_bandit}
\end{subfigure}
\hfill
\begin{subfigure}{.42\textwidth}
  \centering
  \includegraphics[width=0.8\textwidth]{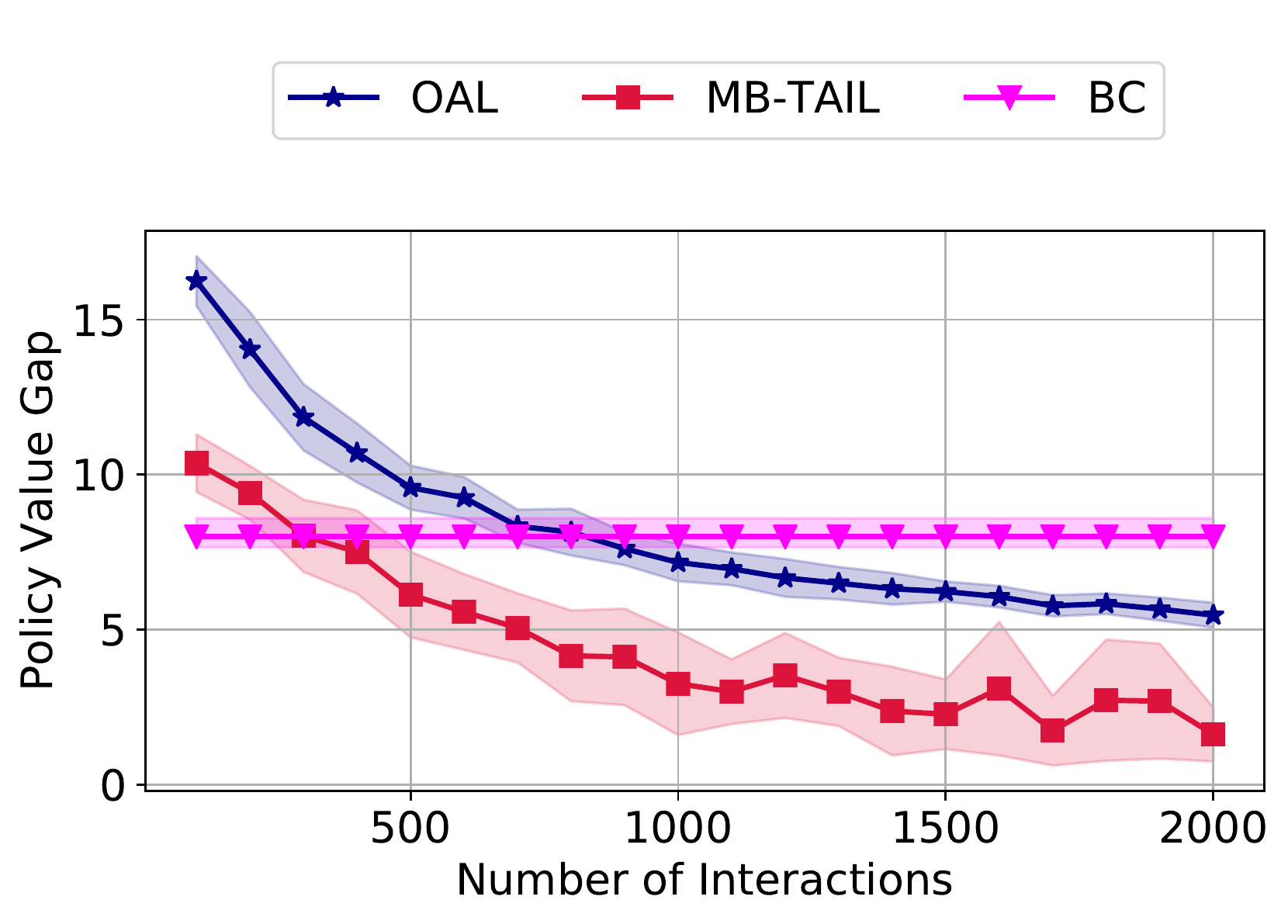}
  \caption{The policy value gap on Reset Cliff.}
  \label{fig:unknown_transition_cliffwalking}
\end{subfigure}
\caption{The policy value gap (i.e., $V^{\piE} - V^{\pi}$) on Standard Imitation and Reset Cliff with different number of interactions.}
\label{fig:unknown_transition_result}
\end{figure}

Empirical results are displayed in Figure~\ref{fig:unknown_transition_result}. Note that BC does not need interaction. Similar to the results shown in Figure \ref{figure:main_results}, BC performs worse than MB-TAIL on Reset Cliff while BC could be better than MB-TAIL on Standard Imitation. Moreover, we see that MB-TAIL outperforms OAL provided with the same number of interactions.

\subsection{Experiment Details}
\label{subsection:experiment_details}
\subsubsection{Known Transition Setting}

All experiments are run with $20$ random seeds. The detailed information on tasks is listed in Table \ref{table:task_information}. All experiments  are run on the machine with $32$ CPU cores, $128$ GB RAM and NVIDIA GeForce RTX $2080$ Ti.

BC directly estimates the expert policy from expert demonstrations. The information on the number of optimization iterations of VAIL, FEM, GTAL and TAIL is summarized in Table \ref{table:num_iterations}. In each iteration, with the recovered reward function, all conventional AIL methods utilize value iteration to solve the optimal policy. As discussed in ~\citep{Zahavy20al_via_frank-wolfe}, the optimization problem of FEM can be solved by Frank Wolfe (FW) algorithm ~\citep{frank1956algorithm}. In particular, the step size of FW is determined by line search. GTAL uses multiplicative weights to solve the outer problem in \eqref{eq:new_algo_max_min}. VAIL and our algorithm TAIL utilize online gradient descent to update the reward function. To utilize the optimization structure, an adaptive step size~\citep{Orabona19a_modern_introduction_to_ol} is implemented for GTAL, VAIL and our algorithm TAIL\footnote{Conclusions about the sample complexity and computational complexity do not change by this adaptive step size.}: 
\begin{align*}
    \eta_{t} = \frac{D}{ \sqrt{\sum_{i=1}^t \lnorm \nabla_{w} f^{(i)} \lp w^{(i)} \rp \rnorm_2^2}},
\end{align*}
where $D = \sqrt{2H |\gS| |\gA|}$ is the diameter of the set $\gW$. After the training process, we evaluate the policy value via exact Bellman update.

\begin{table}[htbp]
\caption{Information about tasks under known transition setting.}
\label{table:task_information}
\centering
{ \small
\begin{tabular}{@{}lllll@{}}
\toprule
Tasks           & Number of states & Number of actions & Horizon & Number of expert trajectories \\ \midrule

Standard Imitation (Figure \ref{fig:bandit_h_log_result})   & 500             & 5 & $10^{1} \to 10^{3}$ & 300 \\ 
\midrule
Standard Imitation (Figure \ref{fig:bandit_m_log_result}) & 500 & 5 &  $10$ &$10^{2} \to 10^{4}$         \\
\midrule
Reset Cliff (Figure \ref{fig:cliffwalking_h_log_result})          & 20              & 5      & $10^{1} \to 10^{3}$          & 5000           \\
\midrule
Reset Cliff (Figure \ref{fig:cliffwalking_m_log_result}) & 5 & 5 & $5$ & $10^2 \to 10^4$
\\
\bottomrule
\end{tabular}%
}
\end{table}

\begin{table}[htbp]
\caption{The number of optimization iterations of different algorithms on Standard Imitation and Reset Cliff.}
\label{table:num_iterations}
\centering
{ \small
\begin{tabular}{@{}lllll@{}}
\toprule
Tasks          & VAIL & FEM & GTAL  &TAIL \\ \midrule
Standard Imitation (Figure \ref{fig:bandit_h_log_result}) & $500$ & $500$ & $500$ & $500$
\\
\midrule
Standard Imitation (Figure \ref{fig:bandit_m_log_result}) & $8000$ & $8000$ & $8000$ & $8000$ \\ 
\midrule
Reset Cliff (Figure \ref{fig:cliffwalking_h_log_result})                      & $4H$ & $300$      & $4H$            & $H$                    \\ \midrule
Reset Cliff (Figure \ref{fig:cliffwalking_m_log_result}) & $20000$ & $20000$ & $20000$ & $20000$
\\ \bottomrule
\end{tabular}%
}
\end{table}

\subsubsection{Unknown Transition Setting}

All experiments are run with $20$ random seeds. Table \ref{table:task_information_unknown_transition_setting} summaries the detailed information on tasks under the unknown transition setting.

In particular, OAL is a model-based method and uses mirror descent (MD)~\citep{beck2003mirror} to optimize policy and reward. The step sizes of MD are set by the results in the theoretical analysis of~\citep{shani21online-al}. During the interaction, OAL maintains an empirical transition model to estimate Q-function for policy optimization. To encourage exploration, OAL adds a bonus function to the Q-function. The bonus used in the theoretical analysis of~\citep{shani21online-al} is too 
large in experiments and hence, OAL requires too many interactions to reach a good and stable performance. Therefore, we simplify their bonus function from $b_{h}^k (s, a)=\sqrt{ \frac{4  |\gS| H^{2} \log \lp 3  |\gS| |\gA| H^{2} n / \delta \rp}{ n_{h}^{k}(s, a) \vee 1}}$ to  $b_{h}^k (s, a)=\sqrt{\frac{ \log \lp  |\gS| |\gA| H n / \delta \rp}{n_{h}^{k}(s, a) \vee 1}}$, where $n$ is the total number of interactions, $\delta$ is the failure probability and $n^{k}_h (s, a)$ is the number of times visiting $(s, a)$ in time step $h$ until episode $k$.

MB-TAIL first establishes the estimator in \eqref{eq:new_estimator_unknown_transition} with half of the environment interactions and learns an empirical transition model by invoking RF-Express~\citep{menard20fast-active-learning} to collect the other half of trajectories. Subsequently, MB-TAIL performs policy and reward optimization with the recovered transition model. In MB-TAIL, the policy and reward optimization steps are the same as TAIL.

\begin{table}[htbp]
\caption{Information about tasks under unknown transition setting.}
\label{table:task_information_unknown_transition_setting}
\centering
{ \small
\begin{tabular}{@{}lllll@{}}
\toprule
Tasks           & Number of states & Number of actions & Horizon & Number of expert trajectories \\ \midrule
Reset Cliff          & 20              & 5      &  20          & 100           \\
\midrule 
Standard Imitation   & 100             & 5 & 10  & 400          \\ \bottomrule
\end{tabular}%
}
\end{table}

\subsection{GAIL}
\label{subsection:additional_results_gail}

\begin{figure*}[htbp]
     \centering
     \begin{subfigure}[b]{0.23\textwidth}
         \centering
         \includegraphics[width=\textwidth]{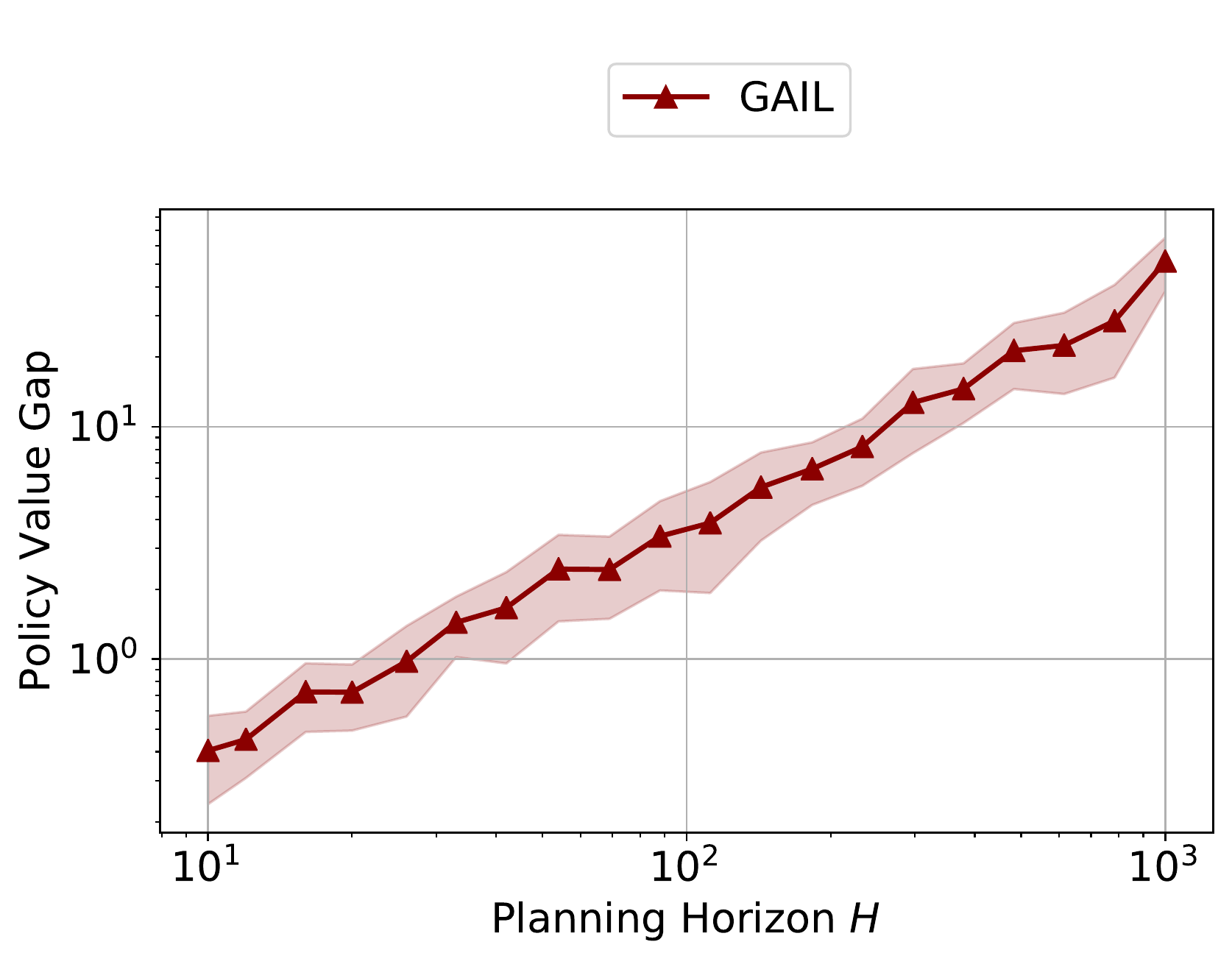}
         \caption{On the planning horizon on Standard Imitation.}
         \label{fig:gail_bandit_h_log_result}
     \end{subfigure}
     \hfill
     \begin{subfigure}[b]{0.23\textwidth}
         \centering
         \includegraphics[width=\textwidth]{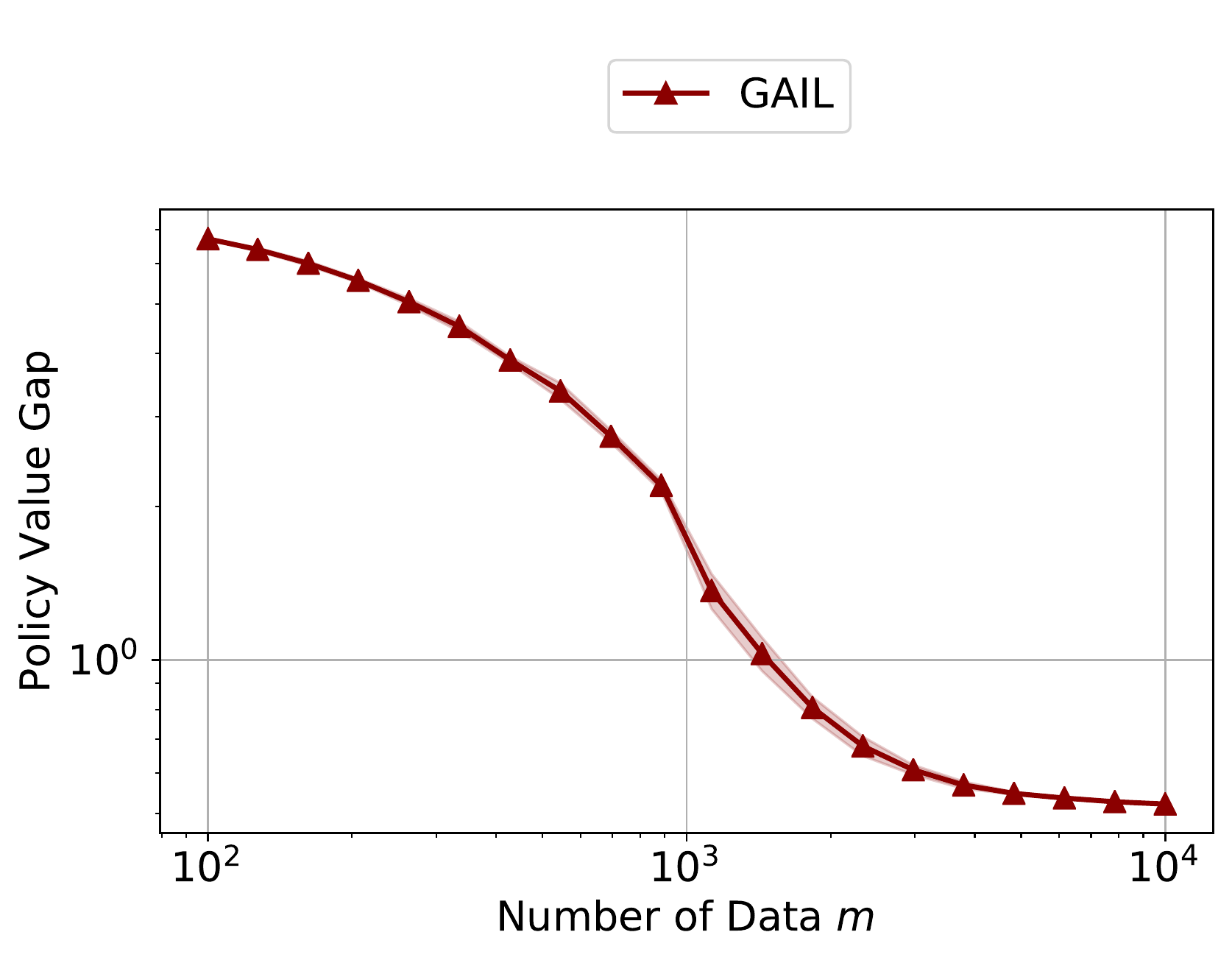}
         \caption{On the expert sample size on Standard Imitation.}
         \label{fig:gail_bandit_m_log_result}
     \end{subfigure}
     \hfill
     \begin{subfigure}[b]{0.23\textwidth}
         \centering
         \includegraphics[width=\textwidth]{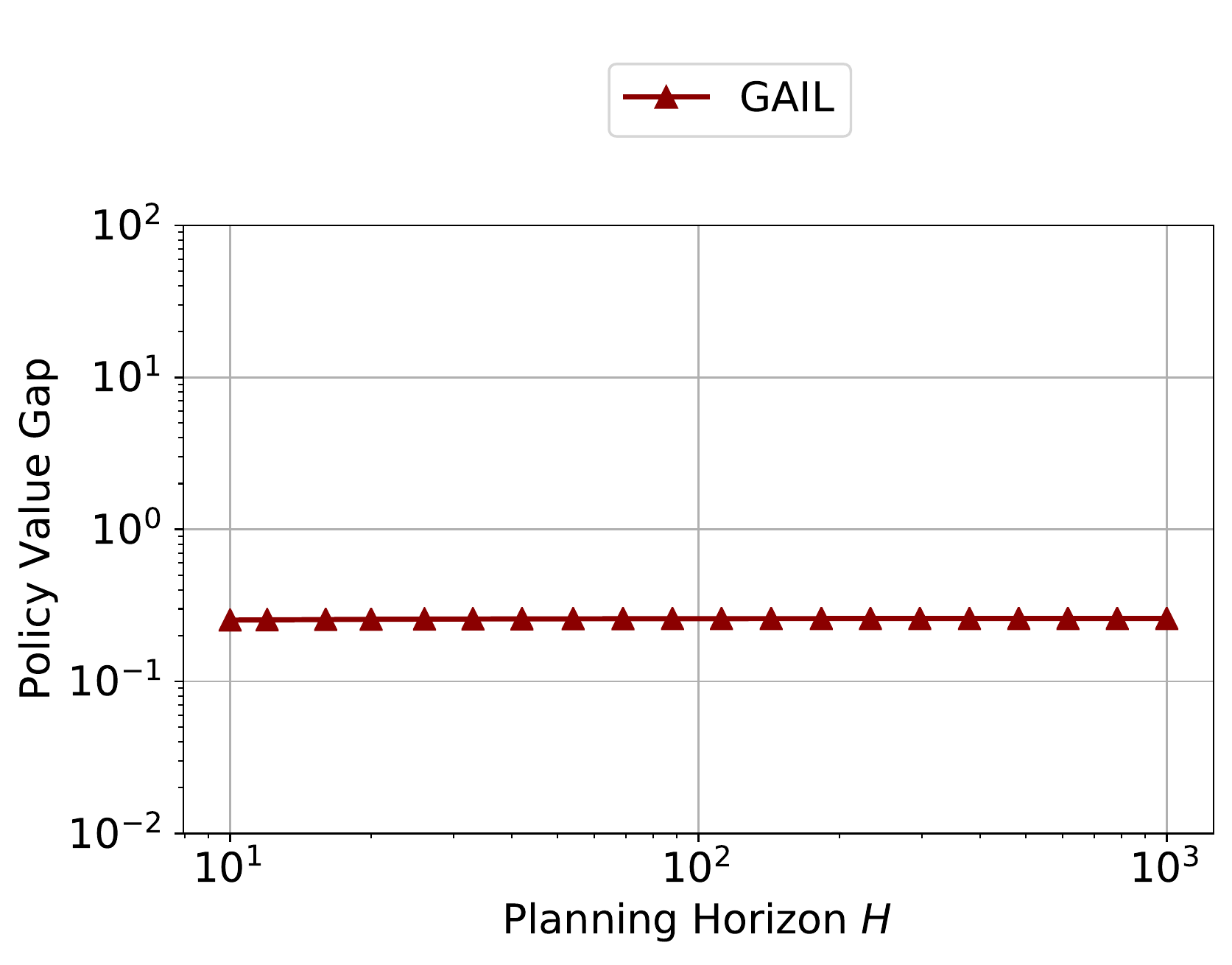}
         \caption{On the planning horizon on Reset Cliff.}
         \label{fig:gail_cliffwalking_h_log_result}
     \end{subfigure}
     \hfill
    \begin{subfigure}[b]{0.23\textwidth}
      \centering
      \includegraphics[width=\textwidth]{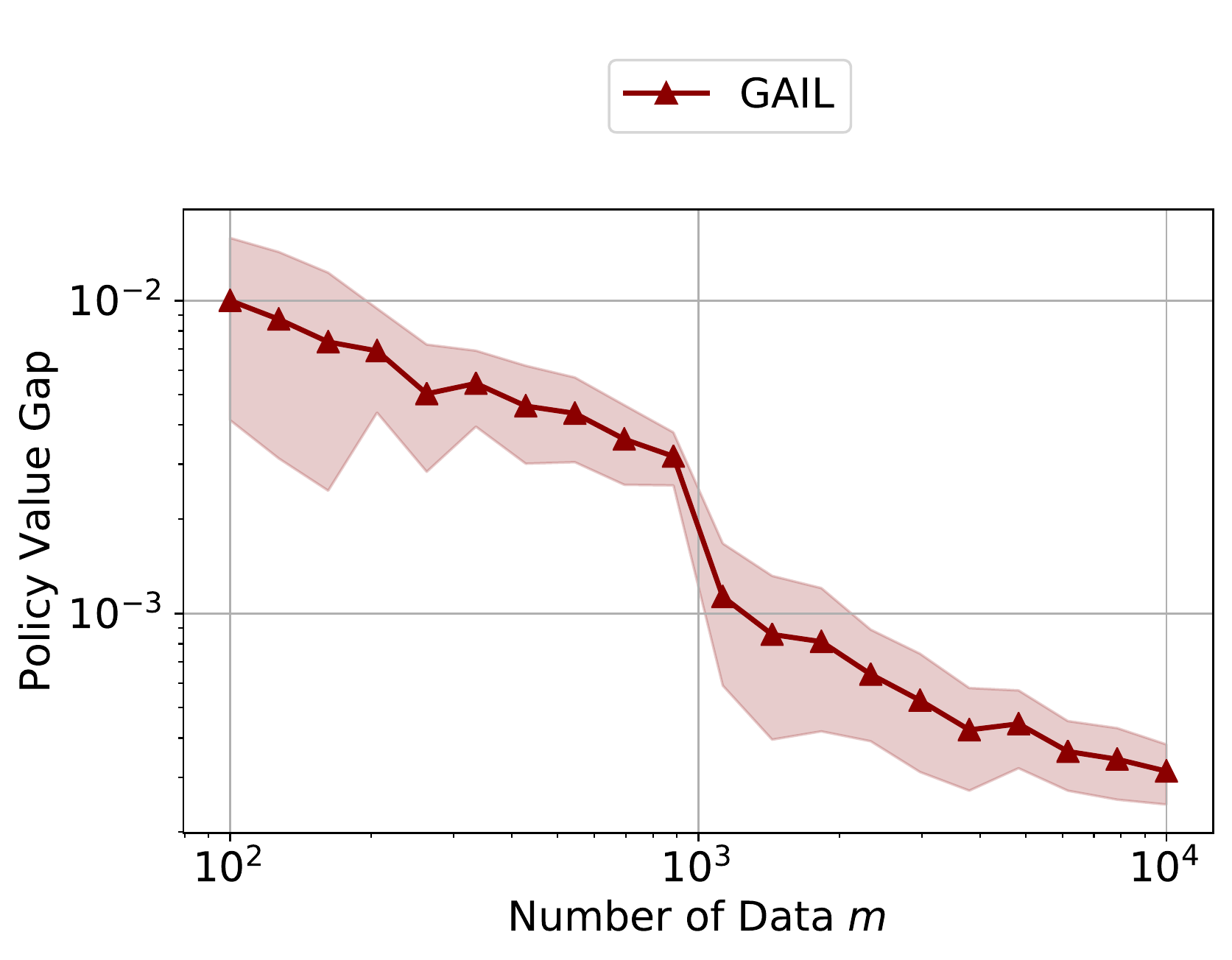}
      \caption{On the expert sample size on Reset Cliff.}
      \label{fig:gail_cliffwalking_m_log_result}
    \end{subfigure}

     \caption{The policy value gap (i.e., $V^{\piE} - V^{\pi}$) of GAIL on Standard Imitation and Reset Cliff. The solid lines are mean of results and the shaded region corresponds to the standard deviation over $20$ random seeds.}
     \label{figure:gail_results}
\end{figure*}

Under the known transition setting, we also test a famous practical AIL method named GAIL~\citep{ho2016gail}. Let $D = (D_1, \cdots, D_H)$ with $D_h: \gS \times \gA \rar [0, 1]$ for $h \in [H]$. The min-max objective of GAIL is shown as follows.
\begin{align}
    \min_{\pi \in \Pi} \max_{D} \sum_{h=1}^H \expect_{(s, a) \sim P^{\piE}_h} \ls \log \lp 1 - D_h (s, a) \rp \rs + \sum_{h=1}^H \expect_{(s, a) \sim P^{\pi}_h} \ls \log \lp D_h (s, a) \rp \rs.
    \label{eq:gail_objective}
\end{align}
\citet{ho2016gail} provided a practical implementation of GAIL under the unknown transition setting. Specifically, GAIL uses stochastic gradient descent ascent (SGDA) to update the policy and reward function alternatively. It is well-known even the full-batch version of SGDA (i.g., GDA) may not converge properly \citep{benaim1999mixed, lin2020gda}. As such, GAIL has no theoretical guarantee about the convergence or sample complexity.

To study the sample complexity of GAIL under the known transition setting, we make a small modification. In particular, we use the closed-form solution to the inner loop problem in \eqref{eq:gail_objective}:
\begin{align*}
    D^*_h (s, a) = \frac{P^{\pi^{(t)}}_h (s, a)}{P^{\pi^{(t)}}_h (s, a) + P^{\piE}_h (s, a)}.
\end{align*}
Then the recovered reward function is 
\begin{align*}
    w^{(t+1)}_h (s, a) = - \log \lp D^*_h (s, a) \rp.
\end{align*}
As for the policy, we use the mirror descent update~\citep{shalev12online-learning}, which is widely applied to solving a saddle point problem:
\begin{align*}
    \pi^{(t+1)}_h (a|s) = \frac{\pi^{(t)}_h (a|s) \exp \lp \eta Q^{\pi^{(t)}}_h (s, a) \rp}{\sum_{a^\prime \in \gA} \pi^{(t)}_h (a^\prime|s) \exp \lp \eta Q^{\pi^{(t)}}_h (s, a^\prime) \rp },
\end{align*}
where $\eta$ is the stepsize and $Q^{\pi^{(t)}}_h (s, a)$ is the action value function of $\pi^{(t)}$ with reward $w^{(t)}$.

The results of GAIL on Standard Imitation and Reset Cliff are plotted in Figure \ref{figure:gail_results}. Compared with results in Figure \ref{figure:main_results}, we see that the performance of GAIL is comparative with other conventional AIL methods such as FEM and GTAL. In particular, there is no  difference in the order dependence of the planning horizon and the expert sample size between GAIL and conventional AIL methods. This is reasonable since all of them follow the state-action distribution matching principle and use the naive estimation in \eqref{eq:estimate_by_count}.

\section{Proof of Results in Section \ref{sec:generalization_of_ail}}
\label{appendix:proof_generalization_ail}

\subsection{Proof of Theorem \ref{theorem:worst_case_sample_complexity_of_vail}}

First, we formally state the result on the worst-case sample complexity for \textsf{VAIL} to achieve a small policy value gap \emph{with high probability}. Notice that this result does not change too much compared with that \emph{in expectation}.

\begin{thm}[High Probability Version of \cref{theorem:worst_case_sample_complexity_of_vail}] \label{theorem:worst_case_sample_complexity_of_vail_high_prob}
For any tabular and episodic MDP, with probability at least $1-\delta$, to obtain an $\varepsilon$-optimal policy (i.e., $V^{\piE} -  V^{\piail} \leq \varepsilon$), \textsf{VAIL} in \eqref{eq:ail} requires at most $\widetilde{\gO} ( |\gS| H^2/\varepsilon^2)$ expert trajectories. 
\end{thm}

\begin{proof}[Proof of \cref{theorem:worst_case_sample_complexity_of_vail} and \cref{theorem:worst_case_sample_complexity_of_vail_high_prob}]

In the following part, we provide proof for both \cref{theorem:worst_case_sample_complexity_of_vail} and \cref{theorem:worst_case_sample_complexity_of_vail_high_prob}. To prove \cref{theorem:worst_case_sample_complexity_of_vail} and \cref{theorem:worst_case_sample_complexity_of_vail_high_prob}, we take two steps. For the first step, we extend \citep[Lemma 1]{xu2020error} from infinite-horizon MDPs to episodic MDPs. Suppose that $\piail$ is the optimal solution of the \textsf{VAIL} objective in \eqref{eq:ail}. We have the following re-formulation for policy value (see \cref{lemma:policy_dual_value}): 
\begin{align*}
    V^{\piail} = \sum_{h=1}^{H} \sum_{(s, a) \in \gS \times \gA} P^{\piail}_h(s, a) r_h(s, a).
\end{align*}
Then, we obtain the following decomposition for the policy value gap:
\begin{align*}
    V^{\piE} - V^{\piail} &= \sum_{h=1}^{H} \sum_{(s, a) \in \gS \times \gA} \lp P^{\piE}_h(s, a) - P^{\piail}_h(s, a) \rp r_h(s, a).
\end{align*}
Recall that $r_h(s, a) \in [0, 1]$ by assumption. It is straightforward to see that for the optimal solution $\piail$ to \textsf{VAIL}'s objective in \eqref{eq:ail}, we have 
\begin{align*}
    \labs  V^{\piE} - V^{\piail} \rabs &\leq \sum_{h=1}^{H}  \sum_{(s, a) \in \gS \times \gA} \labs P^{\piE}_h(s, a) - P^{\piail}_h(s, a) \rabs \\
    &\leq \sum_{h=1}^{H}  \sum_{(s, a) \in \gS \times \gA} \labs P^{\piE}_h(s, a) - \widehat{P}^{\piE}_h(s, a) \rabs + \sum_{h=1}^{H}  \sum_{(s, a) \in \gS \times \gA} \labs  \widehat{P}^{\piE}_h(s, a) - P^{\piail}_h(s, a) \rabs \\
    &\leq 2 \sum_{h=1}^{H}  \sum_{(s, a) \in \gS \times \gA} \labs P^{\piE}_h(s, a) - \widehat{P}^{\piE}_h(s, a) \rabs,
\end{align*}
where the last inequality holds because  
\begin{align*}
 \sum_{h=1}^{H}  \sum_{(s, a) \in \gS \times \gA} \labs  \widehat{P}^{\piE}_h(s, a) - P^{\piail}_h(s, a) \rabs &= \min_{\pi} \sum_{h=1}^{H}  \sum_{(s, a) \in \gS \times \gA} \labs P^{\pi}_h(s, a) - \widehat{P}^{\piE}_h(s, a) \rabs \\
 &\leq \sum_{h=1}^{H}  \sum_{(s, a) \in \gS \times \gA} \labs P^{\piE}_h(s, a) - \widehat{P}^{\piE}_h(s, a) \rabs.
\end{align*}
Notice that $\piE$ is deterministic and hence $P^{\piE}_h (s, a) = \widehat{P}^{\piE}_h (s, a) = 0$ for $a \not= \piE_h(s)$. Then we have that
\begin{align*}
    \labs  V^{\piE} - V^{\piail} \rabs &\leq 2\sum_{h=1}^{H}  \sum_{(s, a) \in \gS \times \gA} \labs P^{\piE}_h(s, a) - \widehat{P}^{\piE}_h(s, a) \rabs
    \\
    &= 2\sum_{h=1}^{H}  \sum_{s \in \gS } \labs P^{\piE}_h(s, \piE_h (s)) - \widehat{P}^{\piE}_h(s, \piE_h (s)) \rabs
    \\
    &= 2\sum_{h=1}^{H}  \sum_{s \in \gS } \labs P^{\piE}_h(s) - \widehat{P}^{\piE}_h(s) \rabs 
\end{align*}
For the second step, we upper bound the estimation error between $P^{\piE}_h(s)$ and $\widehat{P}^{\piE}_h(s)$. We first prove the sample complexity to achieve a small policy value gap with \emph{high probability}. 

\begin{lem}[Concentration Inequality for Total Variation Distance \citep{weissman2003inequalities}]  \label{lemma:l1_concentration}
Let $\gX = \{1, 2, \cdots, |\gX|\}$ be a finite set. Let $P$ be a distribution on $\gX$. Futhermore, let $\widehat{P}$ be the empirical distribution given $m$ i.i.d. samples $x_1, x_2, \cdots, x_m$ from $P$, i.e.,
\begin{align*}
    \widehat{P}(j) = \frac{1}{m} \sum_{i=1}^{m} \mathbb{I} \lb x_i = j \rb.
\end{align*}
Then, with probability at least $1-\delta$, we have that 
\begin{align*}
    \lnorm P - \widehat{P} \rnorm_1 := \sum_{x \in \gX} \labs P(x) - \widehat{P}(x) \rabs \leq \sqrt{\frac{2 |\gX| \ln(1/\delta) }{m}}.
\end{align*}
\end{lem}

It is clear that each $\widehat{P}^{\piE}_h(s)$ is an empirical estimation for $P^{\piE}_h(s)$. By \cref{lemma:l1_concentration}, for any fixed $h$, with probability at least $1-\delta$, we have that 
\begin{align*}
    \sum_{s \in \gS } \labs P^{\piE}_h(s) - \widehat{P}^{\piE}_h(s) \rabs \leq \sqrt{\frac{2 |\gS| \ln(1/\delta) }{m}},
\end{align*}
where $m$ is the number of expert trajectories in the dataset $\gD$. By a union bound over $h = 1, \cdots, H$, with probability at least $1-\delta$,  we have that 
\begin{align*}
    \sum_{h=1}^{H} \sum_{s \in \gS } \labs P^{\piE}_h(s) - \widehat{P}^{\piE}_h(s) \rabs \leq H\sqrt{\frac{2 |\gS| \ln(H/\delta) }{m}}.
\end{align*}
As a result, we obtain that with probability at least $1-\delta$ 
\begin{align*}
     \labs  V^{\piE} - V^{\piail} \rabs \leq 2H\sqrt{\frac{2 |\gS| \ln(H/\delta) }{m}},
\end{align*}
which translates to a sample complexity $\widetilde{\gO}( |\gS| H^2/\varepsilon^2)$ with high probability.

Second, we prove the corresponding sample complexity to achieve a small policy value gap \emph{in expectation}. With \citep[Theorem 1]{han2015minimax}, we can upper bound the expected $\ell-1$ risk of the maximum likelihood estimation.
\begin{align*}
    \forall h \in [H], \, \expect\ls \lnorm \widehat{P}^{\piE}_h - P^{\piE}_h   \rnorm_{1} \rs \leq \sqrt{ \frac{|\gS| - 1}{m}},
\end{align*}
where the expectation is taken w.r.t the randomness of expert demonstrations. For the expected policy value gap, we have 
\begin{align*}
     \expect \ls V^{\piE} - V^{\piail} \rs \leq H \sqrt{ \frac{|\gS| - 1}{m}}, 
\end{align*}
which translates to a sample complexity $\gO ( |\gS| H^2/\varepsilon^2)$ in expectation as in \cref{theorem:worst_case_sample_complexity_of_vail}.

\end{proof}

\subsection{Proof of Claim in Example \ref{example:ail_fail}}

In this part, we formally state and prove the theoretical result in \cref{example:ail_fail}.

\begin{claim} \label{claim:ail_standard_imitation}
Consider the MDP and expert demonstration configuration in \cref{example:ail_fail}. We have that $\piail (a^{1}|s^{1}) \in [0.8, 1.0], \piail (a^{1}|s^{2}) = 1$ are all globally optimal solutions of \eqref{eq:ail}. The largest policy value gap among all optimal solutions is $0.1$.  
\end{claim}

\begin{proof}[Proof of \cref{claim:ail_standard_imitation}]
The empirical distribution is calculated as
\begin{align*}
    \widehat{P}^{\piE} (s^{1}, a^{1}) = 0.4, \widehat{P}^{\piE} (s^{1}, a^{2}) = 0.0,
    \widehat{P}^{\piE} (s^{2}, a^{1}) = 0.6, \widehat{P}^{\piE} (s^{2}, a^{2}) = 0.0.
\end{align*}
With the above empirical distribution, we can obtain \textsf{VAIL}'s objective.
\begin{align*}
    &\quad \labs \widehat{P}^{\piE} (s^{1}, a^{1})  - \rho (s^{1}) \pi (a^{1}|s^{1})  \rabs + \labs \widehat{P}^{\piE} (s^{1}, a^{2}) - \rho (s^{1}) \lp 1-\pi (a^{1}|s^{1}) \rp \rabs
    \\
    &+ \labs \widehat{P}^{\piE} (s^{2}, a^{1}) - \rho (s^{2}) \pi (a^{1}|s^{2}) \rabs + \labs \widehat{P}^{\piE} (s^{2}, a^{2}) - \rho (s^{2}) \lp 1-\pi (a^{1}|s^{2}) \rp \rabs
    \\
    &= \labs 0.4 - 0.5 \pi (a^{1}|s^{1})  \rabs + \labs 0 - 0.5 \lp 1-\pi (a^{1}|s^{1}) \rp  \rabs + \labs 0.6 - 0.5 \pi (a^{1}|s^{2}) \rabs + \labs 0 - 0.5 \lp 1-\pi (a^{1}|s^{2}) \rp \rabs
    \\
    &= \labs 0.4 - 0.5 \pi (a^{1}|s^{1})  \rabs + 0.5 \lp 1-\pi (a^{1}|s^{1}) \rp + 1.1 -  \pi (a^{1}|s^{2}). 
\end{align*}
Notice that the optimization variables $\pi (a^{1}|s^{1}), \pi (a^{1}|s^{2})$ are independent and we can consider the optimization problem for each optimization variable. For $\pi (a^{1}|s^{1})$, we have that
\begin{align*}
    \piail (a^{1}|s^{1}) \in \argmin_{\pi (a^{1}|s^{1}) \in [0, 1]} \labs 0.4 - 0.5 \pi (a^{1}|s^{1})  \rabs - 0.5 \pi (a^{1}|s^{1}). 
\end{align*}
We apply \cref{lem:single_variable_opt_condition} with $a = 0.5$ and $c = 0.4$. We get that $\piail (a^{1}|s^{1}) \in [0.8, 1.0]$ is the optimal solution. For $\pi (a^{1}|s^{2})$, we have that
\begin{align*}
    \piail (a^{1}|s^{2}) \in \argmin_{\pi (a^{1}|s^{2}) \in [0, 1]} - \pi (a^{1}|s^{2}). 
\end{align*}
It is easy to see that $\piail (a^{1}|s^{2}) = 1$ is the optimal solution, which completes the proof.
\end{proof}

\subsection{Proof of Proposition \ref{proposition:ail_policy_value_gap_standard_imitation}}

\begin{figure}[htbp]
\centering
\includegraphics[width=0.7\linewidth]{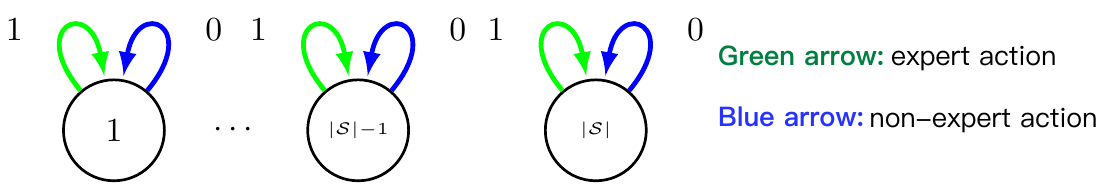}  
\caption{\textsf{Standard Imitation} MDPs corresponding to \cref{asmp:standard_imitation}.}
\label{fig:bandit}
\end{figure}

In this part, we extend the result in \cref{claim:ail_standard_imitation} to the Standard Imitation MDPs shown in \cref{fig:bandit}. In Standard Imitation, all states are absorbing, $a^{1}$ is the expert action (in green) and $a^{2}$ is the non-expert action (in blue). The initial state distribution is denoted as $\rho$.    \cref{proposition:ail_policy_value_gap_standard_imitation} indicates that the largest policy value gap of \textsf{VAIL} \emph{equals} half of the estimation error and formally demonstrates the weak convergence issue of \textsf{VAIL}.

\begin{proof}[Proof of \cref{proposition:ail_policy_value_gap_standard_imitation}]
Notice that each state is absorbing in Standard Imitation and thus, $P^{\pi}_h (s) = \rho (s), \forall s \in \gS, h \in [H]$. Then we obtain
\begin{align*}
  &\quad \min_{\pi \in \Pi} \sum_{h=1}^{H} \sum_{(s, a) \in \gS \times \gA} | P^{\pi}_h(s, a) - \widehat{P}^{\piE}_h(s, a) |
  \\
  &= \min_{\pi \in \Pi} \sum_{h=1}^{H} \sum_{(s, a) \in \gS \times \gA} | \rho (s) \pi_h (a|s) - \widehat{P}^{\piE}_h(s, a) |
  \\
  &= \min_{\pi \in \Pi} \sum_{h=1}^{H} \sum_{s \in \gS } \lp | \rho (s) \pi_h (a^{1}|s) - \widehat{P}^{\piE}_h(s, a^{1}) | + \rho (s) \pi_h (a^{2}|s) \rp
  \\
  &= \min_{\pi \in \Pi} \sum_{h=1}^{H} \sum_{s \in \gS } \lp | \rho (s) \pi_h (a^{1}|s) - \widehat{P}^{\piE}_h(s) | + \rho (s) \lp 1- \pi_h (a^{1}|s) \rp \rp
  \\
  &= \min_{\pi \in \Pi} \sum_{h=1}^{H} \sum_{s \in \gS } \lp | \rho (s) \pi_h (a^{1}|s) - \widehat{P}^{\piE}_h(s) | - \rho (s) \pi_h (a^{1}|s) \rp. 
\end{align*}
Since the optimization variables $\pi_h (a^{1}|s)$ for different $s \in \gS, h \in [H]$ are independent, we can consider the optimization problem for each $s \in \gS, h \in [H]$ individually. For each $h \in [H]$ and $s \in \gS$,
\begin{align*}
    \min_{\pi_h (a^{1}|s) \in [0, 1]} | \rho (s) \pi_h (a^{1}|s) - \widehat{P}^{\piE}_h(s) | - \rho (s) \pi_h (a^{1}|s). 
\end{align*}
For any state $s \in \gS^1_h = \{s \in \gS: \widehat{P}^{\piE}_h (s) < \rho (s) \} $, with \cref{lem:single_variable_opt_condition}, we have that $\piail_h (a^{1}|s) \in [\widehat{P}^{\piE}_h (s) / \rho (s), 1]$ are all optimal solutions. On the other hand, for any state $s \in \lp \gS^1_h \rp^c$, the optimization problem is reduced to
\begin{align*}
    \min_{\pi_h (a^{1}|s) \in [0, 1]}  \widehat{P}^{\piE}_h(s) - 2\rho (s) \pi_h (a^{1}|s).
\end{align*}
It is easy to see that the optimal solution is $\pi_h (a^{1}|s) = 1$. Therefore, for each time step $h$, $\piail_h (a^1|s) \in [\widehat{P}^{\piE}_h(s) / \rho (s), 1], \forall s \in \gS_h^1$ and $\piail_h (a^1|s) = 1, \forall s \in (\gS_h^1)^c$ are all optimal solutions of \eqref{eq:ail}.

For the policy value gap, according to the dual representation of policy value, we have that
\begin{align*}
    V^{\piE} - V^{\piail} &= \sum_{h=1}^{H} \sum_{(s, a) \in \gS \times \gA}  \lp P^{\piE}_h(s, a) - P^{\piail}_h(s, a) \rp r_h (s, a)
    \\
    &= \sum_{h=1}^{H} \sum_{s \in \gS}  P^{\piE}_h(s, a^{1}) - P^{\piail}_h(s, a^{1}) 
    \\
    &= \sum_{h=1}^{H} \sum_{s \in \gS } P^{\piE}_h(s)  - \rho(s) \piail_h( a^{1}|s).
\end{align*}
Among all optimal solutions, the largest policy value gap is obtained at $\forall h \in [H], \piail_h (a^1|s) = \widehat{P}^{\piE}_h(s) / \rho (s), \forall s \in \gS_h^1; \piail_h (a^1|s) = 1, \forall s \in (\gS_h^1)^c$. The largest policy value gap is
\begin{align*}
    V^{\piE} - V^{\piail} &= \sum_{h=1}^{H} \sum_{s \in \gS_h^1} P^{\piE}_h(s)  - \rho(s) \piail_h( a^{1}|s) = \sum_{h=1}^{H} \sum_{s \in \gS_h^1} P^{\piE}_h(s)  - \widehat{P}^{\piE}_h(s). 
\end{align*}
Next, we connect the term $\sum_{s \in \gS_h^1} P^{\piE}_h(s)  - \widehat{P}^{\piE}_h(s)$ with the $\ell_1$-norm estimation error. Notice that for each time step $h \in [H]$, $\sum_{s \in \gS} P^{\piE}_h(s) = \sum_{s \in \gS} \widehat{P}^{\piE}_h(s)$ = 1. Then we have that
\begin{align*}
    \sum_{s \in \gS^1_h} P^{\piE}_h(s)  - \widehat{P}^{\piE}_h(s) = \sum_{s \in (\gS^1_h)^c} \widehat{P}^{\piE}_h(s) - P^{\piE}_h(s).  
\end{align*}
Furthermore, we obtain
\begin{align}
     \sum_{h=1}^H \lnorm \widehat{P}^{\piE}_h - P^{\piE}_h   \rnorm_1 &= \sum_{h=1}^H \sum_{s \in \gS} \labs P^{\piE}_h(s) - \widehat{P}^{\piE}_h (s)  \rabs \nonumber
     \\
     &= \sum_{h=1}^H \sum_{s \in \gS^1_h} P^{\piE}_h(s)  - \widehat{P}^{\piE}_h(s) + \sum_{s \in (\gS^1_h)^c} \widehat{P}^{\piE}_h(s) - P^{\piE}_h(s) \nonumber
     \\
     &= 2 \sum_{h=1}^H \sum_{s \in \gS^1_h} P^{\piE}_h(s)  - \widehat{P}^{\piE}_h(s), \label{eq:vail_standard_imitation_policy_value_gap_half_estimation_error}
\end{align}
where the penultimate equality follows that $\gS^1_h = \{s \in \gS: \widehat{P}^{\piE}_h (s) < \rho (s) \}$. Finally, we get that
\begin{align*}
    V^{\piE} - V^{\piail} = \sum_{h=1}^{H} \sum_{s \in \gS_h^1} P^{\piE}_h(s)  - \widehat{P}^{\piE}_h(s) = \frac{1}{2}  \sum_{h=1}^H \lnorm \widehat{P}^{\piE}_h - P^{\piE}_h   \rnorm_1.
\end{align*}
\end{proof}

\subsection{Proof of Proposition \ref{prop:lower_bound_vail}}

\begin{proof}

To prove \cref{prop:lower_bound_vail}, we make two steps. First, we connect the policy value gap with the estimation error with the help of \cref{proposition:ail_policy_value_gap_standard_imitation}. Consider the Standard Imitation MDP in \cref{asmp:standard_imitation}, given estimation $\widehat{P}^{\piE}_h$, for each time step $h \in [H]$, recall the definition of $\gS_h^1 := \{s \in \gS: \widehat{P}^{\piE}_h (s) < \rho (s)   \}$. We construct a policy set $\Pi^{\ail}$ defined as
\begin{align*}
   \Pi^{\ail}: = \lb \pi \in \Pi: \forall h \in [H], \forall s \in \gS_h^1, \piail_h (a^1|s) \in [\widehat{P}^{\piE}_h(s) / \rho (s), 1]; \forall s \in (\gS_h^1)^c,  \piail_h (a^1|s) =1  \rb. 
\end{align*}
With \cref{proposition:ail_policy_value_gap_standard_imitation}, we have that $\Pi^{\ail}$ is the set of all globally optimal solutions of \textsf{VAIL}'s objective \eqref{eq:ail}. Note that \textsf{VAIL} outputs a policy $\piail$ uniformly sampled from $\Pi^{\ail}$. Same with the proof of \cref{proposition:ail_policy_value_gap_standard_imitation}, we have that
\begin{align*}
    V^{\piE} - V^{\piail} &= \sum_{h=1}^{H} \sum_{s \in \gS } P^{\piE}_h(s)  - \rho(s) \piail_h( a^{1}|s)
    \\
    &= \sum_{h=1}^{H} \sum_{s \in \gS_h^1 } P^{\piE}_h(s)  - \rho(s) \piail_h( a^{1}|s) .
\end{align*}
Taking expectation w.r.t the uniformly random selection of $\piail$ on both sides yields that
\begin{align*}
    V^{\piE} - \expect_{\piail \sim \text{Unif} (\Pi^{\ail})} \ls V^{\piail} \rs &= \expect_{\piail \sim \text{Unif} (\Pi^{\ail})} \ls \sum_{h=1}^{H} \sum_{s \in \gS_h^1 } P^{\piE}_h(s)  - \rho(s) \piail_h( a^{1}|s) \rs 
    \\
    &=  \sum_{h=1}^{H} \sum_{s \in \gS_h^1 } P^{\piE}_h(s) - \rho (s) \expect_{\piail_h( a^{1}|s) \sim \text{Unif} \lp [\widehat{P}^{\piE}_h(s) / \rho (s), 1] \rp } \ls \piail_h( a^{1}|s) \rs
    \\
    &= \frac{1}{2} \sum_{h=1}^{H} \sum_{s \in \gS_h^1} P^{\piE}_h(s)  - \widehat{P}^{\piE}_h(s).
\end{align*}
Combined with \eqref{eq:vail_standard_imitation_policy_value_gap_half_estimation_error}, we have that
\begin{align*}
    V^{\piE} - \expect_{\piail \sim \text{Unif} (\Pi^{\ail})} \ls V^{\piail} \rs = \frac{1}{2} \sum_{h=1}^{H} \sum_{s \in \gS_h^1} P^{\piE}_h(s)  - \widehat{P}^{\piE}_h(s) = \frac{1}{4} \sum_{h=1}^H \lnorm \widehat{P}^{\piE}_h - P^{\piE}_h   \rnorm_1.
\end{align*}
We further take the expectation over the randomness of expert demonstrations on both sides.
\begin{align*}
    V^{\piE} - \expect \ls \expect_{\piail \sim \text{Unif} (\Pi^{\ail})} \ls V^{\piail} \rs \rs = \frac{1}{4} \sum_{h=1}^H \expect \ls \lnorm \widehat{P}^{\piE}_h - P^{\piE}_h   \rnorm_1 \rs.
\end{align*}
Second, we apply the lower bound of expected $\ell_1$ risk of \citep[Corollary 9]{kamath2015learning} and have that
\begin{align*}
    V^{\piE} - \expect \ls \expect_{\piail \sim \text{Unif} (\Pi^{\ail})} \ls V^{\piail} \rs \rs \geq \frac{1}{4} H \sqrt{\frac{2 \lp \vert \gS \vert - 1 \rp}{\pi m}}.
\end{align*}
To obtain an $\varepsilon$-optimal policy (i.e., $V^{\piE} - \expect[ V^{\piail}] \leq \varepsilon$), in expectation, \textsf{VAIL} requires at least $\Omega(|\gS| H^2/\varepsilon^2)$ expert trajectories.

\end{proof}

\subsection{Proof of Theorem \ref{theorem:bc_deterministic}}

First, we formally state the result on the sample complexity for BC to achieve an $\varepsilon$-optimal policy \emph{with high probability}. This result is similar to \emph{in expectation} bound.

\begin{thm}[High Probability Version of \cref{theorem:bc_deterministic}]  \label{theorem:bc_deterministic_high_prob}
For any tabular and episodic MDP with deterministic transitions, with probability at least $1-\delta$, to obtain an $\varepsilon$-optimal policy (i.e., $V^{\piE} - V^{\pibc} \leq \varepsilon$), BC as in \eqref{eq:bc} requires at most ${\widetilde{\gO}}(|\gS| H/\varepsilon)$ expert trajectories.  
\end{thm}

\begin{proof}[Proof of \cref{theorem:bc_deterministic} and \cref{theorem:bc_deterministic_high_prob}]

In the following part, we provide proof for both \cref{theorem:bc_deterministic} and \cref{theorem:bc_deterministic_high_prob}. To prove \cref{theorem:bc_deterministic} and \cref{theorem:bc_deterministic_high_prob}, we make two steps. First, we show that when the transition function is deterministic, the policy value gap of BC comes from non-visited states in the first step.

Suppose that $\pibc$ is the minimizer of BC objective in \cref{eq:bc}. Then we have that
\begin{align*}
    &\quad V^{\piE} - V^{\pibc}
    \\
    &= \expect_{s_1 \sim \rho (\cdot)} \ls V^{\piE}_1 (s_1) - V^{\pibc}_1 (s_1)   \rs
    \\
    &= \expect_{s_1 \sim \rho (\cdot)} \ls \indict \lp s_1 \notin \gS_1 (\gD) \rp \lp V^{\piE}_1 (s_1) - V^{\pibc}_1 (s_1) \rp   \rs + \expect_{s_1 \sim \rho (\cdot)} \ls \indict \lp s_1 \in \gS_1 (\gD) \rp \lp V^{\piE}_1 (s_1) - V^{\pibc}_1 (s_1) \rp   \rs
    \\
    &= \expect_{s_1 \sim \rho (\cdot)} \ls \indict \lp s_1 \notin \gS_1 (\gD) \rp \lp V^{\piE}_1 (s_1) - V^{\pibc}_1 (s_1) \rp   \rs. 
\end{align*}
Since the expert policy and transition function are deterministic, the trajectories, started with the visited initial states, are fully covered in the expert demonstrations. Hence, the policy value gap on these trajectories is zero. This is our key observation for deterministic MDPs. Recall that $\gS_1(\gD)$ is the set of visited states in time step $1$ from expert dataset $\gD$. Then we have that
\begin{align*}
  V^{\piE} - V^{\pibc} &= \expect_{s_1 \sim \rho (\cdot)} \ls \indict \lp s_1 \notin \gS_1 (\gD) \rp \lp V^{\piE}_1 (s_1) - V^{\pibc}_1 (s_1) \rp   \rs
    \\
    &\leq H \expect_{s_1 \sim \rho (\cdot)} \ls \indict \lp s_1 \notin \gS_1 (\gD) \rp   \rs, 
\end{align*}
which is tighter than the result in \citep{rajaraman2020fundamental} since their result holds for general MDPs with stochastic transitions. Notice that $\expect_{s_1 \sim \rho (\cdot)} \ls \indict \lp s_1 \notin \gS_1 (\gD) \rp   \rs = \sum_{s \in \gS} \rho (s) \indict \lp s_1 \notin \gS_1 (\gD) \rp $ is the \emph{missing mass} of the distribution of $\rho$ given $m$ i.i.d. samples; see \cref{defn:missing_mass} for the definition of missing mass, which is from \citep[Defintion A.1]{rajaraman2020fundamental}.

\begin{defn}[Missing Mass \citep{rajaraman2020fundamental}] \label{defn:missing_mass}
Let $\gX = \{1, 2, \cdots, |\gX|\}$ be a finite set. Let $P$ be some distribution on $\gX$. Furthermore, let $X^{m} = (X_1, \cdots, X_m)$ be $m$ i.i.d. random variables from $P$.  Let $\mathfrak{n}_x(X^{m}) = \sum_{i=1}^{m} \mathbb{I}\{ X_i =  x\}$ be the number of times the element $x$ was observed in these random variables. Then, $\mathfrak{m}_0(P, X^{m}) = \sum_{x \in \gX} P(x) \mathbb{I} \{ n_x(X^{m}) = 0 \}$ is called missing mass, which means the probability mass contributed by elements never observed in $X^{m}$.
\end{defn}

Second, we need to upper bound the missing mass in the first time step. We first prove the sample complexity to achieve a small policy value gap \emph{with high probability}. To this end, we leverage the following concentration inequality \citep[Lemma A.3]{rajaraman2020fundamental}.
\begin{lem}[Concentration Inequality for Missing Mass \citep{rajaraman2020fundamental}]  \label{lemma:missing_mass_one_step}
Let $\gX = \{1, 2, \cdots, |\gX|\}$ be a finite set. Let $P$ be some distribution on $\gX$. Furthermore, let $X^{m} = (X_1, \cdots, X_m)$ be $m$ i.i.d. random variables from $P$. For any $\delta \in (0, 1/10]$, with probability at least $1-\delta$, we have 
\begin{align*}
\mathfrak{m}_0(P, X^{m}) := \sum_{x \in \gX} P(x) \mathbb{I} \{ n_x(X^{m}) = 0 \} \leq \frac{4|\gX|}{9m}  + \frac{3\sqrt{|\gX|} \log (1/\delta)}{m}.
\end{align*}
\end{lem}
With \cref{lemma:missing_mass_one_step}, we obtain that with probability at least $1-\delta$, 
\begin{align*}
    V^{\piE} - V^{\pibc} \leq H \lp \frac{4|\gS|}{9m}  + \frac{3\sqrt{|\gS|} \log (1/\delta)}{m} \rp,
\end{align*}
which translates to the sample complexity $\widetilde{\gO}(|\gS| H/\varepsilon)$ with high probability.

We continue to prove the sample complexity to achieve a small policy value gap \emph{in expectation}. We have that
\begin{align*}
    \expect \ls V^{\piE} - V^{\pibc} \rs \leq H \expect \ls \expect_{s_1 \sim \rho (\cdot)} \ls \indict \lp s_1 \notin \gS_1 (\gD) \rp   \rs \rs. 
\end{align*}
The outer expectation is taken w.r.t the randomness of expert demonstrations. For RHS, we have 
\begin{align}
    \expect \ls \expect_{s_1 \sim \rho (\cdot)} \ls \indict \lp s_1 \notin \gS_1 (\gD) \rp   \rs \rs &= \expect_{s_1 \sim \rho (\cdot)} \ls \expect \ls \indict \lp s_1 \notin \gS_1 (\gD) \rp \rs   \rs \nonumber
    \\
    &= \sum_{s \in \gS } \rho (s) \sP \lp  s \notin \gS_1 (\gD) \rp \nonumber
    \\
    &=  \sum_{s \in \gS } \rho (s) \lp 1 - \rho (s) \rp^m \nonumber
    \\
    &\leq \vert \gS \vert \max_{x \in [0, 1]} x (1-x)^m \nonumber
    \\
    &\leq \frac{|\gS|}{em}. \label{eq:expected_missing_mass_upper_bound} 
\end{align}
In the last inequality, we consider the optimization problem $\max_{x \in [0, 1]} f(x) = x (1-x)^m$. Here $f'(x) = (1-x)^{m-1} (1- (m+1)x)$. It is easy to see that the maximum is achieved at $x = 1/(m+1)$. Hence,
\begin{align*}
  \max_{x \in [0, 1]} x (1-x)^m = \frac{1}{m} \lp 1 - \frac{1}{m+1} \rp^{m+1} \leq \frac{1}{em}.   
\end{align*}
Finally, we have $ V^{\piE} - \expect [ V^{\pibc}] \leq (|\gS|H) / (em)$, which translates to the sample complexity $\gO(|\gS| H/\varepsilon)$ as in \cref{theorem:bc_deterministic}.

\end{proof}

\subsection{Reset Cliff and Useful Properties}
\label{appendix:reset_cliff_and_ail_properties}
In this part, we first give a detailed introduction of a family of MDPs called Reset Cliff shown in \cref{subsec:ail_generalize_well}. Then we present some properties of \textsf{VAIL}, which are useful in proving the results in \cref{subsec:ail_generalize_well}.

\begin{figure}[htbp]
\centering
\includegraphics[width=0.9\linewidth]{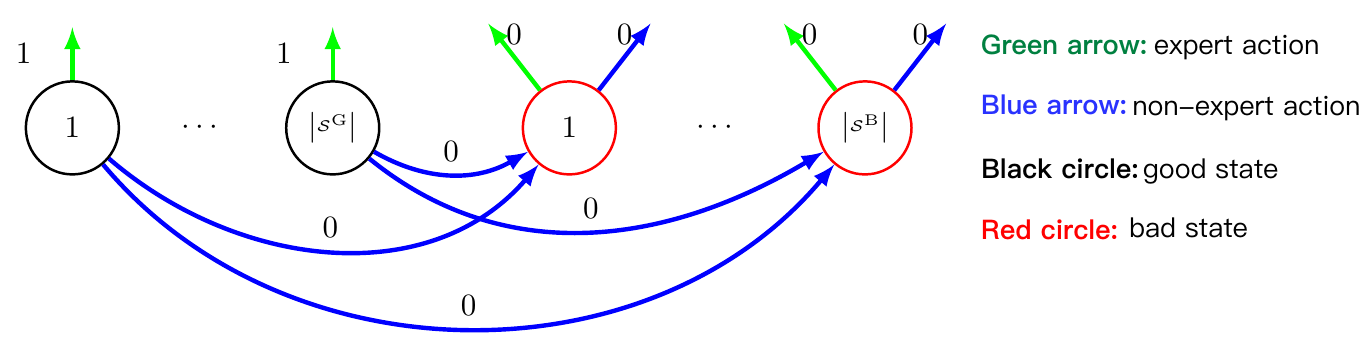}
\caption{\textsf{Reset Cliff} MDPs corresponding to \cref{asmp:reset_cliff}.}
\label{fig:reset_cliff}
\end{figure}

The Reset Cliff MDPs (refer to \cref{asmp:reset_cliff}) are illustrated in \cref{fig:reset_cliff}. Their properties are re-stated as follows.
\begin{itemize}
    \item In Reset Cliff, the state space is divided into the set of good states (shown in black circle) and the set of bad states (shown in red circle). That is $\gS = \goodS \cup \badS, \goodS \cap \badS = \emptyset$.
    \item The action space is denoted as $\gA$, in which $a^1$ is expert action (shown in green arrow) and the others are non-expert actions (shown in blue arrow).
    \item The agent gets $+1$ reward only by taking expert action $a^1$ on good states. For other cases, the agents gets 0 reward.
    \item On a good state, when the agent takes expert action $a^1$, then it transits into good states. Otherwise, the agent transits into bad states. Formally, $\forall h \in [H], s \in \goodS, a \in \gA \setminus \{a^1\}, \sum_{s^\prime \in \goodS} P_h (s^\prime|s, a^1) = 1, \sum_{s^\prime \in \badS} P_h (s^\prime|s, a) = 1$. Besides, we assume that $\forall h \in [H], \forall s, s^\prime \in \goodS, P_h (s^\prime |s, a^1) > 0 $.
    
    \item On a bad state, no matter which action is taken, the agent always goes to bad states. That is, $\forall h \in [H], s \in \badS, a \in \gA, \sum_{s^\prime \in \badS} P_h (s^\prime |s, a) = 1$.
\end{itemize}
       
For Reset Cliff MDPs, we know the expert policy never visits bad states. Thus, we have the following fact. 
\begin{fact}    \label{fact:ail_estimation}
For any tabular and episodic MDP satisfying \cref{asmp:reset_cliff}, considering any unbiased estimation $\widehat{P}^{\piE}_h (s, a)$, we have that 
\begin{align*}
   &\forall h \in [H], \forall s \in \badS, \forall a \in \gA, \quad  \widehat{P}^{\piE}_h(s) = 0,  \widehat{P}^{\piE}_h(s, a) = 0, \\
   &\forall h \in [H], \quad \sum_{s \in \goodS} \widehat{P}^{\piE}_h(s, a^{1}) = 1.0, \\
   &\forall h \in [H], \forall s \in \goodS, \forall a \ne a^{1}, \quad  \widehat{P}^{\piE}_h(s, a) = 0.0. 
\end{align*}
\end{fact}

Then we continue to present some useful properties of \textsf{VAIL} on Reset Cliff MDPs, which will be applied in the proof of results in \cref{subsec:ail_generalize_well}. Recall that 
\begin{align*}
  \piail = \argmin_{\pi} \sum_{h=1}^{H} \sum_{(s, a) \in \gS \times \gA} | P^{\pi}_h(s, a) - \widehat{P}^{\piE}_h(s, a) |.
\end{align*}
The first lemma states that on Reset Cliff, in each time step, $\piail$ takes the expert action on some good state with a positive probability.
\begin{lem}
\label{lem:condition_for_ail_optimal_solution}
For any tabular and episodic MDP satisfying \cref{asmp:reset_cliff}, suppose that $\piail$ is the optimal solution of \eqref{eq:ail}. Then $\forall h \in [H]$, $\exists s \in \goodS$, $\piail_h (a^{1} |s) > 0$.
\end{lem}

\begin{proof}
This proof is based on contradiction. Assume that the original statement is false: there exists a policy $\piail$, which is the optimal solution of \eqref{eq:ail}, such that $\exists h \in [H]$, $\forall s \in \goodS$, $\piail_h (a^{1}|s) = 0$. Let $h$ denote the smallest time step index such that $\forall s \in \goodS, \piail_{h} (a^{1}|s) = 0$. It also implies that $\forall s \in \goodS, \sum_{a \in \gA \setminus \{a^1 \}} \piail_{h} (a|s) = 1$.

We construct another policy $\widetilde{\pi}^{\operatorname{AIL}}$. $\widetilde{\pi}^{\operatorname{AIL}}$ is the same as $\piail$ in the first $h-1$ steps. In time step $h$, $\widetilde{\pi}^{\operatorname{AIL}} (a^{1}|s) = 1, \forall s \in \goodS$. Here we compare objective values of $\piail$ and $\widetilde{\pi}^{\operatorname{AIL}}$. Since $\piail$ is the same as $\widetilde{\pi}^{\operatorname{AIL}}$ in the first $h-1$ steps, their objective values are the same in the first $h-1$ steps. We only need to compare VAIL's objectives of $\piail$ and $\widetilde{\pi}^{\operatorname{AIL}}$ from time step $h$.

In time step $h$, notice that $P^{\widetilde{\pi}^{\operatorname{AIL}}}_h (s) =P^{\piail}_h (s)$, we obtain
\begin{align*}
    &\quad \text{Loss}_{h}(\piail) \\
    &= \sum_{(s, a)} \labs \widehat{P}^{\piE}_h (s, a) - P^{\piail}_h (s, a)  \rabs \\
    &= \sum_{s \in \goodS} \ls \labs \widehat{P}^{\piE}_h (s, a^{1}) - P^{\piail}_h (s, a^{1})  \rabs + \sum_{a \ne a^{1}} \labs \widehat{P}^{\piE}_h (s, a) - P^{\piail}_h (s, a)  \rabs \rs   + \sum_{s \in \badS} \sum_{a} \labs \widehat{P}^{\piE}_h (s, a) - P^{\piail}_h (s, a)  \rabs  \\
    &= \sum_{s \in \goodS} \ls \labs \widehat{P}^{\piE}_h (s, a^{1}) - 0  \rabs + \sum_{a \ne a^{1}} \labs 0 - P^{\piail}_h (s, a)  \rabs \rs   + \sum_{s \in \badS} \sum_{a} \labs 0 - P^{\piail}_h (s, a)  \rabs \\
    &=  \sum_{s \in \goodS} \lp \widehat{P}^{\piE}_h (s) + P^{\piail}_h (s) \rp + \sum_{s \in \badS}  P^{\piail}_h (s),
    \\
    &\quad \text{Loss}_{h}(\widetilde{\pi}^{\operatorname{AIL}}) \\
    &= \sum_{(s, a)} \labs \widehat{P}^{\piE}_h (s, a) - P^{\widetilde{\pi}^{\operatorname{AIL}}}_h (s, a)  \rabs \\
    &= \sum_{s \in \goodS} \ls \labs \widehat{P}^{\piE}_h (s, a^{1}) - P^{\widetilde{\pi}^{\operatorname{AIL}}}_h (s, a^{1})  \rabs + \sum_{a \ne a^{1}} \labs \widehat{P}^{\piE}_h (s, a) - P^{\widetilde{\pi}^{\operatorname{AIL}}}_h (s, a)  \rabs \rs   + \sum_{s \in \badS} \sum_{a} \labs \widehat{P}^{\piE}_h (s, a) - P^{\widetilde{\pi}^{\operatorname{AIL}}}_h (s, a)  \rabs  \\
    &=  \sum_{s \in \goodS} \ls \labs \widehat{P}^{\piE}_h (s, a^{1}) - P^{\widetilde{\pi}^{\operatorname{AIL}}}_h (s, a^{1})  \rabs + \sum_{a \ne a^{1}} \labs 0 - 0 \rabs \rs + \sum_{s \in \badS} \sum_{a} \labs 0 - P^{\piail}_h (s, a)  \rabs \\
    &= \sum_{s \in \goodS} \labs \widehat{P}^{\piE}_h (s) - P^{\piail}_h (s) \rabs + \sum_{s \in \badS}  P^{\piail}_h (s).
\end{align*}
Then we have
\begin{align*}
    \text{Loss}_{h}(\widetilde{\pi}^{\operatorname{AIL}}) - \text{Loss}_{h}(\piail) = \sum_{s \in \goodS} \labs \widehat{P}^{\piE}_h (s) - P^{\piail}_h (s) \rabs - \widehat{P}^{\piE}_h (s) - P^{\piail}_h (s) < 0,  
\end{align*}
where the last strict inequality follows that there always exists $s \in \goodS$ such that $\widehat{P}^{\piE}_h (s)>0, P^{\piail}_h (s) >0$. This is because $h$ is the smallest time step index such that $\forall s \in \goodS, \piail_{h} (a^1|s) = 0$ and $\forall h \in [H], \forall s, s^\prime \in \goodS, P_h (s^\prime |s, a^1) > 0$. Hence, $\forall s \in \goodS, P^{\piail}_h (s) >0$. For time step $h^\prime$ where $h+1 \leq h^\prime \leq H$,
\begin{align*}
    &\quad \text{Loss}_{h^\prime}(\piail) \\
    &=  \sum_{(s, a)} \labs \widehat{P}^{\piE}_{h^\prime} (s, a) - P^{\piail}_{h^\prime} (s, a)  \rabs  \\
    &= \sum_{s \in \goodS} \sum_{a}  \labs \widehat{P}^{\piE}_{h^\prime} (s, a) - P^{\piail}_{h^\prime} (s, a)  \rabs   + \sum_{s \in \badS} \sum_{a} \labs \widehat{P}^{\piE}_{h^\prime} (s, a) - P^{\piail}_{h^\prime} (s, a)  \rabs  \\
    &=  \sum_{s \in \goodS} \sum_{a}  \labs \widehat{P}^{\piE}_{h^\prime} (s, a) - 0 \rabs   + \sum_{s \in \badS} \sum_{a} \labs 0 - P^{\piail}_{h^\prime} (s, a)  \rabs \\
    &= \sum_{s \in \goodS} \widehat{P}^{\piE}_{h^\prime} (s) + \sum_{s \in \badS}  P^{\piail}_{h^\prime} (s) = 2,
\end{align*}
which is the maximal value of VAIL's objective at each time step. Thus, we have that $\text{Loss}_{h^\prime}(\widetilde{\pi}^{\operatorname{AIL}}) < \text{Loss}_{h^\prime}(\piail)$.

To summarize, we construct a policy $\widetilde{\pi}^{\operatorname{AIL}}$ whose VAIL's objective is strictly smaller than that of $\piail$. It contradicts with the fact that $\piail$ is the optimal solution of VAIL's objective. Hence the original statement is true and we finish the proof. 

\end{proof}

For any fixed unbiased estimation $\widehat{P}^{\piE}_h (s, a)$, we define a set of states $\gS ( \widehat{P}^{\piE}_H  ) := \{ s \in \gS, \widehat{P}^{\piE}_H (s) > 0  \}$. The following lemma states that in the last time step, \textsf{VAIL}'s policy takes the expert action on each $s \in \gS ( \widehat{P}^{\piE}_H  )$ with a positive probability. Note this positive probability may not be 1 due to the \dquote{weak convergence} issue.

\begin{lem}
\label{lem:general_condition_ail_policy_at_last_step}
Consider any tabular and episodic MDP satisfying \cref{asmp:reset_cliff}. For any fixed unbiased estimation $\widehat{P}^{\piE}_h (s, a)$, we define a set of states $\gS ( \widehat{P}^{\piE}_H  ) := \{ s \in \gS, \widehat{P}^{\piE}_H (s) > 0  \}$. Suppose that $\piail = (\piail_1, \cdots, \piail_H)$ is the optimal solution of \eqref{eq:ail}, then $\forall s \in \gS ( \widehat{P}^{\piE}_H  ), \piail_{H} \lp a^{1} |s \rp > 0$. 
\end{lem}

\begin{proof}
With \cref{lem:n_vars_opt_greedy_structure}, if $\piail = (\piail_1, \cdots, \piail_{H-1}, \piail_{H})$ is the optimal solution, then fixing $(\piail_1, \cdots, \piail_{H-1})$, $\piail_{H}$ is also optimal w.r.t VAIL's objective. Furthermore, since $P^{\piail}_{H} (s)$ is independent of $\piail_H$, we have
\begin{align*}
    \piail_{H} & \in \argmin_{\pi_{H}}  \sum_{(s, a)} \labs P^{\piail}_H(s) \pi_{H}(a|s) - \widehat{P}^{\piE}_H(s, a) \rabs.
\end{align*}
Recall that $\gS ( \widehat{P}^{\piE}_H  ) := \{ s \in \gS, \widehat{P}^{\piE}_H (s) > 0  \}$, we have 
\begin{align*}
   \piail_{H} &\in \argmin_{\pi_{H}}  \sum_{s \in \gS ( \widehat{P}^{\piE}_H  )} \sum_{a \in \gA} \labs P^{\piail}_H(s) \pi_{H}(a|s) - \widehat{P}^{\piE}_H(s, a) \rabs + \sum_{s \notin \gS ( \widehat{P}^{\piE}_H  )} \sum_{a \in \gA} \labs P^{\piail}_H(s) \pi_H(a|s) - \widehat{P}^{\piE}_H(s, a) \rabs
   \\
   &= \argmin_{\pi_{H}}  \sum_{s \in \gS ( \widehat{P}^{\piE}_H  )} \lp \labs P^{\piail}_H(s) \pi_H ( a^{1} |s) - \widehat{P}^{\piE}_H(s, a^{1} ) \rabs + \sum_{a \ne a^{1} } \labs P^{\piail}_H(s) \pi_H (a|s) - \widehat{P}^{\piE}_H(s, a) \rabs \rp
   \\
   &\quad + \sum_{s \notin \gS ( \widehat{P}^{\piE}_H  )} \sum_{a \in \gA } \labs P^{\piail}_H(s) \pi_H(a|s) - \widehat{P}^{\piE}_H(s, a) \rabs.
\end{align*}
Since $\widehat{P}^{\piE}_H(s, a) = 0$ for $a \not= a^{1}$, we obtain
\begin{align*}
\piail_{H} &\in \argmin_{\pi_{H}}  \sum_{s \in \gS ( \widehat{P}^{\piE}_H  )} \lp \labs P^{\piail}_H(s) \pi_H ( a^{1} |s) - \widehat{P}^{\piE}_H(s, a^{1}) \rabs +   P^{\piail}_H(s) \lp 1 - \pi_H (a^{1} |s) \rp \rp
   \\
   &\quad + \sum_{s \notin \gS ( \widehat{P}^{\piE}_H  )} \sum_{a \in \gA } \labs P^{\piail}_H(s) \pi_H(a|s) - \widehat{P}^{\piE}_H(s, a) \rabs
   \\
   &= \argmin_{\pi_{H}}  \sum_{s \in \gS ( \widehat{P}^{\piE}_H  )} \lp \labs P^{\piail}_H(s) \pi_H ( a^{1} |s) - \widehat{P}^{\piE}_H(s, a^{1} ) \rabs -   P^{\piail}_H(s) \pi_H ( a^{1} |s)  \rp
   \\
   &\quad + \sum_{s \notin \gS ( \widehat{P}^{\piE}_H  )} \sum_{a \in \gA } \labs P^{\piail}_H(s) \pi_H(a|s) - \widehat{P}^{\piE}_H(s, a) \rabs.
\end{align*}
The last equation follows that $P^{\piail}_H(s)$ is independent of $\pi_H$. Note that for different $s \in \gS ( \widehat{P}^{\piE}_H  )$, $\pi_H (a^{1} |s)$ are independent by the tabular formulation. Thus, we can consider the optimization problem for each $s \in \gS ( \widehat{P}^{\piE}_H  )$ separately. For each $s \in \gS ( \widehat{P}^{\piE}_H  )$, we have
\begin{align*}
    \piail_{H} (a^{1} |s) = \argmin_{\pi_H (a^{1} |s) \in [0, 1]} \labs P^{\piail}_H(s) \pi_H ( a^{1} |s) - \widehat{P}_H(s, a^{1}) \rabs -   P^{\piail}_H(s) \pi_H ( a^{1} |s).
\end{align*}
For the above one-dimension optimization problem, \cref{lem:single_variable_opt_condition} claims that the optimal solution must be positive, i.e., $\piail_{H} (a^{1} |s)  > 0$. Thus, we finish the proof if we can verify the conditions in \cref{lem:single_variable_opt_condition}. 

In the following part, we verify the conditions of \cref{lem:single_variable_opt_condition} by setting $a = P^{\piail}_H(s), c = \widehat{P}_H(s, a^{1})$. Since $\piail$ is the optimal solution of VAIL's objective, with \cref{lem:condition_for_ail_optimal_solution}, we have that $\forall h \in [H]$, $\exists s \in \goodS$, $\piail_h (a^1 |s) > 0$. Combined with the assumption that $\forall h \in [H], s, s^\prime \in \goodS, P_h (s^\prime |s, a^1) > 0$, we have that $P^{\piail}_H(s) > 0, \forall s \in \goodS$. Based on the definition, for each $s \in \gS ( \widehat{P}^{\piE}_H  )$, $\widehat{P}_H^{\piE}(s, a^{1}) > 0$. Now the conditions of \cref{lem:single_variable_opt_condition} are verified and we obtain that $\piail_H (a^{1} |s) > 0, \forall s \in \gS ( \widehat{P}^{\piE}_H  )$. 
\end{proof}

\subsection{Proof of Claim in Example \ref{example:ail_success}}
\label{appendix:warm_up_of_proposition_reset_cliff}

In this part, we formally state and prove the theoretical result in \cref{example:ail_success}, which is a simplified version of \cref{prop:ail_general_reset_cliff}. 

\begin{claim} \label{claim:ail_reset_cliff}
Consider the MDP and expert demonstration configuration in \cref{example:ail_success}. Suppose that $\piail$ is the optimal solution of \eqref{eq:ail}, then for each time step $h \in [2]$, $\piail_{h} (a^{1}|s) = \piE_h (a^{1}|s) = 1, \forall s \in \{s^{1}, s^{2} \}$.  
\end{claim}

Let us briefly discuss the proof idea. Since the objective in \eqref{eq:ail} involves multi-stage optimization problems, it is common to use the dynamic programming (DP) technique to show the structure of the optimal solutions; see examples in the famous book \citep{bertsekas2012dynamic}.  

\begin{proof}
First of all, recall that there are three states $(s^{1}, s^{2}, s^{3})$ and two actions $(a^{1}, a^{2})$. In particular, $s^{1}$ and $s^{2}$ are good states while $s^{3}$ is a bad absorbing state. Suppose $H=2$ and $\rho = (0.5, 0.5, 0.0)$.  Moreover, the expert policy always takes action $a^{1}$. The agent is provided only 2 expert trajectories: $\tr_1 = (s^{1}, a^{1}) \rar (s^{1}, a^{1}) $ and $\tr_2 = (s^{1}, a^{1}) \rar (s^{2}, a^{1})$.  

Let us compute the empirical state-action distribution:
\begin{align*}
    \widehat{P}^{\piE}_{1}(s^{1}, a^{1}) = \RED{1.0}, \widehat{P}^{\piE}_{1}(s^{2}, a^{1}) = 0.0, \widehat{P}^{\piE}_{1}(s^{3}, a^{1}) = 0.0, \\
    \widehat{P}^{\piE}_{1}(s^{1}, a^{2}) = 0.0, \widehat{P}^{\piE}_{1}(s^{2}, a^{2}) = 0.0, \widehat{P}^{\piE}_{1}(s^{3}, a^{2}) = 0.0, \\
    \widehat{P}^{\piE}_{2}(s^{1}, a^{1}) = \RED{0.5}, \widehat{P}^{\piE}_{2}(s^{2}, a^{1}) = \RED{0.5}, \widehat{P}^{\piE}_{2}(s^{3}, a^{1}) = 0.0, \\
    \widehat{P}^{\piE}_{2}(s^{1}, a^{2}) = 0.0, \widehat{P}^{\piE}_{2}(s^{2}, a^{2}) = 0.0, \widehat{P}^{\piE}_{2}(s^{3}, a^{2}) = 0.0.
\end{align*}
Define the single-stage loss function in time step $h$ as 
\begin{align*}
    \text{Loss}_h (\pi) = \sum_{(s, a) \in \gS \times \gA} \labs P^{\pi}_h(s, a)  - \widehat{P}^{\piE}_h(s, a) \rabs .
\end{align*}
Then, we can define the \dquote{cost-to-go} function:
\begin{align*}
    \ell_h (\pi) = \sum_{t=h}^{H} \text{Loss}_t (\pi)  = \sum_{t=h}^{H} \sum_{(s, a) \in \gS \times \gA} \labs P^{\pi}_t(s, a)  - \widehat{P}^{\piE}_t(s, a) \rabs .
\end{align*}
As $\piail = (\piail_1, \piail_2)$ is the optimal solution of \eqref{eq:ail}, with \cref{lem:n_vars_opt_greedy_structure}, fixing $\piail_1$, $\piail_2$ is optimal w.r.t to VAIL's objective. Notice that $P^{\piail}_1$ and $\text{Loss}_1$ are independent of $\piail_2$, we have that
\begin{align*}
    \piail_2 \in \argmin_{\pi_2} \ell_2 (\pi_2)
\end{align*}
With a slight abuse of notation, we use $P^{\pi}_2$ denote the distribution induce by $(\piail_1, \pi_2)$ for any optimization variable $\pi_2$. For \cref{example:ail_success}, in the last time step $h = 2$, we have that 
\begin{align*}
    \ell_2 (\pi_2) &= \sum_{(s, a) \in \gS \times \gA} \labs P^{\pi}_2(s, a)  - \widehat{P}^{\piE}_2(s, a) \rabs  \\
    &= \labs P^{\pi}_2(s^{1}, a^{1}) - \widehat{P}^{\piE}_2(s^{1}, a^{1}) \rabs + \labs P^{\pi}_2(s^{2}, a^{1}) - \widehat{P}^{\piE}_2(s^{2}, a^{1}) \rabs + \labs P^{\pi}_2(s^{3}, a^{1}) - \widehat{P}^{\piE}_2(s^{3}, a^{1}) \rabs  \\
    &\quad + \labs P^{\pi}_2(s^{1}, a^{2}) - \widehat{P}^{\piE}_2(s^{1}, a^{2}) \rabs  + \labs P^{\pi}_2(s^{2}, a^{2}) - \widehat{P}^{\piE}_2(s^{2}, a^{2}) \rabs  + \labs P^{\pi}_2(s^{3}, a^{2}) - \widehat{P}^{\piE}_2(s^{3}, a^{2}) \rabs \\
    &= \labs P^{\pi}_2(s^{1}) \pi_2(a^{1}|s^{1}) - 0.5 \rabs +  \labs P^{\pi}_2(s^{2}) \pi_2(a^{1}|s^{2}) - 0.5 \rabs + \labs P^{\pi}_2(s^{3}) \pi_2(a^{1}|s^{3}) - 0.0 \rabs \\
    &\quad + \labs P^{\pi}_2(s^{1}) \pi_2(a^{2}|s^{1}) - 0.0 \rabs +  \labs P^{\pi}_2(s^{2}) \pi_2(a^{2}|s^{2}) - 0.0 \rabs + \labs P^{\pi}_2(s^{3}) \pi_2(a^{2}|s^{3}) - 0.0 \rabs \\
    &= \labs P^{\pi}_2(s^{1}) \pi_2(a^{1} | s^{1}) - 0.5 \rabs +  \labs P^{\pi}_2(s^{2}) \pi_2(a^{1}|s^{2}) - 0.5 \rabs + P^{\pi}_2(s^{3})  \\
    &\quad +  P^{\pi}_2(s^{1}) (1 - \pi_2(a^{1} | s^{1}))  + P^{\pi}_2(s^{2}) (1 - \pi_2(a^{1} | s^{2})).
\end{align*}
Note that $\pi_2$ is the optimization variable for $\ell_2 (\pi_2)$ while $P_2^{\pi}(s^{1}) = P_2^{\piail}(s^{1}), P_2^{\pi}(s^{2}) = P_2^{\piail}(s^{2}), P_2^{\pi}(s^{3}) = P_2^{\piail}(s^{3})$ are independent of $\pi_2$. We obtain
\begin{align*}
    \piail_2 &\in \argmin_{\pi_2} \ell_2 (\pi_2)
    \\
    &= \argmin_{\pi_2} \labs P_2^{\piail}(s^{1}) \pi_2(a^{1} | s^{1}) - 0.5 \rabs +  \labs P_2^{\piail}(s^{2}) \pi_2(a^{1}|s^{2}) - 0.5 \rabs + P^{\piail}_2(s^{3})  \\
    &\quad +  P^{\piail}_2(s^{1}) (1 - \pi_2(a^{1} | s^{1}))  + P^{\piail}_2(s^{2}) (1 - \pi_2(a^{1} | s^{2}))
    \\
    &= \argmin_{\pi_2} \labs P_2^{\piail}(s^{1}) \pi_2(a^{1} | s^{1}) - 0.5 \rabs - P^{\piail}_2(s^{1}) \pi_2(a^{1} | s^{1})  +  \labs P_2^{\piail}(s^{2}) \pi_2(a^{1}|s^{2}) - 0.5 \rabs \\
    &\quad - P^{\piail}_2(s^{2})  \pi_2(a^{1} | s^{2}). 
\end{align*}

Note that we only have two free optimization variables: $\pi_2(a^{1}|s^{1})$ and $\pi_2(a^{1}|s^{2})$ and they are independent. Then we obtain
\begin{align*}
    & \piail_2 (a^1|s^1) \in \argmin_{\pi_2 (a^1|s^1) \in [0, 1]} \labs P_2^{\piail}(s^{1}) \pi_2(a^{1} | s^{1}) - 0.5 \rabs - P^{\piail}_2(s^{1}) \pi_2(a^{1} | s^{1}),
    \\
    &\piail_2 (a^1|s^2) \in \argmin_{\pi_2 (a^1|s^2) \in [0, 1]} \labs P_2^{\piail}(s^{2}) \pi_2(a^{1}|s^{2}) - 0.5 \rabs - P^{\piail}_2(s^{2})  \pi_2(a^{1} | s^{2}).
\end{align*}
We first consider $\piail_2 (a^1|s^1)$ and we want to argue that $\pi_2(a^{1} | s^{1}) = 1$ is the optimal solution. We can directly prove this claim for this specific example but we have a more powerful lemma in \cref{appendix:technical_lemmas}. In particular, \cref{lem:mn_variables_opt_unique} claims that $\pi_2(a^{1} | s^{1}) = 1$ is the unique globally optimal solution. Similarly, we also have that $\piail_2 (a^1|s^2) = 1$. This finishes the proof in time step $h=2$.

 In the following part, we check the conditions of \cref{lem:mn_variables_opt_unique}. We apply \cref{lem:mn_variables_opt_unique} with $m = n = 1$, $c_1 = 0.5$, $a_{11} = P_2^{\piail}(s^{1})$ and $d_1 = P^{\piail}_2(s^{1})$.  \cref{lem:condition_for_ail_optimal_solution} implies that $\exists s \in \{s^1, s^2 \}$, $\piail_1 (a^1|s) > 0$ and hence we have $a_{11} = P_2^{\piail}(s^{1}) > 0$. Besides, $P_2^{\piail}(s^{1}) \leq 0.5 = c_1$, where the equality holds if and only if $\piail_2 (a^1|s^1) = 1, \piail_2 (a^1|s^2) = 1$. By \cref{lem:mn_variables_opt_unique}, we have that $\piail_2 (a^1|s^1) = 1$.

Then we consider the policy optimization in time step $h=1$. With \cref{lem:single_variable_opt_condition}, we have that fixing $\piail_2$, $\piail_1$ is optimal w.r.t VAIL's objective.
\begin{align*}
    \piail_1 \in \argmin_{\pi_1} \ell_1 (\pi_1) = \argmin_{\pi_1} \text{Loss}_1 (\pi_1) + \text{Loss}_2 (\pi_1). 
\end{align*}
We have proved that $\piail_2 (a^1|s^1) = 1, \piail_2 (a^1|s^2) = 1$ and plug it into $\text{Loss}_2 (\pi_1)$.
\begin{align*}
    \text{Loss}_2 (\pi_1) &= \labs P^{\pi}_2(s^{1}) - 0.5 \rabs + \labs P^{\pi}_2(s^{2})  - 0.5 \rabs + P^{\pi}_2(s^{3}) \\
    &= \labs P^{\pi}_1(s^{1}) \pi_1(a^{1}|s^{1}) P_1(s^{1}|s^{1}, a^{1})  + P^{\pi}_1(s^{2}) \pi_1(a^{1}|s^{2}) P_1(s^{1}|s^{2}, a^{1})  -0.5 \rabs  \\
    &\quad + \labs P^{\pi}_1(s^{1}) \pi_1(a^{1}|s^{1}) P_1(s^{2}|s^{1}, a^{1})  + P^{\pi}_1(s^{2}) \pi_1(a^{1}|s^{2}) P_1(s^{2}|s^{2}, a^{1})  -0.5 \rabs \\
    &\quad + P^{\pi}_1(s^{1}) \pi_1(a^{2}|s^{1})  P_1(s^{3}|s^{1}, a^{2}) + P^{\pi}_1(s^{2}) \pi_1(a^{2}|s^{2})  P_1(s^{3}|s^{2}, a^{2}) \\
    &= 2 \labs 0.25 \pi_1(a^{1} | s^{1}) + 0.25 \pi_1(a^{1} | s^{2}) - 0.5 \rabs + 0.5(1- \pi_1(a^{1} |s^{1})) + 0.5(1- \pi_1(a^{1} |s^{2})) \\
    &=  (1.0 - 0.5 \pi_1(a^{1} | s^{1}) - 0.5 \pi_1(a^{1} | s^{2}) ) - 0.5 \pi_1(a^{1} | s^{2}) - 0.5 \pi_1(a^{1} | s^{2}) + 1.0 \\
    &= 2.0 - \pi_1(a^{1} | s^{1}) - \pi_1(a^{1} | s^{2}),
\end{align*}
which has a unique globally optimal solution at $\pi_1(a^{1} | s^{1}) = 1.0$ and $\pi_1(a^{1} | s^{2}) = 1.0$. For $\text{Loss}_1 (\pi_1)$, 
\begin{align*}
    \text{Loss}_1 (\pi_1) &= \labs P^{\pi}_1 (s^{1}) - \rho (s^{1}) \pi_1 (a^1|s^1) \rabs + \rho (s^{1}) \lp 1 - \pi_1 (a^1|s^1) \rp + \rho (s^{2})
    \\
    &= \labs 1 - 0.5 \pi_1 (a^1|s^1) \rabs + 0.5 (1-\pi_1 (a^1|s^1)) + 0.5
    \\
    &= 2 - \pi_1 (a^1|s^1),
\end{align*}
which has a globally optimal solution at $\pi_1(a^{1} | s^{1}) = 1.0$ and $\pi_1(a^{1} | s^{2}) = 1.0$. By \cref{lem:unique_opt_solution_condition}, we have that $\pi_1(a^{1} | s^{1}) = \pi_1(a^{1} | s^{2}) = 1.0$ is the unique globally optimal solution of the joint objective $ \text{Loss}_1 (\pi_1) + \text{Loss}_2 (\pi_1)$. Recall that $\piail_1 \in \argmin_{\pi_1} \text{Loss}_1 (\pi_1) + \text{Loss}_2 (\pi_1)$. Hence it holds that $\piail_1(a^{1} | s^{1}) = \piail_1(a^{1} | s^{2}) = 1.0$. This finishes the proof in time step $h=1$. 
\end{proof}
In \cref{example:ail_success}, the estimator in the last time step happens to equal the true distribution, i.e., $\widehat{P}^{\piE}_{2}(s^{1}, a^{1}) = P^{\piE}_{2}(s^{1}, a^{1})$ and $\widehat{P}^{\piE}_{2}(s^{2}, a^{1}) = P^{\piE}_{2}(s^{2}, a^{1})$. Therefore, we can prove that $\piE_H$ is the unique globally optimal solution of \eqref{eq:ail}. We remark that in general, we cannot prove that in the last time step $h=H$, $\piE_H$ is the unique globally optimal solution of \eqref{eq:ail} due to the \dquote{weak convergence} issue discussed in \cref{sec:generalization_of_ail}.

\subsection{Proof of Proposition \ref{prop:ail_general_reset_cliff}}

Since the objective in \eqref{eq:ail} is a multi-stage optimization problem, we leverage backward induction to analyze its optimal solution step by step. In particular, we generalize the proof idea in \cref{claim:ail_reset_cliff} in \cref{example:ail_success}. The main intuition is that if the agent does not select the expert action, it goes to a bad absorbing state and suffers a huge loss for future state-action distribution matching. This implies the expert action is expected to be the optimal solution. With assumed transitions, we further prove that the optimal solution is unique in the first $H-1$ time steps.

\begin{proof}
The proof is based on backward induction. Suppose that $\piail = (\piail_1, \cdots, \piail_{H})$ is the optimal solution of \eqref{eq:ail}.  Define the single-stage loss function in time step $h$ as 
\begin{align*}
    \text{Loss}_h (\pi) = \sum_{(s, a) \in \gS \times \gA} \labs P^{\pi}_h(s, a)  - \widehat{P}^{\piE}_h(s, a) \rabs .
\end{align*}
\RED{First, we consider the base case (3 pages).} We aim to prove that $\piail_{H-1} (a^{1}|s) =  1, \forall s \in \goodS$. By \cref{lem:n_vars_opt_greedy_structure}, with fixed $(\piail_1, \cdots, \piail_{H-2}, \piail_{H})$, $\piail_{H-1}$ is optimal w.r.t the \textsf{VAIL} objective in \eqref{eq:ail}. This is direct from the global optimality condition. Furthermore, with fixed $(\piail_1, \cdots, \piail_{H-2}, \piail_{H})$, the state-action distribution losses from time step $1$ to $H-2$ are independent of $\pi_{H-1}$. Therefore, we have
\begin{align*}
\piail_{H-1} \in \argmin_{\pi_{H-1}} \text{Loss}_{H-1} (\pi_{H-1}) + \text{Loss}_{H} (\pi_{H-1}).
\end{align*}
In the following part, we will prove that $\piail_{H-1} (a^{1}|s) =  1, \forall s \in \goodS$ is the \emph{unique} optimal solution of the optimization problem $\min_{\pi_{H-1}} \text{Loss}_{H-1} (\pi_{H-1}) + \text{Loss}_{H} (\pi_{H-1})$. Our strategy is to prove that $\piail_{H-1} (a^{1}|s) =  1, \forall s \in \goodS$ is the optimal solution of $\min_{\pi_{H-1}} \text{Loss}_{H-1} (\pi_{H-1})$ and the \emph{unique} optimal solution of $\min_{\pi_{H-1}} \text{Loss}_{H} (\pi_{H-1})$. As a consequence, $\piail_{H-1} (a^{1}|s) =  1, \forall s \in \goodS$ is the unique optimal solution in time step $H-1$; see also \cref{lem:unique_opt_solution_condition}. We prove two terms separately. 
\begin{itemize}
    \item \BLUE{Term 1.}  We consider $\text{Loss}_{H-1} (\pi_{H-1})$. 
    \begin{align*}
        &\quad \text{Loss}_{H-1} (\pi_{H-1})
        \\
        &= \sum_{s \in \gS} \sum_{a \in \gA} \labs \widehat{P}^{\piE}_{H-1} (s, a) - P^{\piail}_{H-1} (s) \pi_{H-1} (a|s) \rabs
        \\
        &= \sum_{s \in \goodS}  \ls \labs \widehat{P}^{\piE}_{H-1} (s, a^{1}) - P^{\piail}_{H-1} (s) \pi_{H-1} (a|s^{1}) \rabs +  \sum_{a \ne a^{1}} \labs \widehat{P}^{\piE}_{H-1} (s, a) - P^{\piail}_{H-1} (s) \pi_{H-1} (a|s)  \rabs \rs \\
        &\quad +  \sum_{s \in \badS} \sum_{a \in \gA} \labs \widehat{P}^{\piE}_{H-1} (s, a) - P^{\piail}_{H-1} (s) \pi_{H-1} (a|s) \rabs \\
        &= \sum_{s \in \goodS} \lp \labs \widehat{P}^{\piE}_{H-1} (s) - P^{\piail}_{H-1} (s) \pi_{H-1} (a^{1}|s)  \rabs + P^{\piail}_{H-1} (s) \lp 1 - \pi_{H-1} (a^{1}|s) \rp \rp + \sum_{s \in \badS} P^{\piail}_{H-1} (s). 
    \end{align*}
    The last equation follows that the expert policy is deterministic and hence 1) $\forall s \in \goodS, \widehat{P}^{\piE}_{H-1} (s, a^{1}) = \widehat{P}^{\piE}_{H-1} (s)$; 2) $\forall a \in \gA \setminus \{a^{1} \}, \widehat{P}^{\piE}_{H-1} (s, a) = 0$; 3) $\forall s \in \badS, \widehat{P}^{\piE}_{H-1} (s) = 0$. Notice that $P^{\piail}_{H-1} (s)$ is fixed and independent of $\pi_{H-1}$, so we can obtain the following optimization problem: 
    \begin{align*}
        &\quad \argmin_{\pi_{H-1}} \mathrm{Loss}_{H-1} (\pi_{H-1})
        \\
        &= \argmin_{\pi_{H-1}} \sum_{s \in \goodS} \labs \widehat{P}^{\piE}_{H-1} (s) - P^{\piail}_{H-1} (s) \pi_{H-1} (a^{1}|s)  \rabs - P^{\piail}_{H-1} (s) \pi_{H-1} (a^{1}|s).
    \end{align*}
    Since the optimization variables $\pi_{H-1} (a^{1}|s)$ for different $s \in \goodS$ are independent, we can consider the above optimization problem for each $s \in \goodS$ individually.
    \begin{align*}
        \argmin_{\pi_{H-1} (a^{1}|s) \in [0, 1]} \labs \widehat{P}^{\piE}_{H-1} (s) - P^{\piail}_{H-1} (s) \pi_{H-1} (a^{1}|s)  \rabs - P^{\piail}_{H-1} (s) \pi_{H-1} (a^{1}|s).
    \end{align*}
    For this one-dimension optimization problem, we can use \cref{lem:single_variable_opt} to show that $\piail_{H-1} (a^{1}|s) = 1$ is the optimal solution. Consequently, we obtain that $\piail_{H-1} (a^{1}|s) = 1, \forall s \in \goodS$ is the optimal solution of $\mathrm{Loss}_{H-1} (\pi_{H-1})$.
    \item \BLUE{Term 2.} We consider the \textsf{VAIL}'s loss in step $H$. Recall the definition of $\gS (\widehat{P}^{\piE}_H) := \{ s \in \gS: \widehat{P}^{\piE}_H (s) > 0  \}$. Note that on non-visited state $s \notin \gS (\widehat{P}^{\piE}_H)$, we have that $\widehat{P}^{\piE}_{H} (s) = 0$. Then we obtain
    \begin{align*}
        &\quad \text{Loss}_{H} (\pi_{H-1})
        \\
        &= \sum_{s \in \gS} \sum_{a \in \gA} \labs \widehat{P}^{\piE}_{H} (s, a) - P^{\piail}_{H} (s, a) \rabs
        \\
        &= \sum_{s \in \goodS} \sum_{a \in \gA} \labs \widehat{P}^{\piE}_{H} (s, a) - P^{\piail}_{H} (s, a) \rabs + \sum_{s \in \badS} P^{\piail}_H (s)  
        \\
        &= \sum_{s \in \gS (\widehat{P}^{\piE}_H)} \lp \labs \widehat{P}^{\piE}_{H} (s) - P^{\piail}_{H} (s, a^{1}) \rabs + \sum_{a\in \gA \setminus \{a^1\}} P^{\piail}_H (s, a) \rp + \sum_{s \in \goodS \text{ and } s \notin \gS (\widehat{P}^{\piE}_H)} P^{\piail}_H (s)
        \\
        &\quad + \sum_{s \in \badS} P^{\piail}_H (s).
    \end{align*}
    Readers may notice that here we slightly abuse the notation: we use $P^{\piail}_{H} (s, a), P^{\piail}_{H} (s)$ to denote the distributions induced by $(\piail_1, \cdots, \piail_{H-2}, \pi_{H-1}, \piail_{H})$ for optimization variable $\pi_{H-1}$.  With the \dquote{transition flow equation}, we have that
    \begin{align*}
        \forall s \in \goodS, P^{\piail}_{H} (s) &= \sum_{s^\prime \in \gS} \sum_{a \in \gA} P^{\piail}_{H-1} (s^\prime) \pi_{H-1} (a|s^\prime) P_{H-1} (s | s^\prime, a)
        \\
        &=  \sum_{s^\prime \in \goodS} P^{\piail}_{H-1} (s^\prime) \pi_{H-1} (a^{1}|s^\prime) P_{H-1} (s | s^\prime, a^{1}).
    \end{align*}
    Recall that when the agent takes a non-expert action,  it transits into bad states. Therefore, the probability of visiting bad states in time step $H$ arises from two parts. One is the probability of visiting bad states in time step $H-1$ and the other is the probability of visiting good states and taking non-expert actions in time step $H-1$. Accordingly, we obtain 
    \begin{align*}
        \sum_{s \in \badS} P^{\piail}_H (s) &= \sum_{s \in \badS} P^{\piail}_{H-1} (s) + \sum_{s^\prime \in \goodS} P^{\piail}_{H-1} (s^\prime) \lp \sum_{a \in \gA \setminus \{a^{1}\}} \pi_{H-1} (a|s^\prime) \rp
        \\
        &= \sum_{s \in \badS} P^{\piail}_{H-1} (s) + \sum_{s^\prime \in \goodS} P^{\piail}_{H-1} (s^\prime) \lp 1 - \pi_{H-1} (a^{1}|s^\prime) \rp.
    \end{align*}
    Plugging the above two equations into $\text{Loss}_{H} (\pi_{H-1})$ yields
    \begin{align*}
        &\quad \text{Loss}_{H} (\pi_{H-1})
        \\
        &= \sum_{s \in \gS (\widehat{P}^{\piE}_H)} \labs \widehat{P}^{\piE}_{H} (s) - \lp \sum_{s^\prime \in \goodS} P^{\piail}_{H-1} (s^\prime) \pi_{H-1} (a^{1}|s^\prime) P_{H-1} (s | s^\prime, a^{1})  \rp \piail_{H} (a^{1}|s) \rabs
        \\
        &\quad + \sum_{s \in \gS (\widehat{P}^{\piE}_H)} \lp \sum_{s^\prime \in \goodS} P^{\piail}_{H-1} (s^\prime) \pi_{H-1} (a^{1}|s^\prime) P_{H-1} (s | s^\prime, a^{1})  \rp \lp 1- \piail_{H} (a^{1}|s) \rp 
        \\
        &\quad + \sum_{s \in \goodS \text{ and } s \notin \gS (\widehat{P}^{\piE}_H)} \lp \sum_{s^\prime \in \goodS} P^{\piail}_{H-1} (s^\prime) \pi_{H-1} (a^{1}|s^\prime) P_{H-1} (s | s^\prime, a^{1}) \rp
        \\
        &\quad +  \sum_{s \in \badS} P^{\piail}_{H-1} (s) + \sum_{s^\prime \in \goodS} P^{\piail}_{H-1} (s^\prime) \lp 1 - \pi_{H-1} (a^{1}|s^\prime) \rp
        \\
        &= \sum_{s \in \gS (\widehat{P}^{\piE}_H)} \labs \widehat{P}^{\piE}_{H} (s) - \sum_{s^\prime \in \goodS} P^{\piail}_{H-1} (s^\prime)  P_{H-1} (s | s^\prime, a^{1}) \piail_{H} (a^{1}|s) \pi_{H-1} (a^{1}|s^\prime) \rabs
        \\
        &\quad + \sum_{s^\prime \in \goodS} \lp \sum_{s \in \gS (\widehat{P}^{\piE}_H)} P^{\piail}_{H-1} (s^\prime) P_{H-1} (s | s^\prime, a^{1}) \lp 1- \piail_{H} (a^{1}|s) \rp    \rp \pi_{H-1} (a^{1}|s^\prime)
        \\
        &\quad + \sum_{s^\prime \in \goodS} \lp \sum_{s \in \goodS \text{ and } s \notin \gS (\widehat{P}^{\piE}_H)} P^{\piail}_{H-1} (s^\prime) P_{H-1} (s | s^\prime, a^{1}) \rp \pi_{H-1} (a^{1}|s^\prime) - \sum_{s^\prime \in \goodS} P^{\piail}_{H-1} (s^\prime) \pi_{H-1} (a^{1}|s^\prime)
        \\
        &\quad + \sum_{s \in \badS} P^{\piail}_{H-1} (s) + \sum_{s^\prime \in \goodS} P^{\piail}_{H-1} (s^\prime)
    \end{align*}
    Then, we merge the terms that are linear w.r.t $\pi_{H-1} (a^{1}|s^\prime)$, i.e., the second, third and forth terms in RHS.
    \begin{align*}
        & \quad \sum_{s^\prime \in \goodS} \lp \sum_{s \in \gS (\widehat{P}^{\piE}_H)} P^{\piail}_{H-1} (s^\prime) P_{H-1} (s | s^\prime, a^{1}) \lp 1- \piail_{H} (a^{1}|s) \rp    \rp \pi_{H-1} (a^{1}|s^\prime)
        \\
        &\quad + \sum_{s^\prime \in \goodS} \lp \sum_{s \in \goodS \text{ and } s \notin \gS (\widehat{P}^{\piE}_H)} P^{\piail}_{H-1} (s^\prime) P_{H-1} (s | s^\prime, a^{1}) \rp \pi_{H-1} (a^{1}|s^\prime) - \sum_{s^\prime \in \goodS} P^{\piail}_{H-1} (s^\prime) \pi_{H-1} (a^{1}|s^\prime)
        \\
        &= \sum_{s^\prime \in \goodS} \Bigg( \sum_{s \in \gS (\widehat{P}^{\piE}_H)} P^{\piail}_{H-1} (s^\prime) P_{H-1} (s | s^\prime, a^{1}) - \sum_{s \in \gS (\widehat{P}^{\piE}_H)} P^{\piail}_{H-1} (s^\prime) P_{H-1} (s | s^\prime, a^{1}) \piail_{H} (a^{1}|s)
        \\
        &\quad + \sum_{s \in \goodS \text{ and } s \notin \gS (\widehat{P}^{\piE}_H)} P^{\piail}_{H-1} (s^\prime) P_{H-1} (s | s^\prime, a^{1}) - P^{\piail}_{H-1} (s^\prime)   \Bigg) \pi_{H-1} (a^{1}|s^\prime)
        \\
        &= \sum_{s^\prime \in \goodS} \Bigg( \sum_{s \in \goodS} P^{\piail}_{H-1} (s^\prime) P_{H-1} (s | s^\prime, a^{1}) - \sum_{s \in \gS (\widehat{P}^{\piE}_H)} P^{\piail}_{H-1} (s^\prime) P_{H-1} (s | s^\prime, a^{1}) \piail_{H} (a^{1}|s)
        \\
        &\quad - P^{\piail}_{H-1} (s^\prime)   \Bigg) \pi_{H-1} (a^{1}|s^\prime)
        \\
        &= - \sum_{s^\prime \in \goodS} \lp \sum_{s \in \gS (\widehat{P}^{\piE}_H)} P^{\piail}_{H-1} (s^\prime) P_{H-1} (s | s^\prime, a^{1}) \piail_{H} (a^{1}|s)  \rp \pi_{H-1} (a^{1}|s^\prime),
    \end{align*}
    where in the last equation we use the fact that $\sum_{s \in \goodS} P_{H-1}(s|s^\prime, a^\prime) = 1$ so that the first term and the third term are canceled. Plugging the above equation into $\text{Loss}_{H} (\pi_{H-1})$ yields
    \begin{equation}    \label{eq:proof_vail_reset_cliff_1}
    \begin{split}
        &\quad \text{Loss}_{H} (\pi_{H-1})
        \\
        &= \sum_{s \in \gS (\widehat{P}^{\piE}_H)} \labs \widehat{P}^{\piE}_{H} (s) - \sum_{s^\prime \in \goodS} P^{\piail}_{H-1} (s^\prime)  P_{H-1} (s | s^\prime, a^{1}) \piail_{H} (a^{1}|s) \pi_{H-1} (a^{1}|s^\prime) \rabs
        \\
        & \quad - \sum_{s^\prime \in \goodS} \lp \sum_{s \in \gS (\widehat{P}^{\piE}_H)} P^{\piail}_{H-1} (s^\prime) P_{H-1} (s | s^\prime, a^{1}) \piail_{H} (a^{1}|s) \rp \pi_{H-1} (a^{1}|s^\prime)
        \\
        & \quad + \sum_{s \in \badS} P^{\piail}_{H-1} (s) + \sum_{s^\prime \in \goodS} P^{\piail}_{H-1} (s^\prime).     
    \end{split}
    \end{equation}
    Notice that $P^{\piail}_{H-1} (s)$ is independent of $\pi_{H-1}$ and then we have
    \begin{align*}
        &\quad \argmin_{\pi_{H-1}} \text{Loss}_{H} (\pi_{H-1})
        \\
        &= \argmin_{\pi_{H-1}} \sum_{s \in \gS (\widehat{P}^{\piE}_H)} \labs \widehat{P}^{\piE}_{H} (s) - \sum_{s^\prime \in \goodS} P^{\piail}_{H-1} (s^\prime)  P_{H-1} (s | s^\prime, a^{1}) \piail_{H} (a^{1}|s) \pi_{H-1} (a^{1}|s^\prime) \rabs
        \\
        &\qquad - \sum_{s^\prime \in \goodS} \lp \sum_{s \in \gS (\widehat{P}^{\piE}_H)} P^{\piail}_{H-1} (s^\prime) P_{H-1} (s | s^\prime, a^{1}) \piail_{H} (a^{1}|s) \rp \pi_{H-1} (a^{1}|s^\prime).
    \end{align*}
    For this type optimization problem, we apply \cref{lem:mn_variables_opt_unique} to show that $\forall s \in \goodS, \pi_{H-1}(a^{1} | s) = 1$ is the unique optimal solution. In particular, we verify the conditions of \cref{lem:mn_variables_opt_unique} by defining the following terms:
    \begin{align*}
        &m = \labs \gS (\widehat{P}^{\piE}_H)  \rabs, n = \labs \goodS \rabs, \forall s \in \gS (\widehat{P}^{\piE}_H), c(s) = \widehat{P}^{\piE}_{H} (s),
        \\
        & \forall s \in \gS (\widehat{P}^{\piE}_H), s^\prime \in \goodS, A (s, s^\prime) = P^{\piail}_{H-1} (s^\prime)  P_{H-1} (s | s^\prime, a^{1}) \piail_{H} (a^{1}|s),
        \\
        & \forall s^\prime \in \goodS, d(s^\prime) = \sum_{s \in \gS (\widehat{P}^{\piE}_H)} P^{\piail}_{H-1} (s^\prime) P_{H-1} (s | s^\prime, a^{1}) \piail_{H} (a^{1}|s). 
    \end{align*}
    To help us verify the conditions in \cref{lem:mn_variables_opt_unique}, we note that 
    \cref{lem:condition_for_ail_optimal_solution} implies that if $\piail$ is the optimal solution, then $\forall h \in [H]$, $\exists s \in \goodS$, $\piail_{h} (a^{1}|s) > 0$. Intuitively, in each time step, $\piail$ always takes the expert action on some good state with a positive probability. Combined with the reachable assumption that $\forall h \in [H], s, s^\prime \in \goodS, P_h (s^\prime| s, a^1) > 0$, we have that 
    \begin{align*}
        \forall s^\prime \in \goodS, s \in \gS (\widehat{P}^{\piE}_H), P^{\piail}_{H-1} (s^\prime)  > 0, P_{H-1} (s | s^\prime, a^{1}) > 0.
    \end{align*}
    With \cref{lem:general_condition_ail_policy_at_last_step}, we have that $\forall s \in \gS (\widehat{P}^{\piE}_H), \piail_{H} (a^{1}|s) > 0$. Hence we have that $A > 0$, where $>$ means element-wise comparison. Besides, we have that
    \begin{align*}
         & \sum_{s \in  \gS (\widehat{P}^{\piE}_H) } \sum_{s^\prime \in \goodS} A (s, s^\prime)  \leq \sum_{s \in  \gS (\widehat{P}^{\piE}_H) } \sum_{s^\prime \in \goodS} P^{\piail}_{H-1} (s^\prime)  P_{H-1} (s | s^\prime, a^{1}) \leq  1
         \\
         & \quad = \sum_{s \in  \gS (\widehat{P}^{\piE}_H) } \widehat{P}^{\piE}_{H} (s) = \sum_{s \in  \gS (\widehat{P}^{\piE}_H) } c(s) .
    \end{align*}
    For each $s^\prime \in \goodS$, it holds that 
    \begin{align*}
        \sum_{s \in \gS (\widehat{P}^{\piE}_H)} A (s, s^\prime) =  \sum_{s \in \gS (\widehat{P}^{\piE}_H)} P^{\piail}_{H-1} (s^\prime)  P_{H-1} (s | s^\prime, a^{1}) \piail_{H} (a^{1}|s) = d (s^\prime). 
    \end{align*}
    Thus, we have verified conditions in \cref{lem:mn_variables_opt_unique}. With \cref{lem:mn_variables_opt_unique}, we obtain that $\piail_{H-1} (a^{1}|s) = 1, \forall s \in \goodS$ is the \emph{unique} optimal solution of $\mathrm{Loss}_{H} (\pi_{H-1})$. 
\end{itemize}
Therefore, $\piail_{H-1} (a^{1}|s) = 1, \forall s \in \goodS$ is the \emph{unique} globally optimal solution of $\min_{\pi_{H-1}} \text{Loss}_{H-1} (\pi_{H-1}) + \text{Loss}_{H} (\pi_{H-1})$ and we finish the proof of the base case.

\RED{Second, we prove the induction step (3 pages).} The main proof strategy is similar to what we have used in the proof of the base case but is more tricky. We assume that for step $h^\prime = h+1, h+2, \cdots, H-1$, $\piail_{h^\prime} (a^{1}|s) = 1, \forall s \in \goodS$. We aim to prove that for step $h$, $\piail_{h} (a^{1}|s) = 1, \forall s \in \goodS$. By \cref{lem:n_vars_opt_greedy_structure}, we have that with fixed $(\piail_1, \cdots, \piail_{h-1}, \piail_{h+1}, \cdots, \piail_{H})$, $\piail_h$ is the optimal solution of the \textsf{VAIL}'s objective in \eqref{eq:ail}. This is direct from the global optimality condition. Moreover, note that $P^{\piail}_{t} (s, a)$ for $t \in [h-1]$ is fixed and independent of $\piail_{h}$ under this case. Therefore, we only need to consider the \textsf{VAIL}'s  state-action distribution matching losses from step $h$ to $H$. That is, we only need to prove that $\piail_{h} (a^{1}|s) = 1, \forall s \in \goodS$ is the \emph{unique} optimal solution of the losses from step $h$ to $H$.

Recall that the single-stage loss function in time step $h$ is
\begin{align*}
    \text{Loss}_h (\pi) = \sum_{(s, a) \in \gS \times \gA} \labs P^{\pi}_h(s, a)  - \widehat{P}^{\piE}_h(s, a) \rabs .
\end{align*}
By backward induction, we have three types of losses: 1) the single-stage loss in time step $h$; 2) the single-stage loss in time step $h < h^\prime \leq H-1$; 3) the single-stage loss in time step $H$. We need to prove that $\forall s \in \goodS, \pi_{h}(a^{1} | s) = 1$ is optimal for each cases. Furthermore, we will show that $\forall s \in \goodS, \pi_{h}(a^{1} | s) = 1$ is the unique optimal solution for case 2) and case 3), which proves the uniqueness of the optimal solution of the total losses. 

\begin{itemize}
    \item \BLUE{Term 1.}  For time step $h$, we have that
    \begin{align*}
        \text{Loss}_{h} (\pi_{h}) &= \sum_{s \in \gS} \sum_{a \in \gA} \labs \widehat{P}^{\piE}_{h} (s, a) - P^{\piail}_{h} (s, a) \rabs
        \\
        &= \sum_{s \in \goodS}  \lp \labs \widehat{P}^{\piE}_{h} (s, a^1) - P^{\piail}_{h} (s, a^1) \rabs + \sum_{a \in \gA \setminus \{ a^1\}} P^{\piail}_{h} (s, a)  \rp + \sum_{s \in \badS} P^{\piail}_{h} (s)  
        \\
        &= \sum_{s \in \goodS} \lp \labs \widehat{P}^{\piE}_{h} (s) - P^{\piail}_{h} (s) \pi_{h} (a^{1}|s)  \rabs + P^{\piail}_{h} (s) \lp 1 - \pi_{h} (a^{1}|s) \rp \rp + \sum_{s \in \badS} P^{\piail}_{h} (s). 
    \end{align*}
    Readers may notice that here we slightly abuse the notation and use $P^{\piail}_{h} (s, a)$ to denote the distribution induced by $(\piail_1, \piail_2, \cdots, \pi_h)$. Notice that $P^{\piail}_{h} (s)$ is independent of $\pi_{h}$, then we have that
    \begin{align*}
        &\quad \argmin_{\pi_{h}} \mathrm{Loss}_{h} (\pi_{h})
        \\
        &= \argmin_{\pi_{h}} \sum_{s \in \goodS} \labs \widehat{P}^{\piE}_{h} (s) - P^{\piail}_{h} (s) \pi_{h} (a^{1}|s)  \rabs - P^{\piail}_{h} (s) \pi_{h} (a^{1}|s).
    \end{align*}
    Since the optimization variables $\pi_{h} (a^{1}|s)$ for different $s \in \goodS$ are independent, we can consider the above optimization problem for each $s \in \goodS$ individually.
    \begin{align*}
        \argmin_{\pi_{h} (a^{1}|s) \in [0, 1]} \labs \widehat{P}^{\piE}_{h} (s) - P^{\piail}_{h} (s) \pi_{h} (a^{1}|s)  \rabs - P^{\piail}_{h} (s) \pi_{h} (a^{1}|s).
    \end{align*}
    For this one-dimension optimization problem, we can show that $\pi_h(a^{1}|s) = 1$ is the optimal solution by \cref{lem:single_variable_opt}. Thus, we obtain that $\piail_{h} (a^{1}|s) = 1, \forall s \in \goodS$ is the optimal solution of $\mathrm{Loss}_{h} (\pi_{h})$.
    \item \BLUE{Term 2.} Next, we consider \textsf{VAIL}'s objective values in time step $h^\prime$ where $h+1 \leq h^\prime \leq H-1$. Since $\piail_{h^\prime} (a^{1}|s) = 1, \forall s \in \goodS$, \textsf{VAIL}'s objective value regarding $\pi_h$ in time step $h^\prime$ is formulated as
    \begin{align*}
        \text{Loss}_{h^\prime} (\pi_{h}) &= \sum_{s \in \gS} \sum_{a \in \gA} \labs \widehat{P}^{\piE}_{h^\prime} (s, a) - P^{\piail}_{h^\prime} (s, a) \rabs
        \\
        &= \sum_{s \in \goodS} \sum_{a \in \gA} \labs \widehat{P}^{\piE}_{h^\prime} (s, a) - P^{\piail}_{h^\prime} (s) \piail_{h^\prime} (a|s)  \rabs + \sum_{s \in \badS} \sum_{a \in \gA} P^{\piail}_{h^\prime} (s, a)
        \\
        &= \sum_{s \in \goodS} \labs \widehat{P}^{\piE}_{h^\prime} (s, a^1) - P^{\piail}_{h^\prime} (s, a^1) \rabs + \sum_{s \in \badS} P^{\piail}_{h^\prime} (s)  
        \\
        &= \sum_{s \in \goodS} \labs \widehat{P}^{\piE}_{h^\prime} (s) - P^{\piail}_{h^\prime} (s) \rabs + \sum_{s \in \badS} P^{\piail}_{h^\prime} (s).
    \end{align*}
    With a little abuse of notation, we use $P^{\piail}_{h^\prime} (s)$ to denote the distribution induced by $(\piail_1, \piail_2, \cdots, \pi_h, \piail_{h+1}, \cdots, \piail_{h^\prime})$. Note that only through taking the expert action on good states, the agent could transit into good states. With the \dquote{transition flow equation}, we have 
    \begin{align*}
        \forall s \in \goodS, P^{\piail}_{h^\prime} (s) &= \sum_{s^\prime \in \gS} \sum_{a \in \gA} P^{\piail}_h (s^\prime) \pi_h (a|s^\prime) \sP^{\piail} \lp s_{h^\prime} = s |s_h = s^\prime, a_h = a^{1} \rp  
        \\
        &= \sum_{s^\prime \in \goodS} P^{\piail}_h (s^\prime) \pi_h (a^{1}|s^\prime) \sP^{\piail} \lp s_{h^\prime} = s |s_h = s^\prime, a_h = a^{1} \rp.
    \end{align*}
    Notice that the conditional probability $\sP^{\piail} \lp s_{h^\prime} = s |s_h = s^\prime, a_h = a^{1} \rp$ is independent of $\pi_h$. Besides, as for each $h^\prime = h+1, h+2, \cdots, H-1$, $\piail_{h^\prime} (a^{1}|s) = 1, \forall s \in \goodS$, the visitation probability of bad states in step $h^\prime$ comes from two parts in step $h$. One is the visitation probability of bad states in step $h$. The other is the probability of visiting good states and taking non-expert actions in step $h$. We obtain
    \begin{align*}
        \sum_{s \in \badS} P^{\piail}_{h^\prime} (s) &= \sum_{s \in \badS} P^{\piail}_{h} (s) + \sum_{s^\prime \in \goodS} \sum_{a \in \gA \setminus \{a^1 \}} P^{\piail}_h (s^\prime)  \pi_h (a|s^\prime)
        \\
        &= \sum_{s \in \badS} P^{\piail}_{h} (s) + \sum_{s^\prime \in \goodS} P^{\piail}_h (s^\prime) \lp 1 - \pi_h (a^{1}|s^\prime) \rp .
    \end{align*}
    Plugging the above two equations into $\text{Loss}_{h^\prime} (\pi_{h})$ yields
    \begin{align*}
        \text{Loss}_{h^\prime} (\pi_{h}) &= \sum_{s \in \goodS} \labs \widehat{P}^{\piE}_{h^\prime} (s) - \sum_{s^\prime \in \goodS} P^{\piail}_h (s^\prime) \pi_h (a^{1}|s^\prime) \sP^{\piail} \lp s_{h^\prime} = s |s_h = s^\prime, a_h = a^{1} \rp \rabs
        \\
        &\quad + \sum_{s \in \badS} P^{\piail}_{h} (s) + \sum_{s^\prime \in \goodS} P^{\piail}_h (s^\prime) \lp 1 - \pi_h (a^{1}|s^\prime) \rp. 
    \end{align*}
    This equation is similar to \eqref{eq:proof_vail_reset_cliff_1} in the proof of the base case. Note that $P^{\piail}_h (s)$ is independent of $\pi_h$ and then we have that
    \begin{align*}
        & \quad \argmin_{\pi_h} \text{Loss}_{h^\prime} (\pi_{h})
        \\
        &= \argmin_{\pi_h} \sum_{s \in \goodS} \labs \widehat{P}^{\piE}_{h^\prime} (s) - \sum_{s^\prime \in \goodS} P^{\piail}_h (s^\prime)  \sP^{\piail} \lp s_{h^\prime} = s |s_h = s^\prime, a_h = a^{1} \rp \pi_h (a^{1}|s^\prime) \rabs
        \\
        & \quad - \sum_{s^\prime \in \goodS} P^{\piail}_h (s^\prime) \pi_h (a^{1}|s^\prime).   
    \end{align*}
    For this type optimization problem, we can again use \cref{lem:mn_variables_opt_unique} to prove that $\forall s \in \goodS, \pi_h(a^{1} | s) = 1$ is the unique global optimal solution. To check the conditions in \cref{lem:mn_variables_opt_unique}, we define 
    \begin{align*}
        & m = n = \labs \goodS \rabs, \forall s \in \goodS, c(s) = \widehat{P}^{\piE}_{h^\prime} (s), \\
        & \forall s, s^\prime \in \goodS, A (s, s^\prime) = P^{\piail}_h (s^\prime)  \sP^{\piail} \lp s_{h^\prime} = s |s_h = s^\prime, a_h = a^{1} \rp,
        \\
        & \forall s^\prime \in \goodS, d(s^\prime) = P^{\piail}_h (s^\prime). 
    \end{align*}
    To help verify \cref{lem:mn_variables_opt_unique}, we note that  \cref{lem:condition_for_ail_optimal_solution} implies that if $\piail$ is the optimal solution, then $\forall h \in [H]$, $\exists s \in \goodS$, $\piail_{h} (a^{1}|s) > 0$. Combined with the reachable assumption that $\forall h \in [H], s, s^\prime \in \goodS, P_h (s^\prime |s, a^1) > 0$, we have that 
    \begin{align*}
        \forall s, s^\prime \in \goodS, P^{\piail}_{h} (s^\prime)  > 0, \sP^{\piail} \lp s_{h^\prime} = s |s_h = s^\prime, a_h = a^{1} \rp > 0.
    \end{align*}
    Then we can obtain that $A > 0$ where $>$ means element-wise comparison. Besides, we have that
    \begin{align*}
        & \sum_{s \in \goodS} c (s) = 1 \geq \sum_{s \in \goodS} \sum_{s^\prime \in \goodS} P^{\piail}_h (s^\prime)  \sP^{\piail} \lp s_{h^\prime} = s |s_h = s^\prime, a_h = a^{1} \rp = \sum_{s \in \goodS} \sum_{s^\prime \in \goodS} A (s, s^\prime).
    \end{align*}
    For each $s^\prime \in \goodS$, we further have that 
    \begin{align*}
        \sum_{s \in \goodS} A (s, s^\prime) = \sum_{s \in \goodS} P^{\piail}_h (s^\prime)  \sP^{\piail} \lp s_{h^\prime} = s |s_h = s^\prime, a_h = a^{1} \rp = P^{\piail}_h (s^\prime) = d(s^\prime).  
    \end{align*}
    Thus, we have verified conditions in \cref{lem:mn_variables_opt_unique}. By Lemma \ref{lem:mn_variables_opt_unique}, we obtain that $\piail_{h} (a^{1}|s) = 1, \forall s \in \goodS$ is the \emph{unique} optimal solution of $\mathrm{Loss}_{h^\prime} (\pi_{h})$ for each time step $h^\prime$, where $h+1 \leq h^\prime \leq H-1$.
    \item \BLUE{Term 3.} Finally, we consider the last time step $H$. Recall the definition that $\gS (\widehat{P}^{\piE}_H) = \{s \in \gS, \widehat{P}^{\piE}_H (s) > 0  \}$. \textsf{VAIL}'s loss in step $H$ is formulated as 
    \begin{align*}
        &\quad \text{Loss}_{H} (\pi_h)
        \\
        &= \sum_{s \in \gS} \sum_{a \in \gA} \labs \widehat{P}^{\piE}_{H} (s, a) - P^{\piail}_{H} (s, a) \rabs
        \\
        &= \sum_{s \in \goodS} \sum_{a \in \gA} \labs \widehat{P}^{\piE}_{H} (s, a) - P^{\piail}_{H} (s, a) \rabs + \sum_{s \in \badS} P^{\piail}_H (s)
        \\
        &=\sum_{s \in \gS (\widehat{P}^{\piE}_H)} \lp \labs \widehat{P}^{\piE}_{H} (s, a^1) - P^{\piail}_{H} (s, a^1) \rabs + \sum_{a \ne a^{1}} P^{\piail}_{H} (s, a) \rp + \sum_{s \in \goodS \text{ and } s \notin \gS (\widehat{P}^{\piE}_H)} \sum_{a \in \gA} P^{\piail}_{H} (s, a)
        \\
        &\quad  + \sum_{s \in \badS} P^{\piail}_H (s)
        \\
        &= \sum_{s \in \gS (\widehat{P}^{\piE}_H)} \lp \labs \widehat{P}^{\piE}_{H} (s) - P^{\piail}_{H} (s, a^{1}) \rabs + P^{\piail}_H (s) (1 - \piail_{H} (a^{1}|s)) \rp + \sum_{s \in \goodS \text{ and } s \notin \gS (\widehat{P}^{\piE}_H)} P^{\piail}_H (s)
        \\
        &\quad + \sum_{s \in \badS} P^{\piail}_H (s), 
    \end{align*}
    where we slightly abuse the notation and use $P^{\piail}_{H} (s, a), P^{\piail}_{H} (s)$ to denote the distributions induced by $(\piail_1, \piail_2, \cdots, \pi_h, \piail_{h+1}, \cdots, \piail_{H})$. Similarly, with the \dquote{transition flow equation}, we have
    \begin{align*}
         \forall s \in \goodS, P^{\piail}_{H} (s) &= \sum_{s^\prime \in \gS} \sum_{a \in \gA} P^{\piail}_{h} (s^\prime) \pi_{h} (a|s^\prime) \sP^{\piail} (s_H = s | s_h = s^\prime, a_h = a)
        \\
        &= \sum_{s^\prime \in \goodS} P^{\piail}_{h} (s^\prime) \pi_{h} (a^{1}|s^\prime) \sP^{\piail} (s_H = s | s_h = s^\prime, a_h = a^{1}).  
    \end{align*}
    Notice that for time step $h^\prime = h+1, h+2, \cdots, H-1$, $\piail_{h^\prime} (a^{1}|s) = 1, \forall s \in \goodS$. Then we have 
    \begin{align*}
        \sum_{s \in \badS} P^{\piail}_H (s) &= \sum_{s \in \badS} P^{\piail}_{h} (s) + \sum_{s^\prime \in \goodS} \sum_{a \in \gA \setminus \{a^1\}} P^{\piail}_{h} (s^\prime)  \pi_{h} (a|s^\prime)
        \\
        &= \sum_{s \in \badS} P^{\piail}_{h} (s) + \sum_{s^\prime \in \goodS} P^{\piail}_{h} (s^\prime) \lp 1 - \pi_{h} (a^{1}|s^\prime) \rp.
    \end{align*}
   
    Plugging the above equation into $\text{Loss}_{H} (\pi_h)$ yields that
    \begin{align*}
        \text{Loss}_{H} (\pi_h) &= \sum_{s \in \gS (\widehat{P}^{\piE}_H)} \labs \widehat{P}^{\piE}_{H} (s) - \sum_{s^\prime \in \goodS} P^{\piail}_{h} (s^\prime) \sP^{\piail} (s_H = s | s_h = s^\prime, a_h = a^{1}) \piail_H (a^{1}|s) \pi_h (a^{1}|s^\prime)    \rabs
        \\
        &\quad - \sum_{s^\prime \in \goodS} \lp \sum_{s \in \gS (\widehat{P}^{\piE}_H)} P^{\piail}_{h} (s^\prime) \sP^{\piail} (s_H = s | s_h = s^\prime, a_h = a^{1}) \piail_H (a^{1}|s) \rp \pi_h (a^{1}|s^\prime)
        \\
        &\quad + \sum_{s \in \badS} P^{\piail}_{h} (s) + \sum_{s^\prime \in \goodS} P^{\piail}_h (s^\prime). 
    \end{align*}
    This equation is similar to \eqref{eq:proof_vail_reset_cliff_1} in the proof of the base case.  Since $P^{\piail}_h (s)$ is independent of $\pi_h$, we have that
    \begin{align*}
        & \quad \argmin_{\pi_h} \text{Loss}_{H} (\pi_h)
        \\
        &= \argmin_{\pi_h} \sum_{s \in \gS (\widehat{P}^{\piE}_H)} \labs \widehat{P}^{\piE}_{H} (s) - \sum_{s^\prime \in \goodS} P^{\piail}_{h} (s^\prime) \sP^{\piail} (s_H = s | s_h = s^\prime, a_h = a^{1}) \piail_H (a^{1}|s) \pi_h (a^{1}|s^\prime)    \rabs
        \\
        &\quad - \sum_{s^\prime \in \goodS} \lp \sum_{s \in \gS (\widehat{P}^{\piE}_H)} P^{\piail}_{h} (s^\prime) \sP^{\piail} (s_H = s | s_h = s^\prime, a_h = a^{1}) \piail_H (a^{1}|s) \rp \pi_h (a^{1}|s^\prime).
    \end{align*}
    For this type optimization problem, we can again use \cref{lem:mn_variables_opt_unique} to prove that $\forall s \in \goodS, \piail_h(a^{1} | s) = 1$ is the unique globally optimal solution. To check conditions in \cref{lem:mn_variables_opt_unique}, we define 
    \begin{align*}
        & m = \labs \gS (\widehat{P}^{\piE}_H) \rabs, n = \labs \goodS \rabs, \forall s \in \gS (\widehat{P}^{\piE}_H), c(s) = \widehat{P}^{\piE}_{H} (s), 
        \\
        & \forall s \in \gS (\widehat{P}^{\piE}_H), s^\prime \in \goodS, A (s, s^\prime) = P^{\piail}_{h} (s^\prime) \sP^{\piail} (s_H = s | s_h = s^\prime, a_h = a^{1}) \piail_H (a^{1}|s),
        \\
        & \forall s^\prime \in \goodS, d(s^\prime) =  \sum_{s \in \gS (\widehat{P}^{\piE}_H)} P^{\piail}_{h} (s^\prime) \sP^{\piail} (s_H = s | s_h = s^\prime, a_h = a^{1}) \piail_H (a^{1}|s). 
    \end{align*}
    Similarly, we have that
    \begin{align*}
        & A > 0, \sum_{s \in \gS (\widehat{P}^{\piE}_H)} \sum_{s^\prime \in \goodS} A (s, s^\prime) \leq \sum_{s \in \gS (\widehat{P}^{\piE}_H)} \sum_{s^\prime \in \goodS} P^{\piail}_{h} (s^\prime) \sP^{\piail} (s_H = s | s_h = s^\prime, a_h = a^{1}) \leq 1
        \\
        & \qquad \qquad =  \sum_{s \in \gS (\widehat{P}^{\piE}_H)} c(s),
        \\
        &\forall s^\prime \in \goodS, \sum_{s \in \gS (\widehat{P}^{\piE}_H)} A (s, s^\prime) =  \sum_{s \in \gS (\widehat{P}^{\piE}_H)} P^{\piail}_{h} (s^\prime) \sP^{\piail} (s_H = s | s_h = s^\prime, a_h = a^{1}) \piail_H (a^{1}|s) = d(s^\prime).  
    \end{align*}
    Thus we have verified conditions in \cref{lem:mn_variables_opt_unique}. By \cref{lem:mn_variables_opt_unique}, we obtain $\piail_h(a^{1} | s) = 1, \forall s \in \goodS$ is the unique optimal solution of $\min_{\pi_h} \text{Loss}_{H} (\pi_h)$.
\end{itemize}
Thus, we finish the induction proof and the whole proof is done. 

\end{proof}

\subsection{Proof of Theorem \ref{theorem:ail_reset_cliff}}

We first formally present the sample complexity of \textsf{VAIL} to achieve a small policy value gap \emph{with high probability} on Reset Cliff. This result is similar to Theorem \ref{theorem:ail_reset_cliff}.

\begin{thm}[High Probability Version of \cref{theorem:ail_reset_cliff}]     
\label{theorem:ail_reset_cliff_high_prob}
For each tabular and episodic MDP satisfying \cref{asmp:reset_cliff}, with probability at least $1-\delta$, to obtain an $\varepsilon$-optimal policy (i.e., $V^{\piE} - V^{\piail} \leq \varepsilon$), \textsf{VAIL} requires at most ${\widetilde{\gO}}(|\gS|/\varepsilon^2)$ expert trajectories.
\end{thm}

We discuss the proof idea here. \cref{prop:ail_general_reset_cliff} indicates \textsf{VAIL} exactly recovers the expert policy in the first $H-1$ time steps. Thus, we infer that the policy value gap of $\piail$ only arises from the decision errors in the last time step. Similar to the proof of \cref{theorem:worst_case_sample_complexity_of_vail}, we can utilize $\ell_1$-norm concentrations inequality to upper bound the policy value gap in the last time step.

\begin{proof}[Proof of \cref{theorem:ail_reset_cliff} and \cref{theorem:ail_reset_cliff_high_prob}]
Similar to the proof of \cref{theorem:worst_case_sample_complexity_of_vail}, we can upper bound the policy value gap with the state-action distribution discrepancy. 
\begin{align*}
    \labs  V^{\piE} - V^{\piail} \rabs &\leq \sum_{h=1}^{H}  \sum_{(s, a) \in \gS \times \gA} \labs P^{\piE}_h(s, a) - P^{\piail}_h(s, a) \rabs.
\end{align*}
We apply \cref{prop:ail_general_reset_cliff} with the maximum likelihood estimation $\widehat{P}^{\piE}_h (s, a)$. From \cref{prop:ail_general_reset_cliff}, we have that for any $h \in [H-1]$, $\piail_{h} (a^{1}|s) = \piE_h (a^{1}|s) = 1, \forall s \in \goodS$. Therefore, $\piE$ and $\piail$ never visit bad states and for any $h \in [H-1]$, $P^{\piE}_h(s, a) = P^{\piail}_h(s, a)$. As a result, the policy value gap is upper bounded by the state-action distribution discrepancy in the last time step.
\begin{align*}
    \labs  V^{\piE} - V^{\piail} \rabs &\leq \sum_{(s, a) \in \gS \times \gA} \labs P^{\piE}_H (s, a) - P^{\piail}_H (s, a) \rabs
    \\
    &\leq \sum_{(s, a) \in \gS \times \gA} \labs P^{\piE}_H (s, a) - \widehat{P}^{\piE}_H (s, a) \rabs + \labs \widehat{P}^{\piE}_H (s, a) - P^{\piail}_H (s, a)  \rabs.
\end{align*}
Since $\piail = (\piail_1, \cdots, \piail_H)$ is the optimal solution of VAIL's objective in \eqref{eq:ail}, by \cref{lem:n_vars_opt_greedy_structure}, it holds that with fixed $(\piail_1, \cdots, \piail_{H-1})$, $\piail_{H}$ is also optimal w.r.t VAIL's objective. From \cref{prop:ail_general_reset_cliff}, we know that $(\piail_1 (a^{1}|s), \cdots, \piail_{H-1} (a^{1}|s)) = (\piE_1 (a^{1}|s), \cdots, \piE_{H-1} (a^{1}|s)), \forall s \in \goodS$ and thus $P^{\piail}_H (s) = P^{\piE}_H (s), \forall s \in \gS$. Then with fixed $(\piail_1, \cdots, \piail_{H-1})$, VAIL's objective is formulated as
\begin{align*}
    \piail_{H} &\in \argmin_{\pi_H} \sum_{(s, a) \in \gS \times \gA} \labs \widehat{P}^{\piE}_H (s, a) - P^{\piail}_H (s, a) \rabs
    \\
    &= \argmin_{\pi_H} \sum_{(s, a) \in \gS \times \gA} \labs \widehat{P}^{\piE}_H (s, a) - P^{\piE}_H (s) \pi_H (a|s)  \rabs, 
\end{align*}
where we slightly abuse the notation and use $P^{\piail}_H (s, a)$ to denote the distribution induced by $(\piail_1, \cdots, \piail_{H-1}, \pi_H)$ temporally. Since $\piail_{H} \in  \argmin_{\pi_H} \sum_{(s, a) \in \gS \times \gA} \labs \widehat{P}^{\piE}_H (s, a) - P^{\piE}_H (s) \pi_H (a|s)  \rabs$, we have
\begin{align*}
    \sum_{(s, a) \in \gS \times \gA} \labs \widehat{P}^{\piE}_H (s, a) - P^{\piail}_H (s, a)  \rabs &=\sum_{(s, a) \in \gS \times \gA} \labs \widehat{P}^{\piE}_H (s, a) - P^{\piE}_H (s) \piail_H (a|s)  \rabs
    \\
    &\leq \sum_{(s, a) \in \gS \times \gA} \labs \widehat{P}^{\piE}_H (s, a) - P^{\piE}_H (s) \piE_H (a|s)  \rabs.
\end{align*}
Then we obtain
\begin{align}
    \labs  V^{\piE} - V^{\piail} \rabs &\leq \sum_{(s, a) \in \gS \times \gA} \labs P^{\piE}_H (s, a) - \widehat{P}^{\piE}_H (s, a) \rabs + \labs \widehat{P}^{\piE}_H (s, a) - P^{\piail}_H (s, a)  \rabs \nonumber
    \\
    &\leq 2 \sum_{(s, a) \in \gS \times \gA} \labs P^{\piE}_H (s, a) - \widehat{P}^{\piE}_H (s, a) \rabs = 2\sum_{s \in \gS} \labs \widehat{P}^{\piE}_H (s) - P^{\piE}_H (s)  \rabs. \label{eq:vail_reset_cliff_value_gap_last_step_estimation_error} 
\end{align}
First, we prove the sample complexity required to achieve a small policy value gap \emph{with high probability}. With \cref{lemma:l1_concentration}, with probability at least $1-\delta$, we have
\begin{align}
\label{eq:value_gap_ail_policy}
    \labs  V^{\piE} - V^{\piail} \rabs \leq 2 \sum_{s \in \gS} \labs  \widehat{P}^{\piE}_H (s) - P^{\piE}_H (s)  \rabs \leq 2\sqrt{\frac{2 \vert \gS \vert \ln (1/\delta)}{m}},
\end{align}
which translates to a sample complexity $\widetilde{\gO} \lp \vert \gS \vert / \varepsilon^2 \rp$.

Finally, we prove the sample complexity required to achieve a small policy value gap \emph{in expectation}. We apply \citep[Theorem 1]{han2015minimax} and have that
\begin{align*}
    V^{\piE} - \expect \ls V^{\piail} \rs \leq 2 \expect \ls \lnorm \widehat{P}^{\piE}_H (\cdot) - P^{\piE}_H (\cdot) \rnorm_{1} \rs \leq 2 \sqrt{ \frac{|\gS| - 1}{m}},
\end{align*}
which translates to a sample complexity $\gO \lp \vert \gS \vert / \varepsilon^2 \rp$ as in \cref{theorem:ail_reset_cliff}.
\end{proof}

\section{Proof of Results in Section \ref{sec:beyond_vanilla_ail}}
\label{appendix:proof_beyond_vanilla_ail}

\subsection{Proof of Theorem \ref{theorem:final_sample_complexity}}
\label{appendix:proof_of_theorem:final_sample_complexity}

Before we prove Theorem \ref{theorem:final_sample_complexity}, we first state three key lemmas: Lemma \ref{lemma:regret_of_ogd}, and Lemma \ref{lemma:approximate-minimax}, and \cref{lemma:sample_complexity_of_new_estimator_known_transition}.
\begin{lem}\label{lemma:regret_of_ogd}
Consider the adversarial imitation learning approach displayed in Algorithm \ref{algo:main_aglorithm}, then we have
\begin{align*}
    \sum_{t=1}^T f^{(t)} \lp w^{(t)} \rp - \min_{w \in \gW} \sum_{t=1}^T f^{(t)} (w) \leq 2H \sqrt{2 |\gS| |\gA| T},
\end{align*}
where $f^{(t)}(w) = \sum_{h=1}^{H} \sum_{(s, a) \in \gS \times \gA} w_h(s, a) ( P^{\pi^{(t)}}_h (s, a) - \widetilde{P}^{\piE}_h (s, a) )$.
\end{lem}

Refer to \cref{appendix:proof_lemma_regret_of_ogd} for the proof. Basically, \cref{lemma:regret_of_ogd} is a direct consequence of the regret bound of online gradient descent \citep{shalev12online-learning}.

\begin{lem} \label{lemma:approximate-minimax}
Consider the transition-aware adversarial imitation learning approach displayed in Algorithm \ref{algo:main_aglorithm} and $\widebar{\pi}$ is the output policy, then we have
\begin{align*}
\sum_{h=1}^H \lnorm P^{\widebar{\pi}}_h - \widetilde{P}^{\piE}_h \rnorm_{1} \leq \min_{\pi \in \Pi} \sum_{h=1}^H \lnorm P^{\pi}_h - \widetilde{P}^{\piE}_h \rnorm_{1} + 2 H \sqrt{\frac{2 |\gS| |\gA|}{T}} + \varepsilon_{\mathrm{opt}}. 
\end{align*}
\end{lem}

Refer to \cref{appendix:proof_lemma_approximate_minimax} for the proof. In particular, \cref{lemma:approximate-minimax} provides the guarantee of online gradient descent for the approximate saddle point optimization (i.e., the min-max optimization in \eqref{eq:new_algo_max_min}).

\begin{lem}
\label{lemma:sample_complexity_of_new_estimator_known_transition}
Consider $\gD$ is randomly divided into two subsets, i.e., $\gD = \gD_{1} \cup \gD_{1}^c$ with $\labs \gD_1 \rabs = \labs \gD_1^{c} \rabs = m / 2$. Fix $\varepsilon \in (0, H)$ and $\delta \in (0, 1)$; suppose $H \geq 5$. Consider the estimator in \eqref{eq:new_estimator}, if the number of trajectories ($m$) satisfies
\begin{align*}
  m \succsim   \frac{  | \gS | H^{3/2}}{\varepsilon} \log\lp  \frac{ |\gS| H}{\delta} \rp ,
\end{align*}
then with probability at least $1-\delta$, we have $\sum_{h=1}^H \Vert \widetilde{P}^{\piE}_h - P^{\piE}_h  \Vert_1 \leq \varepsilon$.
\end{lem}

Refer to \cref{appendix:proof_lemma_sample_complexity_of_new_estimator_known_transition} for the proof. The proof relies on the fine-grained analysis of \citep[Lemma A.12]{rajaraman2020fundamental}. Specifically, \cref{lemma:sample_complexity_of_new_estimator_known_transition} indicates a better sample complexity of the new estimator in \eqref{eq:new_estimator}.

\begin{proof}[Proof of Theorem \ref{theorem:final_sample_complexity}]
Let $\widebar{\pi}$ be the policy output by Algorithm \ref{algo:main_aglorithm}. With Lemma \ref{lemma:approximate-minimax}, we establish the upper bound on the $\ell_1$ deviation between $P^{\widebar{\pi}}_h (s, a)$ and $\widetilde{P}^{\piE}_h(s, a)$. 
\begin{align*}
    \sum_{h=1}^H \lnorm P^{\widebar{\pi}}_h  - \widetilde{P}^{\piE}_h \rnorm_{1} \leq \min_{\pi \in \Pi} \sum_{h=1}^H \lnorm P^{\pi}_h  - \widetilde{P}^{\piE}_h \rnorm_{1} + 2 H \sqrt{\frac{2 |\gS| |\gA|}{T}} + \varepsilon_{\mathrm{opt}}. 
\end{align*}
Since $\piE \in \Pi$, we further obtain that
\begin{align*}
    \sum_{h=1}^H \lnorm P^{\widebar{\pi}}_h  - \widetilde{P}^{\piE}_h \rnorm_{1} \leq \sum_{h=1}^H \lnorm P^{\piE}_h - \widetilde{P}^{\piE}_h \rnorm_{1} + 2 H \sqrt{\frac{2 |\gS| |\gA|}{T}} + \varepsilon_{\mathrm{opt}}.
\end{align*}
By \cref{lemma:sample_complexity_of_new_estimator_known_transition}, fix $\varepsilon \in (0, H)$ and $\delta \in (0, 1)$, when the number of trajectories in $\gD$ satisfies that $m \succsim | \gS | H^{3/2} \log\lp  |\gS| H / \delta \rp / \varepsilon $, with probability at least $1-\delta$, we have
\begin{align*}
    \sum_{h=1}^H \lnorm P^{\piE}_h - \widetilde{P}^{\piE}_h \rnorm_{1} \leq \frac{\varepsilon}{8}.
\end{align*}
Moreover, with $T \succsim |\gS| |\gA| H^2 / \varepsilon^2$ and $\varepsilon_{\mathrm{opt}} \leq \varepsilon / 2$, we can obtain that
\begin{align*}
    \sum_{h=1}^H \lnorm P^{\widebar{\pi}}_h - \widetilde{P}^{\piE}_h \rnorm_{1} \leq \frac{\varepsilon}{8} + \frac{\varepsilon}{4} + \varepsilon_{\mathrm{opt}} \leq \frac{7\varepsilon}{8}.
\end{align*}
Finally, with the dual representation of policy value, we can upper bound the policy value gap by the state-action distribution error.
\begin{align*}
    \left\vert V^{\piE} - V^{\bar{\pi}} \right\vert &= \left\vert \sum_{h=1}^H \sum_{(s, a) \in \gS \times \gA} \lp P^{\piE}_h (s, a) - P^{\widebar{\pi}}_h (s, a) \rp r_h (s, a) \right\vert
    \\
    &\leq \sum_{h=1}^H \lnorm P^{\piE}_h - P^{\widebar{\pi}}_h \rnorm_1
    \\
    &\leq \sum_{h=1}^H \lnorm P^{\piE}_h - \widetilde{P}^{\piE}_h \rnorm_1 + \sum_{h=1}^H \lnorm \widetilde{P}^{\piE}_h - P^{\widebar{\pi}}_h \rnorm_1
    \\
    &\leq \frac{\varepsilon}{8} + \frac{7\varepsilon}{8} = \varepsilon.
\end{align*}
\end{proof}

\subsection{Proof of Proposition \ref{prop:connection}}
\label{subsec:proof_of_proposition_connection}

\begin{proof}

Let $\widetilde{P}^{\piE}_h(s, a)$ be an expert state-action distribution estimator and $\widehat{\gP}$ be a transition model learned by a reward-free method. We define the following two events.
\begin{align*}
    &E_{\mathrm{EST}}:= \lb \sum_{h=1}^H \lnorm \widetilde{P}^{\piE}_h - P^{\piE}_h  \rnorm_{1} \leq \varepsilon_{\mathrm{EST}} \rb,
    \\
    & E_{\mathrm{RFE}}:= \lb \forall r: \gS \times \gA \rar [0, 1], \pi \in \Pi, \left\vert V^{\pi, \gP, r} - V^{\pi, \widehat{\gP}, r} \right\vert \leq \varepsilon_{\mathrm{RFE}} \rb. 
\end{align*}
According to assumption $(a)$ and $(b)$, we have that $\sP \lp E_{\mathrm{EST}}  \rp \geq 1 - \delta_{\mathrm{EST}}$ and $\sP \lp E_{\mathrm{RFE}}  \rp \geq 1 - \delta_{\mathrm{RFE}}$. Applying union bound yields
\begin{align*}
    \sP \lp E_{\mathrm{EST}}  \cap E_{\mathrm{RFE}} \rp \geq 1 - \delta_{\mathrm{EST}} -  \delta_{\mathrm{RFE}}.
\end{align*}
The following analysis is established on the event $E_{\mathrm{EST}}  \cap E_{\mathrm{RFE}}$. Let $\widebar{\pi}$ be the output of Algorithm \ref{algo:framework}.
\begin{align*}
    \left\vert V^{\piE, \gP} - V^{\widebar{\pi}, \gP} \right\vert \leq \left\vert V^{\piE, \gP} - V^{\widebar{\pi}, \widehat{\gP}} \right\vert + \left\vert V^{\widebar{\pi}, \widehat{\gP}} - V^{\widebar{\pi}, \gP} \right\vert \leq \left\vert V^{\piE, \gP} - V^{\widebar{\pi}, \widehat{\gP}} \right\vert + \varepsilon_{\mathrm{RFE}}. 
\end{align*}
The last inequality follows the event $E_{\mathrm{RFE}}$. Then we consider the error $\vert V^{\piE, \gP} - V^{\widebar{\pi}, \widehat{\gP}} \vert$. From the dual form of the policy value in \cref{lemma:policy_dual_value}, we have that
\begin{align*}
    \left\vert V^{\piE, \gP} - V^{\widebar{\pi}, \widehat{\gP}} \right\vert &= \left\vert \sum_{h=1}^H \sum_{(s, a) \in \gS \times \gA} \lp P^{\piE, \gP}_h (s, a) - P^{\widebar{\pi}, \widehat{\gP}}_h (s, a) \rp r_h (s, a)  \right\vert \leq \sum_{h=1}^H \lnorm P^{\piE, \gP}_h - P^{\widebar{\pi}, \widehat{\gP}}_h   \rnorm_1,
\end{align*}
where $P^{\widebar{\pi}, \widehat{\gP}}_h (s, a)$ is the state-action distribution of the policy $\widebar{\pi}$ under the transition model $\widehat{\gP}$. Then we get that
\begin{align*}
    \sum_{h=1}^H \lnorm P^{\piE, \gP}_h - P^{\widebar{\pi}, \widehat{\gP}}_h   \rnorm_1 & \leq \sum_{h=1}^H \lnorm P^{\piE, \gP}_h - \widetilde{P}^{\piE}_h   \rnorm_1 + \sum_{h=1}^H \lnorm \widetilde{P}^{\piE}_h - P^{\widebar{\pi}, \widehat{\gP}}_h   \rnorm_1
    \\
    &\leq \varepsilon_{\mathrm{EST}} + \sum_{h=1}^H \lnorm \widetilde{P}^{\piE}_h - P^{\widebar{\pi}, \widehat{\gP}}_h   \rnorm_1. 
\end{align*}
The last inequality follows the event $E_{\mathrm{EST}}$. Combining the above three inequalities yields
\begin{align*}
    \left\vert V^{\piE, \gP} - V^{\widebar{\pi}, \gP} \right\vert \leq \sum_{h=1}^H \lnorm \widetilde{P}^{\piE}_h - P^{\widebar{\pi}, \widehat{\gP}}_h   \rnorm_1 + \varepsilon_{\mathrm{EST}} +  \varepsilon_{\mathrm{RFE}}.
\end{align*}
According to assumption $(c)$, with the estimator $\widetilde{P}^{\piE}_h (s, a)$ and transition model $\widehat{\gP}$, algorithm B solves the projection problem in \eqref{eq:l1_norm_imitation_with_estimator} up to an error $\varepsilon_{\mathrm{AIL}}$ and $\widebar{\pi}$ is the output of the algorithm B. Formally,
\begin{align*}
    \sum_{h=1}^H \lnorm \widetilde{P}^{\piE}_h - P^{\widebar{\pi}, \widehat{\gP}}_h   \rnorm_1 \leq \min_{\pi \in \Pi} \sum_{h=1}^H \lnorm \widetilde{P}^{\piE}_h - P^{\pi, \widehat{\gP}}_h   \rnorm_1 + \varepsilon_{\mathrm{AIL}}.
\end{align*}
Then we get that
\begin{align*}
    \left\vert V^{\piE, \gP} - V^{\widebar{\pi}, \gP} \right\vert &\leq \sum_{h=1}^H \lnorm \widetilde{P}^{\piE}_h - P^{\widebar{\pi}, \widehat{\gP}}_h   \rnorm_1 + \varepsilon_{\mathrm{EST}} +  \varepsilon_{\mathrm{RFE}}
    \\
    &\leq \min_{\pi \in \Pi} \sum_{h=1}^H \lnorm \widetilde{P}^{\piE}_h - P^{\pi, \widehat{\gP}}_h  \rnorm_1 +  \varepsilon_{\mathrm{AIL}} + \varepsilon_{\mathrm{EST}} +  \varepsilon_{\mathrm{RFE}}
    \\
    &\overset{(1)}{\leq} \sum_{h=1}^H \lnorm \widetilde{P}^{\piE}_h - P^{\piE, \widehat{\gP}}_h   \rnorm_1 +  \varepsilon_{\mathrm{AIL}} + \varepsilon_{\mathrm{EST}} +  \varepsilon_{\mathrm{RFE}}
    \\
    &\leq \sum_{h=1}^H \lnorm \widetilde{P}^{\piE}_h - P^{\piE, \gP}_h   \rnorm_1 + \sum_{h=1}^H \lnorm P^{\piE, \gP}_h - P^{\piE, \widehat{\gP}}_h   \rnorm_1  +  \varepsilon_{\mathrm{AIL}} + \varepsilon_{\mathrm{EST}} +  \varepsilon_{\mathrm{RFE}}
    \\
    &\overset{(2)}{\leq} \sum_{h=1}^H \lnorm P^{\piE, \gP}_h - P^{\piE, \widehat{\gP}}_h   \rnorm_1  +  \varepsilon_{\mathrm{AIL}} + 2\varepsilon_{\mathrm{EST}} +  \varepsilon_{\mathrm{RFE}},
\end{align*}
where inequality $(1)$ holds since $\piE \in \Pi$ and inequality $(2)$ follows the event $E_{\mathrm{EST}}$. With the dual representation of $\ell_1$-norm, we have that
\begin{align*}
    \sum_{h=1}^H \lnorm P^{\piE, \gP}_h - P^{\piE, \widehat{\gP}}_h \rnorm_1 &= \max_{w \in \gW} \sum_{h=1}^H w_h (s, a) \lp P^{\piE, \gP}_h (s, a) - P^{\piE, \widehat{\gP}}_h (s, a)  \rp
    \\
    &= \max_{w \in \gW} V^{\piE, \gP, w} - V^{\piE, \widehat{\gP}, w} \leq \varepsilon_{\mathrm{RFE}},
\end{align*}
where $\gW = \{w: \|w \|_{\infty} \leq 1 \}$, $V^{\piE, \widehat{\gP}, w}$ is the value of policy $\piE$ with the transition model $\widehat{\gP}$ and reward function $w$. The last inequality follows the event $E_{\mathrm{RFE}}$. Then we prove that
\begin{align*}
    \left\vert V^{\piE, \gP} - V^{\widebar{\pi}, \gP} \right\vert \leq  2\varepsilon_{\mathrm{EST}} + 2 \varepsilon_{\mathrm{RFE}} + \varepsilon_{\mathrm{AIL}}. 
\end{align*}

\end{proof}

\subsection{Proof of Theorem \ref{theorem:sample-complexity-unknown-transition}}
\label{subsection:proof-of-theorem-sample-complexity-unknown-transition}

Before we prove \cref{theorem:sample-complexity-unknown-transition}, we first explain the modified estimator in \eqref{eq:new_estimator_unknown_transition}. In particular, we demonstrate it is an unbiased estimator under the unknown transition setting and present its sample complexity and interaction complexity. Then, we review the theoretical guarantee of the RF-Express algorithm.

We consider the decomposition of $P_h^{\piE} (s, a)$.
\begin{align*} 
P_h^{\piE}(s, a) &= {\sum_{\tr_h \in \Tr_h^{\gD_1}} \sP^{\piE}(\tr_h) \indict\lb \tr_h(s_h, a_h) = (s, a) \rb} + {\sum_{\tr_h \notin \Tr_h^{\gD_1}} \sP^{\piE}(\tr_h) \indict\lb \tr_h(s_h, a_h) = (s, a) \rb}
\\
&= {\sum_{\tr_h \in \Tr_h^{\gD_1}} \sP^{\pi}(\tr_h) \indict\lb \tr_h(s_h, a_h) = (s, a) \rb} + {\sum_{\tr_h \notin \Tr_h^{\gD_1}} \sP^{\piE}(\tr_h) \indict\lb \tr_h(s_h, a_h) = (s, a) \rb},
\end{align*}
where $\pi \in \Pi_{\text{BC}} \lp \gD_1 \rp$ and the last equality follows Lemma \ref{lemma:unknown-transition-unbiased-estimation}. Recall the definition of the new estimator.
\begin{align*}
\widetilde{P}_h^{\piE} (s, a) = {\frac{\sum_{\tr_h \in \gD^\prime_{\mathrm{env}}} \indict \{ \tr_h (s_h, a_h) = (s, a), \tr_h \in \Tr_h^{\gD_1} \}}{|\gD^\prime_{\mathrm{env}}|}} + {\frac{  \sum_{\tr_h \in \gD_1^c}  \indict\{ \tr_h (s_h, a_h) = (s, a), \tr_h \not\in \Tr_h^{\gD_1}  \} }{|\gD_1^c|}},
\end{align*}
where $\gD^\prime_{\mathrm{env}}$ is the dataset collected by the policy $\pi \in \Pi_{\text{BC}} (\gD_1)$. Notice that the two terms in RHS are Monte Carlo estimations of ${\sum_{\tr_h \in \Tr_h^{\gD_1}} \sP^{\pi}(\tr_h) \indict\lb \tr_h(s_h, a_h) = (s, a) \rb}$ and ${\sum_{\tr_h \notin \Tr_h^{\gD_1}} \sP^{\piE}(\tr_h) \indict\lb \tr_h(s_h, a_h) = (s, a) \rb}$ based on the dataset $\gD^\prime_{\mathrm{env}}$ and $\gD_1^c$, respectively. Therefore, $\widetilde{P}_h^{\piE} (s, a)$ is an unbiased estimation of $P_h^{\piE}(s, a)$.

\begin{lem} \label{lemma:unknown-transition-unbiased-estimation}
We define $\Pi_{\text{BC}} \lp \gD_1 \rp$ as the set of policies, each of which takes expert action on states contained in $\gD_{1}$. For each $\pi \in \Pi_{\text{BC}} \lp \gD_{1} \rp$, $\forall h \in [H]$ and $(s, a) \in \gS \times \gA$, we have
\begin{align*}
    &\sum_{\tr_h \in \Tr_h^{\gD_1}} \sP^{\piE}(\tr_h) \indict\lb \tr_h(s_h, a_h) = (s, a) \rb = \sum_{\tr_h \in \Tr_h^{\gD_1}} \sP^{\pi}(\tr_h) \indict\lb \tr_h(s_h, a_h) = (s, a) \rb.
\end{align*}
\end{lem}

\begin{proof}
Let $\Pi_{\text{BC}} \lp \gD_1 \rp$ denote the set of policies, each of which exactly takes expert action on states contained in $\gD_{1}$. Fix $\pi \in \Pi_{\text{BC}} \lp \gD_{1} \rp$, $h \in [H]$ and $(s, a) \in \gS \times \gA$, we consider the probability $\sP^{\piE} \lp \tr_h \rp$ of a truncated trajectory $\tr_h \in \Tr^{\gD_{1}}_h$. Since $\pi$ exactly takes expert action on states contained in $\gD_{1}$, we have
\begin{align*}
    &\quad \sP^{\piE}(\tr_h) \\
    &= \rho (\tr_h(s_1)) \piE_1 \lp \tr_h(a_1)| \tr(s_1) \rp \prod_{\ell=1}^{h-1}  P_{\ell} \lp \tr_h(s_{\ell+1}) | \tr_h(s_{\ell}), \tr_h(a_{\ell}) \rp \piE_{\ell+1} \lp \tr_h(a_{\ell+1}) | \tr_h(s_{\ell+1}) \rp
    \\
    &= \rho (\tr_h(s_1)) \pi_1 \lp \tr_h(a_1)| \tr(s_1) \rp \prod_{\ell=1}^{h-1}  P_{\ell} \lp \tr_h(s_{\ell+1}) | \tr_h(s_{\ell}), \tr_h(a_{\ell}) \rp \pi_{\ell+1} \lp \tr_h(a_{\ell+1}) | \tr_h(s_{\ell+1}) \rp
    \\
    &= \sP^{\pi}(\tr_h).
\end{align*}
Therefore, we obtain that
\begin{align*}
    \sum_{\tr_h \in \Tr_h^{\gD_1}} \sP^{\piE}(\tr_h) \indict\lb \tr_h(s_h, a_h) = (s, a) \rb = \sum_{\tr_h \in \Tr_h^{\gD_1}} \sP^{\pi}(\tr_h) \indict\lb \tr_h(s_h, a_h) = (s, a) \rb,
\end{align*}
which completes the proof.
\end{proof}

The sample complexity and interaction complexity of the estimator \eqref{eq:new_estimator_unknown_transition} are given in the following \cref{lemma:sample_complexity_of_new_estimator_unknown_transition}.

\begin{lem} \label{lemma:sample_complexity_of_new_estimator_unknown_transition}
Given expert dataset $\gD$ and $\gD$ is divided into two equal subsets, i.e., $\gD = \gD_{1} \cup \gD_{1}^c$ with $\labs \gD_1 \rabs = \labs \gD_1^{c} \rabs = m / 2$. Fix $\pi \in \Pi_{\text{BC}} \lp \gD_1 \rp$, let $\gD^\prime_{\mathrm{env}}$ be the dataset collected by $\pi$ and $|\gD^\prime_{\mathrm{env}} | = n^\prime$. Fix $\varepsilon \in (0, 1)$ and $\delta \in (0, 1)$; suppose $H \geq 5$. Consider the estimator $\widetilde{P}^{\piE}_h$ shown in \eqref{eq:new_estimator_unknown_transition}, if the number of expert trajectories ($m$) and the number of interaction trajectories in $\gD^\prime_{\mathrm{env}}$ for estimation ($n^\prime$) satisfy
\begin{align*}
    m \succsim   \frac{| \gS | H^{3/2}  }{\varepsilon} \log\lp  \frac{|\gS| H}{\delta} \rp, \; n^\prime \succsim \frac{ | \gS |H^{2}}{\varepsilon^2} \log\lp  \frac{|\gS| H}{\delta} \rp,
\end{align*}
then with probability at least $1-\delta$, we have
\begin{align*}
    \sum_{h=1}^H \lnorm \widetilde{P}^{\piE}_h - P^{\piE}_h  \rnorm_{1} \leq \varepsilon.
\end{align*}
\end{lem}

Refer to \cref{appendix:proof_lemma_sample_complexity_of_new_estimator_unknown_transition} for the proof. The proof is based on \cref{lemma:sample_complexity_of_new_estimator_known_transition} and \cref{lemma:unknown-transition-unbiased-estimation}.

Next, we state the theoretical guarantee of \textnormal{RF-Express} algorithm~\citep{menard20fast-active-learning}, which corresponds to assumption $(a)$ in Proposition \ref{prop:connection}.

\begin{thm}[Theorem 1 in ~\citep{menard20fast-active-learning}] \label{theorem:rf_express_sample_complexity}
Fix $\varepsilon \in \lp 0, 1 \rp$ and $\delta \in (0, 1)$. Consider the RF-Express algorithm and $\widehat{\gP}$ is the empirical transition function built on the collected trajectories, if the number of trajectories collected by RF-Express ($n$) satisfies 
\begin{align*}
    n \succsim  \frac{ |\gS| |\gA| H^{3} }{\varepsilon^2}    \lp |\gS| + \log\lp\frac{|\gS| H}{\delta} \rp \rp.
\end{align*}
Then with probability at least $1-\delta$, for any policy $\pi$ and any bounded reward function $w$ between $[-1, 1]$, we have\footnote{This is implied by the stopping rule in RF-Express algorithm and Lemma 1 in \citep{menard20fast-active-learning}.} $| V^{\pi, \gP, w} - V^{\pi, \widehat{\gP}, w} | \leq {\varepsilon}/{2}$; furthermore, for any bounded reward function $w$ between $[-1, 1]$, we have $ \max_{\pi \in \Pi} V^{\pi, w} \leq V^{\widehat{\pi}_{w}^{*}, w} + \varepsilon$, where $\widehat{\pi}_{w}^{*}$ is the optimal policy under empirical transition function $\widehat{\gP}$ and reward function $w$.
\end{thm}

In the following part, we formally prove \cref{theorem:sample-complexity-unknown-transition}. The proof combines \cref{prop:connection}, \cref{lemma:sample_complexity_of_new_estimator_unknown_transition}, and \cref{theorem:rf_express_sample_complexity}.

\begin{proof}[Proof of \cref{theorem:sample-complexity-unknown-transition}]
When the number of trajectories collected by \textnormal{RF-Express} satisfies
\begin{align*}
    n \succsim  \frac{ |\gS| |\gA| H^{3} }{\varepsilon^2}    \lp |\gS| + \log\lp\frac{|\gS| H}{\delta} \rp \rp,
\end{align*}
for any policy $\pi \in \Pi$ and reward function $w : \gS \times \gA \rar [0, 1]$, with probability at least $1-\delta/2$, $| V^{\pi, \gP, w} - V^{\pi, \widehat{\gP}, w} | \leq \varepsilon / 16 = \varepsilon_{\text{RFE}}$. In a word, the assumption $(a)$ in Proposition \ref{prop:connection} holds with $\delta_{\mathrm{RFE}} = \delta / 2$ and $\varepsilon_{\mathrm{RFE}} = \varepsilon / 16$.

Secondly, we note that the ku $(b)$ in Proposition \ref{prop:connection} holds by Lemma \ref{lemma:sample_complexity_of_new_estimator_unknown_transition}. More concretely, if the expert sample complexity and interaction complexity satisfies
\begin{align*}
    m \succsim   \frac{ | \gS | H^{3/2} }{\varepsilon} \log\lp  \frac{|\gS| H}{\delta} \rp, \; n^\prime \succsim \frac{ | \gS | H^{2}}{\varepsilon^2} \log\lp  \frac{|\gS| H}{\delta} \rp,
\end{align*}
with probability at least $1-\delta/2$, $\sum_{h=1}^H \Vert \widetilde{P}^{\piE}_h - P^{\piE}_h  \Vert_{1} \leq \varepsilon / 16 = \varepsilon_{\text{EST}}$. Hence, the assumption $(b)$ in Proposition \ref{prop:connection} holds with $\delta_{\mathrm{EST}} = \delta / 2$ and $\varepsilon_{\mathrm{EST}} = \varepsilon / 16$.

Thirdly, we aim to verify that the assumption $(c)$ in Proposition \ref{prop:connection} holds with $\widetilde{P}^{\piE}_h (s, a)$ and $\widehat{\gP}$. With the dual representation of $\ell_1$-norm and the minimax theorem, we get that
\begin{align*}
    \min_{\pi \in \Pi} \sum_{h=1}^H \lnorm P^{\pi, \widehat{\gP}}_h - \widetilde{P}^{\piE}_h \rnorm_{1} = - \min_{w \in \gW} \max_{\pi \in \Pi} \sum_{h=1}^H \sum_{(s, a) \in \gS \times \gA} w_h (s, a) \lp P^{\pi, \widehat{\gP}}_h (s, a) -  \widetilde{P}^{\piE}_h(s, a) \rp.
\end{align*}
Recall that $w^{(t)}$ is the reward function inferred by MB-TAIL in the iteration $t$. Then we have
\begin{align*}
    &\quad \min_{w \in \gW} \max_{\pi \in \Pi} \sum_{h=1}^H \sum_{(s, a) \in \gS \times \gA} w_h (s, a) \lp P^{\pi, \widehat{\gP}}_h (s, a) -  \widetilde{P}^{\piE}_h(s, a) \rp
    \\
    &\leq \max_{\pi \in \Pi} \sum_{h=1}^H \sum_{(s, a) \in \gS \times \gA} \lp \frac{1}{T} \sum_{t=1}^T w_h^{(t)} (s, a) \rp \lp P^{\pi, \widehat{\gP}}_h (s, a) -  \widetilde{P}^{\piE}_h(s, a) \rp
    \\
    &=  \max_{\pi \in \Pi} \frac{1}{T} \sum_{t=1}^T \sum_{h=1}^H \sum_{(s, a) \in \gS \times \gA}   w_h^{(t)} (s, a) \lp P^{\pi, \widehat{\gP}}_h (s, a) -  \widetilde{P}^{\piE}_h(s, a) \rp
    \\
    &\leq  \frac{1}{T} \sum_{t=1}^T \max_{\pi \in \Pi} \sum_{h=1}^H \sum_{(s, a) \in \gS \times \gA}   w_h^{(t)} (s, a) \lp P^{\pi, \widehat{\gP}}_h (s, a) -  \widetilde{P}^{\piE}_h(s, a) \rp
    \\
    &\leq \frac{1}{T} \sum_{t=1}^T \sum_{h=1}^H \sum_{(s, a) \in \gS \times \gA}   w_h^{(t)} (s, a) \lp P^{\pi^{(t)}, \widehat{\gP}}_h (s, a) -  \widetilde{P}^{\piE}_h(s, a) \rp + \varepsilon_{\text{opt}}.
\end{align*}
In the last inequality, the policy $\pi^{(t)}$ is the nearly optimal policy w.r.t $w^{(t)}$ and $\widehat{\gP}$ up to an error of $\varepsilon_{\text{opt}}$. Then we have that
\begin{align*}
    &\quad \min_{\pi \in \Pi} \sum_{h=1}^H \lnorm P^{\pi, \widehat{\gP}}_h - \widetilde{P}^{\piE}_h \rnorm_{1}
    \\
    &\geq -\frac{1}{T} \sum_{t=1}^T \sum_{h=1}^H \sum_{(s, a) \in \gS \times \gA}   w_h^{(t)} (s, a) \lp P^{\pi^{(t)}, \widehat{\gP}}_h (s, a) -  \widetilde{P}^{\piE}_h(s, a) \rp - \varepsilon_{\text{opt}}
    \\
    &\geq \frac{1}{T} \max_{w \in \gW} \sum_{t=1}^T \sum_{h=1}^H \sum_{(s, a) \in \gS \times \gA}   w_h (s, a) \lp P^{\pi^{(t)}, \widehat{\gP}}_h (s, a) -  \widetilde{P}^{\piE}_h(s, a) \rp - \varepsilon_{\text{opt}} - 2H \sqrt{\frac{2 \vert \gS \vert \vert \gA \vert}{T}}. 
\end{align*}
Note that the reward function $w^{(t)}$ is updated by online projected gradient descent with objective function $f^{(t)} (w) = \sum_{h=1}^H \sum_{(s, a) \in \gS \times \gA}   w_h (s, a) ( P^{\pi^{(t)}, \widehat{\gP}}_h (s, a) -  \widetilde{P}^{\piE}_h(s, a) )$. Hence, the last inequality follows Lemma \ref{lemma:regret_of_ogd}. Then we have that
\begin{align*}
    &\quad \min_{\pi \in \Pi} \sum_{h=1}^H \lnorm P^{\pi, \widehat{\gP}}_h - \widetilde{P}^{\piE}_h \rnorm_{1}
    \\
    &\geq \frac{1}{T} \max_{w \in \gW} \sum_{t=1}^T \sum_{h=1}^H \sum_{(s, a) \in \gS \times \gA}   w_h (s, a) \lp P^{\pi^{(t)}, \widehat{\gP}}_h (s, a) -  \widetilde{P}^{\piE}_h(s, a) \rp - \varepsilon_{\text{opt}} - 2H \sqrt{\frac{2 \vert \gS \vert \vert \gA \vert}{T}}
    \\
    &=  \max_{w \in \gW}  \sum_{h=1}^H \sum_{(s, a) \in \gS \times \gA}   w_h (s, a) \lp \frac{1}{T} \sum_{t=1}^T P^{\pi^{(t)}, \widehat{\gP}}_h (s, a) -  \widetilde{P}^{\piE}_h(s, a) \rp - \varepsilon_{\text{opt}} - 2H \sqrt{\frac{2 \vert \gS \vert \vert \gA \vert}{T}}
    \\
    &= \max_{w \in \gW}  \sum_{h=1}^H \sum_{(s, a) \in \gS \times \gA}   w_h (s, a) \lp P^{\widebar{\pi}, \widehat{\gP}}_h (s, a) -  \widetilde{P}^{\piE}_h(s, a) \rp - \varepsilon_{\text{opt}} - 2H \sqrt{\frac{2 \vert \gS \vert \vert \gA \vert}{T}}
    \\
    &= \sum_{h=1}^H \lnorm P^{\widebar{\pi}, \widehat{\gP}}_h - \widetilde{P}^{\piE}_h \rnorm_{1} - \varepsilon_{\text{opt}} - 2H \sqrt{\frac{2 \vert \gS \vert \vert \gA \vert}{T}}.
\end{align*}
When $\varepsilon_{\text{opt}} \leq \varepsilon / 2$ and $T \succsim |\gS| |\gA| H^2 / \varepsilon^2$ such that $2 H \sqrt{2 |\gS| |\gA| / T} \leq \varepsilon / 4$, we have that
\begin{align*}
    \sum_{h=1}^H \lnorm P^{\widebar{\pi}, \widehat{\gP}}_h - \widetilde{P}^{\piE}_h \rnorm_{1} - \min_{\pi \in \Pi} \sum_{h=1}^H \lnorm P^{\pi, \widehat{\gP}}_h - \widetilde{P}^{\piE}_h \rnorm_{1} \leq \frac{3\varepsilon}{4}  = \varepsilon_{\mathrm{AIL}}.
\end{align*}
Therefore, the assumption $(c)$ in Proposition \ref{prop:connection} holds with $\varepsilon_{\mathrm{AIL}} = 3\varepsilon / 4H$. Now, we summarize the conditions what we have obtained.
\begin{itemize}
    \item The assumption $(a)$ in Proposition \ref{prop:connection} holds with $\delta_{\mathrm{RFE}} = \delta / 2$ and $\varepsilon_{\mathrm{RFE}} = \varepsilon / 16$.
    \item The assumption $(b)$ in Proposition \ref{prop:connection} holds with $\delta_{\mathrm{EST}} = \delta / 2$ and $\varepsilon_{\mathrm{EST}} = \varepsilon / 16$.
    \item The assumption $(c)$ in Proposition \ref{prop:connection} holds with $\varepsilon_{\mathrm{AIL}} = 3\varepsilon / 4$. 
\end{itemize}
Applying Proposition \ref{prop:connection} finishes the proof. With probability at least $1-\delta$,
\begin{align*}
    V^{\piE} - V^{\widebar{\pi}} \leq 2 \varepsilon_{\mathrm{RFE}} + 2 \varepsilon_{\mathrm{EST}} + \varepsilon_{\mathrm{AIL}} = \varepsilon.  
\end{align*}

\end{proof}

\section{Discussion}
\label{appendix:discussion}

In this section, we discuss some theoretical results in the main paper. 

\subsection{Non-convexity of VAIL}
\label{appendix:vail_objective_is_non_convex}

In this part, we give an example to show that \textsf{VAIL}'s objective in \eqref{eq:ail} is non-convex. Our construction is based on the example in \citep{agarwal2020pg}, in which the authors showed that policy optimization for infinite-horizon tabular MDPs is a non-convex problem.

\begin{claim}   \label{claim:vail_non_convex}
For tabular and episodic MDPs, there exists an instance such that the objective of VAIL in \eqref{eq:ail} is non-convex. 
\end{claim}

\begin{figure}[htbp]
    \centering
    \includegraphics[width=0.6\linewidth]{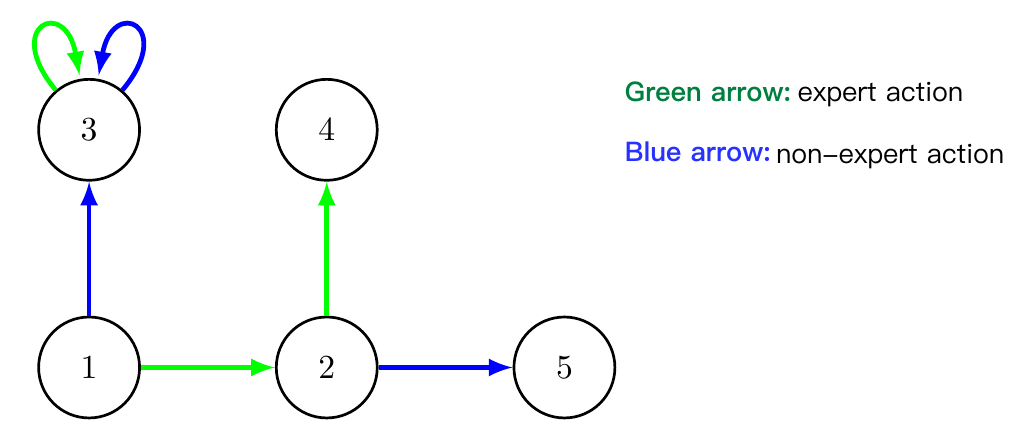}
    \caption{A simple example to show that \textsf{VAIL}'s objective in \eqref{eq:ail} is non-convex.}
    \label{fig:non_convex}
\end{figure}

\begin{proof}[Proof of \cref{claim:vail_non_convex}]
Our construction is shown in \cref{fig:non_convex}. In particular, there are 5 states $(s^{1}, s^{2}, s^{3}, s^{4}, s^{5})$ and two actions $(a^{1}, a^{2})$. Each arrow shows a deterministic transition. The initial state is $s^1$ and the planning horizon is $2$. Assume  $a^{1}$ is the expert action and there is only one expert trajectory: $(s^{1}, a^{1}) \rar (s^{2}, a^{1})$. We can calculate the empirical distribution:
\begin{align*}
    \widehat{P}^{\piE}_{1}(s^{1}, a^{1}) = 1.0, \quad  \widehat{P}^{\piE}_{2}(s^{2}, a^{1}) = 1.0
\end{align*}
Let us use the following notations: $x := \pi_1(a^{1} | s^{1})$ and $y := \pi_2(a^{1} | s^{2})$. In time step $h=1$, we have 
\begin{align*}
    \ell_1 &= \sum_{(s, a)} \labs P^{\pi}_1(s, a) - \widehat{P}^{\piE}_1(s, a) \rabs \\
    &= \labs P^{\pi}_{1}(s^{1}, a^{1}) - \widehat{P}^{\piE}_{1} (s^{1}, a^{1}) \rabs + \labs P^{\pi}_{1} (s^{1}, a^{2}) - \widehat{P}^{\piE}_{1}(s^{1}, a^{2}) \rabs \\
    &= \labs x - 1 \rabs + \labs 1-x - 0 \rabs = 2(1 - x).
\end{align*}
In time step $h=2$, we have 
\begin{align*}
    \ell_2 &= \sum_{(s, a)} \labs P^{\pi}_2(s, a) - \widehat{P}^{\piE}_2(s, a) \rabs \\
    &= \labs P^{\pi}_{2}(s^{2}, a^{1}) - \widehat{P}^{\piE}_2(s^{2}, a^{1}) \rabs +  \labs P^{\pi}_{2}(s^{2}, a^{2}) - \widehat{P}^{\piE}_2(s^{2}, a^{2}) \rabs \\
    &\quad + \labs P^{\pi}_{2}(s^{3}, a^{1}) - \widehat{P}^{\piE}_2(s^{3}, a^{1}) \rabs + \labs P^{\pi}_{2}(s^{3}, a^{2}) - \widehat{P}^{\piE}_2(s^{3}, a^{2}) \rabs \\
    &= \labs xy - 1 \rabs + \labs x(1-y) - 0 \rabs + (1-x)  \\
    &= 1 - xy + x - xy + 1 - x = 2(1 - xy).
\end{align*}
Thus, we have that 
\begin{align*}
    f(x, y) = \ell_1 + \ell_2 = 2 \lp 2 - x - xy \rp.
\end{align*}
Furthermore, we can compute that 
\begin{align*}
    \nabla f(x, y) = \bvec{-2 - 2y \\ -2x}, \\
    \nabla^2 f(x, y) = \bvec{0 & -2 \\ -2 & 0}.
\end{align*}
Since $\nabla^2 f(x, y)$ is not a PSD, we claim that $f(x, y)$ is non-convex w.r.t. $(x, y)$.

\end{proof}

\subsection{VAIL with Subsampled Trajectories}
\label{appendix:further_discssuion_of_example_success_subsampling}

In this part, we discuss the generalization of \textsf{VAIL} with \emph{subsampled} expert trajectories. Before discussion, we point out that VAIL cannot recover the expert policy with any subsampled trajectories even on Reset Cliff. What really matters for VAIL is the state-action pairs with large time steps. If these state-action pairs are masked, we cannot expect VAIL to generalize well. This claim can be easily verified on Gym MuJoCo locomotion tasks. Hence, we mainly focus on the following example to illustrate the key idea under the subsampling case. 

\begin{example}  \label{example:subsample_trajectory}
Consider an example similar to \cref{example:ail_success}. We consider the same state space, action space, initial state distribution and transition function as in \cref{example:ail_success}. Different from \cref{example:ail_success}, we consider the horizon length $H=3$. The agent is provided only 2 expert trajectories $\tr_1 = (s^{1}, a^{1}) \rar (s^{1}, a^{1}) \rar (s^{1}, a^{1}) $ and $\tr_2 = (s^{1}, a^{1}) \rar (s^{2}, a^{1}) \rar (s^{2}, a^{1})$. We subsample expert trajectories so that \emph{the data in the time step $h=1$ is masked (i.e., missing)}. This operation is similar to the subsampling procedure in \citep{ho2016gail, Kostrikov19dac}. 
\end{example}

\begin{claim} \label{claim:ail_reset_cliff_subsample}
Consider the MDP and expert demonstrations in \cref{example:subsample_trajectory}. Suppose that $\piail$ is the optimal solution of \eqref{eq:ail}, then for each time step $h \in [3]$, $\piail_{h} (a^{1}|s) = \piE_h (a^{1}|s) = 1, \forall s \in \{s^{1}, s^{2} \}$.  
\end{claim}

\begin{rem}
\cref{claim:ail_reset_cliff_subsample} indicates that if the expert trajectories are properly subsampled, VAIL can still recover the expert policy. We remark that this conclusion does not hold for the case where the horizon is 2 and we mask the state-action pair in the first time step. This is because the objective in the second time step is too weak to provide effective guidance for the policy optimization in the first time step. Instead, the conclusion holds for the case where the horizon is 3 considered in \cref{example:subsample_trajectory}. This implies that to retain good performance with subsampled trajectories, we must post constraints on the subsampling rate and subsampling interval.
\end{rem}

Before proving \cref{claim:ail_reset_cliff_subsample}, we first prove a useful claim which is similar to \cref{lem:condition_for_ail_optimal_solution}.

\begin{claim}
\label{claim:condition_for_ail_optimal_solution}
Consider the MDP and expert demonstrations in \cref{example:subsample_trajectory}. Suppose that $\piail$ is the optimal solution of \eqref{eq:ail}, then for each time step $h \in [3]$, $\exists s \in \goodS, \piail_h (a^{1}|s) > 0$. 
\end{claim}

\begin{proof}
The empirical state-action distribution with subsampled expert demonstrations are formulated as follows. Note that the data in time step $h=1$ is masked and thus the corresponding empirical distribution is uniform distribution.
\begin{align*}
    \widehat{P}^{\piE}_{1}(s^{1}, a^{1}) = \frac{1}{6}, \widehat{P}^{\piE}_{1}(s^{2}, a^{1}) = \frac{1}{6}, \widehat{P}^{\piE}_{1}(s^{3}, a^{1}) = \frac{1}{6}, \\
    \widehat{P}^{\piE}_{1}(s^{1}, a^{2}) = \frac{1}{6}, \widehat{P}^{\piE}_{1}(s^{2}, a^{2}) = \frac{1}{6}, \widehat{P}^{\piE}_{1}(s^{3}, a^{2}) = \frac{1}{6}, \\
    \widehat{P}^{\piE}_{2}(s^{1}, a^{1}) = \RED{0.5}, \widehat{P}^{\piE}_{2}(s^{2}, a^{1}) = \RED{0.5}, \widehat{P}^{\piE}_{2}(s^{3}, a^{1}) = 0.0, \\
    \widehat{P}^{\piE}_{2}(s^{1}, a^{2}) = 0.0, \widehat{P}^{\piE}_{2}(s^{2}, a^{2}) = 0.0, \widehat{P}^{\piE}_{2}(s^{3}, a^{2}) = 0.0, \\
    \widehat{P}^{\piE}_{3}(s^{1}, a^{1}) = \RED{0.5}, \widehat{P}^{\piE}_{3}(s^{2}, a^{1}) = \RED{0.5}, \widehat{P}^{\piE}_{3}(s^{3}, a^{1}) = 0.0, \\
    \widehat{P}^{\piE}_{3}(s^{1}, a^{2}) = 0.0, \widehat{P}^{\piE}_{3}(s^{2}, a^{2}) = 0.0, \widehat{P}^{\piE}_{3}(s^{3}, a^{2}) = 0.0.
\end{align*}
We first prove that for the first time step, $\exists s \in \goodS, \piail_1 (a^{1}|s) > 0$. The proof is based on contradiction.

We assume that the original statement is false and $\forall s \in \goodS$, $\piail_1 (a^{1}|s) = 0$. We construct another policy $\widetilde{\pi}^{\ail}$. In time steps $h=1$ and $h=2$, $\forall s \in \goodS, \widetilde{\pi}^{\ail}_h (a^{1}|s) = 1$. We compare \textsf{VAIL}'s objectives under $\piail$ and $\widetilde{\pi}^{\ail}$. We first consider $\piail$. It is easy to compute the state-action distribution induced by $\piail$.
\begin{align*}
    &P^{\piail}_{1}(s^{1}, a^{1}) = 0.0, P^{\piail}_{1}(s^{2}, a^{1}) = 0.0, P^{\piail}_{1}(s^{3}) = 0.0, \\
    &P^{\piail}_{1}(s^{1}, a^{2}) = 0.5, P^{\piail}_{1}(s^{2}, a^{2}) = 0.5, \\
    &P^{\piail}_{2}(s^{1}) = 0.0, P^{\piail}_{2}(s^{2}) = 0.0, P^{\piail}_{2}(s^{3}) = 1.0, \\
    &P^{\piail}_{3}(s^{1}) = 0.0, P^{\piail}_{3}(s^{2}) = 0.0, P^{\piail}_{3}(s^{3}) = 1.0.
\end{align*}
Recall the definition of the single-stage loss function in time step $h$.
\begin{align*}
    \text{Loss}_h (\pi) = \sum_{(s, a) \in \gS \times \gA} \labs P^{\pi}_h(s, a)  - \widehat{P}^{\piE}_h(s, a) \rabs .
\end{align*}
It is direct to compute that $\text{Loss}_1 (\piail) = 4/3, \text{Loss}_2 (\piail) = \text{Loss}_3 (\piail) = 2$.

Second, we consider \textsf{VAIL}'s objective of $\widetilde{\pi}^{\ail}$.
\begin{align*}
     &P^{\widetilde{\pi}^{\ail}}_{1}(s^{1}, a^{1}) = 0.5, P^{\widetilde{\pi}^{\ail}}_{1}(s^{2}, a^{1}) = 0.5, P^{\widetilde{\pi}^{\ail}}_{1}(s^{3}) = 0.0, \\
    &P^{\widetilde{\pi}^{\ail}}_{1}(s^{1}, a^{2}) = 0.0, P^{\widetilde{\pi}^{\ail}}_{1}(s^{2}, a^{2}) = 0.0.
\end{align*}
Similarly, we have $\text{Loss}_1 (\widetilde{\pi}^{\ail}) = 4/3$. We proceed to consider time step $h=2$. By \dquote{transition flow equation}, it holds that
\begin{align*}
    &P^{\widetilde{\pi}^{\ail}}_{2}(s^{1}) = P^{\widetilde{\pi}^{\ail}}_{1}(s^{1}, a^{1}) P_1 (s^{1}|s^{1}, a^{1}) +  P^{\widetilde{\pi}^{\ail}}_{1}(s^{2}, a^{1}) P_1 (s^{1}|s^{2}, a^{1}) = 0.5,
    \\
    &P^{\widetilde{\pi}^{\ail}}_{2}(s^{2}) = P^{\widetilde{\pi}^{\ail}}_{1}(s^{1}, a^{1}) P_1 (s^{2}|s^{1}, a^{1}) +  P^{\widetilde{\pi}^{\ail}}_{1}(s^{2}, a^{1}) P_1 (s^{2}|s^{2}, a^{1}) = 0.5,
    \\
    & P^{\widetilde{\pi}^{\ail}}_{2}(s^{3}) = 0. 
\end{align*}
Then we can calculate that
\begin{align*}
    \text{Loss}_2 (\widetilde{\pi}^{\ail}) &= \labs 0.5 - P^{\widetilde{\pi}^{\ail}}_{2}(s^{1}) \widetilde{\pi}^{\ail}_2( a^{1}|s^{1})  \rabs + P^{\widetilde{\pi}^{\ail}}_{2}(s^{1})  \widetilde{\pi}^{\ail}_2 (a^{2}|s^{1})
    \\
    &\quad + \labs 0.5 - P^{\widetilde{\pi}^{\ail}}_{2}(s^{2}) \widetilde{\pi}^{\ail}_{2} (a^{1}|s^{2})  \rabs + P^{\widetilde{\pi}^{\ail}}_{2}(s^{2})  \widetilde{\pi}^{\ail}_{2} (a^{2}|s^{2})
    \\
    &= \labs 0.5 - 0.5 \widetilde{\pi}^{\ail}_2( a^{1}|s^{1})  \rabs + 0.5 - 0.5  \widetilde{\pi}^{\ail}_2 (a^{1}|s^{1})
    \\
    &\quad + \labs 0.5 - 0.5 \widetilde{\pi}^{\ail}_{2} (a^{1}|s^{2})  \rabs + 0.5 - 0.5 \widetilde{\pi}^{\ail}_{2} (a^{1}|s^{2})
    \\
    &= 2 - \widetilde{\pi}^{\ail}_2 (a^{1}|s^{1}) - \widetilde{\pi}^{\ail}_{2} (a^{1}|s^{2}) = 0.  
\end{align*}
Therefore, we have that $\text{Loss}_2 (\widetilde{\pi}^{\ail}) < 2 = \text{Loss}_2 (\pi^{\ail})$. For time step $h=3$, note that $2$ is the maximal value of the single-stage loss function and $\text{Loss}_3 (\widetilde{\pi}^{\ail}) \leq 2 = \text{Loss}_3 (\pi^{\ail})$. In a word, we construct policy $\widetilde{\pi}^{\ail}$ whose \textsf{VAIL}'s objective is strictly smaller than that of $\piail$. This contradicts with the fact that $\piail$ is the optimal solution of \eqref{eq:ail} and thus the original statement is true. That is, $\exists s \in \goodS$, $\piail_1 (a^{1}|s) > 0$.

We continue to consider time steps $h=2$. With \cref{lem:n_vars_opt_greedy_structure}, fixing $\piail_1$, $\piail_2$ and $\piail_3$ is also optimal solution w.r.t \textsf{VAIL}'s objective. Formally,
\begin{align}
    (\piail_2, \piail_3) &\in \argmin_{\pi_2, \pi_3} \text{Loss}_1 (\piail_1) + \text{Loss}_2 (\piail_1, \pi_2) + \text{Loss}_3 (\piail_1, \pi_2, \pi_3) \nonumber
    \\
    &= \argmin_{\pi_2, \pi_3} \text{Loss}_2 (\piail_1, \pi_2) + \text{Loss}_3 (\piail_1, \pi_2, \pi_3). \label{eq:piail_2_and_piail_3_are_optimal}  
\end{align}
The proof is also based on contradiction. We assume that $\forall s \in \goodS$, $\piail_2 (a^{1}|s) = 0$. We construct another policy $(\widetilde{\pi}^{\ail}_2, \piail_3)$: $\forall s \in \goodS$, $\widetilde{\pi}^{\ail}_2 (a^{1}|s) = 1$. On the one hand,
\begin{align*}
    \text{Loss}_2 (\piail_1, \piail_2) = 2, \text{Loss}_3 (\piail_1, \piail_2, \piail_3) = 2.
\end{align*}
On the other hand,
\begin{align*}
    \text{Loss}_2 (\piail_1, \widetilde{\pi}^{\ail}_2) &= \labs 0.5 - P^{\piail}_{2}(s^{1}) \widetilde{\pi}^{\ail}_2( a^{1}|s^{1})  \rabs + P^{\piail}_{2}(s^{1})  \widetilde{\pi}^{\ail}_2 (a^{2}|s^{1})
    \\
    &+ \labs 0.5 - P^{\piail}_{2}(s^{2}) \widetilde{\pi}^{\ail}_{2} (a^{1}|s^{2})  \rabs + P^{\piail}_{2}(s^{2})  \widetilde{\pi}^{\ail}_{2} (a^{2}|s^{2}) + P^{\piail}_{2}(s^{3})
    \\
    &= \labs 0.5 - P^{\piail}_{2}(s^{1}) \rabs + \labs 0.5 - P^{\piail}_{2}(s^{2}) \rabs + P^{\piail}_{2}(s^{3})
    \\
    &= 2 - 2 \lp P^{\piail}_{2}(s^{1}) + P^{\piail}_{2}(s^{2})  \rp.
\end{align*}
We have proved for time step $h=1$, $\exists s \in \goodS$, $\piail_1 (a^{1}|s) > 0$. Therefore, it holds that $P^{\piail}_{2}(s^{1}) > 0, P^{\piail}_{2}(s^{2}) > 0$ and $\text{Loss}_2 (\piail_1, \widetilde{\pi}^{\ail}_2) < \text{Loss}_2 (\piail_1, \piail_2)$. Besides, it is obvious that $\text{Loss}_3 (\piail_1, \widetilde{\pi}^{\ail}_2, \piail_3) \leq \text{Loss}_3 (\piail_1, \piail_2, \piail_3) = 2$. In a word, we construct another policy $(\widetilde{\pi}^{\ail}_2, \piail_3)$ such that
\begin{align*}
    \text{Loss}_2 (\piail_1, \widetilde{\pi}^{\ail}_2) + \text{Loss}_3 (\piail_1, \widetilde{\pi}^{\ail}_2, \piail_3) < \text{Loss}_2 (\piail_1, \piail_2) + \text{Loss}_3 (\piail_1, \piail_2, \piail_3),
\end{align*}
which contradicts with the fact in \eqref{eq:piail_2_and_piail_3_are_optimal}. Hence, the original statement is true and $\exists s \in \goodS$, $\piail_2 (a^{1}|s) > 0$.

Finally, we consider the last time step. Similarly, we have that
\begin{align*}
    \piail_3 &\in \argmin_{\pi_3} \text{Loss}_3 (\piail_1, \piail_2, \pi_3)
    \\
    &= \argmin_{\pi_3} \labs 0.5 - P^{\piail}_{3}(s^{1}) \pi_3 ( a^{1}|s^{1})  \rabs + P^{\piail}_{3}(s^{1})  \pi_3 (a^{2}|s^{1})
    \\
    &\quad + \labs 0.5 - P^{\piail}_{3}(s^{2}) \pi_{3} (a^{1}|s^{2})  \rabs + P^{\piail}_{3}(s^{2})  \pi_3 (a^{2}|s^{2}) + P^{\piail}_{3}(s^{3})
    \\
    &=\argmin_{\pi_3} \labs 0.5 - P^{\piail}_{3}(s^{1}) \pi_3 ( a^{1}|s^{1})  \rabs - P^{\piail}_{3}(s^{1})  \pi_3 (a^{1}|s^{1})
    \\
    &\quad + \labs 0.5 - P^{\piail}_{3}(s^{2}) \pi_{3} (a^{1}|s^{2})  \rabs - P^{\piail}_{3}(s^{2})  \pi_3 (a^{1}|s^{2}).
\end{align*}
We have proved that for time steps $h=1$ and $h=2$, $\exists s \in \goodS$, $\piail_h (a^{1}|s) > 0$. Hence $P^{\piail}_{3}(s^{1}) > 0$ and $P^{\piail}_{3}(s^{2}) > 0$. With \cref{lem:single_variable_opt_condition}, we have that $\piail_3 (a^{1}|s^1) > 0$ and $\piail_3 (a^{1}|s^2) > 0$. We finish the whole proof. 
\end{proof}

\begin{proof}[Proof of \cref{claim:ail_reset_cliff_subsample}]
We first compute the empirical state-action distribution with subsampled expert demonstrations. Note that the data in time step $h=1$ is masked and thus the corresponding empirical distribution is uniform distribution.
\begin{align*}
    \widehat{P}^{\piE}_{1}(s^{1}, a^{1}) = \frac{1}{6}, \widehat{P}^{\piE}_{1}(s^{2}, a^{1}) = \frac{1}{6}, \widehat{P}^{\piE}_{1}(s^{3}, a^{1}) = \frac{1}{6}, \\
    \widehat{P}^{\piE}_{1}(s^{1}, a^{2}) = \frac{1}{6}, \widehat{P}^{\piE}_{1}(s^{2}, a^{2}) = \frac{1}{6}, \widehat{P}^{\piE}_{1}(s^{3}, a^{2}) = \frac{1}{6}, \\
    \widehat{P}^{\piE}_{2}(s^{1}, a^{1}) = \RED{0.5}, \widehat{P}^{\piE}_{2}(s^{2}, a^{1}) = \RED{0.5}, \widehat{P}^{\piE}_{2}(s^{3}, a^{1}) = 0.0, \\
    \widehat{P}^{\piE}_{2}(s^{1}, a^{2}) = 0.0, \widehat{P}^{\piE}_{2}(s^{2}, a^{2}) = 0.0, \widehat{P}^{\piE}_{2}(s^{3}, a^{2}) = 0.0, \\
    \widehat{P}^{\piE}_{3}(s^{1}, a^{1}) = \RED{0.5}, \widehat{P}^{\piE}_{3}(s^{2}, a^{1}) = \RED{0.5}, \widehat{P}^{\piE}_{3}(s^{3}, a^{1}) = 0.0, \\
    \widehat{P}^{\piE}_{3}(s^{1}, a^{2}) = 0.0, \widehat{P}^{\piE}_{3}(s^{2}, a^{2}) = 0.0, \widehat{P}^{\piE}_{3}(s^{3}, a^{2}) = 0.0.
\end{align*}

Recall the definition of the single-stage loss function $\text{Loss}_h (\pi)$ and the \dquote{cost-to-go} function $\ell_h (\pi)$ in time step $h$ 
\begin{align*}
    &\text{Loss}_h (\pi) = \sum_{(s, a) \in \gS \times \gA} \labs P^{\pi}_h(s, a)  - \widehat{P}^{\piE}_h(s, a) \rabs,
    \\
    &\ell_h (\pi) = \sum_{t=h}^{H} \text{Loss}_t (\pi)  = \sum_{t=h}^{H} \sum_{(s, a) \in \gS \times \gA} \labs P^{\pi}_t(s, a)  - \widehat{P}^{\piE}_t(s, a) \rabs .
\end{align*}

We perform a similar analysis to that in \cref{example:ail_success}. As $\piail = (\piail_1, \piail_2, \piail_3)$ is the optimal solution of \eqref{eq:ail}, with \cref{lem:n_vars_opt_greedy_structure}, fixing $(\piail_1, \piail_2)$, $\piail_3$ is optimal w.r.t to VAIL's objective. Notice that $P^{\piail}_1$ and $P^{\piail}_2$ are independent of $\piail_3$, so we have that
\begin{align*}
    \piail_3 &\in \argmin_{\pi_3} \text{Loss}_3 (\pi_3)
    \\
    &= \argmin_{\pi_3} \sum_{(s, a) \in \gS \times \gA} \labs P^{\pi}_3(s, a)  - \widehat{P}^{\piE}_3(s, a) \rabs  
    \\
    &= \argmin_{\pi_3}  \labs P^{\pi}_3 (s^{1}) \pi_3 (a^{1} | s^{1}) - 0.5 \rabs +  \labs P^{\pi}_3(s^{2}) \pi_3(a^{1}|s^{2}) - 0.5 \rabs + P^{\pi}_3(s^{3})  \\
    &\quad +  P^{\pi}_3(s^{1}) (1 - \pi_3(a^{1} | s^{1}))  + P^{\pi}_3(s^{2}) (1 - \pi_3(a^{1} | s^{2}))
\end{align*}

With a slight abuse of notation, we use $P^{\pi}_3$ denote the distribution induce by $(\piail_1, \piail_2, \pi_3)$ for any optimization variable $\pi_3$. Note that $\pi_3$ is the optimization variable for $\text{Loss}_3 (\pi_3)$ while $P_3^{\pi}(s^{1}) = P_3^{\piail}(s^{1})$, and $P_3^{\pi}(s^{2}) = P_3^{\piail}(s^{2}), P_3^{\pi}(s^{3}) = P_3^{\piail}(s^{3})$ are independent of $\pi_3$. We obtain
\begin{align*}
    \piail_3 &\in \argmin_{\pi_3} \labs P_3^{\piail}(s^{1}) \pi_3(a^{1} | s^{1}) - 0.5 \rabs - P^{\piail}_3(s^{1}) \pi_3(a^{1} | s^{1})  +  \labs P_3^{\piail}(s^{2}) \pi_3(a^{1}|s^{2}) - 0.5 \rabs \\
    &\quad - P^{\piail}_3(s^{2})  \pi_3(a^{1} | s^{2}).
\end{align*}
We only have two free optimization variables: $\pi_3(a^{1}|s^{1})$ and $\pi_3(a^{1}|s^{2})$ and they are independent. Then we obtain
\begin{align*}
    & \piail_3 (a^1|s^1) \in \argmin_{\pi_3 (a^1|s^1) \in [0, 1]} \labs P_3^{\piail}(s^{1}) \pi_3(a^{1} | s^{1}) - 0.5 \rabs - P^{\piail}_3(s^{1}) \pi_3(a^{1} | s^{1}),
    \\
    &\piail_3 (a^1|s^2) \in \argmin_{\pi_3 (a^1|s^2) \in [0, 1]} \labs P_3^{\piail}(s^{2}) \pi_3(a^{1}|s^{2}) - 0.5 \rabs - P^{\piail}_3(s^{2})  \pi_3(a^{1} | s^{2}).
\end{align*}
With \cref{claim:condition_for_ail_optimal_solution} and $\rho (s_1) = \rho (s_2) > 0$, it holds that $P_3^{\piail}(s^{1}) > 0$ and $P_3^{\piail}(s^{2}) > 0$. With \cref{lem:mn_variables_opt_unique}, we have that $\piail_3 (a^1|s^1) = 1$ and $\piail_3 (a^1|s^2) = 1$ are the unique optimal solutions of the above two problems, respectively. This finishes the proof in time step $h=3$.

Then we consider the policy optimization in time step $h=2$. With \cref{lem:single_variable_opt_condition}, we have that fixing $(\piail_1, \piail_3)$, $\piail_2$ is optimal w.r.t \textsf{VAIL}'s objective. Note that \textsf{VAIL}'s objective in time step $h=1$ is fixed, so we have 
\begin{align*}
    \piail_2 \in \argmin_{\pi_2} \ell_1 (\pi_2) = \argmin_{\pi_2} \text{Loss}_1 (\pi_2) + \text{Loss}_2 (\pi_2) +  \text{Loss}_3 (\pi_2) = \argmin_{\pi_2} \text{Loss}_2 (\pi_2) +  \text{Loss}_3 (\pi_2). 
\end{align*}
We have proved that $\piail_3 (a^1|s^1) = \piail_3 (a^1|s^2) = 1$ and plug it into $\text{Loss}_3 (\pi_2)$.
\begin{align*}
    \text{Loss}_3 (\pi_2) &= \labs P^{\pi}_3(s^{1}) - 0.5 \rabs + \labs P^{\pi}_3(s^{2})  - 0.5 \rabs + P^{\pi}_3(s^{3}) = 2.0 - \pi_2 (a^{1} | s^{1}) - \pi_2 (a^{1} | s^{2}),
\end{align*}
which has a unique globally optimal solution at $\pi_2 (a^{1} | s^{1}) = 1.0$ and $\pi_2 (a^{1} | s^{2}) = 1.0$. For $\text{Loss}_2 (\pi_2)$, we have
\begin{align*}
    \piail_2 &\in \argmin_{\pi_2} \text{Loss}_2 (\pi_2)
    \\
    &= \argmin_{\pi_2} \sum_{(s, a) \in \gS \times \gA} \labs P^{\pi}_2(s, a)  - \widehat{P}^{\piE}_2(s, a) \rabs  
    \\
    &= \argmin_{\pi_2}  \labs P^{\pi}_2 (s^{1}) \pi_2 (a^{1} | s^{1}) - 0.5 \rabs +  \labs P^{\pi}_2(s^{2}) \pi_2(a^{1}|s^{2}) - 0.5 \rabs + P^{\pi}_2(s^{3})  \\
    &\quad +  P^{\pi}_2(s^{1}) (1 - \pi_2(a^{1} | s^{1}))  + P^{\pi}_2(s^{2}) (1 - \pi_2(a^{1} | s^{2})).
\end{align*}
Here we use $P^{\pi}_2(s, a)$ and $P^{\pi}_2(s)$ to denote the distributions induced by $(\piail_1, \pi_2)$. Note that $\pi_2$ is the optimization variable for $\text{Loss}_2 (\pi_2)$ while $P_2^{\pi}(s^{1}) = P_2^{\piail}(s^{1}), P_2^{\pi}(s^{2}) = P_2^{\piail}(s^{2})$, and $P_2^{\pi}(s^{3}) = P_2^{\piail}(s^{3})$ are independent of $\pi_2$. Then we have that
\begin{align*}
    \piail_2 &\in \argmin_{\pi_2} \labs P_2^{\piail}(s^{1}) \pi_2(a^{1} | s^{1}) - 0.5 \rabs - P^{\piail}_2(s^{1}) \pi_2(a^{1} | s^{1})  +  \labs P_2^{\piail}(s^{2}) \pi_2(a^{1}|s^{2}) - 0.5 \rabs \\
    &\quad - P^{\piail}_2(s^{2})  \pi_2(a^{1} | s^{2}).
\end{align*}
With \cref{lem:single_variable_opt}, we have that $\piail_2 (a^1|s^1) = 1, \piail_2 (a^1|s^2) = 1$ is the optimal solution of $\text{Loss}_2 (\pi_2)$. Thus, $\piail_2 (a^1|s^1) = 1, \piail_2 (a^1|s^2) = 1$ is the unique optimal solution of optimization problem $\argmin_{\pi_2} \text{Loss}_2 (\pi_2) +  \text{Loss}_3 (\pi_2)$. We finish the proof in time step $h=2$.

Finally, we consider the policy optimization in time step $h=1$. With \cref{lem:single_variable_opt_condition}, we have that fixing $(\piail_2, \piail_3)$, $\piail_1$ is optimal w.r.t \textsf{VAIL}'s objective. 
\begin{align*}
    \piail_1 \in \argmin_{\pi_1} \ell_1 (\pi_1) = \argmin_{\pi_1} \text{Loss}_1 (\pi_1) + \text{Loss}_2 (\pi_1) +  \text{Loss}_3 (\pi_1) . 
\end{align*}
Note that we have proved that $\piail_2 (a^1|s^1) = \piail_2 (a^1|s^2) = \piail_3 (a^1|s^1) = \piail_3 (a^1|s^2) = 1$ and plug it into the above equation. We use $P^{\pi}$ to denote the distribution induced by $(\pi_1, \piail_2, \piail_3)$.
\begin{align*}
    \text{Loss}_1 (\pi_1) &= \labs P^{\pi}_1(s^{1}, a^{1}) - \frac{1}{6}  \rabs + \labs P^{\pi}_1(s^{1}, a^{2}) - \frac{1}{6}  \rabs + \labs P^{\pi}_1(s^{2}, a^{1}) - \frac{1}{6}  \rabs + \labs P^{\pi}_1(s^{2}, a^{2}) - \frac{1}{6}  \rabs + \frac{1}{3}
    \\
    &= \labs \frac{1}{2} \pi_1 (a^{1}|s^{1}) - \frac{1}{6} \rabs + \labs \frac{1}{2} \pi_1 (a^{1}|s^{1}) - \frac{1}{3} \rabs + \labs \frac{1}{2} \pi_1 (a^{1}|s^{2}) - \frac{1}{6} \rabs + \labs \frac{1}{2} \pi_1 (a^{1}|s^{2}) - \frac{1}{3} \rabs + \frac{1}{3},  
    \\
    \text{Loss}_2 (\pi_1) &= \labs P^{\pi}_2(s^{1}) - 0.5 \rabs + \labs P^{\pi}_2(s^{2})  - 0.5 \rabs + P^{\pi}_2(s^{3})
    \\
    &= \labs \frac{1}{4} \lp \pi_1 (a^{1}|s^{1}) + \pi_1 (a^{1}|s^{2}) \rp - 0.5 \rabs + \labs \frac{1}{4} \lp \pi_1 (a^{1}|s^{1}) + \pi_1 (a^{1}|s^{2}) \rp - 0.5 \rabs \\
    &\quad +\frac{1}{2} \lp 2 - \pi_1 (a^{1}|s^{1}) - \pi_1 (a^{1}|s^{2})   \rp,
    \\
    &= 2 - \pi_1 (a^1|s^1) - \pi_1 (a^1|s^2),  
    \\
    \text{Loss}_3 (\pi_1) &= \labs P^{\pi}_3(s^{1}) - 0.5 \rabs + \labs P^{\pi}_3(s^{2})  - 0.5 \rabs + P^{\pi}_3(s^{3})
    \\
    &= \labs \frac{1}{2} \lp P^{\pi}_2 (s^{1}) + P^{\pi}_2 (s^{1}) \rp - 0.5 \rabs + \labs  \frac{1}{2} \lp P^{\pi}_2 (s^{1}) + P^{\pi}_2 (s^{1}) \rp  - 0.5 \rabs + P^{\pi}_2(s^{3})
    \\
    &= \labs \frac{1}{4} \lp \pi_1 (a^{1}|s^{1}) + \pi_1 (a^{1}|s^{2}) \rp - 0.5 \rabs + \labs \frac{1}{4} \lp \pi_1 (a^{1}|s^{1}) + \pi_1 (a^{1}|s^{2}) \rp - 0.5 \rabs \\
    &\quad +
    \frac{1}{2} \lp 2 - \pi_1 (a^{1}|s^{1}) - \pi_1 (a^{1}|s^{2})   \rp
    \\
    &= 2 - \pi_1 (a^1|s^1) - \pi_1 (a^1|s^2). 
\end{align*}
Combining the above three equations yields that
\begin{align*}
     &\quad \argmin_{\pi_1} \text{Loss}_1 (\pi_1) + \text{Loss}_2 (\pi_1) +  \text{Loss}_3 (\pi_1)
     \\
     &= \argmin_{\pi_1} \labs \frac{1}{2} \pi_1 (a^{1}|s^{1}) - \frac{1}{6} \rabs + \labs \frac{1}{2} \pi_1 (a^{1}|s^{1}) - \frac{1}{3} \rabs + \labs \frac{1}{2} \pi_1 (a^{1}|s^{2}) - \frac{1}{6} \rabs + \labs \frac{1}{2} \pi_1 (a^{1}|s^{2}) - \frac{1}{3} \rabs
     \\
     &\quad - 2\pi_1 (a^1|s^1) - 2\pi_1 (a^1|s^2).
\end{align*}
Note that $\pi_1 (a^1|s^1)$ and $\pi_1 (a^1|s^2)$ are independent and we can view the optimization problem individually.
\begin{align*}
    \argmin_{\pi_1 (a^{1}|s^{1}) \in [0, 1]} \labs \frac{1}{2} \pi_1 (a^{1}|s^{1}) - \frac{1}{6} \rabs + \labs \frac{1}{2} \pi_1 (a^{1}|s^{1}) - \frac{1}{3} \rabs - 2\pi_1 (a^1|s^1).  
\end{align*}
This is a piece-wise linear function and it is direct to see that $\piail_1 (a^1|s^1)=1$ is the unique optimal solution. In the same way, we can also prove that $\piail_1 (a^1|s^2)=1$ is the unique optimal solution. Therefore, we finish the proof in the step $h=1$. 

\end{proof}

\subsection{VAIL with Approximately Optimal Solutions}
\label{appendix:proof_of_the_approximate_solution_in_VAIL}

Here we consider the generalization of an approximately optimal solution of \textsf{VAIL}'s objective instead of the exactly optimal solution discussed in \cref{sec:generalization_of_ail}. In particular, for a policy $\pi$, given estimation $\widehat{P}^{\piE}_h(s, a)$, \textsf{VAIL}'s objective is formulated as
\begin{align*}
    \min_{\pi \in \Pi} f(\pi) := \sum_{h=1}^H \text{Loss}_h (\pi) =  \sum_{h=1}^{H} \sum_{(s, a) \in \gS \times \gA} | P^{\pi}_h(s, a) - \widehat{P}^{\piE}_h(s, a) |.
\end{align*}
Here $\text{Loss}_h (\pi) = \sum_{(s, a) \in \gS \times \gA} | P^{\pi}_h(s, a) - \widehat{P}^{\piE}_h(s, a) |$. Suppose that we can get an $\varepsilon_{\ail}$-approximately optimal solution $\widebar{\pi}$ instead of the exact optimal solution $\piail$. More specifically, it holds that
\begin{align*}
    f (\widebar{\pi}) \leq \min_{\pi \in \Pi} f(\pi) + \varepsilon_{\ail} = f(\piail) + \varepsilon_{\ail} .  
\end{align*}
We consider the generalization of $\widebar{\pi}$ on Standard Imitation and Reset Cliff. Note that the analysis of the approximately optimal solution on Standard Imitation is straightforward. To see this, through the reduction analysis, we can directly plug the optimization error into the final policy value gap.  
\begin{align*}
    V^{\piE} - V^{\widebar{\pi}} &= \sum_{h=1}^{H} \sum_{(s, a) \in \gS \times \gA} \lp P^{\piE}_h(s, a) - P^{\widebar{\pi}}_h(s, a) \rp r_h(s, a)
    \\
    &\leq \sum_{h=1}^{H} \lnorm P^{\piE}_h(\cdot, \cdot) - P^{\widebar{\pi}}_h(\cdot, \cdot)  \rnorm_{1}
    \\
    &\leq  \sum_{h=1}^{H} \lnorm P^{\piE}_h(\cdot, \cdot) - \widehat{P}^{\piE}_h(\cdot, \cdot)  \rnorm_{1} +  \sum_{h=1}^{H} \lnorm \widehat{P}^{\piE}_h(\cdot, \cdot) - P^{\widebar{\pi}}_h(\cdot, \cdot)  \rnorm_{1}
    \\
    &\leq \sum_{h=1}^{H} \lnorm P^{\piE}_h(\cdot, \cdot) - \widehat{P}^{\piE}_h(\cdot, \cdot)  \rnorm_{1} +  \min_{\pi \in \Pi} \sum_{h=1}^{H} \lnorm \widehat{P}^{\piE}_h(\cdot, \cdot) - P^{\pi}_h(\cdot, \cdot)  \rnorm_{1} + \varepsilon_{\ail}
    \\
    &\leq 2\sum_{h=1}^{H} \lnorm P^{\piE}_h(\cdot, \cdot) - \widehat{P}^{\piE}_h(\cdot, \cdot)  \rnorm_{1} + \varepsilon_{\ail}. 
\end{align*}

Thus, it is straightforward to obtain the following theoretical guarantee. 
\begin{thm}[Sample Complexity of Approximate VAIL]  \label{theorem:sample_complexity_approximate_vail}
For any tabular and episodic MDP, assume $\widebar{\pi}$ is an $\varepsilon_{\ail}$-approximately optimal solution of \eqref{eq:ail}. To obtain an $\varepsilon$-optimal policy (i.e., $V^{\piE} - \expect[V^{\widebar{\pi}}] \leq \varepsilon$), in expectation, when $\varepsilon_{\ail} \leq \varepsilon/2$, VAIL requires at most $\gO(|\gS| H^2/\varepsilon^2)$ expert trajectories. 
\end{thm}

However, the analysis of the approximately optimal solution on Reset Cliff is non-trivial. On Reset Cliff, we hope to obtain a tight horizon-free sample complexity as in \cref{theorem:ail_reset_cliff} and thus we cannot apply the reduction analysis. Here we briefly discuss the proof idea. To achieve a horizon-free sample complexity, we first develop a sharp analysis to measure the distance between the approximately optimal solution and the exactly optimal solution in \cref{prop:ail_general_reset_cliff_approximate_solution}. With \cref{prop:ail_general_reset_cliff_approximate_solution}, we can upper bound the policy value gap of the approximately optimal solution and obtain the corresponding sample complexity.

First, we present a useful property of \textsf{VAIL}'s objective on Reset Cliff.

\begin{lem}
\label{lemma:ail_policy_ail_objective_equals_expert_policy_ail_objective}
Consider the Reset Cliff MDP satisfying \cref{asmp:reset_cliff}. Suppose that $\piail$ is the optimal solution of \eqref{eq:ail}, then we have that $f(\piail) = f(\piE)$.
\end{lem}
Refer to \cref{appendix:proof_lemma:ail_policy_ail_objective_equals_expert_policy_ail_objective} for the proof. The following proposition demonstrates that the distance between the approximately optimal solution and the exactly optimal solution can be upper bounded by the optimization error. 

\begin{prop}
\label{prop:ail_general_reset_cliff_approximate_solution}
Consider any tabular and episodic MDP satisfying \cref{asmp:reset_cliff}. The candidate policy set is defined as $\Pi^{\text{opt}} = \{ \pi \in \Pi: \forall h \in [H], \exists s \in \goodS, \pi_h (a^{1}|s) > 0 \}$. Given expert state-action distribution estimation $\widehat{P}^{\piE}_{H}$, suppose that $\widebar{\pi} \in \Pi^{\text{opt}}$ is an $\varepsilon_{\ail}$-approximately optimal solution of \eqref{eq:ail}, when $\vert \gD \vert \geq 1$, we have the following approximate optimality condition almost surely:
\begin{align*}
    c (\widebar{\pi}) \lp \sum_{h=1}^{H} \sum_{\ell=1}^{h-1} \sum_{s \in \goodS} P^{\widebar{\pi}}_{\ell} (s) \lp 1 - \widebar{\pi}_{\ell} (a^{1}|s)  \rp + \sum_{s \in  \gS^{\widebar{\pi}}_H}  P^{\widebar{\pi}}_{H} (s) \lp \min\{1,  \widehat{P}^{\piE}_{H} (s) / P^{\piE}_H (s)\} -  
    \widebar{\pi}_H (a^1|s)  \rp  \rp \leq \varepsilon_{\ail},
\end{align*}
where $c (\pi) := \min_{1 \leq \ell < h \leq H, s, s^\prime \in \goodS} \{ \sP^{\pi} \lp s_{h} = s |s_{\ell} = s^\prime, a_{\ell} = a^{1} \rp \}$ and $\gS^{\pi}_H  = \{s \in \goodS, \pi_H (a^1|s) \leq \min \{1,  \widehat{P}^{\piE}_{H} (s) / P^{\piE}_H (s) \}  \}$. Note that $c(\pi) > 0$ for any $\pi \in \Pi^{\text{opt}}$.
\end{prop}

This proof is rather technical and is deferred to \cref{appendix:proof_prop:ail_general_reset_cliff_approximate_solution}. We explain \cref{prop:ail_general_reset_cliff_approximate_solution} by connecting it with \cref{prop:ail_general_reset_cliff}. In particular, if $\varepsilon_{\ail} = 0$, we can show that the optimality condition in  \cref{prop:ail_general_reset_cliff_approximate_solution} reduces to the one in \cref{prop:ail_general_reset_cliff}. To see this, for each $h \in [H-1]$ and $s \in \goodS$, since $c(\widebar{\pi}) > 0$ and $P^{\widebar{\pi}}_h (s) > 0$, we must have $\widebar{\pi}_{h}(a^{1} | s) = 1$ for all $h \in [H-1]$ while there exists many optimal solutions in the last step policy optimization.

Equipped with \cref{prop:ail_general_reset_cliff} and \cref{prop:ail_general_reset_cliff_approximate_solution}, we can obtain the horizon-free sample complexity for the approximately optimal solution of \textsf{VAIL} in \cref{theorem:ail_approximate_reset_cliff}.

\begin{thm}[Horizon-free Sample Complexity of Approximate \textsf{VAIL} on \textsf{Reset Cliff}]
\label{theorem:ail_approximate_reset_cliff}
For each tabular and episodic MDP satisfying \cref{asmp:reset_cliff}, the candidate policy set is defined as $\Pi^{\text{opt}} = \{ \pi \in \Pi: \forall h \in [H], \exists s \in \goodS, \pi_h (a^{1}|s) > 0 \}$. Suppose that $\widebar{\pi} \in \Pi^{\text{opt}} $ is an $\varepsilon_{\ail}$-approximately optimal solution of \eqref{eq:ail}, to obtain an $\varepsilon$-optimal policy (i.e., $V^{\piE} - \expect[V^{\widebar{\pi}}] \leq \varepsilon$), in expectation, when $\varepsilon_{\ail}  \leq \varepsilon/8 \cdot  c(\widebar{\pi})$, \textsf{VAIL} requires at most ${\gO}(|\gS|/\varepsilon^2)$ expert trajectories. Here
$c (\pi) := \min_{1 \leq \ell < h \leq H, s, s^\prime \in \goodS} \{ \sP^{\pi} \lp s_{h} = s |s_{\ell} = s^\prime, a_{\ell} = a^{1} \rp \}$.
\end{thm}

\begin{thm}[High Probability Version of \cref{theorem:ail_approximate_reset_cliff}]
\label{theorem:ail_approximate_high_prob_reset_cliff}
For each tabular and episodic MDP satisfying \cref{asmp:reset_cliff}, the candidate policy set is defined as $\Pi^{\text{opt}} = \{ \pi \in \Pi: \forall h \in [H], \exists s \in \goodS, \pi_h (a^{1}|s) > 0 \}$. Suppose that $\widebar{\pi} \in \Pi^{\text{opt}} $ is an $\varepsilon_{\ail}$-approximately optimal solution of \eqref{eq:ail}, with probability at least $1-\delta$, to obtain an $\varepsilon$-optimal policy (i.e., $V^{\piE} - V^{\widebar{\pi}} \leq \varepsilon$), when $ \varepsilon_{\ail}  \leq \varepsilon/8 \cdot  c(\widebar{\pi})$, \textsf{VAIL} requires at most ${\widetilde{\gO}}(|\gS|/\varepsilon^2)$ expert trajectories.
\end{thm}

\begin{proof}[Proof of \cref{theorem:ail_approximate_reset_cliff} and \cref{theorem:ail_approximate_high_prob_reset_cliff}]
With a fixed estimation, we consider \textsf{VAIL}'s objective.
\begin{align*}
    \min_{\pi \in \Pi} \sum_{h=1}^{H} \sum_{(s, a) \in \gS \times \gA} | P^{\pi}_h(s, a) - \widehat{P}^{\piE}_h(s, a) |.
\end{align*}
Suppose that $\widebar{\pi}$ is $\varepsilon_{\ail}$-optimal w.r.t the above objective. We construct an optimal solution $\piail$ in the following way.

\begin{itemize}
    \item By \cref{prop:ail_general_reset_cliff}, we have $\forall h \in [H-1], s \in \goodS, \piail_{h} (a^{1}|s) = \piE_{h} (a^{1}|s)$.
    
    \item For the last time step $H$, we defined a set of states $\gS_H^1 := \{s \in \goodS: \widehat{P}^{\piE}_H (s) < P^{\piE}_H (s)   \}$. The policy in the last time step is defined as $\forall s \in \gS_H^1$, $\piail_{H} (a^{1}|s) = \widehat{P}^{\piE}_H (s) / P^{\piE}_H (s)$ and $\forall s \in \goodS \setminus \gS_H^1$, $\piail_{H} (a^{1}|s) = 1$. In a word, $\forall s \in \goodS$, $\piail_{H} (a^{1}|s) = \min \{\widehat{P}^{\piE}_H (s) / P^{\piE}_H (s),1  \}$.
    
    \item For simplicity of analysis, we also define the policy on bad states although $\piail$ never visit bad states. $\forall h \in [H], s \in \badS, \piail_{h} (\cdot|s) = \widebar{\pi}_{h} (\cdot|s)$.   
\end{itemize}

We first verify that $\piail$ is the optimal solution of \eqref{eq:ail}. With \cref{prop:ail_general_reset_cliff}, we have that $\forall h \in [H-1], s \in \goodS, \piail_{h} (a^{1}|s) = \piE_{h} (a^{1}|s)$ is the \emph{unique} optimal solution of \eqref{eq:ail}. Furthermore, with fixed $ \piail_{h} (a^{1}|s) = \piE_{h} (a^{1}|s), \forall h \in [H-1], s \in \goodS$, \textsf{VAIL}'s objective from time step $1$ to $H-1$ is fixed. Therefore, it suffices to verify that with fixed $ \piail_{h} (a^{1}|s) = \piE_{h} (a^{1}|s), \forall h \in [H-1], s \in \goodS$, $\piail_{H}$ is optimal w.r.t the \textsf{VAIL}'s objective in the last time step. Thus, we take the policy in the last time step as optimization variables and consider \textsf{VAIL}'s objective in the last time step.
\begin{align*}
    & \quad \min_{\pi_H} \sum_{(s, a) \in \gS \times \gA} | P^{\piail}_H (s) \pi_H (a|s) - \widehat{P}^{\piE}_H (s, a) |
    \\
    &=  \min_{\pi_H} \sum_{(s, a) \in \gS \times \gA} | P^{\piE}_H (s) \pi_H (a|s) - \widehat{P}^{\piE}_H (s, a) |
    \\
    &=  \min_{\pi_H} \sum_{s \in \goodS } \lp  | P^{\piE}_H (s) \pi_H (a^{1}|s) - \widehat{P}^{\piE}_H (s) | + P^{\piE}_H (s) \lp 1- \pi_H (a^{1}|s) \rp  \rp
    \\
    &= \min_{\pi_H} \sum_{s \in \goodS } \lp  | P^{\piE}_H (s) \pi_H (a^{1}|s) - \widehat{P}^{\piE}_H (s) | - P^{\piE}_H (s)  \pi_H (a^{1}|s)  \rp .
\end{align*}
We can view the optimization problem for each $s \in \goodS$ individually.
\begin{align*}
    \forall s \in \goodS, \min_{\pi_H (a^{1}|s) \in [0, 1]}   | P^{\piE}_H (s) \pi_H (a^{1}|s) - \widehat{P}^{\piE}_H (s) | - P^{\piE}_H (s)  \pi_H (a^{1}|s) .
\end{align*}
For $s \in \gS_H^1$, by \cref{lem:single_variable_opt_condition}, we have that $ \piail_{H} (a^{1}|s) = \widehat{P}^{\piE}_H (s) / P^{\piE}_H (s)$ is the optimal solution. For $s \in \goodS \setminus \gS_H^1$, by \cref{lem:single_variable_opt}, we have that $ \piail_{H} (a^{1}|s) =1$ is the optimal solution. Therefore, we show that $\piail_{H}$ is also optimal w.r.t the \textsf{VAIL}'s objective in the last time step and hence $\piail$ is the optimal solution of \eqref{eq:ail}.

Now we consider the policy value gap of $\widebar{\pi}$.
\begin{align}
\label{eq:value_gap_decomposition_approximate_solution}
    V^{\piE} - V^{\widebar{\pi}} = V^{\piE} - V^{\piail} + V^{\piail} -  V^{\widebar{\pi}}. 
\end{align}
By \eqref{eq:value_gap_ail_policy} in the proof of \cref{theorem:ail_reset_cliff}, we have that
\begin{align}
\label{eq:ail_policy_piE_value_gap}
    V^{\piE} - V^{\piail} \leq 2 \sum_{s \in \gS} \labs \widehat{P}^{\piE}_H (s) - P^{\piE}_H (s)  \rabs.
\end{align}
Then we consider the policy value gap between $\piail$ and $\widebar{\pi}$. With the dual representation of policy value, we get that
\begin{align*}
    V^{\piail} -  V^{\widebar{\pi}} &= \sum_{h=1}^{H-1} \sum_{(s, a) \in \gS \times \gA} \lp P^{\piail}_h (s, a) - P^{\widebar{\pi}}_h (s, a)  \rp r_h (s, a) + \sum_{(s, a) \in \gS \times \gA} \lp P^{\piail}_H (s, a) - P^{\widebar{\pi}}_H (s, a)  \rp r_H (s, a)
    \\
    &\leq \sum_{h=1}^{H-1} \lnorm P^{\piail}_h (\cdot, \cdot) - P^{\widebar{\pi}}_h (\cdot, \cdot) \rnorm_{1} + \sum_{(s, a) \in \gS \times \gA} \lp P^{\piail}_H (s, a) - P^{\widebar{\pi}}_H (s, a)  \rp r_H (s, a), 
\end{align*}
where we use $P^{\pi}_h (\cdot, \cdot)$ denote the state-action distribution induced by $\pi$. Recall the definition of $\gS^{\pi}_H  = \{s \in \goodS, \pi_H (a^1|s) \leq \min \{1,  \widehat{P}^{\piE}_{H} (s) / P^{\piE}_H (s) \}  \}$ introduced in \cref{prop:ail_general_reset_cliff_approximate_solution}. For the second term in RHS, we have
\begin{align*}
    &\quad \sum_{(s, a) \in \gS \times \gA} \lp P^{\piail}_H (s, a) - P^{\widebar{\pi}}_H (s, a)  \rp r_H (s, a)
    \\
    &= \sum_{s \in \goodS}  \lp P^{\piail}_H (s, a^{1}) - P^{\widebar{\pi}}_H (s, a^{1})  \rp
    \\
    &= \sum_{s \in \goodS}  \lp P^{\piail}_H (s) \piail_{H} \lp a^{1}|s \rp - P^{\widebar{\pi}}_H (s) \widebar{\pi}_H (a^{1}|s)  \rp
    \\
    &=  \sum_{s \in \goodS}   \lp P^{\piail}_H (s) - P^{\widebar{\pi}}_H (s) \rp  \piail_{H} \lp a^{1}|s \rp + \sum_{s \in \goodS} P^{\widebar{\pi}}_H (s) \lp \piail_{H} \lp a^{1}|s \rp -   \widebar{\pi}_H (a^{1}|s) \rp
    \\
    &\leq \lnorm P^{\piail}_H (\cdot) - P^{\widebar{\pi}}_H (\cdot) \rnorm_1 + \sum_{s: s \in \goodS, \piail_{H} \lp a^{1}|s \rp \geq   \widebar{\pi}_H (a^{1}|s)} P^{\widebar{\pi}}_H (s) \lp \piail_{H} \lp a^{1}|s \rp -   \widebar{\pi}_H (a^{1}|s) \rp
    \\
    &= \lnorm P^{\piail}_H (\cdot) - P^{\widebar{\pi}}_H (\cdot) \rnorm_1 + \sum_{s \in \gS^{\widebar{\pi}}_H} P^{\widebar{\pi}}_H (s) \lp \piail_{H} \lp a^{1}|s \rp -   \widebar{\pi}_H (a^{1}|s) \rp,
\end{align*}
where we use $P^{\pi}_h (\cdot)$ to denote the state distribution induced by $\pi$. Plugging the above inequality into the policy value gap yields
\begin{align*}
    V^{\piail} -  V^{\widebar{\pi}} &\leq \sum_{h=1}^{H-1} \lnorm P^{\piail}_h (\cdot, \cdot) - P^{\widebar{\pi}}_h (\cdot, \cdot) \rnorm_{1} + \lnorm P^{\piail}_H (\cdot) - P^{\widebar{\pi}}_H (\cdot) \rnorm_1  + \sum_{s \in \gS^{\widebar{\pi}}_H} P^{\widebar{\pi}}_H (s) \lp \piail_{H} \lp a^{1}|s \rp -   \widebar{\pi}_H (a^{1}|s) \rp.
\end{align*}
With \cref{lemma:state_dist_discrepancy}, we have that
\begin{align*}
    &\quad \sum_{h=1}^{H-1} \lnorm P^{\piail}_h (\cdot, \cdot) - P^{\widebar{\pi}}_h (\cdot, \cdot) \rnorm_{1} + \lnorm P^{\piail}_H (\cdot) - P^{\widebar{\pi}}_H (\cdot) \rnorm_1 
    \\
    &\leq \sum_{h=1}^{H-1} \lnorm P^{\piail}_h (\cdot) - P^{\widebar{\pi}}_h (\cdot) \rnorm_{1} + \sum_{h=1}^{H-1} \expect_{s \sim P^{\widebar{\pi}}_h (\cdot)} \ls \lnorm \piail_h (\cdot|s) - \widebar{\pi}_h (\cdot|s) \rnorm_1 \rs + \lnorm P^{\piail}_H (\cdot) - P^{\widebar{\pi}}_H (\cdot) \rnorm_1
    \\
    &= \sum_{h=1}^{H} \lnorm P^{\piail}_h (\cdot) - P^{\widebar{\pi}}_h (\cdot) \rnorm_{1} + \sum_{h=1}^{H-1} \expect_{s \sim P^{\widebar{\pi}}_h (\cdot)} \ls \lnorm \piail_h (\cdot|s) - \widebar{\pi}_h (\cdot|s) \rnorm_1 \rs
    \\
    &= \sum_{h=1}^{H} \sum_{\ell=1}^{h-1} \expect_{s \sim P^{\widebar{\pi}}_\ell (\cdot)} \ls \lnorm \piail_\ell (\cdot|s) - \widebar{\pi}_\ell (\cdot|s) \rnorm_1 \rs + \sum_{h=1}^{H-1} \expect_{s \sim P^{\widebar{\pi}}_h (\cdot)} \ls \lnorm \piail_h (\cdot|s) - \widebar{\pi}_h (\cdot|s) \rnorm_1 \rs
    \\
    &\leq 2  \sum_{h=1}^{H} \sum_{\ell=1}^{h-1} \expect_{s \sim P^{\widebar{\pi}}_\ell (\cdot)} \ls \lnorm \piail_\ell (\cdot|s) - \widebar{\pi}_\ell (\cdot|s) \rnorm_1 \rs.
\end{align*}
Then we have that
\begin{align*}
    V^{\piail} -  V^{\widebar{\pi}} &\leq 2  \sum_{h=1}^{H} \sum_{\ell=1}^{h-1} \expect_{s \sim P^{\widebar{\pi}}_\ell (\cdot)} \ls \lnorm \piail_\ell (\cdot|s) - \widebar{\pi}_\ell (\cdot|s) \rnorm_1 \rs +  \sum_{s \in \gS^{\widebar{\pi}}_H} P^{\widebar{\pi}}_H (s) \lp \piail_{H} \lp a^{1}|s \rp -   \widebar{\pi}_H (a^{1}|s) \rp.
\end{align*}

Notice that $\piail$ agrees with $\widebar{\pi}$ on bad states and we obtain
\begin{align*}
     \expect_{s \sim P^{\widebar{\pi}}_{\ell} (\cdot)} \ls \lnorm \piail_{\ell} (\cdot|s) - \widebar{\pi}_{\ell} (\cdot|s) \rnorm_1 \rs &= \sum_{s \in \goodS } P^{\widebar{\pi}}_{\ell} (s)    \lnorm \piail_{\ell} (\cdot|s) - \widebar{\pi}_{\ell} (\cdot|s) \rnorm_1
    \\
    &=  \sum_{s \in \goodS } P^{\widebar{\pi}}_{\ell} (s) \lp \labs \piail_{\ell} (a^{1}|s) - \widebar{\pi}_{\ell} (a^{1}|s)  \rabs + \sum_{a \in \gA \setminus \{a^{1}\}} \widebar{\pi}_{\ell} (a|s)  \rp
    \\
    &= 2 \sum_{s \in \goodS } P^{\widebar{\pi}}_{\ell} (s)    \lp 1 - \widebar{\pi}_{\ell} (a^{1}|s) \rp.
\end{align*}
In the penultimate inequality, we use the fact that $\forall \ell \in [H-1], s \in \goodS, \piail_{\ell} (a^{1}|s) = 1$. Then we have that
\begin{align*}
    V^{\piail} -  V^{\widebar{\pi}} &\leq 4  \sum_{h=1}^{H} \sum_{\ell=1}^{h-1} \sum_{s \in \goodS } P^{\widebar{\pi}}_{\ell} (s)    \lp 1 - \widebar{\pi}_{\ell} (a^{1}|s) \rp +  \sum_{s \in \gS^{\widebar{\pi}}_H} P^{\widebar{\pi}}_H (s) \lp \piail_{H} \lp a^{1}|s \rp -   \widebar{\pi}_H (a^{1}|s) \rp.
\end{align*}
Then we consider the second term in RHS. For the last time step $H$, notice that $\piail_{H} (a^{1}|s)  = \min\{ \widehat{P}^{\piE}_H (s) / P^{\piE}_H (s), 1  \}$. Then we have
\begin{align*}
    \sum_{s \in \gS^{\widebar{\pi}}_H} P^{\widebar{\pi}}_H (s) \lp \piail_{H} \lp a^{1}|s \rp -   \widebar{\pi}_H (a^{1}|s) \rp \leq \sum_{s \in \gS^{\widebar{\pi}}_H} P^{\widebar{\pi}}_H (s) \lp \min\lb \frac{\widehat{P}^{\piE}_H (s)}{P^{\piE}_H (s)},  1  \rb -   \widebar{\pi}_H (a^{1}|s) \rp. 
\end{align*}
Combing the above inequality with $V^{\piail} -  V^{\widebar{\pi}}$ yields that
\begin{align*}
    V^{\piail} -  V^{\widebar{\pi}} &\leq 4  \sum_{h=1}^{H} \sum_{\ell=1}^{h-1} \sum_{s \in \goodS } P^{\widebar{\pi}}_{\ell} (s)    \lp 1 - \widebar{\pi}_{\ell} (a^{1}|s) \rp + \sum_{s \in \gS^{\widebar{\pi}}_H} P^{\widebar{\pi}}_H (s) \lp \min\lb \frac{\widehat{P}^{\piE}_H (s)}{P^{\piE}_H (s)},  1  \rb -   \widebar{\pi}_H (a^{1}|s) \rp.
\end{align*}
We apply \cref{prop:ail_general_reset_cliff_approximate_solution} and get that
\begin{align*}
    V^{\piail} -  V^{\widebar{\pi}} &\leq 4 \lp \sum_{h=1}^{H} \sum_{\ell=1}^{h-1} \sum_{s \in \goodS } P^{\widebar{\pi}}_{\ell} (s)    \lp 1 - \widebar{\pi}_{\ell} (a^{1}|s) \rp + \sum_{s \in \gS^{\widebar{\pi}}_H} P^{\widebar{\pi}}_H (s) \lp \min\lb \frac{\widehat{P}^{\piE}_H (s)}{P^{\piE}_H (s)},  1  \rb -   \widebar{\pi}_H (a^{1}|s) \rp \rp
    \\
    &\leq \frac{4}{c (\widebar{\pi})} \varepsilon_{\ail}.
\end{align*}

We first consider the sample complexity required to achieve a small policy value gap with high probability. With \eqref{eq:value_gap_decomposition_approximate_solution} and \eqref{eq:ail_policy_piE_value_gap}, we have
\begin{align*}
    V^{\piE} - V^{\widebar{\pi}} &= V^{\piE} - V^{\piail} + V^{\piail} -  V^{\widebar{\pi}}
    \\
    &\leq 2 \sum_{s \in \gS} \labs \widehat{P}^{\piE}_H (s) - P^{\piE}_H (s)  \rabs + \frac{4}{c (\widebar{\pi})} \varepsilon_{\ail}
    \\
    &=  2 \lnorm \widehat{P}^{\piE}_H (\cdot) - P^{\piE}_H (\cdot) \rnorm_{1} + \frac{4}{c (\widebar{\pi})} \varepsilon_{\ail}. 
\end{align*}
By $\ell_1$-norm concentration inequality in \cref{lemma:l1_concentration}, with probability at least $1-\delta$, we have 
\begin{align*}
    V^{\piE} - V^{\widebar{\pi}} \leq 2\sqrt{\frac{2 \vert \gS \vert \ln (1/\delta)}{m}} + \frac{4}{c (\widebar{\pi})} \varepsilon_{\ail}. 
\end{align*}
To achieve an $\varepsilon$-optimal policy, when $4 \varepsilon_{\ail} / c(\widebar{\pi}) \leq \varepsilon/2$, we need a sample complexity of $\widetilde{\gO} (|\gS|/\varepsilon^2)$ at most.

We proceed to upper bound the expected policy value gap between $\widebar{\pi}$ and $\piE$ in expectation, with \eqref{eq:value_gap_decomposition_approximate_solution} and \eqref{eq:ail_policy_piE_value_gap}, we have 
\begin{align*}
    \expect \ls V^{\piE} - V^{\widebar{\pi}} \rs &=  \expect \ls V^{\piE} - V^{\piail} \rs + \expect \ls V^{\piail} -  V^{\widebar{\pi}} \rs
    \\
    &\leq 2 \expect \ls \sum_{s \in \gS} \labs \widehat{P}^{\piE}_H (s) - P^{\piE}_H (s)  \rabs \rs + \frac{4}{c (\widebar{\pi})} \varepsilon_{\ail}
    \\
    &= 2 \expect \ls  \lnorm \widehat{P}^{\piE}_H (\cdot) - P^{\piE}_H (\cdot) \rnorm_{1}  \rs + \frac{4}{c (\widebar{\pi})} \varepsilon_{\ail}, 
\end{align*}
where the expectation is taken w.r.t the randomness of expert demonstrations. We apply the upper bound of the expected $\ell_1$ risk of empirical distribution \citep[Theorem 1]{han2015minimax} and obtain
\begin{align*}
    \expect \ls V^{\piE} - V^{\widebar{\pi}} \rs \leq 2 \sqrt{ \frac{|\gS| - 1}{m}} + \frac{4}{c (\widebar{\pi})} \varepsilon_{\ail}.
\end{align*}
To achieve an $\varepsilon$-optimal policy (i.e., $\expect[V^{\piE} - V^{\piail}] \leq \varepsilon$), when $4 \varepsilon_{\ail} / c(\widebar{\pi}) \leq \varepsilon/2$, we need a sample complexity of $\gO(|\gS|/\varepsilon^2)$ at most.
\end{proof}

\subsection{Horizon-free Sample Complexity of MIMIC-MD on Reset Cliff}
\label{appendix:horizon_free_complexity_of_mimic_md}

Here we show the horizon-free sample complexity of MIMIC-MD \citep{rajaraman2020fundamental} on Reset Cliff. With the estimator $\widetilde{P}_h^{\piE}$ in \eqref{eq:new_estimator}, MIMIC-MD performs the state-action distribution matching like VAIL.
\begin{align*}
    \min_{\pi \in \Pi_{\text{BC}} \lp \gD_1 \rp} \sum_{h=1}^{H} \sum_{(s, a) \in \gS \times \gA} | P^{\pi}_h(s, a) - \widetilde{P}^{\piE}_h(s, a) |,
\end{align*}
where $\Pi_{\text{BC}} (\gD_1) = \{ \pi \in \Pi: \pi_h (s) = \piE_h (s), \forall h \in [H], s \in \gS_h (\gD_1) \}$ is the set of BC policies on $\gD_1$.

\begin{thm}[High Probability Version of \cref{theorem:ail_mimic_md}]     \label{theorem:ail_mimic_md_high_prob}
For each tabular and episodic MDP satisfying \cref{asmp:reset_cliff}, suppose that $\piail$ is the optimal solution of the above problem and $\vert \gD \vert \geq 2$, with probability at least $1-\delta$, to obtain an $\varepsilon$-optimal policy (i.e., $V^{\piE} - V^{\piail} \leq \varepsilon$), MIMIC-MD requires at most $\min\{ {\widetilde{\gO}}(|\gS|/\varepsilon^2),  {\widetilde{\gO}}(|\gS| \sqrt{H}/\varepsilon)\}$ expert trajectories.
\end{thm}

\begin{proof}[Proof of \cref{theorem:ail_mimic_md} and \cref{theorem:ail_mimic_md_high_prob}]

The proof is mainly based on \cref{prop:ail_general_reset_cliff} and \cref{theorem:ail_reset_cliff}. We apply \cref{prop:ail_general_reset_cliff} with the unbiased estimation in \eqref{eq:new_estimator}. Therefore, with \cref{prop:ail_general_reset_cliff}, we obtain that $\piail$ agrees with the expert policy on good states in the first $H-1$ time steps. Then the policy value gap of $\piail$ only arises from the decision errors in the last time step. Notice that MIMIC-MD and VAIL both perform the state-action distribution. Following the same proof strategy as in \cref{theorem:ail_reset_cliff}, with \eqref{eq:vail_reset_cliff_value_gap_last_step_estimation_error}, we have that
\begin{align*}
    \labs  V^{\piE} - V^{\piail} \rabs \leq 2\sum_{s \in \gS} \labs \widetilde{P}^{\piE}_H (s) - P^{\piE}_H (s)  \rabs = 2\sum_{s \in \gS} \labs \widetilde{P}^{\piE}_H (s, a^{1}) - P^{\piE}_H (s, a^{1})  \rabs.
\end{align*}
Recall that $\Tr_H^{\gD_1} = \lb \tr_H: \tr_H (s_\ell) \in \gS_{\ell} (\gD), \forall \ell \in [H] \rb$ is the set of trajectories along which each state has been visited in $\gD_1$ up to time step $H$. With \eqref{eq:new_estimation_probability_error}, we have
\begin{align*}
    &\quad \labs \widetilde{P}_H^{\pi_E}(s, a^{1}) - P_H^{\piE}(s, a^{1}) \rabs
    \\
    &= \labs \frac{  \sum_{\tr_H \in \gD_1^c}  \indict\{ \tr_H (s_H, a_H) = (s, a^{1}), \tr_H \not\in \Tr_H^{\gD_1}  \} }{|\gD_1^c|} - \sum_{\tr_H \notin \Tr_H^{\gD_1}} \sP^{\piE}(\tr_H) \indict\lb  \tr_H (s_H, a_H) = (s, a^{1})  \rb  \rabs. 
\end{align*}

For a trajectory $\tr_H$, let $E_H^{s}$ be the event that $\tr_H$ agrees with expert policy at state $s$ in the last time step $H$ but is not in $\Tr_H^{\gD_1}$, that is, 
\begin{align*}
    E_H^{s} = \indict\{\tr_H (s_H, a_H) = (s, a^{1}) \, \cap \, \tr_H \notin {\Tr}_H^{\gD_1}\}.
\end{align*}
We consider $E_{H}^{s}$ is measured by the stochastic process induced by the expert policy $\piE$. Accordingly, its probability is denoted as $\sP^{\piE}(E_{H}^{s})$. In fact, we see that $\sP^{\piE}(E^{s}_H)$ is equal to the second term in the RHS of above equation. Moreover, the first term in the RHS of above equation is an empirical estimation for $\sP^{\piE}(E^{s}_H)$. More specifically, let $X (\tr_{H})$ denote the Bernoulli random variable of $\indict\{ \tr_H (s_H, a_H) = (s, a^{1}), \tr_H \not\in \Tr_H^{\gD_1}  \}$. We have that 
\begin{align*}
    \expect_{\gD_1^c} \ls \frac{\sum_{\tr_H \in \gD_1^c} X (\tr_{H}) }{|\gD_1^c|} \rs = \sP^{\piE}(E^{s}_H),
\end{align*}
where the expectation is taken w.r.t the randomness of $\gD_1^c$. We first prove the sample complexity required to obtain a small policy value gap in expectation. We take expectation w.r.t the randomness of $\gD_1^c$ on both sides.
\begin{align*}
    \expect_{\gD_1^c} \ls \labs \widetilde{P}_H^{\pi_E}(s, a^{1}) - P_H^{\piE}(s, a^{1}) \rabs \rs &= \expect_{\gD_1^c} \ls \labs \frac{  \sum_{\tr_H \in \gD_1^c} X (\tr_{H}) }{|\gD_1^c|} - \sP^{\piE}(E_{H}^{s})   \rabs \rs
    \\
    &\leq \sqrt{ \expect_{\gD_1^c} \ls \lp \frac{  \sum_{\tr_H \in \gD_1^c}  X (\tr_{H}) }{|\gD_1^c|} - \sP^{\piE}(E_{H}^{s})   \rp^2 \rs }, 
\end{align*}
where the last inequality follows the Jensen's inequality. Furthermore, we obtain
\begin{align*}
    \expect_{\gD_1^c} \ls \labs \widetilde{P}_H^{\pi_E}(s, a^{1}) - P_H^{\piE}(s, a^{1}) \rabs \rs &\leq \sqrt{ \Var \ls \frac{  \sum_{\tr_H \in \gD_1^c}  X (\tr_{H}) }{|\gD_1^c|}     \rs } = \sqrt{ \frac{\Var \ls X (\tr_{H})      \rs}{|\gD_1^c|}  } \leq \sqrt{ \frac{\sP^{\piE}(E_{H}^{s})}{|\gD_1^c|}  }.    
\end{align*}
The last inequality follows that for $X \sim \text{Ber} (p)$, $\Var \ls X \rs = p (1-p) \leq p$. Then we have that
\begin{align*}
    \expect_{\gD_1^c} \ls \sum_{s \in \gS} \labs \widetilde{P}_H^{\pi_E}(s, a^{1}) - P_H^{\piE}(s, a^{1}) \rabs \rs \leq \sum_{s \in \gS} \sqrt{ \frac{\sP^{\piE}(E_{H}^{s})}{|\gD_1^c|}  } \leq \sqrt{ \frac{\sum_{s \in \gS} \sP^{\piE}(E_{H}^{s}) \vert \gS \vert}{|\gD_1^c|}  }.  
\end{align*}
The last inequality follows the Cauchy-Schwarz inequality. It remains to upper bound $\sum_{s \in \gS}  \sP^{\piE}(E_{H}^{s})$. To this end, we define the event $G_H^{\gD_1}$: the expert policy $\piE$ visits certain states that are uncovered in $\gD_1$ up to time step $H$. Formally, $G_H^{\gD_1} = \indict\{ \exists h^{\prime} \leq H,  s_{h^{\prime}} \notin \gS_{h^{\prime}} (\gD_1) \}$, where $\gS_{h}(\gD_1)$ is the set of states in $\gD_1$ in time step $h$. Then we have 
\begin{align}
\label{eq:probability_equation}
    \sum_{s \in \gS} \sP^{\piE} \lp E_H^{s}  \rp = \sP (G_H^{\gD_1}),
\end{align}
where the equality is true because $\cup_{s} E_H^{s}$ corresponds to the event that $\piE$ does not visit any trajectory fully covered in $\gD_1$. On the one hand, we have that $\expect_{\gD_1} \ls \sP (G_H^{\gD_1}) \rs \leq 1$. On the other hand, it holds that
\begin{align*}
    \expect_{\gD_1} \ls \sP (G_H^{\gD_1}) \rs &\leq \expect_{\gD_1} \ls \sum_{h=1}^{H} \sum_{s \in \gS} P^{\piE}_h(s) \indict\lb s \notin \gS_h(\gD_1)  \rb \rs
    \\
    &= \sum_{h=1}^{H} \sum_{s \in \gS} P^{\piE}_h(s) \sP \lp s \notin \gS_h(\gD_1)  \rp
    \\
    &= \sum_{h=1}^{H} \sum_{s \in \gS} P^{\piE}_h(s) \lp 1 - P^{\piE}_h(s)  \rp^{m/2}
    \\
    &\leq \frac{2 \vert \gS \vert H}{em},
\end{align*}
where the last inequality follows \eqref{eq:expected_missing_mass_upper_bound}. In a word, we have that $\expect_{\gD_1} \ls \sP (G_H^{\gD_1}) \rs \leq \min \{1,  (2 \vert \gS \vert H) / (em) \}$.

Then we have that
\begin{align*}
    V^{\piE} - \expect \ls V^{\piail} \rs &\leq 2 \expect \ls \sum_{s \in \gS} \labs \widetilde{P}_H^{\pi_E}(s, a^{1}) - P_H^{\piE}(s, a^{1}) \rabs \rs
    \\
    &= 2 \expect_{\gD_1} \ls  \sqrt{ \frac{ \sP (G_H^{\gD_1}) \vert \gS \vert}{|\gD_1^c|}  } \rs
    \\
    &\leq 2 \sqrt{ \frac{ \expect_{\gD_1} \ls \sP (G_H^{\gD_1}) \rs \vert \gS \vert}{|\gD_1^c|}  }
    \\
    &\leq \min \lb 2 \sqrt{\frac{2 \vert \gS \vert}{m}}, 4 \sqrt{\frac{\vert \gS \vert H}{e m^2}} \rb.
\end{align*}
which translates into sample complexity of $\min \{ \gO \lp \vert \gS \vert / \varepsilon^2  \rp, \gO ( \vert \gS \vert \sqrt{H} / \varepsilon  )  \}$.

Second, we prove the sample complexity required to achieve a small policy value gap with high probability. Notice that $\sum_{\tr_H \in \gD_1^c} X (\tr_{H}) / |\gD_1^c|$ is an empirical estimation of $\sP^{\piE}(E_{H}^{s})$. By Chernoff's bound in \cref{lemma:chernoff_bound}, with probability at least $1 - \delta/(2|\gS|)$ with $\delta \in (0, 1)$ (over the randomness of the dataset $\gD_1^c$), for each $s \in \gS$,
\begin{align*}
    &\quad \labs \frac{  \sum_{\tr_H \in \gD_1^c}  \indict\{ \tr_H (s_H, a_H) = (s, a^{1}), \tr_H \not\in \Tr_H^{\gD_1}  \} }{|\gD_1^c|} - \sum_{\tr_H \notin \Tr_H^{\gD_1}} \sP^{\piE}(\tr_H) \indict\lb  \tr_H (s_H, a_H) = (s, a^{1})  \rb  \rabs
    \\
    &\leq \sqrt{\sP^{\piE}(E_{H}^{s})  \frac{3 \log (4 |\gS| /\delta)}{m}}.
\end{align*}
By union bound, with probability at least $1-\delta/2$, we have
\begin{align*}
    \labs  V^{\piE} - V^{\piail} \rabs &\leq 2\sum_{s \in \gS} \labs \widetilde{P}^{\piE}_H (s, a^{1}) - P^{\piE}_H (s, a^{1})  \rabs
    \\
    &\leq  2\sum_{s \in \gS} \sqrt{\sP^{\piE}(E_{H}^{s})  \frac{3 \log (4 |\gS| /\delta)}{m}}
    \\
    &\leq 2 \sqrt{ \lp \sum_{s \in \gS} \sP^{\piE}(E_{H}^{s}) \rp  \frac{3 |\gS| \log (4 |\gS| /\delta)}{m}},
\end{align*}
where the last inequality follows the Cauchy–Schwarz inequality. With \eqref{eq:probability_equation}, with probability at least $1-\delta/2$ (over the randomness of $\gD_1^c$), we have
\begin{align*}
    \labs  V^{\piE} - V^{\piail} \rabs \leq 2 \sqrt{ \sP (G_H^{\gD_1})  \frac{3 |\gS| \log (4 |\gS| /\delta)}{m}} .
\end{align*}
On the one hand, with probability of 1, $\sP (G_H^{\gD_1}) \leq 1$. On the other hand, notice that
\begin{align*}
    \sP (G_H^{\gD_1}) \leq \sum_{h=1}^{H} \sum_{s \in \gS} P^{\piE}_h(s) \indict\lb s \notin \gS_h(\gD_1)  \rb.
\end{align*}
Furthermore, by \cref{lemma:missing_mass_one_step}, with probability at least $1-\delta/2$,
\begin{align*}
    \sum_{h=1}^{H} \sum_{s \in \gS} P^{\piE}_h(s) \indict\lb s \notin \gS_h(\gD_1)  \rb \leq \frac{8|\gS| H}{9m} + \frac{6 \sqrt{|\gS|} H \log(2H/\delta)}{m}.
\end{align*}
In a word, with probability at least $1-\delta/2$ (over the randomness of $\gD_1$), it holds that
\begin{align*}
    \sP (G_H^{\gD_1}) \leq \min \lb 1,  \frac{8|\gS| H}{9m} + \frac{6 \sqrt{|\gS|} H \log(2H/\delta)}{m} \rb.
\end{align*}
By union bound, with probability at least $1-\delta$, we have 
\begin{align*}
    \labs  V^{\piE} - V^{\piail} \rabs &\leq 2\sqrt{ \min \lb 1, \frac{8|\gS| H}{9m} + \frac{6 \sqrt{|\gS|} H \log(2H/\delta)}{m}  \rb \frac{3 |\gS| \log (4 |\gS| H/\delta)}{m}}
    \\
    &\leq 2\min \lb \sqrt{\frac{3 |\gS| \log (4 |\gS| H/\delta)}{m}}, \frac{ |\gS| \sqrt{H}}{m} \log^{1/2}\lp \frac{4|\gS| H}{\delta}  \rp \sqrt{ \frac{8}{3} + 18 \log (2H/\delta)  } \rb,
\end{align*}
which translates to sample complexity of $\min \{ \widetilde{\gO} \lp \vert \gS \vert / \varepsilon^2  \rp, \widetilde{\gO} ( \vert \gS \vert \sqrt{H} / \varepsilon  )  \}$.
\end{proof}

\subsection{Application of FEM with Proposition \ref{prop:connection}}
\label{appendix:discussion_of_prop:connection}

Note the metric (e.g., $\ell_1$-norm) used in the estimation problem (assumption $(b)$) and the optimization problem (assumption $(c)$) is not unique in \cref{prop:connection}. For instance, FEM \citep{pieter04apprentice} uses the $\ell_2$-norm metric in its algorithm but FEM can be also applied under this framework. As a result, the policy value gap becomes $\gO ( \sqrt{\vert \gS \vert \vert \gA \vert} ( \varepsilon_{\text{EST}} + \varepsilon_{\text{RFE}} + H \varepsilon_{\text{AIL}} ) )$.

\begin{claim}
If we apply FEM \citep{pieter04apprentice} and RF-Express \citep{menard20fast-active-learning} in \cref{algo:framework}, the corresponding policy value gap in \cref{prop:connection} is $\gO ( \sqrt{\vert \gS \vert \vert \gA \vert} ( \varepsilon_{\text{EST}} + \varepsilon_{\text{RFE}} + H \varepsilon_{\text{AIL}} ) )$.
\end{claim}

\begin{proof}
To apply FEM under our framework in Algorithm \ref{algo:framework}, the assumption $(b)$ becomes: with probability at least $1-\delta_{\text{EST}}$, 
\begin{align*}
    \sum_{h=1}^H \lnorm \widetilde{P}^{\piE}_h - P^{\piE}_h \rnorm_2 \leq \varepsilon_{\text{EST}}.
\end{align*}
Besides, the assumption $(c)$ becomes: with estimation $\widetilde{P}_h^{\piE} (s, a)$ and transition model $\widehat{\gP}$, the policy $\widebar{\pi}$ output by FEM satisfies
\begin{align*}
    \frac{1}{H} \sum_{h=1}^H \lnorm \widetilde{P}^{\piE}_h - P^{\widebar{\pi}, \widehat{\gP}}_h   \rnorm_2 \leq \min_{\pi \in \Pi} \frac{1}{H} \sum_{h=1}^H \lnorm \widetilde{P}^{\piE}_h - P^{\pi, \widehat{\gP}}_h   \rnorm_2 + \varepsilon_{\mathrm{AIL}}.
\end{align*}
Following the same idea in the proof of Proposition \ref{prop:connection}, we can get that
\begin{align*}
    \left\vert V^{\piE, \gP} - V^{\widebar{\pi}, \gP} \right\vert &\leq \left\vert V^{\piE, \gP} - V^{\widebar{\pi}, \widehat{\gP}} \right\vert + \varepsilon_{\text{RFE}} 
    \\
    &\leq \sum_{h=1}^H \lnorm P^{\piE, \gP}_h  - P^{\widebar{\pi}, \widehat{\gP}}_h  \rnorm_1 + \varepsilon_{\text{RFE}}
    \\
    &\leq \sum_{h=1}^H \lnorm P^{\piE, \gP}_h  - \widetilde{P}^{\piE}_h  \rnorm_1 + \sum_{h=1}^H \lnorm \widetilde{P}^{\piE}_h  - P^{\widebar{\pi}, \widehat{\gP}}_h  \rnorm_1 + \varepsilon_{\text{RFE}}.
\end{align*}
For an arbitrary vector $x \in \reals^n$, we have that $\lnorm x \rnorm_2 \leq \lnorm x \rnorm_1 \leq \sqrt{n} \lnorm x \rnorm_2$. Then we show that
\begin{align*}
    \lnorm P^{\piE, \gP}_h  - \widetilde{P}^{\piE}_h  \rnorm_1 \leq \sqrt{\vert \gS \vert \vert \gA \vert} \lnorm P^{\piE, \gP}_h  - \widetilde{P}^{\piE}_h  \rnorm_2 \leq \sqrt{\vert \gS \vert \vert \gA \vert} \varepsilon_{\text{EST}}.
\end{align*}
Then we continue to consider the policy value gap.
\begin{align*}
    \left\vert V^{\piE, \gP} - V^{\widebar{\pi}, \gP} \right\vert &\leq \sum_{h=1}^H \lnorm \widetilde{P}^{\piE}_h  - P^{\widebar{\pi}, \widehat{\gP}}_h  \rnorm_1 + \sqrt{\vert \gS \vert \vert \gA \vert} \varepsilon_{\text{EST}} + \varepsilon_{\text{RFE}}
    \\
    &\leq \sqrt{\vert \gS \vert \vert \gA \vert} \sum_{h=1}^H \lnorm \widetilde{P}^{\piE}_h  - P^{\widebar{\pi}, \widehat{\gP}}_h  \rnorm_2 + \sqrt{\vert \gS \vert \vert \gA \vert} \varepsilon_{\text{EST}} + \varepsilon_{\text{RFE}}
    \\
    &\leq \sqrt{\vert \gS \vert \vert \gA \vert} \lp \min_{\pi \in \Pi} \sum_{h=1}^H \lnorm \widetilde{P}^{\piE}_h  - P^{\pi, \widehat{\gP}}_h  \rnorm_2 + H \varepsilon_{\text{AIL}} \rp + \sqrt{\vert \gS \vert \vert \gA \vert} \varepsilon_{\text{EST}} + \varepsilon_{\text{RFE}}.
\end{align*}
The last inequality holds since that FEM performs $\ell_2$-norm projection with $\widetilde{P}^{\piE}_h (s, a)$ and $\widehat{\gP}$ up to an error of $\varepsilon_{\text{AIL}}$. Then we have that
\begin{align*}
    &\quad \left\vert V^{\piE, \gP} - V^{\widebar{\pi}, \gP} \right\vert \\
    &\leq \sqrt{\vert \gS \vert \vert \gA \vert} \sum_{h=1}^H \lnorm \widetilde{P}^{\piE}_h  - P^{\piE, \widehat{\gP}}_h  \rnorm_2 + \sqrt{\vert \gS \vert \vert \gA \vert} \lp H \varepsilon_{\text{AIL}} +  \varepsilon_{\text{EST}} \rp + \varepsilon_{\text{RFE}}
    \\
    &\leq \sqrt{\vert \gS \vert \vert \gA \vert} \lp \sum_{h=1}^H \lnorm \widetilde{P}^{\piE}_h  - P^{\piE, \gP}_h  \rnorm_2 + \sum_{h=1}^H \lnorm P^{\piE, \gP}_h  - P^{\piE, \widehat{\gP}}_h  \rnorm_2  \rp + \sqrt{\vert \gS \vert \vert \gA \vert} \lp H \varepsilon_{\text{AIL}} +  \varepsilon_{\text{EST}} \rp + \varepsilon_{\text{RFE}}
    \\
    &\leq \sqrt{\vert \gS \vert \vert \gA \vert} \lp \varepsilon_{\mathrm{EST}} + \sum_{h=1}^H \lnorm P^{\piE, \gP}_h  - P^{\piE, \widehat{\gP}}_h  \rnorm_2  \rp + \sqrt{\vert \gS \vert \vert \gA \vert} \lp H \varepsilon_{\text{AIL}} +  \varepsilon_{\text{EST}} \rp + \varepsilon_{\text{RFE}}
    \\
    &\leq \sqrt{\vert \gS \vert \vert \gA \vert} \sum_{h=1}^H \lnorm P^{\piE, \gP}_h  - P^{\piE, \widehat{\gP}}_h  \rnorm_1 + \sqrt{\vert \gS \vert \vert \gA \vert} \lp H \varepsilon_{\text{AIL}} +  2 \varepsilon_{\text{EST}} \rp + \varepsilon_{\text{RFE}}
    \\
    &\leq \sqrt{\vert \gS \vert \vert \gA \vert} \lp H \varepsilon_{\text{AIL}} +  2 \varepsilon_{\text{EST}} + \varepsilon_{\text{RFE}} \rp + \varepsilon_{\text{RFE}}.
\end{align*}
In the last inequality, we use the dual representation of $\ell_1$-norm and policy value. Furthermore, $\widehat{\gP}$ satisfies that for any policy $\pi \in \Pi$ and reward $r \in \gS \times \gA \rar [0, 1]$, $\vert V^{\pi, \gP, r} - V^{\pi, \widehat{\gP}, r} \vert \leq \varepsilon_{\text{RFE}}$. 

\end{proof}

\begin{rem}
Note that the additional factor $\sqrt{|\gS||\gA|}$ is partially caused by the $\ell_2$-norm. In particular, the original assumption in FEM \citep{pieter04apprentice} is that there exists some $w_h$ such that $r_h(s, a) = w_h^{\top} \phi_h(s, a)$. When $\phi_h(s, a)$ is the one-hot feature used in the tabular MDP in this paper, $w_h(s, a) = r_h(s, a)$. According to our assumption that $r_h(s, a) \in [0, 1]$, such an $w_h$ satisfies $\Vert w_h \Vert_2 \leq \sqrt{|\gS| |\gA|}$, which is different from the assumption $\Vert w_h \Vert_2 \leq 1$ in \citep{pieter04apprentice}. However, this mismatch may not be a big issue since the concentration rate for $\ell_2$-norm metric is faster than $\ell_1$-norm when the estimation error is small.
\end{rem}

\subsection{From Regret Guarantee to Sample Complexity Guarantee}
\label{appendix:from_regret_to_pac}

\citet{shani21online-al} proved a regret guarantee for their OAL algorithm. In particular, \citet{shani21online-al} showed that with probability at least $1 - \delta^\prime$, we have 
\begin{align}   \label{eq:oal_regret}
     \sum_{k=1}^{K}  V^{\piE} - V^{\pi_k}  \leq  \widetilde{\gO}\lp \sqrt{H^4 |\gS|^2 |\gA|K} + \sqrt{H^3 |\gS| |\gA| K^2 /m} \rp,
\end{align}
where $\pi^{k}$ is the policy obtained at episode $k$, $K$ is the number of interaction episodes, and $m$ is the number of expert trajectories. We would like to comment that the second term in \eqref{eq:oal_regret} involves the statistical estimation error about the expert policy. Furthermore, this term reduces to $\widetilde{\gO}(\sqrt{H^2 |\gS| K^2 /m})$ under the assumption that the expert policy is deterministic.

To further convert this regret guarantee to the sample complexity guarantee considered in this paper, we can apply Markov's inequality as suggested by \citep{jin18qlearning}. Concretely, let $\widebar{\pi}$ be the policy that randomly chosen from $\{\pi^{1}, \pi^{2}, \cdots, \pi^{K}\}$ with equal probability, then we have 
\begin{align*}
    \sP \lp V^{\piE} - V^{\widebar{\pi}} \geq \varepsilon \rp \leq \frac{1}{\varepsilon} \expect \ls \frac{1}{K} \sum_{k=1}^{K}  V^{\piE} - V^{\pi_k} \rs \leq \frac{1}{\varepsilon} \lp \widetilde{\gO} \lp \sqrt{ \frac{H^4 |\gS|^2 |\gA|}{K} } + \sqrt{H^2 |\gS| /m}\rp + \delta^\prime H  \rp,
\end{align*}
Therefore, if we set $\delta^{\prime} = \varepsilon \delta / (3H)$, and
\begin{align*}
    K = \widetilde{\gO} \lp \frac{H^4 |\gS|^2 |\gA|}{\varepsilon^2 \delta^2} \rp, \quad m = \widetilde{\gO} \lp \frac{H^2 |\gS|}{\varepsilon^2} \rp,
\end{align*}
we obtain that $\sP ( V^{\piE} - V^{\widebar{\pi}} \geq \varepsilon ) \leq \delta $. As commented in \citep{menard20fast-active-learning}, this transformation leads to a worse dependence on failure probability $\delta$, but the sample complexity dependence on other terms does not change.

\subsection{VAIL with State Abstraction}
\label{appendix:discussion_of_function_approximation}

Notice that the upper bounds of sample complexity discussed in this paper depend on the state space size $\vert \gS \vert$. Besides, the lower bounds \citep[Theorem 6.1, 6.2]{rajaraman2020fundamental} imply that the dependence of $\vert \gS \vert$ is inevitable for all imitation learning algorithms if no additional information is provided. In this part, we discuss that if provided with a set of state abstractions \citep{li2006towards}, how to avoid the dependence of $\vert \gS \vert$ on sample complexity. In particular, state abstractions correspond to the function approximation with a series of piecewise constant functions \citep{chen2019information}.

To be more specific, assume we have access to a set of state abstractions $\{ \phi_h \}_{h=1}^H$, where $\phi_h: \gS \rightarrow \Phi$ for each $h \in [H]$ and $\Phi$ is abstract state space. The size of abstract state space is much smaller than that of original state space, i.e., $|\Phi| \ll |\gS|$. We assume that the state abstractions satisfy the following reward-irrelevant condition \citep{li2006towards}.

\begin{asmp}[Reward-irrelevant]
\label{asmp:reward_irrelevant}
Consider the set of state abstractions $\{ \phi_h \}_{h=1}^H$. For each $h \in [H]$, for any $s^{1}, s^{2} \in \gS$ such that $\phi_h (s^{1}) = \phi_h (s^{2})$, $\forall a \in \gA$, $r_h (s^{1}, a) = r_h (s^{2}, a)$. 
\end{asmp}

We highlight that the reward-irrelevant condition is important for AIL to avoid the dependence on $\vert \gS \vert$; see also \citep{pieter04apprentice, syed07game, liu2021provably}. In particular, the bottleneck of the sample complexity of AIL methods is the estimation of $P^{\piE}_h(s, a)$.  With the set of state abstractions, we can calculate the expert policy value as
\begin{align*}
    V^{\piE} = \sum_{h=1}^H \sum_{(s, a) \in \gS \times \gA} r_h (s, a)  P^{\piE}_h (s, a) = \sum_{h=1}^H \sum_{(x, a) \in \Phi \times \gA} r^{\phi}_h (x, a) P^{\piE, \phi}_h (x, a), 
\end{align*}
where $P^{\pi, \phi}_h$ is the \dquote{abstract state-action distribution}: $P^{\pi, \phi}_h (x, a) = \sP^{\piE} (\phi_h (s_h) = x, a_h = a) = \sum_{s \in \phi_h^{-1} (x)} P^{\pi}_h (s, a)$. With the above formulation, to estimate the expert policy value, we can estimate the \emph{abstract} state-action distribution rather than the original state-action distribution. This may remove the dependence on $\vert \gS \vert$. We present our conjecture as follows.

\begin{conj}[Sample Complexity of \textsf{VAIL} with State Abstraction]
\label{theorem:worst_case_sample_complexity_of_vail_state_abstraction}
For any tabular and episodic MDP, suppose that there exists a set of known state abstractions $\{ \phi_h: \gS \rightarrow \Phi \}_{h=1}^H$ satisfying \cref{asmp:reward_irrelevant}. To obtain an $\varepsilon$-optimal policy (i.e., $V^{\piE} - \expect[ V^{\piail}] \leq \varepsilon$), in expectation,  \textsf{VAIL} requires at most $\gO(|\Phi |H^2/\varepsilon^2)$ expert trajectories. 
\end{conj}

\section{Open Problem}
\label{appendix:open_problem}

To better understand the role of our research, we discuss the following related open problems.

\textbf{Function Approximation.} In this paper, we focus on the tabular MDPs, in which the one-hot feature is used. As a result, the lower bounds in \citep{rajaraman2020fundamental} imply that the dependence of $|\gS|$ is inevitable for all imitation learning algorithms if no additional information is provided. We note that MDPs with \dquote{low-rank} structures allow algorithms (including BC and AIL) to use function approximation to obtain better sample complexity; refer to the related work discussed in \cref{appendix:review_of_previous_work}. Typically, the refined sample complexity is expected to depend on the inherent dimension $d$ rather than $|\gS|$. This direction is orthogonal to our research since we mainly compare algorithms in terms of the horizon $H$, which is usually unrelated to function approximation. Nevertheless, it is interesting to extend our results under the function approximation setting; see the discussion in \cref{appendix:discussion_of_function_approximation}. 

\textbf{Representation Learning.} We firmly believe that GAIL beats FEM and GTAL for MuJoCo tasks because the former uses deep neural networks to learn a good feature representation while the latter uses pre-specified features. However, these methods do not make a big difference under the tabular MDPs in terms of the sample complexity. It would be valuable to investigate this direction under the feature learning framework (see e.g., \citep{uehara2021representation}).

\section{Technical Lemmas and Proofs}
\label{appendix:technical_lemmas}

\subsection{Basic Technical Lemmas}

\begin{lem}   \label{lemma:policy_dual_value}
For tabular and episodic MDP, we have that 
\begin{align*}
    V^{\pi} = \sum_{h=1}^{H} \sum_{(s, a) \in \gS \times \gA} P^{\pi}_h(s, a) r_h(s, a).
\end{align*}
\end{lem}

\begin{proof}
The proof is direct from the definition. 
\end{proof}

\begin{lem}   \label{lemma:state_dist_discrepancy}
For any tabular and episodic MDP, considering two policies $\pi$ and $\pi^\prime$, let $P^{\pi}_h (\cdot)$ and $P^{\pi}_h (\cdot, \cdot)$ denote the state distribution and state-action distribution induced by $\pi$ in time step $h$, respectively. Then we have that
\begin{itemize}
    \item 
    $
    \lnorm P^{\pi}_h (\cdot) - P^{\pi^\prime}_h (\cdot)  \rnorm_{1} \leq \sum_{\ell=1}^{h-1} \expect_{s \sim P^{\pi^\prime}_{\ell} (\cdot)} [\lnorm \pi_{\ell} (\cdot|s) - \pi^\prime_{\ell} (\cdot|s)   \rnorm_1]
    $ when $h \geq 2$.
    \item $\lnorm P^{\pi}_h (\cdot, \cdot) - P^{\pi^\prime}_h (\cdot, \cdot)  \rnorm_{1} \leq \lnorm P^{\pi}_h (\cdot) - P^{\pi^\prime}_h (\cdot)  \rnorm_{1} + \expect_{s \sim P^{\pi^\prime}_h (\cdot)} \ls \lnorm \pi_h (\cdot|s)  - \pi^\prime_h (\cdot|s) \rnorm_1 \rs $.
\end{itemize}
\end{lem}

\begin{proof}
We prove the first statement. It is direct to obtain that $\lnorm P^{\pi}_1 (\cdot) - P^{\pi^\prime}_1 (\cdot)  \rnorm_{1} = \lnorm \rho (\cdot) - \rho (\cdot)  \rnorm_{1}= 0$.

When $h \geq 2$, for any $\ell $ where $1 < \ell \leq h$, we prove the following recursion format.

\begin{align*}
    \lnorm P^{\pi}_\ell (\cdot) - P^{\pi^\prime}_\ell (\cdot)  \rnorm_{1} \leq \lnorm P^{\pi}_{\ell-1} (\cdot) - P^{\pi^\prime}_{\ell-1} (\cdot)  \rnorm_{1} +  \expect_{s \sim P^{\pi^\prime}_{\ell-1} (\cdot)} \ls \lnorm \pi_{\ell-1} (\cdot|s) - \pi^\prime_{\ell-1} (\cdot|s)   \rnorm_1 \rs.
\end{align*}
With the \dquote{transition flow equation}, we have
\begin{align*}
    &\quad \lnorm P^{\pi}_\ell (\cdot) - P^{\pi^\prime}_\ell (\cdot)  \rnorm_{1}
    \\
    &= \sum_{s \in \gS} \labs P^{\pi}_\ell (s) - P^{\pi^\prime}_\ell (s)  \rabs
    \\
    &= \sum_{s \in \gS} \labs \sum_{(s^\prime, a^\prime) \in \gS \times \gA} P^{\pi}_{\ell-1} (s^\prime) \pi_{\ell-1} (a^\prime|s^\prime) P_{\ell}(s|s^\prime, a^\prime) - \sum_{(s^\prime, a^\prime) \in \gS \times \gA} P^{\pi^\prime}_{\ell-1} (s^\prime) \pi^\prime_{\ell-1} (a^\prime|s^\prime) P_{\ell}(s|s^\prime, a^\prime)  \rabs
    \\
    &= \sum_{s \in \gS} \Bigg| \sum_{(s^\prime, a^\prime) \in \gS \times \gA} \lp P^{\pi}_{\ell-1} (s^\prime) - P^{\pi^\prime}_{\ell-1} (s^\prime) \rp \pi_{\ell-1} (a^\prime|s^\prime) P_{\ell}(s|s^\prime, a^\prime) 
    \\
    &\quad + \sum_{(s^\prime, a^\prime) \in \gS \times \gA} P^{\pi^\prime}_{\ell-1} (s^\prime) \lp \pi_{\ell-1} (a^\prime|s^\prime) -\pi^\prime_{\ell-1} (a^\prime|s^\prime) \rp  P_{\ell}(s|s^\prime, a^\prime)  \Bigg|
    \\
    &\leq \sum_{s \in \gS} \sum_{(s^\prime, a^\prime) \in \gS \times \gA} \labs P^{\pi}_{\ell-1} (s^\prime) - P^{\pi^\prime}_{\ell-1} (s^\prime) \rabs \pi_{\ell-1} (a^\prime|s^\prime) P_{\ell}(s|s^\prime, a^\prime)
    \\
    &\quad + \sum_{s \in \gS} \sum_{(s^\prime, a^\prime) \in \gS \times \gA} P^{\pi^\prime}_{\ell-1} (s^\prime) \labs \pi_{\ell-1} (a^\prime|s^\prime) -\pi^\prime_{\ell-1} (a^\prime|s^\prime) \rabs  P_{\ell}(s|s^\prime, a^\prime)
    \\
    &= \lnorm P^{\pi}_{\ell-1} (\cdot) - P^{\pi^\prime}_{\ell-1} (\cdot)  \rnorm_{1} + \expect_{s \sim P^{\pi^\prime}_{\ell-1} (\cdot)} \ls \lnorm \pi_{\ell-1} (\cdot|s) - \pi^\prime_{\ell-1} (\cdot|s)   \rnorm_1 \rs,  
\end{align*}
where we obtain the recursion format. Applying the recursion format with $\lnorm P^{\pi}_1 (\cdot) - P^{\pi^\prime}_1 (\cdot)  \rnorm_{1}=0$ finishes the proof of the first statement.

We continue to prove the second statement.
\begin{align*}
    &\quad \lnorm P^{\pi}_h (\cdot, \cdot) - P^{\pi^\prime}_h (\cdot, \cdot)  \rnorm_{1} 
    \\
    &= \sum_{(s, a) \in \gS \times \gA} \labs P^{\pi}_h (s, a) - P^{\pi^\prime}_h (s, a) \rabs
    \\
    &= \sum_{(s, a) \in \gS \times \gA} \labs P^{\pi}_h (s) \pi_h (a|s) - P^{\pi^\prime}_h (s) \pi^\prime_h (a|s) \rabs
    \\
    &= \sum_{(s, a) \in \gS \times \gA} \labs \lp P^{\pi}_h (s) - P^{\pi^\prime}_h (s) \rp \pi_h (a|s) + P^{\pi^\prime}_h (s) \lp \pi_h (a|s) -   \pi^\prime_h (a|s) \rp \rabs
    \\
    &\leq \sum_{(s, a) \in \gS \times \gA} \labs  P^{\pi}_h (s) - P^{\pi^\prime}_h (s) \rabs \pi_h (a|s) + \sum_{(s, a) \in \gS \times \gA} P^{\pi^\prime}_h (s) \labs \pi_h (a|s) -   \pi^\prime_h (a|s) \rabs
    \\
    &= \lnorm P^{\pi}_h (\cdot) - P^{\pi^\prime}_h (\cdot)  \rnorm_{1} + \expect_{s \sim P^{\pi^\prime}_h (\cdot)} \ls \lnorm \pi_h (\cdot|s)  - \pi^\prime_h (\cdot|s) \rnorm_1 \rs, 
\end{align*}
which proves the second statement.
\end{proof}

\begin{lem}
\label{lem:unique_opt_solution_condition}
Consider the optimization problem: $\min_{x \in [0, 1]^n} f (x) := \sum_{i=1}^m f_i (x)$, where $f_i : [0, 1]^n \rightarrow \reals, \forall i \in [m]$. Suppose that 1) there exists $k \in [m]$ such that $x^*$ is the unique optimal solution to $\min_{x \in [0, 1]^n} f_k (x)$; 2) for each $j \in [m], j \not= k$, $x^*$ is the optimal solution to $\min_{x \in [0, 1]^n} f_j (x)$. Then, $x^*$ is the unique optimal solution to $\min_{x \in [0, 1]^n} f (x)$.
\end{lem}

\begin{proof}
Since $x^*$ is the unique optimal solution to $\min_{x \in [0, 1]^n} f_k (x)$, we have that $\forall x \in [0, 1]^n, x \not= x^*$, $f_k (x^*) < f_k (x)$. Furthermore, for each $j \in [m], j \not= k$, recall that $x^*$ is the optimal solution to $\min_{x \in [0, 1]^n} f_j (x)$. We have that
\begin{align*}
    \forall j \in [m], j \not= k, \forall x \in [0, 1]^n, x \not= x^*, f_j (x^*) \leq f_j (x).  
\end{align*}
Then we derive that $\forall x \in [0, 1]^n, x \not= x^*$, $f(x^*) < f(x)$ and $x$ is the unique optimal solution to $\min_{x \in [0, 1]^n} f (x)$.
\end{proof}

\begin{lem}
\label{lem:n_vars_opt_greedy_structure}
Consider the optimization problem $\min_{x_1, \cdots, x_n} f (x_1, \cdots, x_n)$. Suppose that $x^* = (x^*_1, \cdots, x^*_n)$ is the optimal solution, then $\forall i \in [n]$, $x^*_i$ is the optimal solution to $\min_{x_i} F (x_i) := f (x^*_1, \cdots, x_i, \cdots, x^*_n)$. 
\end{lem}

\begin{proof}
The proof is based on contradiction. Suppose that the original statement is not true. There exists $\widetilde{x}_i \not= x^*_i$ such that
\begin{align*}
    F (\widetilde{x}_i) < F (x^*_i). 
\end{align*}
Consider $\widetilde{x} = (x^*_1, \cdots,\widetilde{x}_i, \cdots, x^*_n)$ which differs from $x^*$ on the $i$-th component. Then we have that
\begin{align*}
    f (x^*_1, \cdots,\widetilde{x}_i, \cdots, x^*_n) = F (\widetilde{x}_i) < F (x^*_i) = f (x^*_1, \cdots, x^*_i, \cdots, x^*_n), 
\end{align*}
which contradicts with the fact that $x^* = (x^*_1, \cdots, x^*_n)$ is the optimal solution to $\min_{x_1, \cdots, x_n} f (x_1, \cdots, x_n)$. Hence, the original statement is true.  
\end{proof}

\begin{lem}
\label{lem:single_variable_opt}
For any constants $a, c \geq 0$, we define the function $f(x) = \vert c - ax \vert - ax$. Consider the optimization problem $\min_{x \in [0, 1]} f(x) $, then $x^* = 1$ is the optimal solution. 
\end{lem}

\begin{proof}
We assume that $x^* = 1$ is not the optimal solution. There exists $\widetilde{x}^* \in [0, 1)$ such that $f(\widetilde{x}^*) < f (x^*)$. That is
\begin{align*}
    \vert c - a\widetilde{x}^* \vert - a\widetilde{x}^* - \vert c - a \vert + a < 0, 
\end{align*}
which implies that $\vert c - a \vert - \vert c - a\widetilde{x}^* \vert > a - a \widetilde{x}^*$. On the other hand, according to the inequality that $\labs p \rabs - \labs q \rabs \leq \labs p-q \rabs$ for $p, q \in \reals$, we have
\begin{align*}
    \vert c - a \vert - \vert c - a\widetilde{x}^* \vert \leq \vert a\widetilde{x}^* - a \vert = a- a\widetilde{x}^*,
\end{align*}
where the last equality follows that $\widetilde{x}^* < 1$. We construct a contradiction. Therefore, the original statement is true.

\end{proof}

\begin{lem}
\label{lem:single_variable_opt_condition}
For any constants $a, c > 0$, we define the function $f(x) = \vert c - ax \vert - ax$. Consider the optimization problem $\min_{x \in [0, 1]} f(x) $, If $x^*$ is the optimal solution, then $x^* > 0$. Furthermore, if $c < a$, then the optimal solutions are $x^* \in [c/a, 1]$.  
\end{lem}

\begin{proof}
To begin with, we prove the first statement. The proof is based on contradiction. We assume that $x = 0$ is the optimal solution. We compare the function value on $x=1$ and $x=0$.
\begin{align*}
    f (0) - f(1) &= c +a - \labs c-a \rabs > 0,
\end{align*}
where the strict inequality follows that $a, c>0$. We obtain that $f(1) < f(0)$, which contradicts with the assumption that $x=0$ is the optimal solution. Therefore, the original statement is true and we finish the proof.

Then we prove the second statement. It is easy to see that
\begin{align*}
    f (x) = \begin{cases}
      c-2ax & x \in [0, \frac{c}{a}), \\
      -c & x \in [\frac{c}{a}, 1].
    \end{cases}
\end{align*}
$f(x)$ is continuous piece-wise linear function. $f(x)$ is strictly decreasing when $x \in [0, c / a)$ and is constant when $x \in [c / a, 1]$. Therefore, we can get that the optimal solutions are $x^* \in [c / a, 1]$. 
\end{proof}

\begin{lem}
\label{lem:single_variable_regularity}
For any constants $a > 0$ and $c \geq 0$, we define the function $f(x) = \vert c - ax \vert - ax$. For any $x \leq \min\{ c/a, 1 \}$, we have $f(x) - f(1) = 2 a ( \min\{ c/a, 1 \} - x)$. 
\end{lem}

\begin{proof}
We consider two cases: $c \geq a$ and $c < a$. When $c \geq a$, the function $f (x)$ at $[0, 1]$ is formulated as $f(x) = c - 2ax$. For any $x \leq \min\{ c/a, 1 \} = 1 $, $f(x) - f(1) = 2a (1-x) = 2 a ( \min\{ c/a, 1 \} - x)$. On the other hand, when $c < a$, the function $f (x)$ at $[0, 1]$ is formulated as
\begin{align*}
    f(x) = \begin{cases}
      c-2ax & x \in [0, \frac{c}{a}), \\
      -c & x \in [\frac{c}{a}, 1].
    \end{cases}
\end{align*}
For any $x \leq \min\{ c/a, 1 \} = c/a$, $f(x) - f(1) = 2 a (c/a-x) = 2 a ( \min\{ c/a, 1 \} - x) $. Therefore, we finish the proof.
\end{proof}

\begin{lem}
\label{lem:mn_variables_opt_unique}
Consider that $A = (a_{ij}) \in \reals^{m \times n}, c \in \reals^{m}, d \in \reals^{n}$ where $a_{ij} > 0$, $\sum_{i=1}^m c_i \geq \sum_{i=1}^m \sum_{j=1}^n a_{ij}$ and for each $j \in [n]$, $\sum_{i=1}^m a_{ij} = d_j$. Consider the following optimization problem:
\begin{align*}
    \min_{x \in [0, 1]^n}f (x) := \lnorm c - A x \rnorm_{1} - d^{\top} x = \sum_{i=1}^m \labs c_i - \sum_{j=1}^n a_{ij} x_j \rabs - \sum_{j=1}^n d_j x_j.
\end{align*}
Then $x^* = \mathbf{1}$ is the unique optimal solution, where $\mathbf{1}$ is the vector that each element is 1. 
\end{lem}

\begin{proof}
For $x = (x_1, \cdots, x_n)$, the function $f(x)$ is formulated as
\begin{align*}
    f (x) = \sum_{i=1}^m \labs c_i - \sum_{j=1}^n a_{ij} x_j \rabs - \sum_{j=1}^n d_j x_j.
\end{align*}
The proof is based on contradiction. We assume that the original statement is not true and there exists $x = (x_1, \cdots, x_n) \not= \mathbf{1}$ such that $x$ is the optimal solution. Let $k \in [n]$ denote some index where $x_k \not= 1$. We construct $\widetilde{x} = \lp \widetilde{x}_1, \cdots, \widetilde{x}_n  \rp \in [0, 1]^n$ in the following way.
\begin{align*}
    \widetilde{x}_j = x_j, \forall j \in [n] \setminus \{k\}, \quad \widetilde{x}_k = 1. 
\end{align*}
We compare the function value of $x$ and $\widetilde{x}$.
\begin{align*}
    f(\widetilde{x}) - f(x) &= \sum_{i=1}^m \lp \labs c_i - \sum_{j=1}^n a_{ij} \widetilde{x}_j \rabs - \labs c_i - \sum_{j=1}^n a_{ij} x_j  \rabs  \rp - d_k (1-x_k)
    \\
    &< \sum_{i=1}^m \lp a_{ik} (1-x_k) \rp - d_k (1-x_k) = 0.
\end{align*}
Here the strict inequality follows the statement that there exists $i^* \in [m]$ such that
\begin{align*}
    \labs c_{i^*} - \sum_{j=1}^n a_{i^* j} \widetilde{x}_j \rabs - \labs c_{i^*} - \sum_{j=1}^n a_{i^* j} x_j  \rabs  <  \labs \lp c_{i^*} - \sum_{j=1}^n a_{i^* j} \widetilde{x}_j \rp - \lp c_{i^*} - \sum_{j=1}^n a_{i^* j} x_j \rp  \rabs=  a_{i^* k} (1-x_k).
\end{align*}
We will prove this statement later. As for $i \in [m], i \not= i^*$, with the inequality that $\labs a \rabs - \labs b \rabs \leq \labs a-b \rabs$ for $a, b \in \reals$, we obtain that
\begin{align*}
    \labs c_i - \sum_{j=1}^n a_{ij} \widetilde{x}_j \rabs - \labs c_i - \sum_{j=1}^n a_{ij} x_j  \rabs  \leq \labs \lp c_i - \sum_{j=1}^n a_{ij} \widetilde{x}_j \rp - \lp c_i - \sum_{j=1}^n a_{ij} x_j \rp  \rabs=  a_{ik} (1-x_k) .
\end{align*}
Hence the strict inequality holds and we construct $\widetilde{x}$ such that $f(\widetilde{x}) < f(x) $, which contradicts with the assumption that $x$ is the optimal solution. Therefore, we prove that the original statement is true and finish the proof.

Now we proceed to prove the statement that there exists $i^* \in [m]$ such that
\begin{align*}
    \labs c_{i^*} - \sum_{j=1}^n a_{i^* j} \widetilde{x}_j \rabs - \labs c_{i^*} - \sum_{j=1}^n a_{i^* j} x_j  \rabs  <  \labs \lp c_{i^*} - \sum_{j=1}^n a_{i^* j} \widetilde{x}_j \rp - \lp c_{i^*} - \sum_{j=1}^n a_{i^* j} x_j \rp  \rabs
\end{align*}
We also prove this statement by contradiction. We assume that for all $i \in [m]$,
\begin{align*}
    \labs c_i - \sum_{j=1}^n a_{ij} \widetilde{x}_j \rabs - \labs c_i - \sum_{j=1}^n a_{ij} x_j  \rabs  \geq \labs \lp c_i - \sum_{j=1}^n a_{ij} \widetilde{x}_j \rp - \lp c_i - \sum_{j=1}^n a_{ij} x_j \rp  \rabs
\end{align*}
According to the inequality that $\labs a \rabs - \labs b \rabs \leq \labs a-b \rabs$ for $a, b \in \reals$, we have
\begin{align*}
    \forall i \in [m], \labs c_i - \sum_{j=1}^n a_{ij} \widetilde{x}_j \rabs - \labs c_i - \sum_{j=1}^n a_{ij} x_j  \rabs  = \labs \lp c_i - \sum_{j=1}^n a_{ij} \widetilde{x}_j \rp - \lp c_i - \sum_{j=1}^n a_{ij} x_j \rp  \rabs
\end{align*}
Furthermore, consider the inequality $\labs a \rabs - \labs b \rabs \leq \labs a-b \rabs$ for $a, b \in \reals$. Notice that the equality holds iff $(b-a) b \leq 0$. Hence we have that
\begin{align*}
     \forall i \in [m], \lp a_{ik} (1-x_k)  \rp  \lp  c_i - \sum_{j=1}^n a_{ij} x_j \rp   \leq 0. 
\end{align*}
Since $\lp a_{ik} (1-x_k)  \rp > 0$, we obtain that
\begin{align*}
    \forall i \in [m], c_i - \sum_{j=1}^n a_{ij} x_j \leq 0. 
\end{align*}
This implies that
\begin{align*}
    \sum_{i=1}^m c_i \leq \sum_{i=1}^m \sum_{j=1}^n a_{ij} x_j < \sum_{i=1}^m \sum_{j=1}^n a_{ij} \leq \sum_{i=1}^m c_i,  
\end{align*}
where the strict inequality follows that $x_k < 1$ and $a_{ij} > 0$. The last inequality follows the assumption of \cref{lem:mn_variables_opt_unique}. Here we find a contradiction that $\sum_{i=1}^m c_i < \sum_{i=1}^m c_i$ and hence the original statement is true.
\end{proof}

\begin{lem}
\label{lem:mn_variables_opt_regularity}
Under the same conditions in \cref{lem:mn_variables_opt_unique}, for any $x \in [0,1]^{n}$, we have that
\begin{align*}
    f (x) - f(x^*) \geq \sum_{j=1}^n \min_{i \in [m]} \{a_{ij} \} (1-x_{j}),
\end{align*}
where $x^* = \mathbf{1}$, which is the vector that each element is 1. 
\end{lem}

\begin{proof}
Recall that $f (x) = \sum_{i=1}^m \vert c_i - \sum_{j=1}^n a_{ij} x_j \vert - \sum_{j=1}^n d_j x_j$. We first claim that when $x \in [0, 1]^n$, $c_i - \sum_{j=1}^n a_{ij} x_j < 0$ does not hold simultaneously for all $i \in [m]$. We prove this claim via contradiction. Assume that there exists $x \in [0, 1]^n$ such that $c_i - \sum_{j=1}^n a_{ij} x_j < 0, \forall i \in [m]$. Then we have that
\begin{align*}
    \sum_{i=1}^m c_i < \sum_{i=1}^m \sum_{j=1}^n a_{ij} x_j \overset{(1)}{\leq} \sum_{i=1}^m \sum_{j=1}^n a_{ij} \overset{(2)}{\leq} \sum_{i=1}^m c_i.   
\end{align*}
The inequality $(1)$ follows that $A > 0$ and $x \in [0, 1]^n$ and the inequality $(2)$ follows that original assumption of \cref{lem:mn_variables_opt_regularity}. Thus we constructs a contradiction, which implies that the original claim is true.

Let $x_{p:q}$ be the shorthand of $(x_p, x_{p+1}, \cdots, x_{q})$ for any $1 \leq p \leq q \leq n$. With telescoping, we have that
\begin{align*}
    f(x) - f(x^*) = \sum_{j=1}^n f(x^*_{1:j-1}, x_{j:n}) - f(x^*_{1:j}, x_{j+1:n}).
\end{align*}
Note that $f(x^*_{1:j-1}, x_{j:n}) $ and $f(x^*_{1:j}, x_{j+1:n})$ only differ in the $j$-th variable. For each $j \in [n]$, with fixed $x^*_1, \cdots, x^*_{j-1}, x_{j+1}, \cdots, x_n \in [0, 1]$, we define one-variable function $F_j (t) = f (x^*_{1:j-1}, t, x_{j+1:n}), \forall t \in [0, 1]$. Notice that $F_j (t)$ is also a continuous piece-wise linear function.

On the one hand, $F_j (t)$ is differentiable at any interior point $t_0$ and it holds that 
\begin{align*}
    F_j^\prime (t_0) = -d_j + \sum_{i=1}^m \indict \lb \lp c_i - \sum_{k=1}^{j-1} a_{ik} x^*_k - a_{ij} t_0 - \sum_{k=j+1}^n a_{ik} x_k \rp < 0  \rb a_{ij} \leq - \min_{i \in [m]} \{ a_{ij} \}.
\end{align*}

The last inequality follows that $\forall x \in [0, 1]^n, c_i - \sum_{j=1}^n a_{ij} x_j \leq 0$ does not hold simultaneously for all $i \in [m]$ and $d_j = \sum_{i=1}^m a_{ij}$. On the other hand, the number of boundary points of $F_j (t)$ is $m$ at most. Let $b_j^1, b_j^2, \cdots, b_j^{n_j}$ denote the boundary point of $F_j (t)$ when $t \in [x_j, x_j^*]$. With fundamental theorem of calculus, we have that
\begin{align*}
   f(x) - f(x^*) &= \sum_{j=1}^n f(x^*_{1:j-1}, x_{j:n}) - f(x^*_{1:j}, x_{j+1:n})
   \\
   &= \sum_{j=1}^n F_j (x_j) - F_j (x_j^*)
   \\
   &= \sum_{j=1}^n \lp F_j (x_j) - F_j (b_j^1) + \sum_{k=1}^{n_j-1} F_j (b_j^k) - F_j (b_j^{k+1}) + F_j (b_j^{n_j}) - F(x_j^*)   \rp  
   \\
   &= - \sum_{j=1}^n \lp \int_{x_j}^{b_j^1} F_j^\prime (t) dt + \sum_{k=1}^{n_j-1} \int_{b_j^k}^{b_j^{k+1}} F_j^\prime (t) dt + \int_{b_j^{n_j}}^{x_j^*} F_j^\prime (t) dt    \rp 
   \\
   &\geq  \sum_{j=1}^n \min_{i \in [m]} \{ a_{ij} \} \lp x_j^* - x_j  \rp =  \sum_{j=1}^n \min_{i \in [m]} \{ a_{ij}\} \lp 1 - x_j  \rp.
\end{align*}

\end{proof}

\subsection{Proof of Technical Lemmas in Appendix \ref{appendix:proof_generalization_ail}, \ref{appendix:proof_beyond_vanilla_ail}, and \ref{appendix:discussion} }

\subsubsection{Proof of Lemma \ref{lemma:regret_of_ogd}}
\label{appendix:proof_lemma_regret_of_ogd}
\begin{proof}

Lemma \ref{lemma:regret_of_ogd} is a direct consequence of the regret bound of online gradient descend \citep{shalev12online-learning}. To apply such a regret bound, we need to verify that 1) the iterate norm $\lnorm w \rnorm_2$ has an upper bound; 2) the gradient norm $\Vert \nabla_{w}  f^{(t)}(w) \Vert_2$ also has an upper bound. The first point is easy to show, i.e., $\lnorm w \rnorm_2 \leq \sqrt{H |\gS| |\gA|}$ by the condition that $w \in \gW = \{ w: \Vert w \Vert_{\infty} \leq 1 \}$. For the second point, let $\widetilde{P}^{1}_h$ and $\widetilde{P}^{2}_h$ be the first and the second part in $\widetilde{P}^{\piE}_h$ defined in \eqref{eq:new_estimator}. Then, 
\begin{align*}
    \lnorm \nabla_{w} f^{(t)} (w) \rnorm_{2} &= \sqrt{\sum_{h=1}^H \sum_{(s, a) \in \gS \times \gA} \lp P^{\pi^{(t)}}_h (s, a) - \widetilde{P}^{\piE}_h (s, a) \rp^2 }
    \\
    &= \sqrt{\sum_{h=1}^H \sum_{(s, a) \in \gS \times \gA} \lp P^{\pi^{(t)}}_h (s, a) - \widetilde{P}^{1}_h(s, a) - \widetilde{P}^{2}_h(s, a) \rp^2 } 
    \\
    &\leq \sqrt{\sum_{h=1}^H  3 \sum_{(s, a) \in \gS \times \gA}   \lp P^{\pi^{(t)}}_h (s, a) \rp^2 + \lp \widetilde{P}^{1}_h(s, a) \rp^2 + \lp \widetilde{P}^{2}_h(s, a)  \rp^2 }
    \\
     &\leq \sqrt{\sum_{h=1}^H  3  \lp \lnorm P^{\pi^{(t)}}_h \rnorm_1 + \lnorm \widetilde{P}^{1}_h \rnorm_1 + \lnorm \widetilde{P}^{2}_h \rnorm_1  \rp }
    \\
    &\leq 2\sqrt{ H },
\end{align*}
where the first inequality follows $(a+b+c)^2 \leq 3(a^2+b^2+c^2)$ and the second inequality is based on that $ x ^2 \leq \vert x \vert$ if $0 \leq x \leq 1$.

Invoking Corollary 2.7 in \citep{shalev12online-learning} with $B = \sqrt{H |\gS| |\gA|}$ and $L = 2 \sqrt{H}$ finishes the proof. 
\end{proof}

\subsubsection{Proof of Lemma \ref{lemma:approximate-minimax}}
\label{appendix:proof_lemma_approximate_minimax}

\begin{proof}
With the dual representation of $\ell_1$-norm, we have
\begin{align*}
    \min_{\pi \in \Pi} \sum_{h=1}^H \lnorm P^{\pi}_h - \widetilde{P}^{\piE}_h \rnorm_{1} = \min_{\pi \in \Pi} \max_{w \in \gW} \sum_{h=1}^H \sum_{(s, a) \in \gS \times \gA} w_{h} (s, a) \lp \widetilde{P}^{\piE}_h(s, a) - P^{\pi}_h (s, a) \rp. 
\end{align*}
Since the above objective is linear w.r.t both $w$ and $P^\pi_h$, invoking the minimax theorem \citep{bertsekas2016nonlinear} yields
\begin{align*}
    &\quad \min_{\pi \in \Pi} \max_{w \in \gW} \sum_{h=1}^H \sum_{(s, a) \in \gS \times \gA} w_{h} (s, a) \lp \widetilde{P}^{\piE}_h(s, a) - P^{\pi}_h (s, a) \rp
    \\
    &= \max_{w \in \gW} \min_{\pi \in \Pi} \sum_{h=1}^H \sum_{(s, a) \in \gS \times \gA} w_{h} (s, a) \lp \widetilde{P}^{\piE}_h(s, a) - P^{\pi}_h (s, a) \rp
    \\
    &= - \min_{w \in \gW} \max_{\pi \in \Pi} \sum_{h=1}^H \sum_{(s, a) \in \gS \times \gA} w_h (s, a) \lp P^{\pi}_h (s, a) -  \widetilde{P}^{\piE}_h(s, a) \rp, 
\end{align*}
where the last step follows the property that for a function $f$, $- \max_{x} f(x) = \min_{x} - f(x)$. Therefore, we have
\begin{align} \label{eq:l1_dual_representation}
    \min_{\pi \in \Pi} \sum_{h=1}^H \lnorm P^{\pi}_h  - \widetilde{P}^{\piE}_h \rnorm_{1} = - \min_{w \in \gW} \max_{\pi \in \Pi} \sum_{h=1}^H \sum_{(s, a) \in \gS \times \gA} w_h (s, a) \lp P^{\pi}_h (s, a) -  \widetilde{P}^{\piE}_h(s, a) \rp.
\end{align}
Then we consider the term $\min_{w \in \gW} \max_{\pi \in \Pi} \sum_{h=1}^H \sum_{(s, a) \in \gS \times \gA} w_h (s, a) \lp P^{\pi}_h (s, a) -  \widetilde{P}^{\piE}_h(s, a) \rp$.
\begin{align*}
    &\quad \min_{w \in \gW} \max_{\pi \in \Pi} \sum_{h=1}^H \sum_{(s, a) \in \gS \times \gA} w_h (s, a) \lp P^{\pi}_h (s, a) -  \widetilde{P}^{\piE}_h(s, a) \rp
    \\
    &\leq \max_{\pi \in \Pi} \sum_{h=1}^H \sum_{(s, a) \in \gS \times \gA} \lp \frac{1}{T} \sum_{t=1}^T w^{(t)}_h (s, a) \rp \lp P^{\pi}_h (s, a) -  \widetilde{P}^{\piE}_h(s, a) \rp
    \\
    &\leq \frac{1}{T} \sum_{t=1}^T \max_{\pi \in \Pi} \sum_{h=1}^H \sum_{(s, a) \in \gS \times \gA} w^{(t)}_h (s, a) \lp P^{\pi}_h (s, a) -  \widetilde{P}^{\piE}_h(s, a) \rp. 
\end{align*}
At iteration $t$, $\pi^{(t)}$ is the approximately optimal policy regarding reward function $w^{(t)}$ with an optimization error of $\varepsilon_{\mathrm{opt}}$. Then we obtain that
\begin{align*}
    &\quad \frac{1}{T} \sum_{t=1}^T \max_{\pi \in \Pi} \sum_{h=1}^H \sum_{(s, a) \in \gS \times \gA} w^{(t)}_h (s, a) \lp P^{\pi}_h (s, a) -  \widetilde{P}^{\piE}_h(s, a) \rp
    \\
    &\leq \frac{1}{T} \sum_{t=1}^T \sum_{h=1}^H \sum_{(s, a) \in \gS \times \gA} w^{(t)}_h (s, a) \lp P^{\pi^{(t)}}_h (s, a) -  \widetilde{P}^{\piE}_h(s, a) \rp + \varepsilon_{\mathrm{opt}}.
\end{align*}
Applying Lemma \ref{lemma:regret_of_ogd} yields that
\begin{align*}
    &\quad \frac{1}{T} \sum_{t=1}^T \sum_{h=1}^H \sum_{(s, a) \in \gS \times \gA} w^{(t)}_h (s, a) \lp P^{\pi^{(t)}}_h (s, a) -  \widetilde{P}^{\piE}_h(s, a) \rp
    \\
    & \leq \min_{w \in \gW} \frac{1}{T} \sum_{t=1}^T \sum_{h=1}^H \sum_{(s, a) \in \gS \times \gA} w_h (s, a) \lp P^{\pi^{(t)}}_h (s, a) -  \widetilde{P}^{\piE}_h(s, a) \rp + 2H \sqrt{ \frac{2 |\gS| |\gA|}{T} }
    \\
    &= \min_{w \in \gW}  \sum_{h=1}^H \sum_{(s, a) \in \gS \times \gA} w_h (s, a) \lp \frac{1}{T} \sum_{t=1}^T P^{\pi^{(t)}}_h (s, a) -  \widetilde{P}^{\piE}_h(s, a) \rp + 2H \sqrt{ \frac{2 |\gS| |\gA|}{T} }
    \\
    &= \min_{w \in \gW}  \sum_{h=1}^H \sum_{(s, a) \in \gS \times \gA} w_h (s, a) \lp P^{\widebar{\pi}}_h (s, a) -  \widetilde{P}^{\piE}_h(s, a) \rp + 2H \sqrt{ \frac{2 |\gS| |\gA|}{T} }.
\end{align*}
Note that $\widebar{\pi}$ is induced by the mean state-action distribution, i.e., $\widebar{\pi}_h (a|s) = \widebar{P}_h(s, a) / \sum_{a} \widebar{P}_h(s, a)$, where $\widebar{P}_h (s, a) = {1}/{T} \cdot \sum_{t=1}^T P^{\pi^{(t)}}_h (s, a)$. Based on Proposition 3.1 in \citep{ho2016gail}, we have that $P^{\widebar{\pi}}_h (s, a) = \widebar{P}_h (s, a)$, and hence the last equation holds. Combined with \eqref{eq:l1_dual_representation}, we have that
\begin{align*}
    &\quad \min_{\pi \in \Pi} \sum_{h=1}^H \lnorm P^{\pi}_h - \widetilde{P}^{\piE}_h \rnorm_{1}
    \\
    &\geq - \min_{w \in \gW}  \sum_{h=1}^H \sum_{(s, a) \in \gS \times \gA} w_h (s, a) \lp P^{\widebar{\pi}}_h (s, a) -  \widetilde{P}^{\piE}_h(s, a) \rp - 2H \sqrt{ \frac{2 |\gS| |\gA|}{T} } - \varepsilon_{\mathrm{opt}}
    \\
    &= \max_{w \in \gW} \sum_{h=1}^H \sum_{(s, a) \in \gS \times \gA} w_h (s, a) \lp  \widetilde{P}^{\piE}_h(s, a) - P^{\widebar{\pi}}_h (s, a)  \rp - 2H \sqrt{ \frac{2 |\gS| |\gA|}{T} } - \varepsilon_{\mathrm{opt}}
    \\
    &= \sum_{h=1}^H \lnorm \widetilde{P}^{\piE}_h - P^{\widebar{\pi}}_h \rnorm_{1} - 2H \sqrt{ \frac{2 |\gS| |\gA|}{T}} - \varepsilon_{\mathrm{opt}},
\end{align*}
where the last step again utilizes the dual representation of $\ell_1$-norm. We complete the proof.

\end{proof}

\subsubsection{Proof of Lemma \ref{lemma:sample_complexity_of_new_estimator_known_transition}}
\label{appendix:proof_lemma_sample_complexity_of_new_estimator_known_transition}

\begin{proof}
Recall the definition of the estimator $\widetilde{P}_h^{\piE}$
\begin{align*} 
\widetilde{P}_h^{\piE}  (s, a) =   { \sum_{\tr_h \in \Tr_h^{\gD_1} } \sP^{\piE}(\tr_h) \indict\lb \tr_h(s_h, a_h) = (s, a)\rb} + {\frac{  \sum_{\tr_h \in \gD_1^c}  \indict\{ \tr_h (s_h, a_h) = (s, a), \tr_h \not\in \Tr_h^{\gD_1}  \} }{|\gD_1^c|}}.
\end{align*}
Our target is to upper bound the estimation error of $\widetilde{P}_h^{\piE} \in \real^{|\gS| \times |\gA|}$:
\begin{align*}
    \sum_{h=1}^H \lnorm \widetilde{P}^{\piE}_h - P^{\piE}_h  \rnorm_{1} = \sum_{h=1}^{H} \sum_{(s, a) \in \gS \times \gA} \labs \widetilde{P}_h^{\pi_E}(s, a) - P_h^{\piE}(s, a) \rabs,
\end{align*}
where $\widetilde{P}_h^{\piE}$ is defined in \eqref{eq:new_estimator}:
\begin{align*}
  \widetilde{P}_h^{\piE}(s, a) =   { \sum_{\tr_h \in \Tr_h^{\gD_1} } \sP^{\piE}(\tr_h) \indict\lb \tr_h(s_h, a_h) = (s, a)\rb}  + {\frac{  \sum_{\tr_h \in \gD_1^c}  \indict\{ \tr_h (s_h, a_h) = (s, a), \tr_h \not\in \Tr_h^{\gD_1}  \} }{|\gD_1^c|}}.
\end{align*}
Recall that $\Tr_h^{\gD_1}$ is the set of trajectories along which each state has been visited in $\gD_1$ up to time step $h$. Similarly, for $P_h^{\piE}$, we have the following decomposition in \eqref{eq:key_decomposition}:
\begin{align*}
    P_h^{\piE}(s, a) = \sum_{\tr_h \in \Tr_h^{\gD_1}} \sP^{\piE}(\tr_h) \indict\lb  \tr_h (s_h, a_h) = (s, a)  \rb + \sum_{\tr_h \notin \Tr_h^{\gD_1}} \sP^{\piE}(\tr_h) \indict\lb  \tr_h (s_h, a_h) = (s, a)  \rb.
\end{align*}

Consequently, we obtain for any $(s, a) \in \gS \times \gA, h \in [H]$, 
\begin{align}
     &\quad  \labs \widetilde{P}_h^{\pi_E}(s, a) - P_h^{\piE}(s, a) \rabs \nonumber \\
     &=   \labs \widetilde{P}_h^{\pi_E}(s, a) - \lp \sum_{\tr_h \in \Tr_h^{\gD_1}} \sP^{\piE}(\tr_h) \indict\lb  \tr_h (s_h, a_h) = (s, a)  \rb + \sum_{\tr_h \notin \Tr_h^{\gD_1}} \sP^{\piE}(\tr_h) \indict\lb  \tr_h (s_h, a_h) = (s, a)  \rb \rp  \rabs \nonumber \\
     &= \labs \frac{  \sum_{\tr_h \in \gD_1^c}  \indict\{ \tr_h (s_h, a_h) = (s, a), \tr_h \not\in \Tr_h^{\gD_1}  \} }{|\gD_1^c|} - \sum_{\tr_h \notin \Tr_h^{\gD_1}} \sP^{\piE}(\tr_h) \indict\lb  \tr_h (s_h, a_h) = (s, a)  \rb  \rabs, \label{eq:new_estimation_probability_error}
\end{align}
where the last equation is based on the fact that the first term in $\widetilde{P}_h^{\pi_E}(s, a)$ and $P_h^{\piE}(s, a)$ is identical. As a result, the estimation error is caused by the unknown expert actions in trajectories that does not fully match with any trajectory in $\gD_1$. Then we obtain that
\begin{align*}
    &\quad \sum_{h=1}^H \lnorm \widetilde{P}^{\piE}_h - P^{\piE}_h  \rnorm_{1}
    \\
    &\leq \sum_{h=1}^H \sum_{(s, a) \in \gS \times \gA} \labs \frac{  \sum_{\tr_h \in \gD_1^c}  \indict\{ \tr_h (s_h, a_h) = (s, a), \tr_h \not\in \Tr_h^{\gD_1}  \} }{|\gD_1^c|} - \sum_{\tr_h \notin \Tr_h^{\gD_1}} \sP^{\piE}(\tr_h) \indict\lb  \tr_h (s_h, a_h) = (s, a)  \rb  \rabs. 
\end{align*}
Next, we invoke Lemma A.12 in \cite{rajaraman2020fundamental} to upper bound the term in RHS. 
\begin{lem}[Lemma A.12 in \cite{rajaraman2020fundamental}]
Fix $\delta \in (0, \min\{1, H/5 \})$, with probability at least $1-\delta$,
\begin{align*}
    & \quad \sum_{h=1}^H \sum_{(s, a) \in \gS \times \gA} \labs \frac{  \sum_{\tr_h \in \gD_1^c}  \indict\{ \tr_h (s_h, a_h) = (s, a), \tr_h \not\in \Tr_h^{\gD_1}  \} }{|\gD_1^c|} - \sum_{\tr_h \notin \Tr_h^{\gD_1}} \sP^{\piE}(\tr_h) \indict\lb  \tr_h (s_h, a_h) = (s, a)  \rb  \rabs
    \\
    &\precsim \frac{\vert \gS \vert H^{3/2}}{m} \lp 1 + \frac{3 \log \lp 2 \vert \gS \vert H / \delta \rp}{\sqrt{\vert \gS \vert}} \rp \sqrt{\log \lp \frac{2 \vert \gS \vert H}{\delta} \rp}. 
\end{align*}
\end{lem}

Then, for any fixed $\delta \in (0, \min\{1, H/5 \})$, with probability at least $1-\delta$,
\begin{align*}
    \sum_{h=1}^H \lnorm \widetilde{P}^{\piE}_h - P^{\piE}_h  \rnorm_{1} \leq \frac{\vert \gS \vert H^{3/2}}{m} \lp 1 + \frac{3 \log \lp 2 \vert \gS \vert H / \delta \rp}{\sqrt{\vert \gS \vert}} \rp \sqrt{\log \lp \frac{2 \vert \gS \vert H }{\delta} \rp}, 
\end{align*}
When $m \succsim \vert \gS \vert H^{3/2} \log \lp H \vert \gS \vert / \delta \rp / \varepsilon$, we have that $\sum_{h=1}^H \Vert \widetilde{P}^{\piE}_h - P^{\piE}_h  \Vert_{1} \leq \varepsilon$. 

\end{proof}

\subsubsection{Proof of Lemma \ref{lemma:sample_complexity_of_new_estimator_unknown_transition}}
\label{appendix:proof_lemma_sample_complexity_of_new_estimator_unknown_transition}

\begin{proof}
We aim to upper bound the estimation error.
\begin{align*}
    \sum_{h=1}^H \lnorm \widetilde{P}^{\piE}_h - P^{\piE}_h  \rnorm_{1}.
\end{align*}
Recall the definition of the estimator $\widetilde{P}^{\piE}_h(s, a)$.
\begin{align*}
    \widetilde{P}_h^{\piE} (s, a) := \frac{\sum_{\tr_h \in \gD^\prime_{\mathrm{env}}} \indict \lb \tr_h (s_h, a_h) = (s, a), \tr_h \in \Tr_h^{\gD_1} \rb }{ | \gD^\prime_{\mathrm{env}} |} + \frac{\sum_{\tr_h \in \gD_1^{c}} \indict \lb \tr_h (s_h, a_h) = (s, a), \tr_h \notin \Tr_h^{\gD_1} \rb}{| \gD_1^{c} |}.
\end{align*}

Similarly, we utilize the decomposition of $P_h^{\piE}(s, a)$ as we have done in the proof of Lemma \ref{lemma:sample_complexity_of_new_estimator_known_transition}.  
\begin{align*}
    P_h^{\piE}(s, a) = \sum_{\tr_h \in \Tr_h^{\gD_1}} \sP^{\piE}(\tr_h) \indict\lb  \tr_h (s_h, a_h) = (s, a)  \rb + \sum_{\tr_h \notin \Tr_h^{\gD_1}} \sP^{\piE}(\tr_h) \indict\lb  \tr_h (s_h, a_h) = (s, a)  \rb.
\end{align*}
Then, for any $h \in [H]$ and $(s, a) \in \gS \times \gA$, we have
\begin{align*}
    &\quad \labs \widetilde{P}^{\piE}_h(s, a) - P^{\piE}_h (s, a)  \rabs
    \\
    &\leq \labs \frac{\sum_{\tr_h \in \gD^\prime_{\mathrm{env}}} \indict \lb \tr_h (s_h, a_h) = (s, a), \tr_h \in \Tr_h^{\gD_1} \rb }{ | \gD^\prime_{\mathrm{env}} |} - \sum_{\tr_h \in \Tr_h^{\gD_1}} \sP^{\piE}(\tr_h) \indict\lb  \tr_h (s_h, a_h) = (s, a)  \rb  \rabs 
    \\
    & \quad + \labs \frac{\sum_{\tr_h \in \gD_1^{c}} \indict \lb \tr_h (s_h, a_h) = (s, a), \tr_h \notin \Tr_h^{\gD_1} \rb}{| \gD_1^{c} |} - \sum_{\tr_h \notin \Tr_h^{\gD_1}} \sP^{\piE}(\tr_h) \indict\lb  \tr_h (s_h, a_h) = (s, a)  \rb  \rabs .
\end{align*}
Thus, we can upper bound the estimation error.
\begin{align*}
    &\quad \sum_{h=1}^H \lnorm \widetilde{P}^{\piE}_h - P^{\piE}_h  \rnorm_{1}
    \\
    &\leq \underbrace{\sum_{h=1}^H \sum_{(s, a) \in \gS \times \gA} \labs \frac{\sum_{\tr_h \in \gD^\prime_{\mathrm{env}}} \indict \lb \tr_h (s_h, a_h) = (s, a), \tr_h \in \Tr_h^{\gD_1} \rb }{ | \gD^\prime_{\mathrm{env}} |} - \sum_{\tr_h \in \Tr_h^{\gD_1}} \sP^{\piE}(\tr_h) \indict\lb  \tr_h (s_h, a_h) = (s, a)  \rb  \rabs}_{\text{Error A}}
    \\
    &\quad + \underbrace{\sum_{h=1}^H \sum_{(s, a) \in \gS \times \gA} \labs \frac{\sum_{\tr_h \in \gD_1^{c}} \indict \lb \tr_h (s_h, a_h) = (s, a), \tr_h \notin \Tr_h^{\gD_1} \rb}{| \gD_1^{c} |} - \sum_{\tr_h \notin \Tr_h^{\gD_1}} \sP^{\piE}(\tr_h) \indict\lb  \tr_h (s_h, a_h) = (s, a)  \rb  \rabs}_{\text{Error B}}. 
\end{align*}
We first analyze the term $\text{Error A}$. Recall that dataset $\gD^\prime_{\mathrm{env}}$ is collected by the policy $\pi \in \Pi_{\text{BC}} \lp \gD_{1} \rp$ with $|\gD^\prime_{\mathrm{env}}| = n^\prime$, and $\sum_{\tr_h \in \gD^\prime_{\mathrm{env}}} \indict \{ \tr_h (s_h, a_h) = (s, a), \tr_h \in \Tr_h^{\gD_1} \} /  | \gD^\prime_{\mathrm{env}} |$ is a maximum likelihood estimator for $\sum_{\tr_h \in \Tr_h^{\gD_1}} \sP^{\piE}(\tr_h) \indict \lb  \tr_h (s_h, a_h) = (s, a)  \rb $. Let ${E^\prime}^{s}_h$ be the event that $\tr_h$ agrees with expert policy at state $s$ in time step $h$ and appears in $\Tr_h^{\gD_1}$. Formally, 
\begin{align*}
    {E^\prime}_h^{s} = \indict\{\tr_h (s_h, a_h) = (s, \piE_h (s)) \cap \tr_h \in \mathbf{Tr}^{\gD_1}_h \}.
\end{align*}
Then we apply Chernoff's bound to upper bound the term $\text{Error A}$.

\begin{lem}[Chernoff's bound \citep{vershynin2018high}]   \label{lemma:chernoff_bound}
Let $\widebar{X} = {1}/{n} \cdot \sum_{i=1}^{n} X_i$, where $X_i$ is a Bernoulli random variable with $\sP(X_i = 1) = p_i$ and $\sP(X_i = 0) = 1 - p_i$ for $i \in [n]$. Furthermore, assume these random variables are independent. Let $\mu = \expect[\widebar{X}] = {1}/{n} \cdot \sum_{i=1}^{n} p_i$. Then for $0 < t \leq 1$, 
\begin{align*}
    \sP\lp  \labs  \widebar{X} - \mu \rabs  \geq t \mu  \rp \leq 2 \exp\lp -\frac{\mu n t^2}{3}  \rp.
\end{align*}
\end{lem}

By Lemma \ref{lemma:chernoff_bound}, for each $s \in \gS$ and $h \in [H]$, with probability at least $1 - \frac{\delta}{2 |\gS| H}$ over the randomness of $\gD ^\prime$, we have
\begin{align*}
    &\quad \labs \frac{\sum_{\tr_h \in \gD^\prime_{\mathrm{env}}} \indict \lb \tr_h (s_h, a_h) = (s, \piE_h(s)), \tr_h \in \Tr_h^{\gD_1} \rb }{ | \gD^\prime_{\mathrm{env}} |}  - \sum_{\tr_h \in \Tr_h^{\gD_1}} \sP^{\piE}(\tr_h) \indict\lb  \tr_h (s_h, a_h) = (s, \piE_h (s))  \rb  \rabs
    \\
    &\leq \sqrt{ \sP^{\piE} \lp {E^\prime}^{s}_h  \rp  \frac{3 \log \lp 4 |\gS| H / \delta \rp}{n^\prime}}.
\end{align*}
By union bound, with probability at least $1-{\delta}/{2}$ over the randomness of $\gD^\prime_{\mathrm{env}}$, we have
\begin{align*}
    \text{Error A} &\leq  \sum_{h=1}^H \sum_{s \in \gS} \sqrt{ \sP^{\piE} \lp {E^\prime}^{s}_h  \rp  \frac{3 \log \lp 4 |\gS| H / \delta \rp}{n^\prime}}
    \\
    &\leq \sum_{h=1}^H \sqrt{|\gS|} \sqrt{\sum_{s \in \gS} \sP^{\piE} \lp {E^\prime}^{s}_h  \rp  \frac{3 \log \lp 4 |\gS| H / \delta \rp}{n^\prime} }
\end{align*}
The last inequality follows the Cauchy-Schwartz inequality. It remains to upper bound $\sum_{s \in \gS}  \sP^{\piE}(E_{h}^{s})$ for all $h \in [H]$. To this end, we define the event ${G^\prime}_h^{\gD_1}$ that expert policy $\piE$ visits states covered in $\gD_1$ up to time step $h$. Formally, ${G^\prime}_h^{\gD_1} = \indict\{ \forall h^{\prime} \leq h,  s_{h^{\prime}} \in \gS_{h^{\prime}} (\gD_1) \}$, where $\gS_{h}(\gD_1)$ is the set of states in $\gD_1$ in time step $h$. Then, for all $h \in [H]$, we have \begin{align*}
    \sum_{s \in \gS} \sP^{\piE} \lp {E^\prime}_h^{s}  \rp = \sP^{\piE}({G^\prime}_h^{\gD_1}) \leq \sP({G^\prime}_1^{\gD_1}).
\end{align*}
The last inequality holds since ${G^\prime}_h^{\gD_1} \subseteq {G^\prime}_1^{\gD_1}$ for all $h \in [H]$. Then we have that
\begin{align*}
    \text{Error A} \leq H \sqrt{\frac{3 |\gS| \log \lp 4 |\gS| H / \delta \rp}{n^\prime}}.
\end{align*}
When the interaction complexity satisfies that $n^\prime \succsim \frac{| \gS | H^{2}}{\varepsilon^2} \log\lp  \frac{|\gS| H}{\delta} \rp$, with probability at least $1-\frac{\delta}{2}$ over the randomness of $\gD^\prime$, we have $\text{Error A} \leq \frac{\varepsilon}{2}$. For the term $\text{Error B}$, we have analyzed it in the proof of Lemma \ref{lemma:sample_complexity_of_new_estimator_known_transition}. When the expert sample complexity satisfies that $m \succsim \frac{|\gS| H^{3/2}}{\varepsilon} \log \lp \frac{|\gS| H}{\delta} \rp$, with probability at least $1-\frac{\delta}{2}$ over the randomness of $\gD$, we have $\text{Error B} \leq \frac{\varepsilon}{2}$. Applying union bound finishes the proof.
\end{proof}

\subsubsection{Proof of Lemma \ref{lemma:ail_policy_ail_objective_equals_expert_policy_ail_objective}}
\label{appendix:proof_lemma:ail_policy_ail_objective_equals_expert_policy_ail_objective}

\begin{proof}
For $h, h^\prime \in [H], h \leq h^\prime$, we use $\pi_{h:h^\prime}$ denote the shorthand of $\lp \pi_h, \pi_{h+1}, \cdots, \pi_{h^\prime} \rp$. From \cref{prop:ail_general_reset_cliff}, we have that $\forall h \in [H-1], s \in \goodS, \piail_{h} (a^{1}|s) = \piE_{h} (a^{1}|s) = 1$. Hence, $\piail$ and $\piE$ never visit bad states. Furthermore, notice that for any time step $h \in [H]$, $\text{Loss}_{h} (\pi)$ only depends on $\pi_{1:h}$. Therefore, we have
\begin{align*}
    \sum_{h=1}^{H-1} \text{Loss}_{h} (\piail) = \sum_{h=1}^{H-1} \text{Loss}_{h} (\piE).
\end{align*}
It remains to prove that $\text{Loss}_{H} (\piail) = \text{Loss}_{H} (\piE)$. From \cref{lem:n_vars_opt_greedy_structure}, fixing $\piail_{1:H-1}$, $\piail_{H}$ is the optimal solution to \textsf{VAIL}'s objective. With fixed $\piail_{1:H-1}$, $\sum_{h=1}^{H-1} \text{Loss}_{h} (\piail)$ is independent of $\piail_{H}$ and thus
\begin{align*}
    \piail_{H} &\in \argmin_{\pi_H}  \text{Loss}_{H} (\piail_{1:H-1}, \pi_H)
    \\
    &=  \argmin_{\pi_H}  \text{Loss}_{H} (\piE_{1:H-1}, \pi_H)
    \\
    &= \argmin_{\pi_H} \sum_{s \in \gS} \sum_{a \in \gA} \labs \widehat{P}^{\piE}_h(s, a) - P^{\piE}_H (s) \pi_H (a|s)  \rabs
    \\
    &= \argmin_{\pi_H} \sum_{s \in \goodS} \labs \widehat{P}^{\piE}_h(s) - P^{\piE}_H (s) \pi_H (a^{1}|s)  \rabs + P^{\piE}_H (s) \lp 1 - \pi_{H} (a^{1}|s) \rp
    \\
    &= \argmin_{\pi_H} \sum_{s \in \goodS} \labs \widehat{P}^{\piE}_h(s) - P^{\piE}_H (s) \pi_H (a^{1}|s)  \rabs - P^{\piE}_H (s) \pi_{H} (a^{1}|s). 
\end{align*}
In the penultimate equality, we use the facts that 1) for each $s \in \goodS$, we have $\widehat{P}^{\piE}_h(s, a^{1}) = \widehat{P}^{\piE}_h(s)$, and $\widehat{P}^{\piE}_h(s, a) = 0, \forall a \in \gA \setminus \{a^{1}\}$; 2) for each $s \in \badS$, $\widehat{P}^{\piE}_h(s) = P^{\piE}_h (s) = 0$. The last equality follows that $P^{\piE}_H (s)$ is independent of $\pi_H$. Since the optimization variables $\pi_H (a^{1}|s)$ for different $s \in \goodS$ are independent, we can view the above optimization problem for each $\pi_H (a^{1}|s)$ individually.
\begin{align*}
    \piail_{H}(a^{1}|s) = \argmin_{\pi_H (a^{1}|s) \in [0, 1]} \labs \widehat{P}^{\piE}_h(s) - P^{\piE}_H (s) \pi_H (a^{1}|s)  \rabs - P^{\piE}_H (s) \pi_{H} (a^{1}|s).
\end{align*}
By \cref{lem:single_variable_opt}, we have that $\piE_H(a^{1}|s) = 1$ is also the optimal solution like $\piail_{H} (a^{1}|s)$. Therefore, we have that 
\begin{align*}
    \text{Loss}_{H} (\piail_{1:H-1}, \piail_H) = \min_{\pi_H} \text{Loss}_{H} (\piail_{1:H-1}, \pi_H)  = \min_{\pi_H} \text{Loss}_{H} (\piE_{1:H-1}, \pi_H) = \text{Loss}_{H} (\piE_{1:H-1}, \piE_H).  
\end{align*}
Finally, we prove that $f (\piail) = f(\piE)$.
\end{proof}

\subsubsection{Proof of Proposition \ref{prop:ail_general_reset_cliff_approximate_solution}}
\label{appendix:proof_prop:ail_general_reset_cliff_approximate_solution}

\begin{proof}
Suppose that $\piail$ is the optimal solution to \eqref{eq:ail}. Since $\widebar{\pi}$ is $\varepsilon_{\ail}$ optimal, we have that
\begin{align*}
    f (\widebar{\pi}) - f(\piail) \leq \varepsilon_{\ail}.
\end{align*}
By \cref{lemma:ail_policy_ail_objective_equals_expert_policy_ail_objective}, it holds that $f(\piail) = f(\piE)$. Furthermore, with the decomposition of $f(\pi)$, we have 
\begin{align}
\label{eq:ail_objective_pi_bar_minus_piE}
    f (\widebar{\pi}) - f(\piail) = f (\widebar{\pi}) - f(\piE) =  \sum_{h=1}^{H} \text{Loss}_{h} (\widebar{\pi}) - \text{Loss}_{h} (\piE) \leq \varepsilon_{\ail}. 
\end{align}
For any $h, h^\prime \in [H]$ with $h \leq h^\prime$, we use $\pi_{h:h^\prime}$ denote the shorthand of $\lp \pi_h, \pi_{h+1}, \cdots, \pi_{h^\prime} \rp$. Note that $\text{Loss}_{h} (\pi)$ only depends on $\pi_{1:h}$ and thus we have
\begin{align*}
    \sum_{h=1}^{H} \text{Loss}_{h} (\widebar{\pi}_{1:h}) - \text{Loss}_{h} (\piE_{1:h}) \leq \varepsilon_{\ail}.
\end{align*}
We defined a policy set $\Pi^{\text{opt}} = \{ \pi \in \Pi: \forall h \in [H], \exists s \in \goodS, \pi_h (a^{1}|s) > 0 \}$ and note that $\widebar{\pi} \in \Pi^{\text{opt}}$. In the following part, we analyze $\sum_{h=1}^{H} \text{Loss}_{h} (\pi_{1:h}) - \text{Loss}_{h} (\piE_{1:h})$ where $\pi \in \Pi^{\text{opt}}$. For each $h \in [H]$, we have the following key composition by telescoping: 
\begin{align}
\label{eq:sum_telescoping}
\boxed{
    \text{Loss}_{h} (\pi_{1:h}) - \text{Loss}_{h} (\piE_{1:h}) = \sum_{\ell=1}^{h} \text{Loss}_{h} (\pi_{1:\ell}, \piE_{\ell+1:h}) - \text{Loss}_{h} (\pi_{1:\ell-1}, \piE_{\ell:h}). 
}
\end{align}
In the following part, we consider two cases: Case I: $ h<H$ and Case II: $ h = H$.

First, we consider Case I and focus on the term $\text{Loss}_{h} (\pi_{1:\ell}, \piE_{\ell+1:h}) - \text{Loss}_{h} (\pi_{1:\ell-1}, \piE_{\ell:h})$. Under Case I, we consider two situations: $\ell = h$ and $\ell < h$.

\begin{itemize}
    \item When $\ell = h$, we consider the term $\text{Loss}_{h} (\pi_{1:h}) - \text{Loss}_{h} (\pi_{1:h-1}, \piE_{h})$. Note that $\pi_{1:h}$ and $(\pi_{1:h-1}, \piE_{h})$ differ in the policy in time step $h$. Take the policy in time step $h$ as variable and we focus on
    \begin{align*}
        g (\pi_{h}) - g(\piE_{h}),
    \end{align*}
    where $g (\pi_{h}) = \text{Loss}_{h} (\pi_{1:h})$ and $g(\piE_{h}) = \text{Loss}_{h} (\pi_{1:h-1}, \piE_{h})$. We formulate $g (\pi_{h}) = \text{Loss}_{h} (\pi_{1:h})$ as
    \begin{align*}
        g (\pi_{h}) &=  \sum_{(s, a) \in \gS \times \gA} | \widehat{P}^{\piE}_h(s, a) - P^{\pi}_h(s, a)  |
        \\
        &= \sum_{s \in \goodS} \sum_{a \in \gA} \labs \widehat{P}^{\piE}_{h} (s, a) - P^{\pi}_{h} (s) \pi_{h} (a|s)  \rabs + \sum_{s \in \badS} \sum_{a \in \gA} P^{\pi}_{h} (s, a)
        \\
        &= \sum_{s \in \goodS} \lp \labs \widehat{P}^{\piE}_{h} (s, a^1) - P^{\pi}_{h} (s) \pi_h(a^1|s) \rabs + P^{\pi}_{h} (s) \lp 1 - \pi_h(a^1|s)  \rp  \rp + \sum_{s \in \badS} P^{\pi}_{h} (s)  
        \\
        &= \sum_{s \in \goodS}\lp \labs \widehat{P}^{\piE}_{h} (s) - P^{\pi}_{h} (s) \pi_h(a^1|s) \rabs + P^{\pi}_{h} (s) \lp 1 - \pi_h(a^1|s)  \rp   \rp + \sum_{s \in \badS} P^{\pi}_{h} (s). 
    \end{align*}
    Note that $P^{\pi}_{h} (s)$ is independent of the policy in time step $h$. Then we have that
    \begin{align*}
       g (\pi_{h}) - g(\piE_{h})  &= \sum_{s \in \goodS}\lp \labs \widehat{P}^{\piE}_{h} (s) - P^{\pi}_{h} (s) \pi_h(a^1|s) \rabs - P^{\pi}_{h} (s)   \pi_h(a^1|s)     \rp \\
       &\quad - \lp \labs \widehat{P}^{\piE}_{h} (s) 
        \quad - P^{\pi}_{h} (s) \piE_h(a^1|s) \rabs - P^{\pi}_{h} (s)   \piE_h(a^1|s)     \rp. 
    \end{align*}
    For each $s \in \goodS$, we apply \cref{lem:single_variable_opt} and obtain that 
    \begin{align}
    \label{eq:case_one_situation_one_result}
        g (\pi_{h}) - g(\piE_{h}) = \text{Loss}_{h} (\pi_{1:h}) - \text{Loss}_{h} (\pi_{1:h-1}, \piE_{h}) \geq 0.
    \end{align}
    \item When $\ell < h$, we consider the term $\text{Loss}_{h} (\pi_{1:\ell}, \piE_{\ell+1:h}) - \text{Loss}_{h} (\pi_{1:\ell-1}, \piE_{\ell:h})$. Notice that $(\pi_{1:\ell}, \piE_{\ell+1:h})$ and $(\pi_{1:\ell-1}, \piE_{\ell:h})$ only differ in the policy in time step $\ell$. Take the policy in time step $\ell$ as variable and we focus on
    \begin{align*}
        g (\pi_{\ell}) - g(\piE_{\ell}),
    \end{align*}
    where $g (\pi_{\ell}) = \text{Loss}_{h} (\pi_{1:\ell}, \piE_{\ell+1:h})$ and $g (\piE_{\ell}) = \text{Loss}_{h} (\pi_{1:\ell-1}, \piE_{\ell:h})$. We can calculate $g (\pi_{\ell})$ as
    \begin{align*}
    g (\pi_{\ell}) &=  \sum_{(s, a) \in \gS \times \gA} | \widehat{P}^{\piE}_h(s, a) - P^{\pi}_h(s, a)  |
    \\
    &= \sum_{s \in \goodS} \sum_{a \in \gA} \labs \widehat{P}^{\piE}_{h} (s, a) - P^{\pi}_{h} (s) \pi_{h} (a|s)  \rabs + \sum_{s \in \badS} \sum_{a \in \gA} P^{\pi}_{h} (s, a)
    \\
    &= \sum_{s \in \goodS} \labs \widehat{P}^{\piE}_{h} (s, a^1) - P^{\pi}_{h} (s, a^1) \rabs + \sum_{s \in \badS} P^{\pi}_{h} (s)  
    \\
    &= \sum_{s \in \goodS} \labs \widehat{P}^{\piE}_{h} (s) - P^{\pi}_{h} (s) \rabs + \sum_{s \in \badS} P^{\pi}_{h} (s).
    \end{align*}
    With a little abuse of notation, we use $P^{\pi}_h(s, a)$ and $P^{\pi}_{h} (s)$ to denote the distributions induced by $(\pi_{1:\ell}, \piE_{\ell+1:h})$. Similar to the proof of \cref{prop:ail_general_reset_cliff}, with the \dquote{transition flow equation}, we have
    \begin{align*}
        \forall s \in \goodS, P^{\pi}_{h} (s) &= \sum_{s^\prime \in \gS} \sum_{a \in \gA} P^{\pi}_{\ell} (s^\prime) \pi_\ell (a|s^\prime) \sP^{\pi} \lp s_{h} = s |s_\ell = s^\prime, a_\ell = a \rp  
        \\
        &= \sum_{s^\prime \in \goodS} P^{\pi}_{\ell} (s^\prime) \pi_{\ell} (a^{1}|s^\prime) \sP^{\pi} \lp s_{h} = s |s_{\ell} = s^\prime, a_{h} = a^{1} \rp.
    \end{align*}
    Notice that the conditional probability $\sP^{\pi} \lp s_{h} = s |s_{\ell} = s^\prime, a_{h} = a^{1} \rp$ is independent of $\pi_\ell$. Besides, for the visitation probability on bad states in time step $h$, we have
    \begin{align*}
        \sum_{s \in \badS} P^{\pi}_{h} (s) &= \sum_{s \in \badS} P^{\pi}_{\ell} (s) + \sum_{s^\prime \in \goodS} \sum_{a \in \gA \setminus \{a^1 \}} P^{\pi}_{\ell} (s^\prime)  \pi_{\ell} (a|s^\prime)
        \\
        &= \sum_{s \in \badS} P^{\pi}_{\ell} (s) + \sum_{s^\prime \in \goodS} P^{\pi}_{\ell} (s^\prime) \lp 1 - \pi_{\ell} (a^{1}|s^\prime) \rp .
    \end{align*}
    Plugging the above two equations into $g (\pi_{\ell})$ yields that
    \begin{align*}
        g (\pi_{\ell})  &= \sum_{s \in \goodS} \labs \widehat{P}^{\piE}_{h} (s) - \sum_{s^\prime \in \goodS} P^{\pi}_{\ell} (s^\prime) \pi_{\ell} (a^{1}|s^\prime) \sP^{\pi} \lp s_{h} = s |s_{\ell} = s^\prime, a_{\ell} = a^{1} \rp \rabs
        \\
        &\quad + \sum_{s \in \badS} P^{\pi}_{\ell} (s) + \sum_{s^\prime \in \goodS} P^{\pi}_{\ell} (s^\prime) \lp 1 - \pi_{\ell} (a^{1}|s^\prime) \rp. 
    \end{align*}
    Notice that $P^{\pi}_{\ell} (s)$ is independent of the policy in time step $\ell$ and we have
    \begin{align*}
    &\quad g (\pi_{\ell}) - g (\piE_{\ell}) \\
    &= \lp \sum_{s \in \goodS} \labs \widehat{P}^{\piE}_{h} (s) - \sum_{s^\prime \in \goodS} P^{\pi}_{\ell} (s^\prime)  \sP^{\pi} \lp s_{h} = s |s_{\ell} = s^\prime, a_{\ell} = a^{1} \rp \pi_{\ell} (a^{1}|s^\prime) \rabs - \sum_{s^\prime \in \goodS} P^{\pi}_{\ell} (s^\prime) \pi_{\ell} (a^{1}|s^\prime)   \rp
    \\
    &\quad - \lp \sum_{s \in \goodS} \labs \widehat{P}^{\piE}_{h} (s) - \sum_{s^\prime \in \goodS} P^{\pi}_{\ell} (s^\prime)  \sP^{\pi} \lp s_{h} = s |s_{\ell} = s^\prime, a_{\ell} = a^{1} \rp \piE_{\ell} (a^{1}|s^\prime) \rabs - \sum_{s^\prime \in \goodS} P^{\pi}_{\ell} (s^\prime) \piE_{\ell} (a^{1}|s^\prime)   \rp .
    \end{align*}
    For this type function, we can use \cref{lem:mn_variables_opt_regularity} to prove that
    \begin{align*}
    g (\pi_{\ell}) - g (\piE_{\ell}) &\geq \sum_{s^\prime \in \goodS} \min_{s \in \goodS} \{ P^{\pi}_{\ell} (s^\prime) \sP^{\pi} \lp s_{h} = s |s_{\ell} = s^\prime, a_{\ell} = a^{1} \rp \} \lp 1 - \pi_{\ell} (a^{1}|s^\prime)  \rp
    \\
    &= \sum_{s^\prime \in \goodS} \min_{s \in \goodS} \{  \sP^{\pi} \lp s_{h} = s |s_{\ell} = s^\prime, a_{\ell} = a^{1} \rp \} P^{\pi}_{\ell} (s^\prime) \lp 1 - \pi_{\ell} (a^{1}|s^\prime)  \rp.
    \end{align*}
    To check conditions in \cref{lem:mn_variables_opt_regularity}, we define
    \begin{align*}
        & m = n = \labs \goodS \rabs, \forall s \in \goodS, c(s) = \widehat{P}^{\piE}_{h} (s), \\
        & \forall s, s^\prime \in \goodS, A (s, s^\prime) = P^{\pi}_{\ell} (s^\prime)  \sP^{\pi} \lp s_{h} = s |s_{\ell} = s^\prime, a_{\ell} = a^{1} \rp,
        \\
        & \forall s^\prime \in \goodS, d(s^\prime) = P^{\pi}_{\ell} (s^\prime). 
    \end{align*}
    Note that $\pi \in \Pi^{\text{opt}} = \{ \pi \in \Pi: \forall h \in [H], \exists s \in \goodS, \pi_h (a^{1}|s) > 0 \}$. Combined with the reachable assumption that $\forall h \in [H], s, s^\prime \in \goodS, P_h (s^\prime |s, a^1) > 0$, we have that
    \begin{align*}
        \forall s, s^\prime \in \goodS, P^{\pi}_{\ell} (s^\prime)  > 0, \sP^{\pi} \lp s_{h} = s |s_\ell = s^\prime, a_\ell = a^{1} \rp > 0.
    \end{align*}
    Then we can obtain that $A > 0$ where $>$ means element-wise comparison. Besides, we have that
    \begin{align*}
        & \sum_{s \in \goodS} c (s) = 1 \geq \sum_{s \in \goodS} \sum_{s^\prime \in \goodS} P^{\pi}_\ell (s^\prime)  \sP^{\pi} \lp s_{h} = s |s_\ell = s^\prime, a_\ell = a^{1} \rp = \sum_{s \in \goodS} \sum_{s^\prime \in \goodS} A (s, s^\prime).
    \end{align*}
    For each $s^\prime \in \goodS$, we further have that 
    \begin{align*}
        \sum_{s \in \goodS} A (s, s^\prime) = \sum_{s \in \goodS} P^{\pi}_\ell (s^\prime)  \sP^{\pi} \lp s_{h} = s |s_\ell = s^\prime, a_\ell = a^{1} \rp = P^{\pi}_\ell (s^\prime) = d(s^\prime).  
    \end{align*}
    Thus, we have verified conditions in \cref{lem:mn_variables_opt_regularity}. By \cref{lem:mn_variables_opt_regularity}, we have that
    \begin{align*}
    g (\pi_{\ell}) - g (\piE_{\ell}) &\geq \sum_{s^\prime \in \goodS} \min_{s \in \goodS} \{  \sP^{\pi} \lp s_{h} = s |s_{\ell} = s^\prime, a_{\ell} = a^{1} \rp \} P^{\pi}_{\ell} (s^\prime) \lp 1 - \pi_{\ell} (a^{1}|s^\prime)  \rp
    \\
    &\geq \min_{s, s^\prime \in \goodS} \{  \sP^{\pi} \lp s_{h} = s |s_{\ell} = s^\prime, a_{\ell} = a^{1} \rp \} \sum_{s^\prime \in \goodS} P^{\pi}_{\ell} (s^\prime) \lp 1 - \pi_{\ell} (a^{1}|s^\prime)  \rp
    \\
    &= c_{\ell, h} \sum_{s^\prime \in \goodS} P^{\pi}_{\ell} (s^\prime) \lp 1 - \pi_{\ell} (a^{1}|s^\prime)  \rp.   
    \end{align*}
    Here $c_{\ell, h} = \min_{s, s^\prime \in \goodS} \{ \sP^{\pi} \lp s_{h} = s |s_{\ell} = s^\prime, a_{\ell} = a^{1} \rp \} > 0$. In conclusion, we prove that for each $\ell < h$,
    \begin{align}
        \text{Loss}_{h} (\pi_{1:\ell}, \piE_{\ell+1:h}) - \text{Loss}_{h} (\pi_{1:\ell-1}, \piE_{\ell:h}) \geq c_{\ell, h} \sum_{s^\prime \in \goodS} P^{\pi}_{\ell} (s^\prime) \lp 1 - \pi_{\ell} (a^{1}|s^\prime)  \rp, \label{eq:case_one_situation_two_result} 
    \end{align}
    where $c_{\ell, h} = \min_{s, s^\prime \in \goodS} \{ \sP^{\pi} \lp s_{h} = s |s_{\ell} = s^\prime, a_{\ell} = a^{1} \rp \} > 0$.
\end{itemize}

Then for Case I where $h < H$, we combine the results in \eqref{eq:case_one_situation_one_result} and \eqref{eq:case_one_situation_two_result} to obtain 
\begin{align}
    \text{Loss}_{h} (\pi_{1:h}) - \text{Loss}_{h} (\piE_{1:h}) &= \sum_{\ell=1}^{h} \text{Loss}_{h} (\pi_{1:\ell}, \piE_{\ell+1:h}) - \text{Loss}_{h} (\pi_{1:\ell-1}, \piE_{\ell:h}) \nonumber
    \\
    &= \text{Loss}_{h} (\pi_{1:h}) - \text{Loss}_{h} (\pi_{1:h-1}, \piE_{h}) + \sum_{\ell=1}^{h-1} \text{Loss}_{h} (\pi_{1:\ell}, \piE_{\ell+1:h}) - \text{Loss}_{h} (\pi_{1:\ell-1}, \piE_{\ell:h}) \nonumber
    \\
    &\geq \sum_{\ell=1}^{h-1} \text{Loss}_{h} (\pi_{1:\ell}, \piE_{\ell+1:h}) - \text{Loss}_{h} (\pi_{1:\ell-1}, \piE_{\ell:h}) \nonumber
    \\
    &\geq \sum_{\ell=1}^{h-1} c_{\ell, h} \sum_{s^\prime \in \goodS} P^{\pi}_{\ell} (s^\prime) \lp 1 - \pi_{\ell} (a^{1}|s^\prime)  \rp, \label{eq:case_one_result} 
\end{align}
where $c_{\ell, h} = \min_{s, s^\prime \in \goodS} \{ \sP^{\pi} \lp s_{h} = s |s_{\ell} = s^\prime, a_{\ell} = a^{1} \rp \}$. The penultimate inequality follows \eqref{eq:case_one_situation_one_result} and the last inequality follows \eqref{eq:case_one_situation_two_result}. 

Second, we consider Case II where $h = H$. By telescoping, we have that
\begin{align}
    &\quad \text{Loss}_{H} (\pi_{1:H}) - \text{Loss}_{H} (\piE_{1:H}) \nonumber
    \\
    &= \sum_{\ell=1}^{H} \text{Loss}_{H} (\pi_{1:\ell}, \piE_{\ell+1:H}) - \text{Loss}_{H} (\pi_{1:\ell-1}, \piE_{\ell:H}) \nonumber
    \\
    &= \text{Loss}_{H} (\pi_{1:H}) - \text{Loss}_{H} (\pi_{1:H-1}, \piE_{H}) + \sum_{\ell=1}^{H-1} \text{Loss}_{H} (\pi_{1:\ell}, \piE_{\ell+1:H}) - \text{Loss}_{H} (\pi_{1:\ell-1}, \piE_{\ell:H}). \label{eq:case_two_telescoping}
\end{align}
Similar to Case I, we also consider two situations: $\ell=H$ and $\ell<H$. We first consider the situation where $\ell<H$, which is similar to the corresponding part under Case I.

\begin{itemize}
    \item When $\ell<H$, we consider $\text{Loss}_{H} (\pi_{1:\ell}, \piE_{\ell+1:H}) - \text{Loss}_{H} (\pi_{1:\ell-1}, \piE_{\ell:H})$. The following analysis is similar to that under Case I. Note that $(\pi_{1:\ell}, \piE_{\ell+1:H})$ and $(\pi_{1:\ell-1}, \piE_{\ell:H})$ only differ in the policy in time step $\ell$. We take the policy in time step $\ell$ as variable and focus on
    \begin{align*}
        g (\pi_{\ell}) - g(\piE_{\ell}),
    \end{align*}
    where $g (\pi_{\ell}) = \text{Loss}_{H} (\pi_{1:\ell}, \piE_{\ell+1:h})$ and $g (\piE_{\ell}) = \text{Loss}_{H} (\pi_{1:\ell-1}, \piE_{\ell:h})$. Similarly, we have 
    \begin{align*}
        g (\pi_{\ell}) = \sum_{s \in \goodS} \labs \widehat{P}^{\piE}_{H} (s) - P^{\pi}_{H} (s) \rabs + \sum_{s \in \badS} P^{\pi}_{H} (s).
    \end{align*}
    With a little abuse of notation, we use $P^{\pi}_h(s)$ to denote the distributions induced by $(\pi_{1:\ell}, \piE_{\ell+1:h})$. With the \dquote{transition flow equation}, it holds that
    \begin{align*}
        &\forall s \in \goodS, P^{\pi}_{H} (s) = \sum_{s^\prime \in \goodS} P^{\pi}_{\ell} (s^\prime) \pi_{\ell} (a^{1}|s^\prime) \sP^{\pi} \lp s_{H} = s |s_{\ell} = s^\prime, a_{h} = a^{1} \rp,
        \\
        & \sum_{s \in \badS} P^{\pi}_{H} (s) = \sum_{s \in \badS} P^{\pi}_{\ell} (s) + \sum_{s^\prime \in \goodS} P^{\pi}_{\ell} (s^\prime) \lp 1 - \pi_{\ell} (a^{1}|s^\prime) \rp .
    \end{align*}
    Plugging the above two equations into $g (\pi_{\ell})$ yields that
    \begin{align*}
        g (\pi_{\ell})  &= \sum_{s \in \goodS} \labs \widehat{P}^{\piE}_{H} (s) - \sum_{s^\prime \in \goodS} P^{\pi}_{\ell} (s^\prime) \pi_{\ell} (a^{1}|s^\prime) \sP^{\pi} \lp s_{H} = s |s_{\ell} = s^\prime, a_{\ell} = a^{1} \rp \rabs
        \\
        &\quad + \sum_{s \in \badS} P^{\pi}_{\ell} (s) + \sum_{s^\prime \in \goodS} P^{\pi}_{\ell} (s^\prime) \lp 1 - \pi_{\ell} (a^{1}|s^\prime) \rp. 
    \end{align*}
    Notice that $P^{\pi}_{\ell} (s)$ is independent of the policy in time step $\ell$ and we have
    \begin{align*}
    &\quad g (\pi_{\ell}) - g (\piE_{\ell})
    \\
    &= \lp \sum_{s \in \goodS} \labs \widehat{P}^{\piE}_{H} (s) - \sum_{s^\prime \in \goodS} P^{\pi}_{\ell} (s^\prime)  \sP^{\pi} \lp s_{H} = s |s_{\ell} = s^\prime, a_{\ell} = a^{1} \rp \pi_{\ell} (a^{1}|s^\prime) \rabs - \sum_{s^\prime \in \goodS} P^{\pi}_{\ell} (s^\prime) \pi_{\ell} (a^{1}|s^\prime)   \rp
    \\
    &- \lp \sum_{s \in \goodS} \labs \widehat{P}^{\piE}_{H} (s) - \sum_{s^\prime \in \goodS} P^{\pi}_{\ell} (s^\prime)  \sP^{\pi} \lp s_{H} = s |s_{\ell} = s^\prime, a_{\ell} = a^{1} \rp \piE_{\ell} (a^{1}|s^\prime) \rabs - \sum_{s^\prime \in \goodS} P^{\pi}_{\ell} (s^\prime) \piE_{\ell} (a^{1}|s^\prime)   \rp .
    \end{align*}
    For this type function in RHS, we can use \cref{lem:mn_variables_opt_regularity} to prove that
    \begin{align*}
    g (\pi_{\ell}) - g (\piE_{\ell}) &\geq \sum_{s^\prime \in \goodS} \min_{s \in \goodS} \{ P^{\pi}_{\ell} (s^\prime) \sP^{\pi} \lp s_{H} = s |s_{\ell} = s^\prime, a_{\ell} = a^{1} \rp \} \lp 1 - \pi_{\ell} (a^{1}|s^\prime)  \rp
    \\
    &= \sum_{s^\prime \in \goodS} \min_{s \in \goodS} \{  \sP^{\pi} \lp s_{H} = s |s_{\ell} = s^\prime, a_{\ell} = a^{1} \rp \} P^{\pi}_{\ell} (s^\prime) \lp 1 - \pi_{\ell} (a^{1}|s^\prime)  \rp.
    \end{align*}
    To check conditions in \cref{lem:mn_variables_opt_regularity}, we define
    \begin{align*}
        & m = n = \labs \goodS \rabs, \forall s \in \goodS, c(s) = \widehat{P}^{\piE}_{H} (s), \\
        & \forall s, s^\prime \in \goodS, A (s, s^\prime) = P^{\pi}_{\ell} (s^\prime)  \sP^{\pi} \lp s_{H} = s |s_{\ell} = s^\prime, a_{\ell} = a^{1} \rp,
        \\
        & \forall s^\prime \in \goodS, d(s^\prime) = P^{\pi}_{\ell} (s^\prime). 
    \end{align*}
    Similar to the analysis under Case I, we obtain that $A > 0$ and
    \begin{align*}
         & \sum_{s \in \goodS} c (s) = 1 \geq \sum_{s \in \goodS} \sum_{s^\prime \in \goodS} P^{\pi}_\ell (s^\prime)  \sP^{\pi} \lp s_{H} = s |s_\ell = s^\prime, a_\ell = a^{1} \rp = \sum_{s \in \goodS} \sum_{s^\prime \in \goodS} A (s, s^\prime),
         \\
         & \forall s^\prime \in \goodS, \sum_{s \in \goodS} A (s, s^\prime) = \sum_{s \in \goodS} P^{\pi}_\ell (s^\prime)  \sP^{\pi} \lp s_{h} = s |s_\ell = s^\prime, a_\ell = a^{1} \rp = P^{\pi}_\ell (s^\prime) = d(s^\prime).
    \end{align*}
    Thus, we have verified conditions in \cref{lem:mn_variables_opt_regularity} and prove that
    \begin{align*}
    g (\pi_{\ell}) - g (\piE_{\ell}) &\geq \sum_{s^\prime \in \goodS} \min_{s \in \goodS} \{  \sP^{\pi} \lp s_{H} = s |s_{\ell} = s^\prime, a_{\ell} = a^{1} \rp \} P^{\pi}_{\ell} (s^\prime) \lp 1 - \pi_{\ell} (a^{1}|s^\prime)  \rp
    \\
    &\geq \min_{s, s^\prime \in \goodS} \{  \sP^{\pi} \lp s_{H} = s |s_{\ell} = s^\prime, a_{\ell} = a^{1} \rp \} \sum_{s^\prime \in \goodS} P^{\pi}_{\ell} (s^\prime)  \lp 1 - \pi_{\ell} (a^{1}|s^\prime)  \rp 
    \\
    &= c_{\ell, H} \sum_{s^\prime \in \goodS} P^{\pi}_{\ell} (s^\prime)  \lp 1 - \pi_{\ell} (a^{1}|s^\prime) \rp. 
    \end{align*}
    Here $c_{\ell, H} = \min_{s, s^\prime \in \goodS} \{ \sP^{\pi} \lp s_{H} = s |s_{\ell} = s^\prime, a_{\ell} = a^{1} \rp \} $. In summary, for $\ell < H$, we prove that
    \begin{align}
    \label{eq:case_two_situation_one_result}
        \text{Loss}_{H} (\pi_{1:\ell}, \piE_{\ell+1:H}) - \text{Loss}_{H} (\pi_{1:\ell-1}, \piE_{\ell:H}) \geq c_{\ell, H} \sum_{s^\prime \in \goodS} P^{\pi}_{\ell} (s^\prime)  \lp 1 - \pi_{\ell} (a^{1}|s^\prime) \rp. 
    \end{align}
    \item When $\ell = H$, we consider the term $\text{Loss}_{H} (\pi_{1:H}) - \text{Loss}_{H} (\pi_{1:H-1}, \piE_{H})$. The analysis under this situation is more complex. Note that $\pi_{1:H}$ and $(\pi_{1:H-1}, \piE_{H})$ only differs in the policy in the last time step $H$. Take the policy in time step $H$ as variable and we focus on
    \begin{align*}
        g (\pi_{H}) - g(\piE_{H}),
    \end{align*}
    where $g (\pi_{H}) = \text{Loss}_{H} (\pi_{1:H})$ and $g(\piE_{H}) = \text{Loss}_{H} (\pi_{1:H-1}, \piE_{H})$. Similarly, we can formulate $g (\pi_{H}) = \text{Loss}_{H} (\pi_{1:H})$ as
    \begin{align*}
        g (\pi_{H}) = \sum_{s \in \goodS}\lp \labs \widehat{P}^{\piE}_{H} (s) - P^{\pi}_{H} (s) \pi_H(a^1|s) \rabs + P^{\pi}_{H} (s) \lp 1 - \pi_H(a^1|s)  \rp   \rp + \sum_{s \in \badS} P^{\pi}_{H} (s).
    \end{align*}
    Note that $P^{\pi}_{H} (s)$ is independent of the policy in time step $H$ and we have that
    \begin{align*}
        g (\pi_{H}) - g(\piE_{H}) &= \sum_{s \in \goodS} \lp \labs \widehat{P}^{\piE}_{H} (s) - P^{\pi}_{H} (s) \pi_H (a^1|s) \rabs - P^{\pi}_{H} (s)   \pi_H (a^1|s)     \rp \\
        &\quad - \lp \labs \widehat{P}^{\piE}_{H} (s) - P^{\pi}_{H} (s) \piE_H (a^1|s) \rabs - P^{\pi}_{H} (s)   \piE_H (a^1|s)     \rp.
    \end{align*}
    Given estimation $\widehat{P}^{\piE}_{H} (s)$, we divide the set of good states into two parts. That is $\goodS = \gS^{\pi}_H \cup  \lp \gS^{\pi}_H \rp^c$ and $\gS^{\pi}_H \cap \lp \gS^{\pi}_H \rp^c = \emptyset$. Here $\gS^{\pi}_H  = \{s \in \goodS, \pi_H (a^1|s) \leq \min\{1, \widehat{P}^{\piE}_{H} (s) / P^{\piE}_H (s) \}  \}$. Therefore, we have that
    \begin{align*}
     g (\pi_{H}) - g(\piE_{H}) &= \sum_{s \in  \gS^{\pi}_H} \lp \labs \widehat{P}^{\piE}_{H} (s) - P^{\pi}_{H} (s) \pi_H (a^1|s) \rabs - P^{\pi}_{H} (s)   \pi_H (a^1|s)     \rp \\
     &\quad - \lp \labs \widehat{P}^{\piE}_{H} (s) - P^{\pi}_{H} (s) \piE_H (a^1|s) \rabs - P^{\pi}_{H} (s)   \piE_H (a^1|s)     \rp
        \\
        &\quad \underbrace{+  \sum_{s \in  \lp \gS^{\pi}_H \rp^c} \lp \labs \widehat{P}^{\piE}_{H} (s) - P^{\pi}_{H} (s) \pi_H (a^1|s) \rabs - P^{\pi}_{H} (s)   \pi_H (a^1|s)     \rp}_{\text{Term I}} \\
        &\quad \underbrace{- \lp \labs \widehat{P}^{\piE}_{H} (s) - P^{\pi}_{H} (s) \piE_H (a^1|s) \rabs - P^{\pi}_{H} (s)   \piE_H (a^1|s)     \rp}_{\text{Term I}}. 
    \end{align*}
    By \cref{lem:single_variable_opt}, we that $\text{Term I} + \text{Term II} \geq 0$. Then we have that
    \begin{align*}
       g (\pi_{H}) - g(\piE_{H})  &\geq \sum_{s \in  \gS^{\pi}_H} \lp \labs \widehat{P}^{\piE}_{H} (s) - P^{\pi}_{H} (s) \pi_H (a^1|s) \rabs - P^{\pi}_{H} (s)   \pi_H (a^1|s)     \rp \\
       &\quad - \lp \labs \widehat{P}^{\piE}_{H} (s) - P^{\pi}_{H} (s) \piE_H (a^1|s) \rabs - P^{\pi}_{H} (s)   \piE_H (a^1|s)     \rp.
    \end{align*}
    For each $s \in  \gS^{\pi}_H$, we consider
    \begin{align*}
        \lp \labs \widehat{P}^{\piE}_{H} (s) - P^{\pi}_{H} (s) \pi_H (a^1|s) \rabs - P^{\pi}_{H} (s)   \pi_H (a^1|s)     \rp - \lp \labs \widehat{P}^{\piE}_{H} (s) - P^{\pi}_{H} (s) \piE_H (a^1|s) \rabs - P^{\pi}_{H} (s)   \piE_H (a^1|s)     \rp.
    \end{align*}
    We aim to apply \cref{lem:single_variable_regularity} to prove that
    \begin{align*}
        &\quad \lp \labs \widehat{P}^{\piE}_{H} (s) - P^{\pi}_{H} (s) \pi_H (a^1|s) \rabs - P^{\pi}_{H} (s)   \pi_H (a^1|s)     \rp - \lp \labs \widehat{P}^{\piE}_{H} (s) - P^{\pi}_{H} (s) \piE_H (a^1|s) \rabs - P^{\pi}_{H} (s)   \piE_H (a^1|s)     \rp
        \\
        &\geq 2 P^{\pi}_{H} (s) \lp \min\{1,  \widehat{P}^{\piE}_{H} (s) / P^{\piE}_H (s)\} -  \pi_H (a^1|s)  \rp.
    \end{align*}
    To check the conditions in \cref{lem:single_variable_regularity}, we define
    \begin{align*}
        c = \widehat{P}^{\piE}_{H} (s), a = P^{\pi}_{H} (s), x = \pi_H (a^1|s).   
    \end{align*}
    It is easy to see that $c \geq 0$. Since $\pi \in \Pi^{\text{opt}} := \{ \pi \in \Pi: \forall h \in [H], \exists s \in \goodS, \pi_h (a^{1}|s) > 0 \}$, combined with the reachable assumption that $\forall h \in [H-1], \forall s, s^\prime \in \goodS, P_{h} (s^\prime|s, a^{1}) > 0$, we have that $a = P^{\pi}_{H} (s) > 0$. According to the definition of $\gS^{\pi}_H$, we have that $x =  \pi_H (a^1|s) \leq \min\{ 1,  \widehat{P}^{\piE}_{H} (s) / P^{\piE}_H (s) \} \leq  \min\{ 1,  \widehat{P}^{\piE}_{H} (s) / P^{\pi}_H (s) \} = \min\{ 1,  c / a \}$. We have verified the conditions in \cref{lem:single_variable_regularity} and obtain that
    \begin{align*}
        &\quad \lp \labs \widehat{P}^{\piE}_{H} (s) - P^{\pi}_{H} (s) \pi_H (a^1|s) \rabs - P^{\pi}_{H} (s)   \pi_H (a^1|s)     \rp - \lp \labs \widehat{P}^{\piE}_{H} (s) - P^{\pi}_{H} (s) \piE_H (a^1|s) \rabs - P^{\pi}_{H} (s)   \piE_H (a^1|s)     \rp
        \\
        &= 2 P^{\pi}_{H} (s) \lp \min\{1,  \widehat{P}^{\piE}_{H} (s) / P^{\pi}_H (s)\} -  \pi_H (a^1|s)  \rp
        \\
        &\geq 2 P^{\pi}_{H} (s) \lp \min\{1,  \widehat{P}^{\piE}_{H} (s) / P^{\piE}_H (s)\} -  \pi_H (a^1|s)  \rp, 
    \end{align*}
    where the last inequality follows that $\forall s \in \goodS, P^{\piE}_H (s) \geq P^{\pi}_H (s)$. Plugging the above inequality into $g (\pi_{H}) - g(\piE_{H})$ yields that
    \begin{align*}
        g (\pi_{H}) - g(\piE_{H}) \geq 2 \sum_{s \in  \gS^{\pi}_H}  P^{\pi}_{H} (s) \lp \min\{1,  \widehat{P}^{\piE}_{H} (s) / P^{\piE}_H (s)\} -  \pi_H (a^1|s)  \rp .
    \end{align*}
    In summary, under this situation, we prove that 
    \begin{align}
    \label{eq:case_two_situation_two_result}
        \text{Loss}_{H} (\pi_{1:H}) - \text{Loss}_{H} (\pi_{1:H-1}, \piE_{H}) \geq  2 \sum_{s \in  \gS^{\pi}_H}  P^{\pi}_{H} (s) \lp \min\{1,  \widehat{P}^{\piE}_{H} (s) / P^{\piE}_H (s)\} -  \pi_H (a^1|s)  \rp . 
    \end{align}
\end{itemize}

Under Case II where $h=H$, with \eqref{eq:case_two_telescoping}, \eqref{eq:case_two_situation_one_result}, and \eqref{eq:case_two_situation_two_result},   we have that
\begin{align}
    &\quad \text{Loss}_{H} (\pi_{1:H}) - \text{Loss}_{H} (\piE_{1:H}) \nonumber
    \\
    &= \text{Loss}_{H} (\pi_{1:H}) - \text{Loss}_{H} (\pi_{1:H-1}, \piE_{H}) + \sum_{\ell=1}^{H-1} \text{Loss}_{H} (\pi_{1:\ell}, \piE_{\ell+1:H}) - \text{Loss}_{H} (\pi_{1:\ell-1}, \piE_{\ell:H}) \nonumber
    \\
    &\geq 2 \sum_{s \in  \gS^{\pi}_H}  P^{\pi}_{H} (s) \lp \min\{1,  \widehat{P}^{\piE}_{H} (s) / P^{\piE}_H (s)\} -  \pi_H (a^1|s)  \rp + \sum_{\ell=1}^{H-1} c_{\ell, H} \sum_{s \in \goodS} P^{\pi}_{\ell} (s)  \lp 1 - \pi_{\ell} (a^{1}|s) \rp, \label{eq:case_two_result} 
\end{align}
where $c_{\ell, H} = \min_{s, s^\prime \in \goodS} \{ \sP^{\pi} \lp s_{H} = s |s_{\ell} = s^\prime, a_{\ell} = a^{1} \rp \} $. The last inequality follows $\eqref{eq:case_two_situation_one_result}$ and $\eqref{eq:case_two_situation_two_result}$.

Finally, we combine the results in \eqref{eq:case_one_result} and $\eqref{eq:case_two_result}$ to obtain that 
\begin{align*}
    f (\pi) - f(\piE) &= \sum_{h=1}^{H} \text{Loss}_{h} (\pi_{1:h}) - \text{Loss}_{h} (\piE_{1:h})
    \\
    &=  \sum_{h=1}^{H-1} \text{Loss}_{h} (\pi_{1:h}) - \text{Loss}_{h} (\piE_{1:H}) + \text{Loss}_{H} (\pi_{1:h}) - \text{Loss}_{H} (\piE_{1:H})
    \\
    &\geq \sum_{h=1}^{H-1} \sum_{\ell=1}^{h-1} c_{\ell, h} \sum_{s \in \goodS} P^{\pi}_{\ell} (s) \lp 1 - \pi_{\ell} (a^{1}|s)  \rp 
    \\
    &\quad + 2 \sum_{s \in  \gS^{\pi}_H}  P^{\pi}_{H} (s) \lp \min\{1,  \widehat{P}^{\piE}_{H} (s) / P^{\piE}_H (s)\} -  \pi_H (a^1|s)  \rp + \sum_{\ell=1}^{H-1} c_{\ell, H} \sum_{s \in \goodS} P^{\pi}_{\ell} (s)  \lp 1 - \pi_{\ell} (a^{1}|s) \rp
    \\
    &= \sum_{h=1}^{H} \sum_{\ell=1}^{h-1} c_{\ell, h} \sum_{s \in \goodS} P^{\pi}_{\ell} (s) \lp 1 - \pi_{\ell} (a^{1}|s)  \rp + 2 \sum_{s \in  \gS^{\pi}_H}  P^{\pi}_{H} (s) \lp \min\{1,  \widehat{P}^{\piE}_{H} (s) / P^{\piE}_H (s)\} -  \pi_H (a^1|s)  \rp, 
\end{align*}
where $c_{\ell, h}  = \min_{s, s^\prime \in \goodS} \{ \sP^{\pi} \lp s_{h} = s |s_{\ell} = s^\prime, a_{\ell} = a^{1} \rp \} $. The penultimate inequality follows \eqref{eq:case_one_result} and \eqref{eq:case_two_result}. In summary, we prove that for any $\pi \in \Pi^{\mathrm{opt}}$, we have
\begin{align*}
    &\quad f (\pi) - f(\piE) \\
    &\geq \sum_{h=1}^{H} \sum_{\ell=1}^{h-1} c_{\ell, h} \sum_{s \in \goodS} P^{\pi}_{\ell} (s) \lp 1 - \pi_{\ell} (a^{1}|s)  \rp + 2 \sum_{s \in  \gS^{\pi}_H}  P^{\pi}_{H} (s) \lp \min\{1,  \widehat{P}^{\piE}_{H} (s) / P^{\piE}_H (s)\} -  \pi_H (a^1|s)  \rp
    \\
    &\geq c (\pi) \lp \sum_{h=1}^{H} \sum_{\ell=1}^{h-1} \sum_{s \in \goodS} P^{\pi}_{\ell} (s) \lp 1 - \pi_{\ell} (a^{1}|s)  \rp + \sum_{s \in  \gS^{\pi}_H}  P^{\pi}_{H} (s) \lp \min\{1,  \widehat{P}^{\piE}_{H} (s) / P^{\piE}_H (s)\} -  \pi_H (a^1|s)  \rp  \rp.
\end{align*}
Here $c (\pi) = \min_{1 \leq \ell < h \leq H} c_{\ell, h} =  \min_{1 \leq \ell < h \leq H, s, s^\prime \in \goodS} \{ \sP^{\pi} \lp s_{h} = s |s_{\ell} = s^\prime, a_{\ell} = a^{1} \rp \}$. Since $\widebar{\pi} \in \Pi^{\mathrm{opt}}$, it holds that
\begin{align*}
    &\quad f (\widebar{\pi}) - f(\piE) \\
    &\geq c (\widebar{\pi}) \lp \sum_{h=1}^{H} \sum_{\ell=1}^{h-1} \sum_{s \in \goodS} P^{\widebar{\pi}}_{\ell} (s) \lp 1 - \widebar{\pi}_{\ell} (a^{1}|s)  \rp + \sum_{s \in  \gS^{\widebar{\pi}}_H}  P^{\widebar{\pi}}_{H} (s) \lp \min\{1,  \widehat{P}^{\piE}_{H} (s) / P^{\piE}_H (s)\} -  
    \widebar{\pi}_H (a^1|s)  \rp  \rp. 
\end{align*}
Combined with \eqref{eq:ail_objective_pi_bar_minus_piE}, we obtain
\begin{align*}
    c (\widebar{\pi}) \lp \sum_{h=1}^{H} \sum_{\ell=1}^{h-1} \sum_{s \in \goodS} P^{\widebar{\pi}}_{\ell} (s) \lp 1 - \widebar{\pi}_{\ell} (a^{1}|s)  \rp + \sum_{s \in  \gS^{\widebar{\pi}}_H}  P^{\widebar{\pi}}_{H} (s) \lp \min\{1,  \widehat{P}^{\piE}_{H} (s) / P^{\piE}_H (s)\} -  
    \widebar{\pi}_H (a^1|s)  \rp  \rp \leq \varepsilon_{\ail},
\end{align*}
which completes the whole proof.

\end{proof}

\end{document}